\documentclass[nonacm, sigconf]{acmart}
\renewcommand\footnotetextcopyrightpermission[1]{} 
\usepackage{subcaption}
\usepackage{makecell}
\usepackage{mathtools}
\usepackage{amsthm}
\usepackage{amssymb}
\usepackage{esvect}
\usepackage{bbm}
\usepackage[export]{adjustbox}
\usepackage{tabu}
\usepackage{diagbox}
\usepackage{array}
\newcolumntype{P}[1]{>{\centering\arraybackslash}p{#1}}
\newcolumntype{M}[1]{>{\centering\arraybackslash}m{#1}}
\usepackage{enumitem}
\usepackage[linesnumbered,ruled]{algorithm2e}

\SetKwInput{kwInit}{Initialization}

\usepackage{rotating, bigstrut}
\theoremstyle{definition}

\acmDOI{}
\begin{document}

\title{Characterizing the Decision Boundary of Deep Neural Networks}

\author{Hamid Karimi}
\affiliation{%
  \institution{Michigan State University}}
\email{karimiha@msu.edu}

\author{Tyler Derr}
\affiliation{%
  \institution{Michigan State University}}
\email{derrtyle@msu.edu}

\author{Jiliang Tang}
\affiliation{%
  \institution{Michigan State University}}
\email{tangjili@msu.edu}

\begin{abstract}

Deep neural networks and in particular, deep neural classifiers have become an integral part of many modern applications. Despite their practical success, we still have limited knowledge of how they work and the demand for such an understanding is evergrowing. In this regard, one crucial aspect of deep neural network classifiers that can help us deepen our knowledge about their decision-making behavior is to investigate their decision boundaries. Nevertheless, this is contingent upon having access to samples populating the areas near the decision boundary. To achieve this,  we propose a novel approach we call \textbf{Deep} \textbf{D}ecision boundary \textbf{I}nstance \textbf{G}eneration (\textbf{DeepDIG}). DeepDIG utilizes a method based on adversarial example generation as an effective way of generating samples near the decision boundary of any deep neural network model. Then, we introduce a set of important principled characteristics that take advantage of the generated instances near the decision boundary to provide multifaceted understandings of deep neural networks. We have performed extensive experiments on multiple representative datasets  across various deep neural network models and characterized their decision boundaries. \emph{The  code is publicly available at \url{https://github.com/hamidkarimi/DeepDIG/}.} 

\end{abstract}


\maketitle

\thispagestyle{plain}
\fancyfoot{}
\section{Introduction}
\label{sec:introduction}
 Thanks to available massive data and high-performance computation technologies (e.g. GPUs), deep neural networks (DNNs) have become ubiquitous models in many decision-making systems. Notwithstanding the high performance that DNNs have brought about in many domains~\cite{karimi-etal-2018-multi,karimi-tang-2019-learning,deception-hamid,e2ecad,karimicongress,karimi2019roadmap,fan2019deep,derr2019deep}, our understating of them is still very limited and lacking in some respects. This is primarily due to the black-box nature of DNNs where the decisions they are making are opaque and elusive.  In this regard, one crucial aspect of DNN classifiers that yet remains fairly unknown is their decision boundaries and its geometrical properties. If we want to continue DNNs' usage for critical applications, understating their decision boundaries and decision regions is essential. This is especially important for safety and security applications such as e.g., self-driving cars~\cite{huval2015empirical} whose deep models are vulnerable to erroneous instances near their decision boundaries~\cite{pei2017deepxplore}.

Compared to other aspects of DNN e.g.,  optimization landscape~\cite{bottou2018optimization}, systematic characterization of the decision boundary of DNNs, despite its importance, is still in the early stages of the study. The main challenge hindering in-depth analysis and investigation of decision boundaries of DNNs is generating instances that are simultaneously close to the decision boundary and resemble real instances\footnote{In this paper, we use the terms \emph{instance}, \emph{sample}, and \emph{example} interchangeably.}. We call such instances as \textit{borderline instances}. The difficulty of generating borderline instances stems from the fact that the input space of DNNs is of high dimension e.g., $\mathbb{R}^{784}$ in the case of simple grayscale MNIST images, which makes searching for instances close to the decision boundary a non-trivial and challenging task. 

\begin{figure}[t]
    \centering
    \includegraphics[width=\columnwidth]{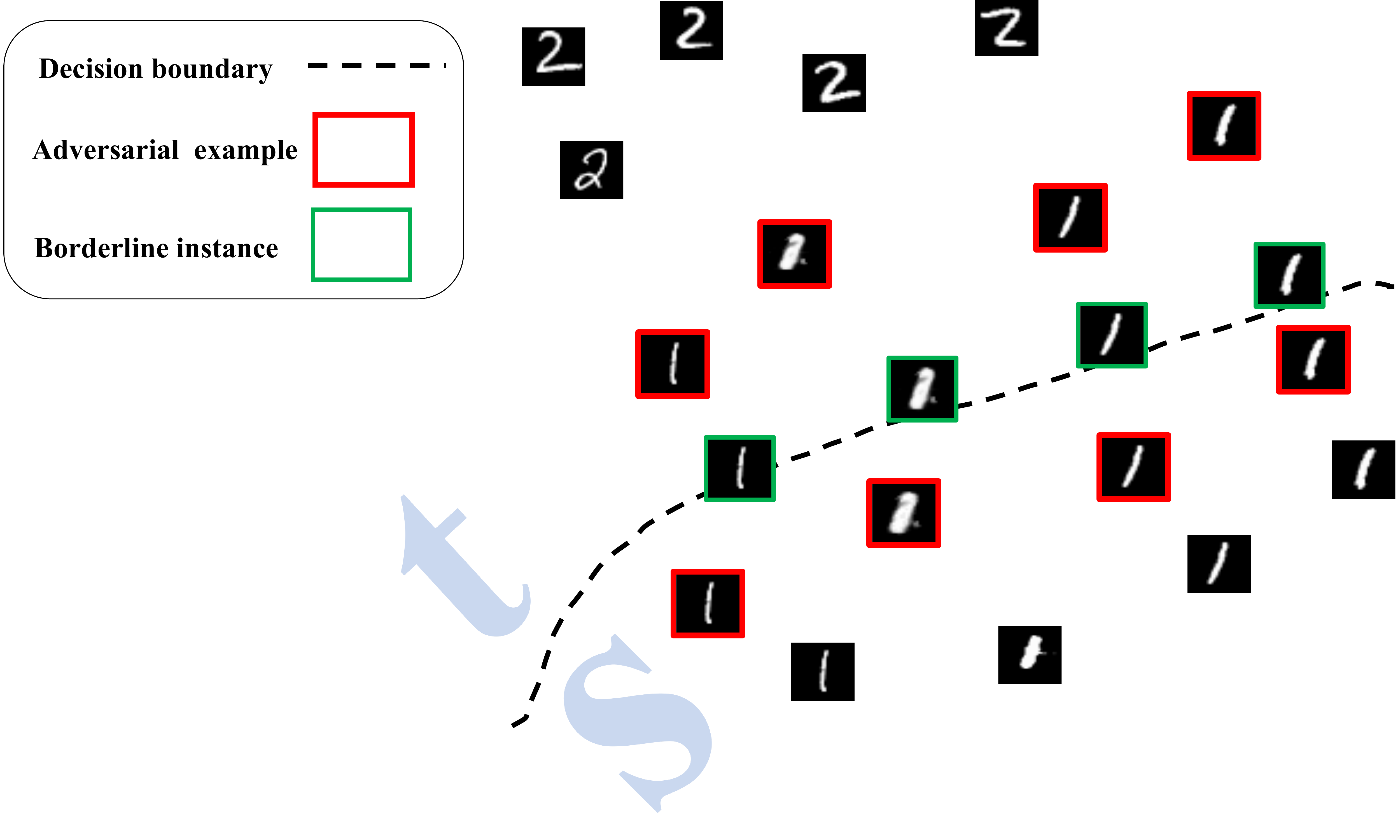}
    \caption{A high-level illustration of Deep Decision boundary Instance Generation (DeepDIG). For a given pre-trained deep neural network model and two classes $s$ and $t$, DeepDIG tries to find instances as close as possible to the decision boundary between the two classes $s$ and $t$.} 
    \label{fig:overalldeepdig}
\end{figure}

To solve this challenge, we propose a novel framework called Deep Decision boundary Instance Generation (DeepDIG) a high-level illustration of which is shown in Figure~\ref{fig:overalldeepdig}. Given two classes of samples as well as a pre-trained DNN model, DeepDIG is optimized to generate borderline instances near a decision boundary between two classes i.e., generating instances whose classification probabilities for two classes are as close as possible. DeepDIG utilizes an autoencoder-based method to generate targeted adversarial examples at the two sides of the decision boundary between two classes and further employs a binary search based algorithm to refine and generate borderline instances. Moreover, we leverage the borderlines instances generated by DeepDIG and investigate two notable characteristics concerning the decision boundary of DNNs. First, we measure the complexity of the decision boundary in the input space. To this end, we measure the classification oscillation along the decision boundary between two classes and devise a novel metric offering us a form of \textit{geometrical complexity} of the decision boundary. Second, we investigate the decision boundary in the embedding space learned by a DNN i.e., we measure the complexity of the decision boundary once it is projected in the embedding space. To this end, we take advantage of the linear separability property of DNNs and propose a metric capturing the complexity of the decision boundary in the embedding space. We found consistency between these two complexity measures.

DeepDIG and further characterization of the decision boundary of DNNs are novel with respect to the existing studies~\cite{He2018DecisionBA,fawzi2018empirical,li2018decision,moosavi2019robustness,yousefzadeh2019investigating,alfarra2020on} in the following ways. First, the previous work investigated the decision boundary merely through the lens of adversarial examples and considered adversarial examples as a type of borderline instances. However, as we show later, given the definition of the decision boundary, adversarial examples while being close to the decision boundary are not borderline instances. In comparison, DeepDIG, while using adversarial example generation, goes beyond adversarial examples and generates instances that \textit{by design} are ensured to be as close as possible to the decision boundary. Second, we do not make any assumption on the DNNs being investigated and DeepDIG can be applied to any pre-trained DNN classifier. Third, instead of investigating the decision boundary of a DNN from the perspective of a single instance and/or its neighborhood, we characterize a decision boundary between two classes as a whole and shed light on its properties by taking advantage of a collection of instances populating that decision boundary.  Through extensive experiments across three datasets, namely MNIST~\cite{lecun1998gradient}, FashionMNIST~\cite{xiao2017fashion}, and CIFAR10~\cite{krizhevsky2009learning}, we verify the working of DeepDIG and investigate various pre-trained DNNs.

In summary, our major contributions are as follows.

\begin{itemize}
    \item We propose a novel framework DeepDIG to generate instances near the decision boundary of a  given pre-trained neural network classifier. 
    \item We present several use-cases of DeepDIG to characterize decision boundaries of  DNNs  which help us to deepen our understating of DNNs. 
\end{itemize}

 The rest of the paper is organized as follows. In Section~\ref{sec:problem}, we present the notations and define the problem. In Section~\ref{sec:framework}, we present the proposed framework DeepDIG. Section~\ref{sec:characteristics} includes how we can use DeepDIG to characterize the decision boundary of a DNN. Experimental settings and details of investigated DNNs, as well as datasets, will be presented in Section~\ref{sec:experiments}. Experimental results and discussions will be presented in Section~\ref{sec:results}. We review the related work in Section~\ref{sec:related} followed by concluding remarks in Section~\ref{sec:conclusion}.

\section{Definitions and Problem Statement}
\label{sec:problem}
In this section, we introduce the basic notations and definitions as well as the problem statement.

\textbf{Notations.}  
Let $f:\mathbb{R}^D \xrightarrow{} \mathbb{R}^c$ denote a pre-trained  $c$-class deep neural network classifier where $D$ is the dimension of input space.  Further, let $\mathcal{F}(x) \in \mathbb{R}^d$ denote the embedding space learned by $f$ where $d$ is the dimension of this space and usually $d \ll D$. We assume that the last layer of $f$ is a $d \times c$ fully connected layer without any non-linear activation function which maps the embeddings to a score vector of size $c$ i.e., $\mathbb{R}^c$. Then, for a  sample $x \in \mathbb{R}^D$, the classification outcome is $\mathcal{C}(x)=\textit{argmax}_{k}f_k (x)$ where $f_k$ is the score of $k{\text{-th}}$ class ($ 1 \leq k \leq c$). We assume scores are calculated by applying the softmax function on the output of last layer of $f$. In other words, $f_k (x_i)$ denotes the prediction probability of classifying $x_i$ as $s$. Finally, let $X=\{x_1, x_2 \cdots x_n\}$ denote a dataset of instances $x_i \in \mathbb{R}^D$ associated with ground-truth labels $Y=\{y_1, y_2 \cdots y_n\}$ where $y_i \in [1,c]$. 

\textbf{Decision Region and Decision Boundary}. The classifier $f$ partitions the space $\mathbb{R}^D$ into $c$ decision regions $r_1, r_2 \cdots r_c$ where for each $x \in r_i$  we have $\mathcal{C}(x)=i$. Now, in line with previous studies~\cite{fawzi2018empirical,li2018decision}, the decision boundary between classes $s$ and $t$ ($t,s \in [1,c]$) is defined as $b_{s,t}=\{v \in \mathbb{R}^D: f_s (v)=f_t (v)\}$. In other words, the deep neural network classifier (and as the matter of fact any other classifier) is ``confused'' about the labels of the instances on the decision boundary between classes $s$ and $t$.

\textbf{Problem Statement.} 
\textit{Given a pre-trained deep neural network model ${f}(.)$, a dataset $X$, and two classes $s$ and $t$, we aim to generate instances near the decision boundary between decision regions $r_s$ and $r_t$. Further, we intend to leverage the borderline instances as well as other generated and original samples to delineate the behavior  of model ${f}(.)$ }

\section{Proposed Framework (DeepDIG)}
\label{sec:framework}

Given a pre-trained DNN, we intend to generate borderline instances satisfying two important criteria: 

\begin{enumerate}[label=(\alph*)]
    \item  They need to be as near as possible to the decision boundary between two classes i.e., their DNN's classification scores (probabilities) be as close as possible. This is basically to follow the definition of decision boundary--Refer to Section~\ref{sec:problem}.
    \item Borderline instances need to be similar to the original (real) instances.
\end{enumerate}

The second criterion is imposed because of two major reasons. First, we are interested in investigating DNNs and their decision boundaries in the presence of realizable and non-random corner cases which in practice can have major safety and security consequences~\cite{pei2017deepxplore,yuan2017adversarial,dube2018high}. Second, essentially a DNN carves out decision regions (and decision boundaries) by learning on its \emph{training data} not other random instances in the space $\mathbb{R}^D$. Therefore, random borderline instances (i.e., those which are not similar to real instances) occupy some parts of the input space which are not practically appealing for further decision boundary characterization in Section~\ref{sec:characteristics}. Note that, in spirit, the second criterion is similar to what that is followed in adversarial example generation~\cite{xu2019adversarial} where adversarial examples are required to be similar to real (benign) samples.

\begin{figure*}[!htb]
    \centering
        \includegraphics[scale=0.125]{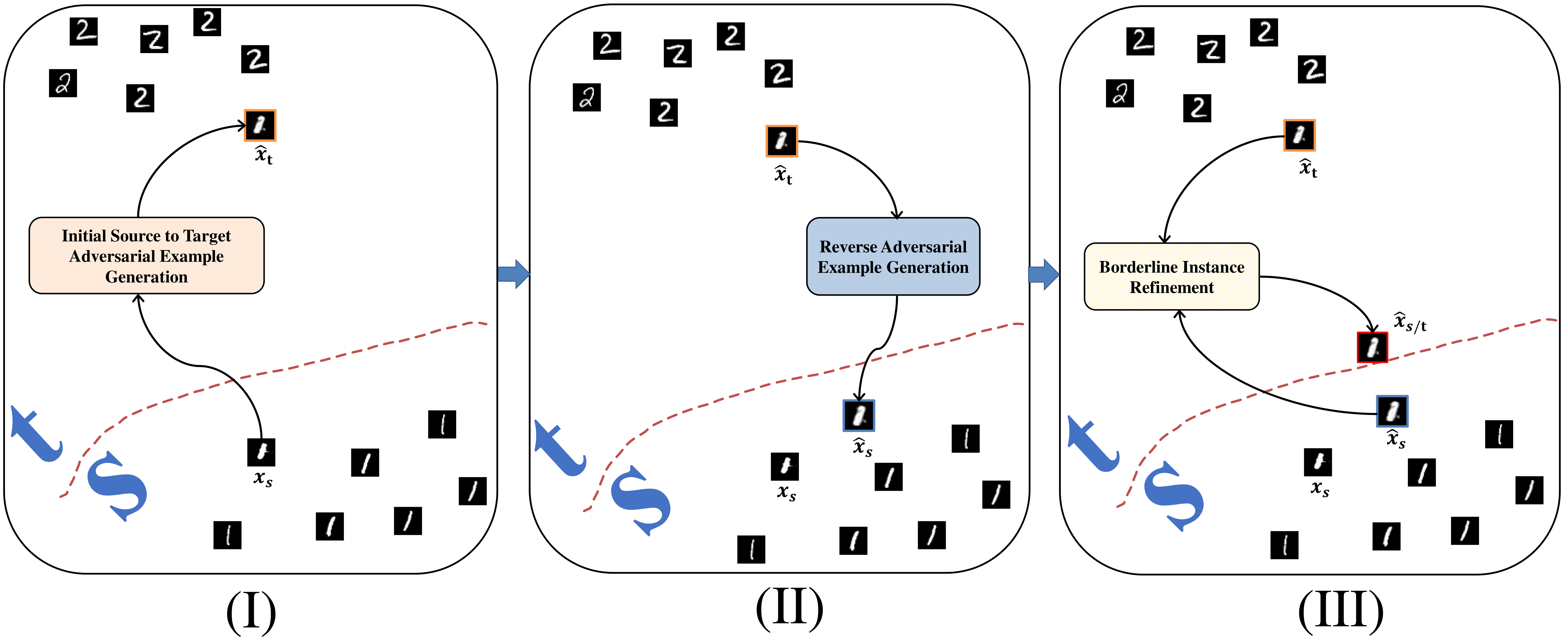}
    \caption{The proposed framework Deep  Decision boundary  Instance Generation (DeepDIG). It consists of three components. In component (I), targeted adversarial examples of source instances are generated ($\hat{x}_t$). In component (II),  from adversarial examples of component (I), a new set of adversarial examples are generated ($\hat{x}_s$) which are classified as $s$. Finally, in component (III), a binary search based algorithm is employed to refine and identify the borderline instances near the decision boundary.    }
    \label{fig:deepdig}
\end{figure*}

To generate borderline samples satisfying the above criteria, we propose the framework Deep Decision boundary Instance
Generation (DeepDIG), which is illustrated in Figure~\ref{fig:deepdig}. As shown in this figure, DeepDIG includes three major components. In the first component, we utilize an autoencoder-based method to generate targeted adversarial instances from a source class to a target class--See Figure~\ref{fig:deepdig} (I). In the second component, we employ another autoencoder-based adversarial instance generation on the first component's adversarial examples and consequently generate new adversarial instances predicted as the source class-- See Figure~\ref{fig:deepdig} (II). Adversarial samples generated in the first and second components of DeepDIG are at the opposite sides of a decision boundary between a source and a target class, and more importantly, these samples are \emph{by design} close to the decision boundary. Hence, in the third component of DeepDIG, we feed these two sets of adversarial samples to a module named Borderline Instance Refinement which based on a binary search algorithm refines and generates borderline instances being sufficiently close to the decision boundary--See Figure~\ref{fig:deepdig} (III). Next, we explain each component in detail.  

\subsection{Component (I): Initial Source to Target 
Adversarial Example Generation}
 One way to obtain samples enjoying the criterion (b) mentioned above is via targeted adversarial examples which are \textit{slightly distorted} versions of real instances and are misclassified by a DNN~\cite{xu2019adversarial}. As will be discussed shortly, targeted adversarial example generation paves the way to meet the criterion (a) as well. Hence, as the first step towards generating borderline instances, we generate targeted adversarial examples from real instances of class $s$ to be misclassified as class $t$. To generate such adversarial examples, we utilize a simple yet effective approach using an autoencoder-based method formulated in the following loss function:

\begin{equation}
    \mathcal{L}_{I} = \sum_{\forall x_s} \bigg(   ||x_s-  A_{1}(x_s)||_2^2 +\alpha \times  CE \big(f(A_{1}(x_s)\big), \vv{{t}}  ) \bigg)
\label{eq:deepdig1}
\end{equation}
\noindent where $x_s$ denotes a sample belonging to class $s$, $A_{1}(.)$ is an autoencoder reproducing its input sample (here $x_s$), $ \vv{{t}}$ is a $c$-dimensional one-hot vector having its $t$-th entry  equal to 1 and the rest to 0, $CE$ is the class entropy loss function\footnote{\url{ https://en.wikipedia.org/wiki/Cross_entropy}}, and $\alpha$ is a hyperparamter controlling the trade-off between reconstruction error and adversarial example generation. The loss function $\mathcal{L}_{I}$ is optimized along with other components of DeepDIG. Also, for convenience, we show the output of $A_{1}(x_s)$ as $\hat{x}_t$ signifying its mis-classification as class $t$.   

Eq.~\ref{eq:deepdig1} has two parts. The first part (reconstruction error) ensures keeping the generated adversarial example, $\hat{x}_t$, as close as possible to the real sample $x_s$ i.e., satisfying the criterion (b). The second part of Eq.~\ref{eq:deepdig1} attempts to misclassify the generated instance i.e., placing it outside the decision region $r_s$. Therefore, optimizing $\mathcal{L}_{I}$ makes the generated adversarial examples close to the decision boundary between two classes as has been shown before as well~\cite{He2018DecisionBA,jaouen2018zonnscan,fawzi2018empirical}. Nevertheless, given the definition of the decision boundary in Section~\ref{sec:problem}, samples $\hat{x}_t$ are not `sufficiently' close to the decision boundary between classes $s$ and $t$ and thus criterion (a) is not fully met yet. Hence, DeepDIG is equipped with two other components to generate proper borderline samples.   

\subsection{Component (II): Reverse Adversarial Example Generation}
As mentioned before, an adversarial example $\hat{x}_t$ is outside of the decision region $r_s$ and is near the decision boundary between classes $s$ and $t$. Aiming at generating samples even closer to the decision boundary, we leverage another targeted adversarial example generation applied on samples $\hat{x}_t$. We call this component \emph{Reverse} Adversarial Example Generation since we generate adversarial examples of the first component's adversarial examples\footnote{Technically, examples generated in component (II) are not \emph{adversarial} since they are correctly classified as $s$. However, for the sake of simplicity in the presentation, we abuse the definition and keep referring to them as \emph{adversarial examples}.}.  The loss function is as follows.

\begin{equation}
    \mathcal{L}_{II} = \sum_{\forall \hat{x}_t} \bigg( ||\hat{x}_t- A_{2}(\hat{x}_t)||_2^2 +\alpha \times  CE \big(f( A_{2}(\hat{x}_t) ), \vv{{s}}  \big)\bigg)
\label{eq:deepdig2}
\end{equation}

\noindent where $A_{2}(.)$ is another autoencoder to reproduce its the input sample here (here $\hat{x}_t$)\footnote{Note that autoencoders $A_1$ ad $A_2$ has the same architecture while they subscripted here to signify their distinct parameters in components (I) and (II) of DeepDIG, respectively.}, $ \vv{{s}}$ is a $c$-dimensional one-hot vector having its  $s$-th entry equal to $1$ and the rest to 0, and $\alpha$ is a hyperparameter controlling the trade-off between reconstruction error and adversarial example generation. $\mathcal{L}_{II}$ is optimized along with other components of DeepDIG. Again for convenience, we show the output of $A_{2}(\hat{x}_t)$ as $\hat{x}_s$ signifying its classification as class $s$.  Next, we explain how to we utilize adversarial examples $\hat{x}_t$ and $\hat{x}_s$ to generate borderline instances that are sufficiently near the decision boundary between classes $s$ and $t$.

\subsection{Component (III): Borderline Instance Refinement}

As mentioned before, the high dimensional nature of input space in DNNs causes a big challenge for generating instances that are simultaneously near the decision boundary and are similar to real instances i.e., satisfying criteria (a) and (b), respectively. More specifically, randomly generating samples in $\mathbb{R}^D$ even for small $D$ (e.g., 100) has an extremely low chance of producing legitimate and similar-to-real instances let alone yielding those near the decision boundary. Moreover, simply perturbing the real instances aiming at finding borderline samples induces a huge number of directions to consider and is prohibitively infeasible. Nevertheless, thanks to components (I) and (II) of DeepDIG, searching for borderline instances is now facilitated. This is because we generate two sets of adversarial examples (i.e., $\hat{x}_s$  and $\hat{x}_t$ through components (I) and (II), respectively) which are \textit{by design} close to a decision boundary between two classes and populates both sides of the decision boundary. More importantly and again by design, they are similar to real instances. Hence, the Borderline Instance Refinement component of DeepDIG employs a binary search algorithm between the trajectory connecting a pair of samples $\hat{x}_s$ and $\hat{x}_t$ aiming at finding the desired borderline instance.  Algorithm~\ref{alg:middle} shows the proposed approach for borderline instance refinement and is explained in the following.

As input, this algorithm takes generated adversarial examples $\hat{x}_t$ and $\hat{x}_s$ belonging to classes $t$ and $s$, respectively, i.e., two instances from distinct sides of the decision boundary of the DNN model $f$--See Figure~\ref{fig:deepdig} (III). The algorithm performs a binary search to find a middle point $\hat{x}_{m}$ whose difference in probabilities belonging to classes $t$ and $s$ is less than a small threshold (e.g., 0.0001) that we denote as $\beta$.  In the algorithm, this is given by $| f(x_m)_s - f(x_m)_t| < \beta$ (line~\ref{alg:beta}). This thresholding is in line with the definition of decision boundary between two classes where instances should have equal classification probabilities for classes $s$ and $t$. In other words, the DNN is `confused' about the class of such instances. We note that  Algorithm~\ref{alg:middle} might fail to find such an instance if the middle point (i.e., $x_m$) deviates from decision regions of classes $s$ or $t$-- See line~\ref{alg:fail}. Nevertheless, the proposed Algorithm~\ref{alg:middle} is empirically quite effective at identifying borderline instances as it will be demonstrated in the experiments (Section~\ref{sec:experiments}).

\begin{algorithm}
\caption{The proposed Borderline Instance 
Refinement algorithm}
\label{alg:middle}
\KwData{\\\hspace{2em} Instances $\hat{x}_s$ and $\hat{x}_t$, threshold $\beta$,\\  \hspace{2em} pre-trained DNN model $f$}
\kwInit{\\  \hspace{2em}       {$x_l =\hat{x}_s$};  {$x_r =\hat{x}_t$};}
\While{True}{
$x_m =\frac{x_l + x_r}{2}$\\
  \uIf{$\mathcal{C}(x_m) = s$}{
    $x_l=x_m$
  }
  \uElseIf{$\mathcal{C}(x_m) = t$}{
    ${x}_r=x_m$
  }
  \Else{
    \textbf{return} {``Fail"} \label{alg:fail}
  }
  \uIf{$| f_s(x_m) - f_t({x_m})| <  \beta$ } 
  {
  \label{alg:beta}
    $\hat{x}_{s/t} = x_m$\\
    \textbf{return} $\hat{x}_{s/t}$ \label{alg:success}
  }

}
\end{algorithm}

\textbf{Remark.} Before introducing the decision boundary characteristics in the next section, we need to clarify a matter. To fully characterize the decision boundary between two classes --say $a$ and $b$-- one needs to generate borderline samples for both $a$ and $b$. More specifically, following our notations and DeepDIG mechanism demonstrated in Figure~\ref{fig:deepdig}, once we apply DeepDIG for ($s$,$t$)=($a$, $b$) and then ($s$,$t$)=($b$, $a$).  Hence, to fully characterize the decision boundary between two classes, we obtain two sets of borderline instances.

\section{Decision Boundary Characteristics}
\label{sec:characteristics}

As mentioned before, one of the challenges of principled and in-depth analysis of the decision boundary of DNNs is the inaccessibility of samples close to the decision boundary which would be similar to real samples as well.  Nevertheless, DeepDIG addresses this challenge and provides us with a systematic way to generate borderline instances near the decision boundary between two classes. This opens us a door to understand and characterize the decision boundary of DNNs in a better way.  To this end, we introduce several metrics
informing us about the different characteristics of the decision boundary of a deep neural network. We group the characterization measures into two distinct groups: decision boundary complexity in the input space (Section~\ref{sec:geometrical}) and decision boundary complexity in the embedding space
 (Section~\ref{sec:linear}).
 
 \subsection{Decision Boundary Complexity in the Input Space}
 \label{sec:geometrical}
As previously shown~\cite{fawzi2018empirical}, DNN classifiers tend to carve out complicated decision regions in the input space to be able to discriminate input samples of different classes. These decision regions are highly non-convex and have highly non-linear decision boundaries. The question is how we can measure the geometrical complexity (non-convexity)  of the decision regions in the input space? Thus far, the practical and systematic investigation of the complexity and non-convexity of decision boundaries (and decision regions) of any DNN (regardless of its architecture, model size, etc)  has been a challenging task because there has not been an efficient method generating samples populating the decision boundary of a DNN. Fortunately, DeepDIG provides us with a method generating borderline samples quite effectively (will be shown in the experiment section). Hence, we now utilize the borderline samples to measure the degree of complexity (or non-convexity) of a decision boundary in the input space. To this end, we devise a new metric described as follows. 

\begin{figure}[!htb]
     \centering
     \includegraphics[width=\columnwidth]{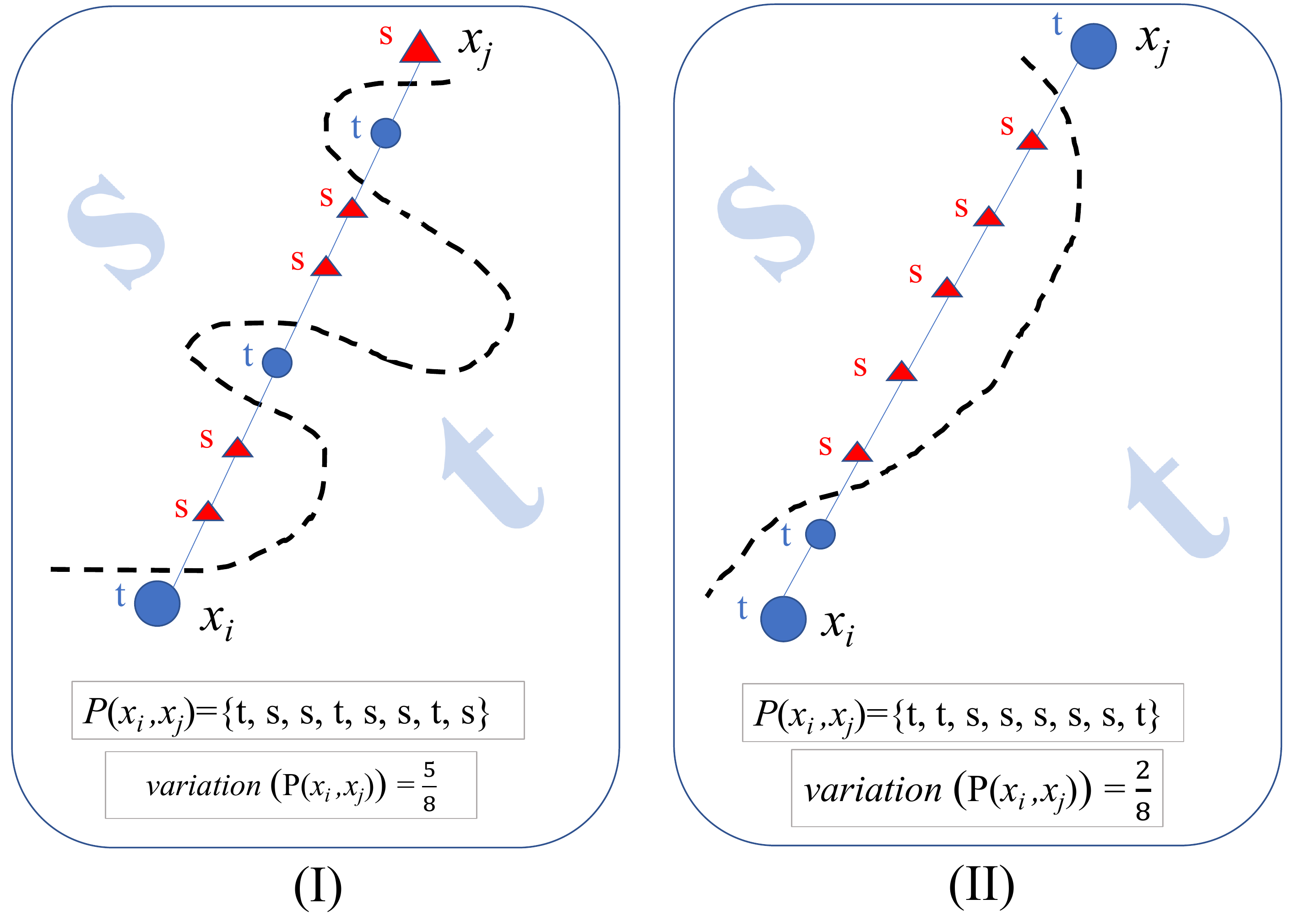}
     \caption{An illustration of capturing geometrical complexity of a decision boundary through measuring the \emph{oscillation} between two decision (classification) regions $r_s$ and $r_t$ for samples on a borderline trajectory. Decision boundary in case (I) is geometrically more complex than that of (II).}
     \label{fig:wiggle}
 
 \end{figure}

 Let $x_i$ and $x_j$ denote two borderline instances for the decision boundary between classes $s$ and $t$.  Further, we define a trajectory $\mathcal{T}(t)$ between $x_i$ and $x_j$  as $\mathcal{T}(t) = t \times x_i + (1-t) \times x_j$ where $ 0 \leq t \leq 1$. Then, for $m$ values of $t$ we interpolate $m$ instances along the trajectory  $\mathcal{T}(t)$ denoted as $\mathcal{I}_{(x_i,x_j)}= \{x_{p_1}, x_{p_2} \cdots x_{p_m}\}$. We retrieve the DNN's classification  outcomes for interpolated samples $\mathcal{I}_{(x_i,x_j)}$ and denote them as $\mathcal{P}_{(x_i,x_j)} = \{ \mathcal{C}(x_{p_1}), \mathcal{C}(x_{p_2}) \cdots \mathcal{C}(x_{p_m})\}$. We define the \textit{oscillation} of classification outcomes of interpolated samples $\mathcal{I}_{(x_i,x_j)}$, denoted as $\mathcal{O}_{(x_i,x_j)}$, as follows.

   \begin{equation}
        \begin{aligned}
            \mathcal{O}_{(x_i,x_j)} = \frac{1}{|\mathcal{I}_{(x_i,x_j)}|}  \sum_{k=1}^{m-1} \mathbbm{1}  (\mathcal{C}(x_{p_k}) \neq  \mathcal{C}(x_{p_{k+1}}) )
        \end{aligned}
        \label{eq:wiggle}
    \end{equation}

    \noindent where $\mathbbm{1}$ denotes the indicator function\footnote{\url{https://en.wikipedia.org/wiki/Indicator_function}}.  $\mathcal{O}_{(x_i,x_j)}$ essentially measures the `variation' along the trajectory connecting two borderline instances $x_i$ and $x_j$. A higher value for this metric indicates more alternating between decision regions $r_s$ and $r_t$ and vice versa. This is pictorially illustrated in Figure~\ref{fig:wiggle}. We can also interpret  $\mathcal{O}$ as a proxy informing us about the \emph{smoothness} of the decision boundary between two classes. 
    
    {}
    Now let $B=\{x_1,x_2 \cdots x_n\}$ denote all borderline instances. For each borderline sample $x_i \in B$, we randomly select $k$ other borderline samples whose set is denoted as $S_{x_i}$. Then, we record the average $\mathcal{O}_{(x_i,x_j)}$ for $x_i \in B$ and $x_j \in S_{x_i}$. Eventually we report the average $\mathcal{O}$ across all borderline samples $x_i$ as the final value of this measure, which we call \textbf{IDC} (\textbf{I}nput space \textbf{D}ecision boundary \textbf{C}omplexity) and is formulated in Eq.~\ref{eq:gcdb}.
 
   \begin{equation}
        \begin{aligned}
            \text{IDC} = \frac{1}{n \times k}  \sum_{x_i \in B}\sum_{x_j \in S_{x_i}}   \mathcal{O}_{(x_i,x_j)}
        \end{aligned}
        \label{eq:gcdb}
    \end{equation}

\subsection{Decision Boundary Complexity in the Embedding Space}
\label{sec:linear}

IDC measure developed in Section~\ref{sec:geometrical} looks into the decision boundary complexity in the input space. In this part, we focus on the decision boundary in the embedding space as defined next.

\textbf{Decision boundary in the embedding space}. We abuse the definition of the decision boundary in Section~\ref{sec:problem} and define the decision boundary in the embedding space as $be_{s,t}=\{z \in \mathbb{R}^d: f_s (f^{-1}(z))=f_t (f^{-1}(z))\}$ where $f^{-1}(z)$ denotes a machinery that returns a sample whose embedding is $z$. We do not have a direct access to $f^{-1}$. Rather, in practice, for a collection of samples $v \in \mathbb{R}^D$ (i.e., training and test samples as well as generated borderline instances) we have pairs of ($v,z$) where through accessing the DNN $f$ we know that $f^{-1}(z)=v$.  

Now two interesting questions emerge regarding the decision boundary in an embedding space learned by a DNN. First, if we project borderline samples in the embedding space will they still be in the area separating two classes? In other words,  will borderline instances be still near the decision boundary in the embedding space? Second, how we can measure the complexity of the decision boundary in the embedding space? To be more specific, does the decision boundary complexity in the input space manifest itself in the embedding space as well? 
Aiming at answering these questions, in this part, we propose two measures quantifying decision boundary characterization in the embedding space. To achieve this, we utilize an intriguing property of DNNs described in the following.

One of the fundamental properties of DNNs is their representation power where through a sophisticated combination of layer-wise and non-linear transformations they can map their complicated high dimensional input data to a low-dimension embedding space. It has been shown and will show in Section~\ref{sec:experiments} that in the embedding space data points from different classes can be linearly separated~\cite{ho2002complexity,mallat2016understanding}. Not that the capability to learn linear separable embeddings by a DNN is closely related to the generalization power of that DNN~\cite{li2018decision}. Hence, should a DNN manage to learn linearly separable embeddings on the training set, it is expected to do so on unseen data samples such as borderline instances\footnote{Note that DeepDIG treats a model $f$ as a pre-trained model whose parameters have been optimized and learned previously. Thus, as far as  model $f$ is concerned, generated borderline samples are still considered unseen data points.}. With this discussion in mind, we train a linear model on embeddings of training samples of classes $s$ and $t$ and the following measures are considered to characterize the decision boundary in the embedding space. We call these measures \textbf{EDC} (\textbf{E}mbedding space \textbf{D}ecision boundary \textbf{C}omplexity).

\begin{itemize}
    \item \textbf{EDC1}. The linear model establishes a hyperplane to separate samples of two classes in the embedding space. This hyperplane acts as a valuable yardstick to characterize the decision boundary in the embedding space. In particular, we measure the average absolute value distance of all borderline instances from the linear model's hyperplane. We call this measure $\text{EDC1}_{\text{Borderline}}$. To contextualize this measure, we also compute it for a held-out test set and denote it as $\text{EDC1}_{\text{Test}}$. If borderline instances are indeed near the decision boundary between two classes in the embedding space, we should expect a higher value for $\text{EDC1}_{\text{Test}}$ than $\text{EDC1}_{\text{Borderline}}$. That is, borderline instances should be closer to the decision boundary. Therefore, through EDC1 we should be able to answer the first question asked above.  
    \item \textbf{EBC2}. To answer the second question asked before, we record the performance (e.g., accuracy) of the trained linear classifier against borderline samples (denoted as $\text{EDC2}_{\text{Borderline}}$) as well as a held-out test set (denoted as $\text{EDC2}_{\text{Test}}$). This measure will complement EDC1 in a sense that allows us to know to what extent samples (borderline samples and an unseen test set) in the embedding space learned by a DNN are linearly separable. Hence, for a more complicated decision boundary in the embedding space,  $\text{EDC2}_{\text{Borderline}}$ will be higher and vice versa.
\end{itemize}{}

As for the linear model, in line with previous studies~\cite{orriols2010documentation,lorena2018complex} we use a linear Support Vector Machine (SVM)~\cite{cristianini2000introduction}. Our linear SVM seeks to find a hyperplane between learned embeddings of two classes $s$ and $t$ according to Eq.~\ref{eq:svm}.
     \begin{equation}
     \begin{aligned}{}
        Minimize_{\mathbf{w},b,\epsilon} \frac{1}{2}||w||^2+ \gamma (\sum_{i=1}^{n}\epsilon_i) \\
        s.t\begin{cases}
    y_i (\mathbf{w} \times \mathcal{F}(x_i)+b) \geq 1- \epsilon_i \\
    \forall i \; \epsilon_i \geq 0 
  \end{cases}
        \end{aligned}
    \label{eq:svm}
    \end{equation}
    \noindent where $\gamma$ is a hyperparmer controlling the error minimization and margin maximization trade-off, $\mathbf{w}$ is the weight vector, $b$ is the bias term, and the $\epsilon_i$s are slack variables that allow a sample to be on the separating hyperplane $\mathbf{w} \times \mathcal{F}(x_i)+b$.  Recall that $\mathcal{F}(x_i)$ is the embedding vector learned by a DNN for an instance $x_i$.

\section{Experimental Settings}
\label{sec:experiments}
To verify the working and usefulness of DeepDIG, we conduct some experiments. In Section~\ref{sec:datasets}, we describe the datasets and pre-trained models developed to investigate DeepDIG. Section~\ref{sec:deepdig-settings} describes the experimental settings for DeepDIG.

We investigate the proposed framework DeepDIG against three datasets, namely MNIST~\cite{lecun1998gradient}, FashionMNIST~\cite{xiao2017fashion}, and CIFAR10~\cite{krizhevsky2009learning}. For each dataset, we train two models whose description can be found in Table~\ref{tab:dnns}. In this table, $CNV(a,b,c)$ denotes the convolution operation with $a$ input channels, $b$ output channels, and kernel size $c \times c$, $ReLU$ is the ReLU  activation function~\cite{nair2010rectified}, $Linear(a,b)$ indicates a fully connected layer with input size $a$ and output size $b$, and $MaxPool(a)$ denotes max pooling of size $a \times a$. For MNIST and FashionMNIST datasets, we use two simple and distinct models, namely a convolutional neural network (CNN) and a fully connected network (FCN). CIFAR10 is a complicated dataset and we use two well-known deep architectures, namely ResNet~\cite{he2016deep} and  GoogleNet~\cite{szegedy2015going}. The building block of the latter is the famous inception network~\cite{szegedy2017inception}. The third column of Table~\ref{tab:dnns} shows the number of trainable parameters. Also, we have included the accuracy of each DNN against the standard test set of its corresponding dataset. Note that the focus in this work is not having DNNs with state-of-the-art performance. Rather, we focus on analyzing a DNN (regardless of its performance) through the lens of its decision boundaries.

We use the PyTorch package~\cite{paszke2019pytorch} to implement DNNs. Each DNN is trained on its standard training set for 40 epochs and a batch size of 64 samples. We use Adam optimizer~\cite{kingma2014adam} to optimize the parameters. The learning rate is set to 0.01 with the decaying rate of 0.99 after every 100 optimization steps. After a model is trained, we save it and utilize it as a pre-trained model for further investigation.
\subsection{Datasets and Deep Neural Networks}
\label{sec:datasets}
\begin{table}[!htb]
\setlength\tabcolsep{1.6pt} 
    \renewcommand{\arraystretch}{0.6} 
    \caption{Description of investigated DNNs for MNIST, FashionMNIST and CIFAR10 datasets}
    \centering
    \begin{tabu}{c|c|c|c}
    \hline
   \textbf{DNN} & \textbf{Architecture} &\textbf{\#P} &\textbf{ \makecell{Test \\ Acc}}  \\
    \hline \hline
      $\text{MNIST}_{\text{CNN}}$ & \makecell{$CNV(1,10,3)$, $MaxPool(2)$, $ReLU$ \\ $CNV(10,10,3)$, $MaxPool(2)$, $ReLU$ \\ $Linear(320,50)$,  $ReLU$, $Linear(50,10)$ } & 14,070 & $98.75$\\ \hline
      $\text{MNIST}_{\text{FCN}}$ & \makecell{$Linear(784,50)$, $ReLU$, $Linear(50,50)$ } & 42,310 & $97.57$\\  \tabucline[1.5pt]{---}
      $\text{FashionMNIST}_{\text{CNN}}$ & \makecell{$CNV(1,10,3)$, $MaxPool(2)$, $ReLU$ \\ $CNV(10,10,3)$, $MaxPool(2)$, $ReLU$ \\ $Linear(320,50)$,  $ReLU$, $Linear(50,10)$ } & 14,070 & $89.11$\\ \hline
      $\text{FashionMNIST}_{\text{FCN}}$ & \makecell{$Linear(784,50)$, $ReLU$, $Linear(50,50)$ } & 42,310 & $88.24$\\  \tabucline[1.5pt]{---}
      $\text{CFIAR10}_{\text{ResNet}}$ &  ResNet~\cite{he2016deep} & 21,282,122  & $82.68$\\ \hline
      $\text{CIFAR10}_{\text{GoogleNet}}$ & GoogleNet~\cite{szegedy2015going} & 6,166,250 & $84.71$\\  \tabucline[1.5pt]{---}
    \end{tabu}
    \label{tab:dnns}
\end{table}{}
\subsection{DeepDIG Experimental Settings}
\label{sec:deepdig-settings}
As described in Section~\ref{sec:framework}, component (I) and (II) utilize an autoencoder to generate adversarial examples. Table~\ref{tab:autoendoers} describes the detail of the utilized autoencoders. Since MNIST and FashionMNIST are similar, we opt for employing the same autoencoder architecture for these two datasets. Each autoencoder consists of two modules: an encoder mapping an input sample to a condensed hidden representation and a decoder mapping back the hidden representation to a reconstruction of the original input sample. Since input samples are images in the pixel range $[0,1]$, we utilize the sigmoid activation function at the end of each decoder\footnote{\url{https://en.wikipedia.org/wiki/Sigmoid_function}}. In Table~\ref{tab:autoendoers}, $TCNV$ denotes transposed convolution operation known as deconvolution as well~\cite{dumoulin2016guide}. Next, we explain the training detail of each component.

\textbf{Component (I).} To optimize component (I), we use samples from the standard train set labeled as class $s$ e.g., all training samples labeled as `Trousers' for FashionMNIST. Out of such samples, we use 80\% for training and the rest as the validation set to tune the hyperparameters. Notably, we use the validation set for finding the optimal value of  the hyperparameter $\alpha$ in Eq.~\ref{eq:deepdig1}. Two criteria are considered to choose the best value for $\alpha$: the success rate of adversarial example generation and the quality of generated adversarial examples  (examples $\hat{x}_t$ in Figure~\ref{fig:deepdig} (I)). To ensure the first criterion, we check the accuracy of adversarial examples against the validation set; the more decline in the accuracy, the better. For the second criterion, we visually inspect the generated examples and ensure they resemble real samples. For all pre-trained DNNs in Table~\ref{tab:dnns}, we found $\alpha=0.8$ as the best choice. We train the adversarial example generation in component (I) for 5000 steps and batch size 128 samples.  Adam optimizer~\cite{kingma2014adam} is used and the learning rate is set to 0.01 with the decaying rate 0.95 after every 1000 steps.

\textbf{Component (II).} The successfully generated adversarial examples in components (I) (i.e., $\hat{x}_t$ samples whose prediction is $t$) are used to optimize component (II). Similar to adversarial example generation in component (I), here we check both the accuracy and the quality of generated examples. Note that since we perform the \emph{reverse} adversarial examples generation (i.e., adversarial examples of adversarial examples), the higher the accuracy is, the better the model is performing. Simulation settings are the same with competent (I) including $\alpha=0.8$ for the loss function in Eq.~\ref{eq:deepdig2}.

\textbf{Component (III).} We run Algorithm~\ref{alg:middle} for all pairs of successfully generated adversarial samples in component (I) and (II) (i.e., $\{(\hat{x}_t,\hat{x}_s) | C(\hat{x}_t)=t, C(\hat{x}_s)=s\})$ aiming at finding borderline samples. We set the threshold $\beta =0.0001$. We believe this value is sufficiently small ensuring the criterion (a) discussed in Section~\ref{sec:framework}. As for the criterion (b) --borderline instances being similar to real examples-- we visually inspect the borderline samples. In fact, as will be demonstrated  in the next section, DeepDIG is capable of meeting both criteria quite effectively.     
\begin{table}[!htb]
\setlength\tabcolsep{1.6pt} 
    \renewcommand{\arraystretch}{1.0} 
    \centering
    \caption{Description of autoencoder models used in components (I) and (II) of DeepDIG }
    \begin{tabu}{c|c|c} \hline
       \textbf{Dataset}  & \textbf{Encoder} & \textbf{Decoder}  \\ \hline \hline
        \makecell{MNIST \\ FashionMNIST} & \makecell{ $Linear(784,100)$,  $ReLU$} & \makecell{ $Linear(100,50)$,  $ReLU$,\\ $Linear(50,784)$,  $Sigmoid$ } \\  \tabucline[1.5pt]{---}
        \makecell{CIFAR10} & \makecell{ $CNV(3,12,4)$, $ReLU$\\ $CNV(12,24,4)$, $ReLU$\\ $CNV(24,48,4)$, $ReLU$} & \makecell{ $TCNV(48,24,4)$, $ReLU$\\ $TCNV(24,12,4)$, $ReLU$\\ $TCNV(12,3,4)$, $Sigmoid$ } \\  \tabucline[1.5pt]{---}
    \end{tabu}
    \label{tab:autoendoers}
\end{table}{}

\section{Results and Discussions}
\label{sec:results}

In this section, we present the experimental results. First, in Section~\ref{sec:deepdig-component-results}, we investigate the working of components (I) and (II) of DeepDIG. In Section~\ref{sec:baselines}, we compare the performance of  DeepDIG with two baseline approaches. Section~\ref{sec:charac-dnns} includes the results of characterizing DNNs in Table~\ref{tab:dnns} for a pair of classes. Finally, in Section~\ref{sec:inter-class}, we present the characterization results for all pair-wise classes in the model $\text{MNIST}_{\text{CNN}}$.

\begin{table*}[!htb]
\setlength\tabcolsep{1.5pt} 
    \renewcommand{\arraystretch}{0.5} 
\caption{Experimental results of investigating components (I) and (II) of DeepDIG for MNIST dataset }
    \centering
        
    \begin{tabular}{c||c|c|c|c|c|c|c|c}
    
    \hline
    ($s$, $t$)& \multicolumn{4}{c|}{(`1', `2')} & \multicolumn{4}{c}{(`2', `1')} \\ \hline
    Component &\multicolumn{2}{c|}{(I)} & \multicolumn{2}{c|}{(II)} & \multicolumn{2}{c|}{ (I)} & \multicolumn{2}{c}{(II)} \\ \hline
     \diagbox{DNN}{Factor}& \makecell{Acc} & \makecell{ Visualisation   } & \makecell{ Acc} & \makecell{Visualisation} & \makecell{Acc} & \makecell{ Visualisation   } & \makecell{Acc } & \makecell{Visualisation}   \\ \hline 
    $\text{MNIST}_{\text{CNN}}$ 
    & 0.0 
    & \makecell{\includegraphics[scale=0.15]{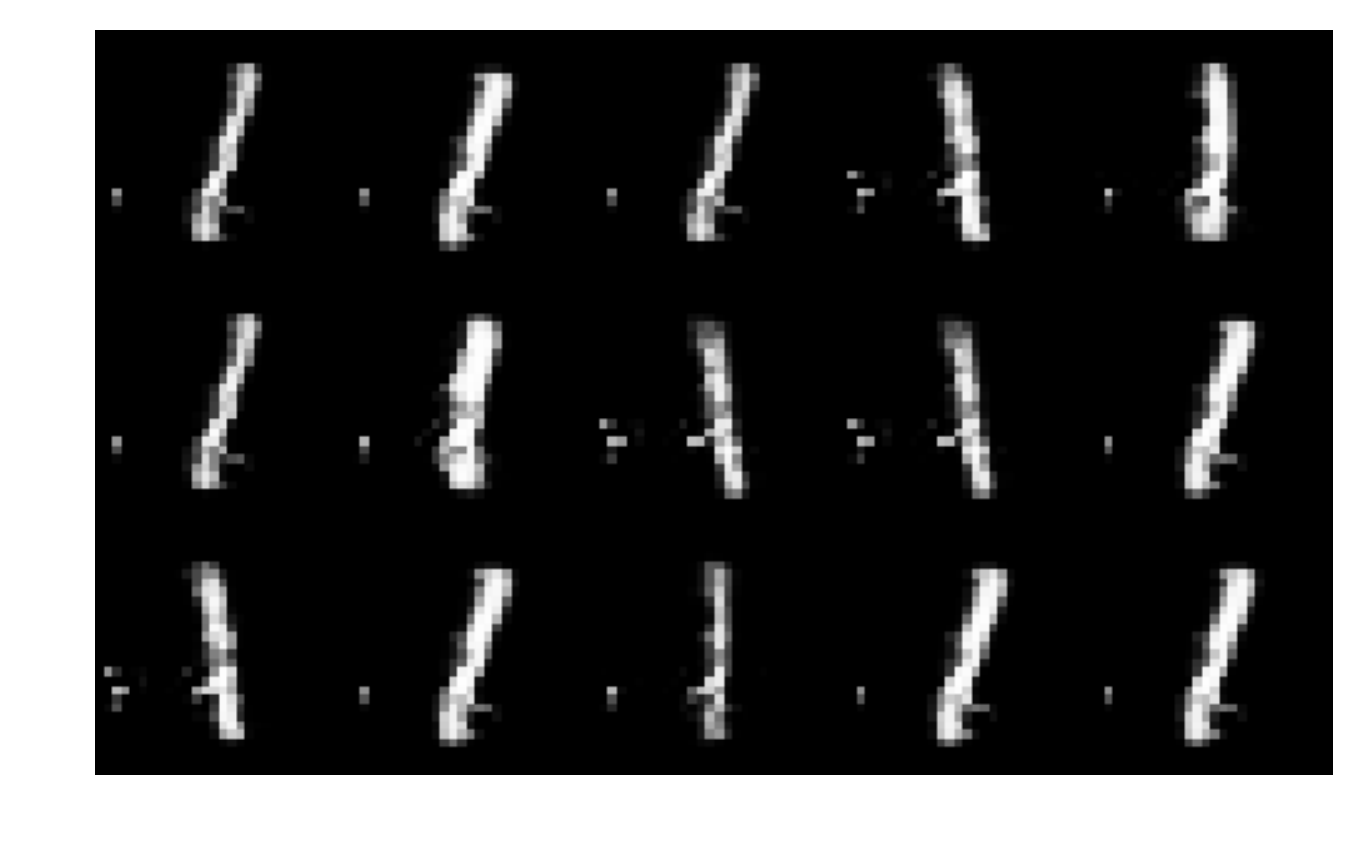}}
    & 1.0 
    & \makecell{\includegraphics[scale=0.15]{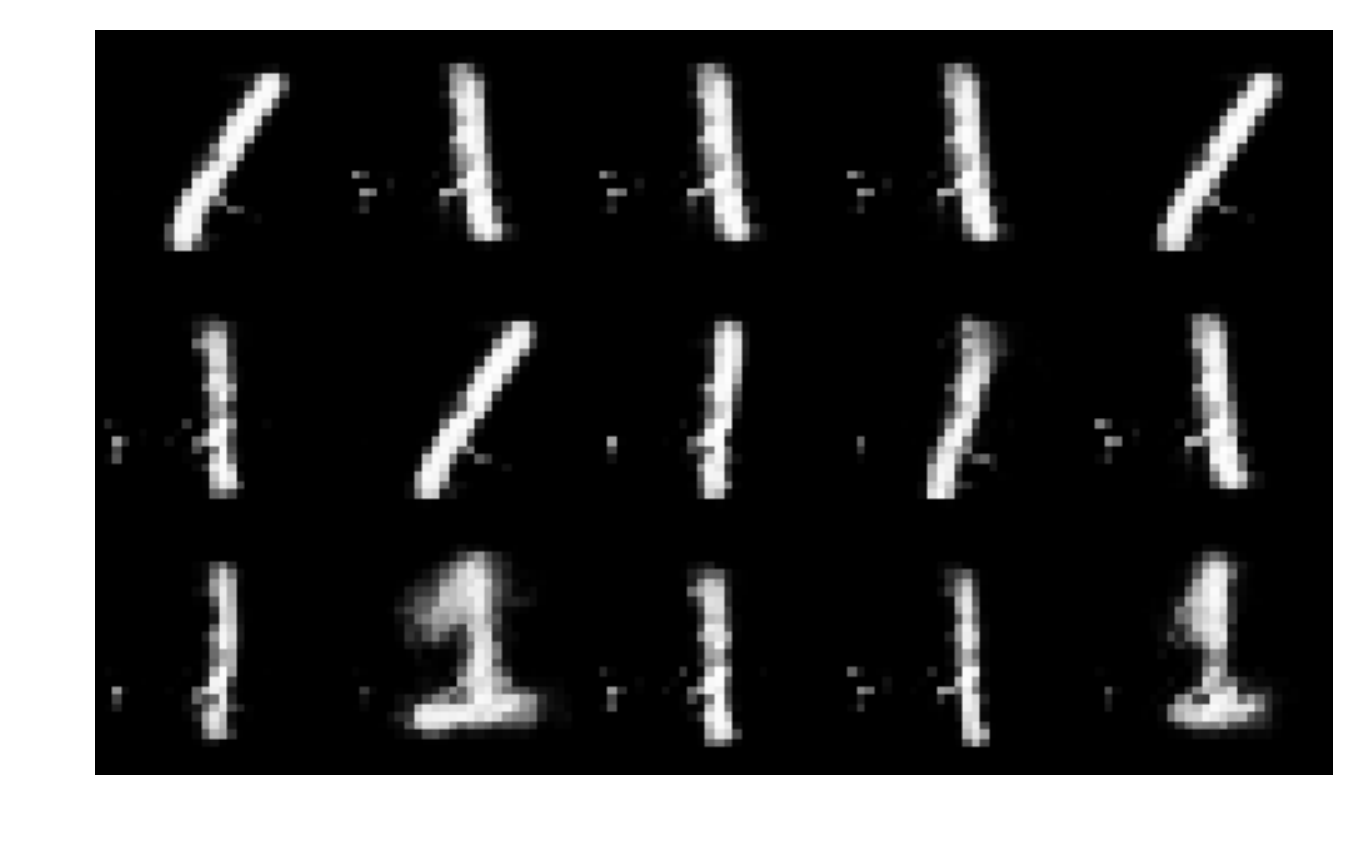}} 
    & 0.0
    & \makecell{\includegraphics[scale=0.15]{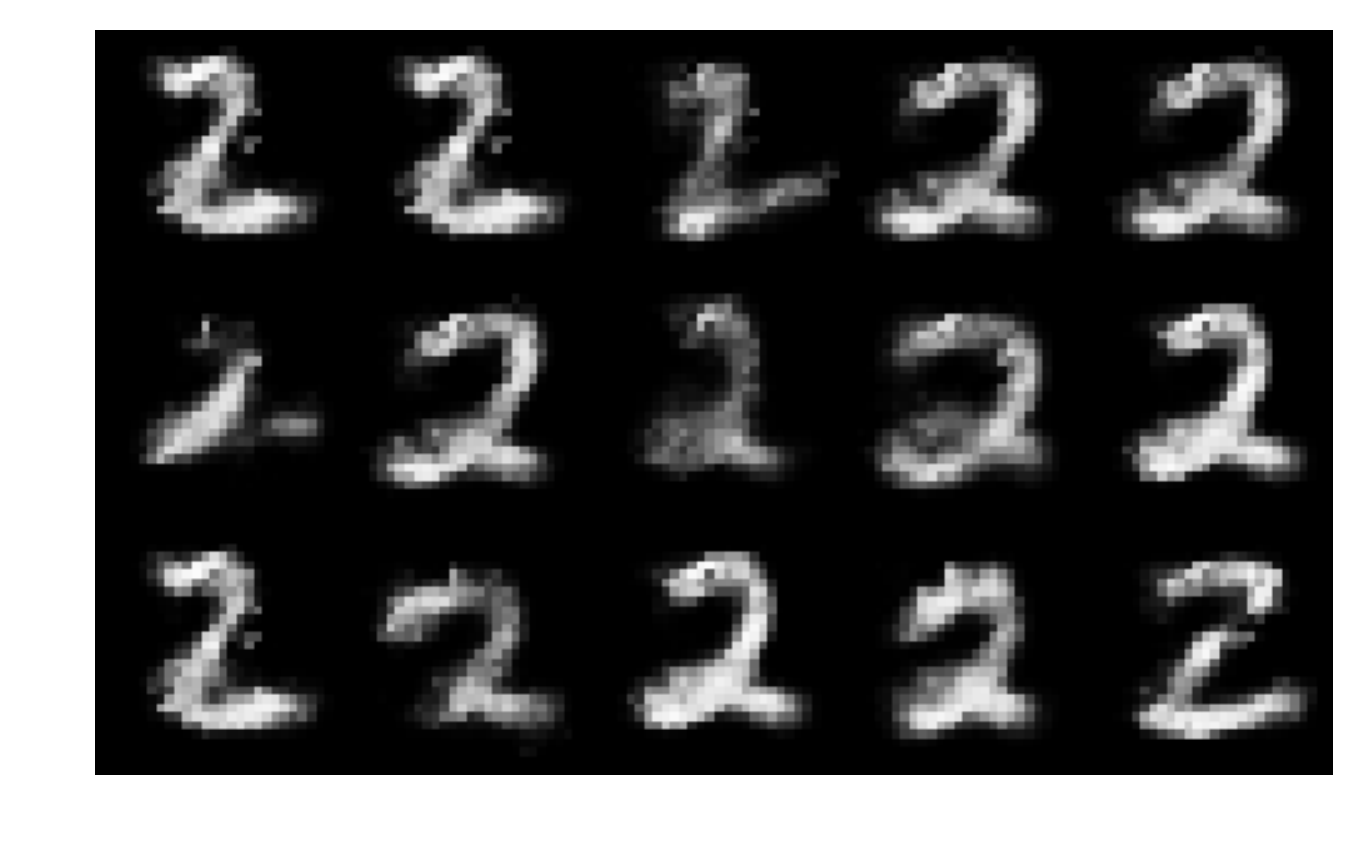}} 
    & 1.0 
    & \makecell{\includegraphics[scale=0.15]{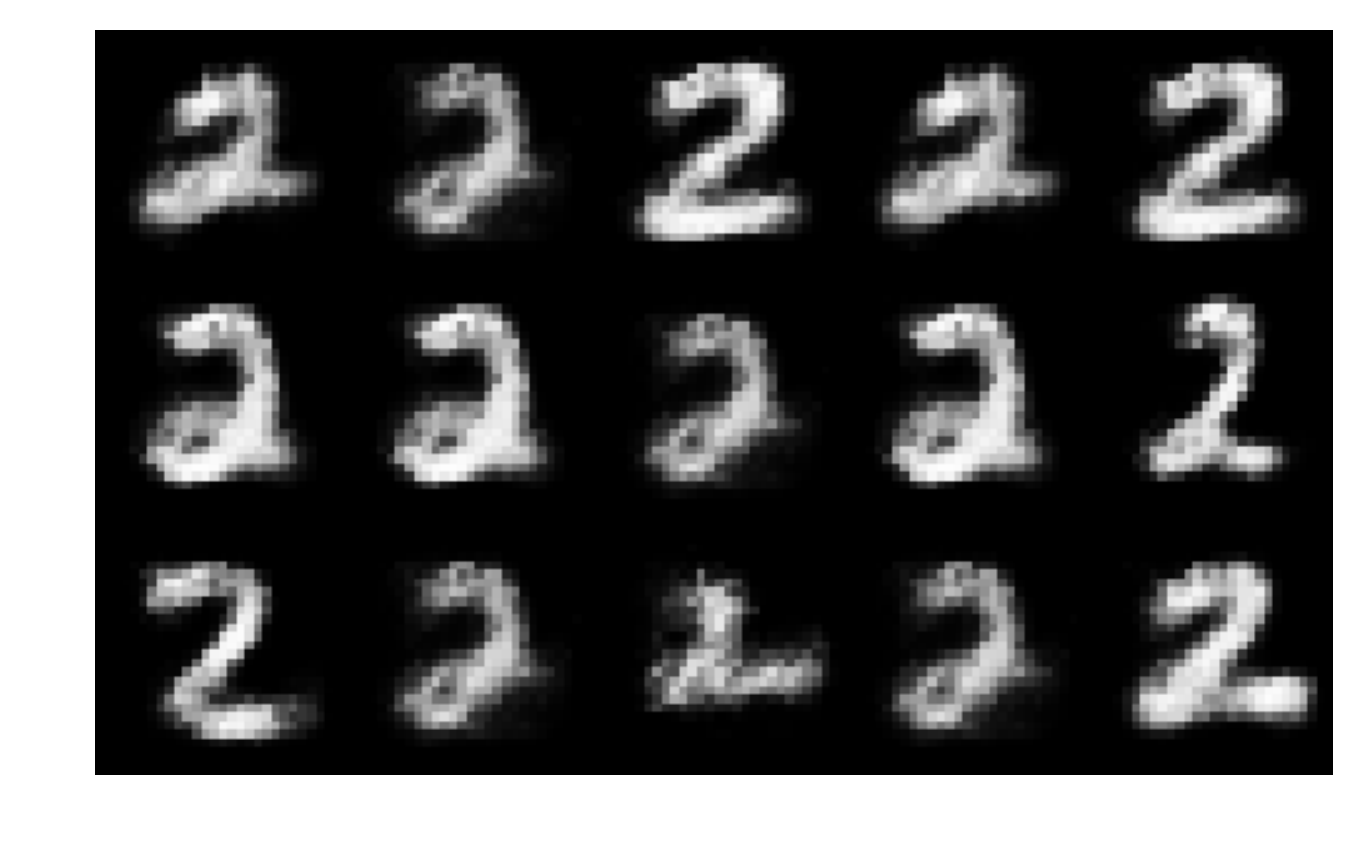}} \\ \hline
 $\text{MNIST}_{\text{FCN}}$ 
    & 0.0 
    & \makecell{\includegraphics[scale=0.15]{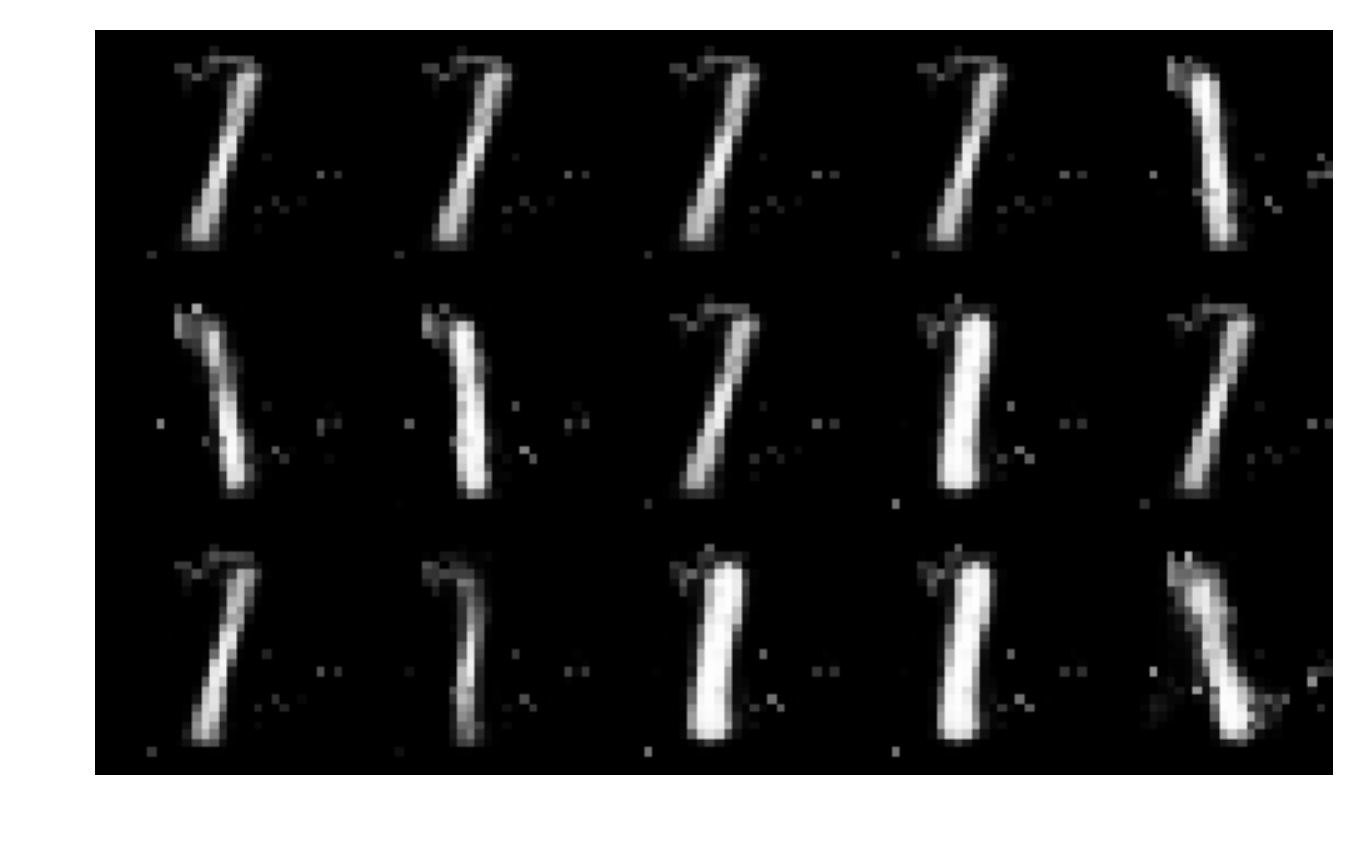}}
    & 0.99 
    & \makecell{\includegraphics[scale=0.15]{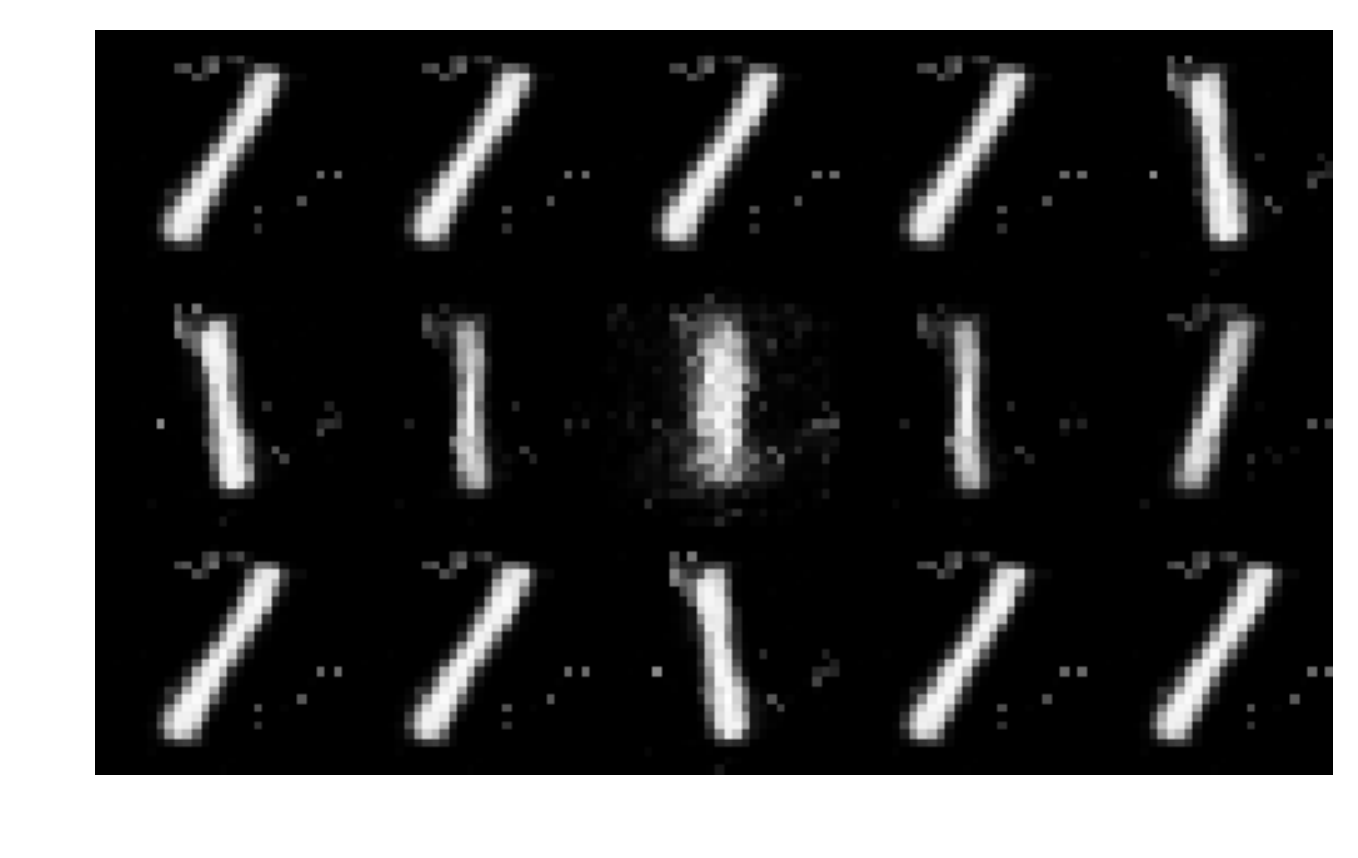}} 
    & 0.0 
    & \makecell{\includegraphics[scale=0.15]{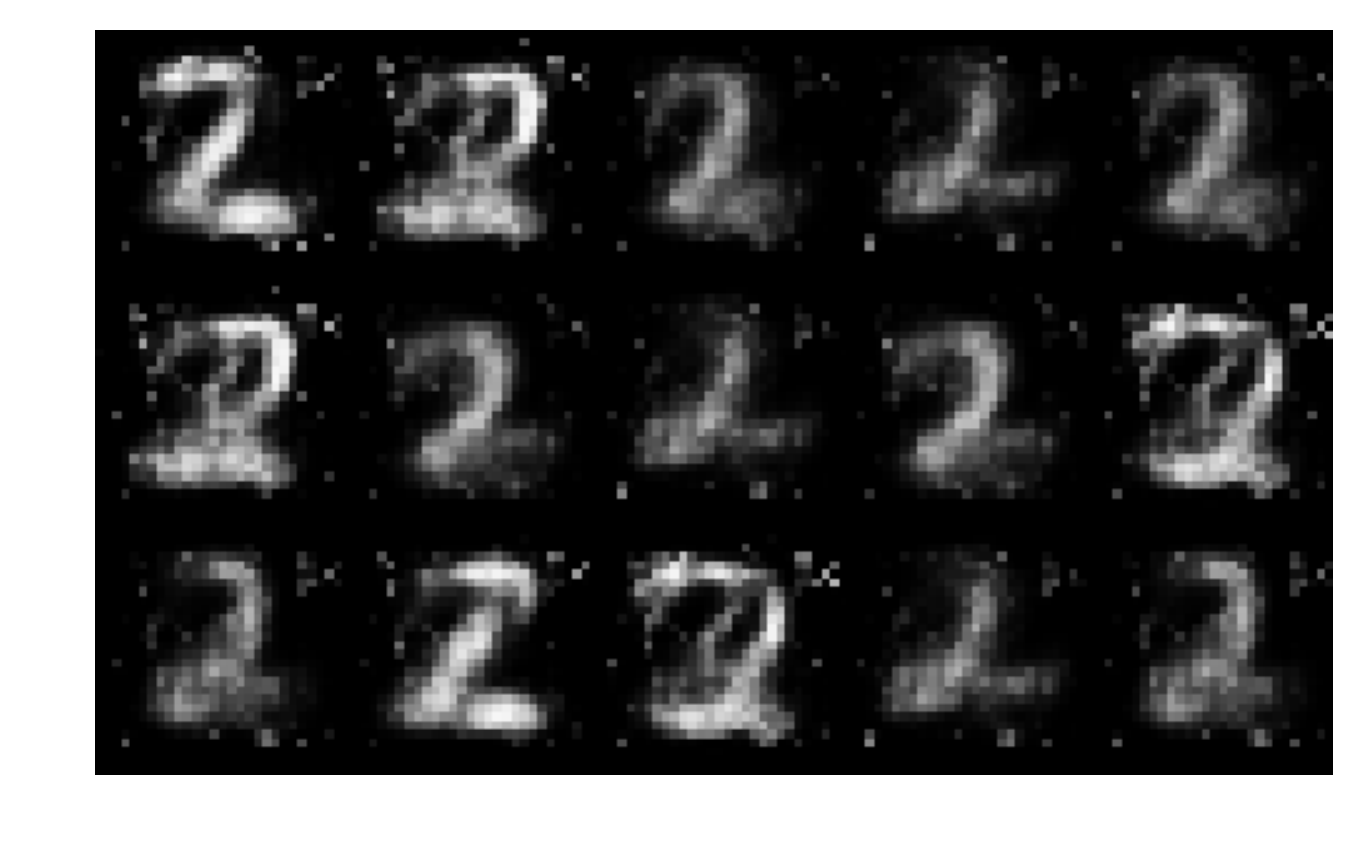}} 
    & 1.0 
    & \makecell{\includegraphics[scale=0.15]{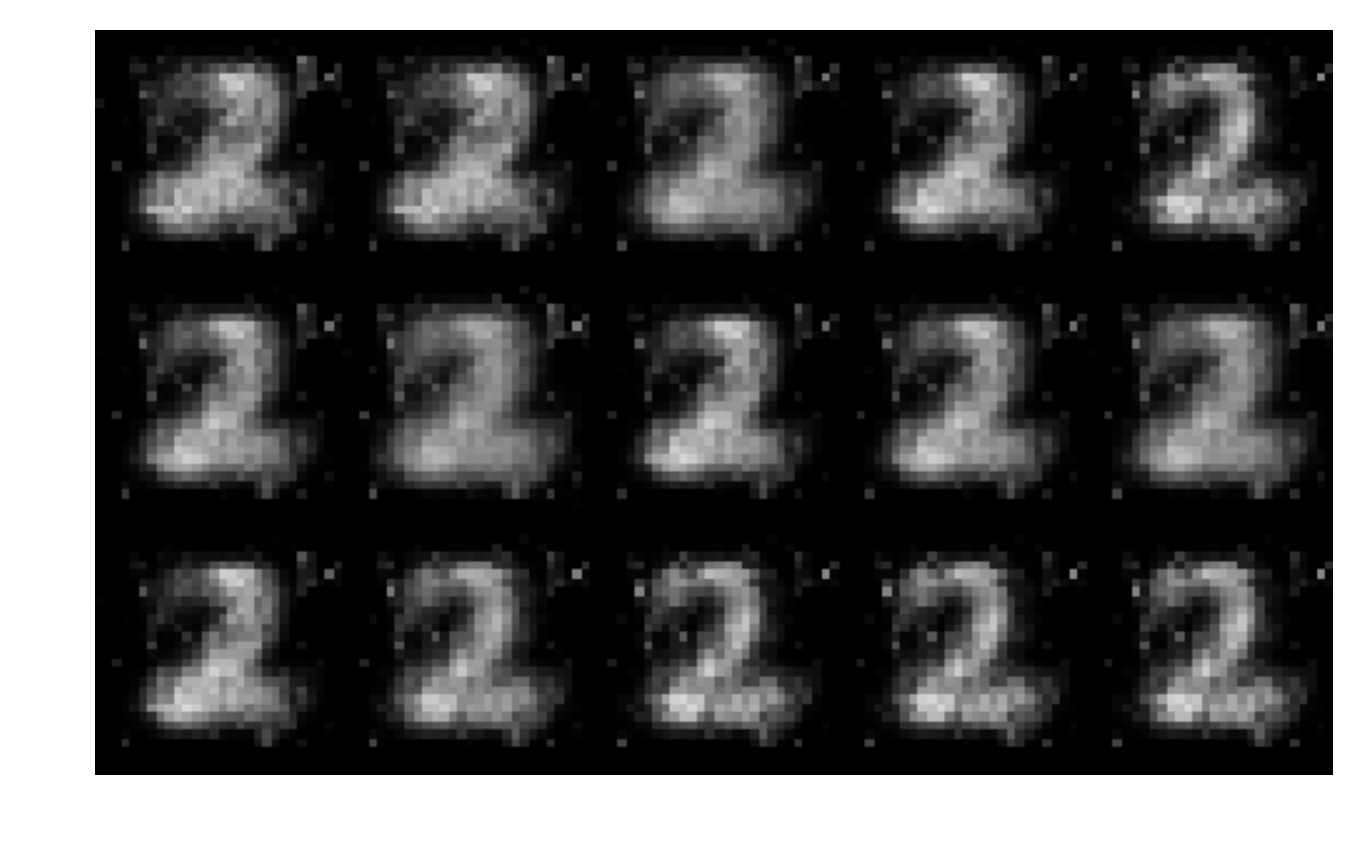}} \\ \hline
    \end{tabular}
    \label{tab:components12mnsit}
\end{table*}{}

\begin{table*}[!htb]
    \setlength\tabcolsep{1.5pt} 
    \renewcommand{\arraystretch}{0.5} 
\caption{Experimental results of investigating components (I) and (II) of DeepDIG for FashionMNIST dataset }
    \centering
        
    \begin{tabular}{c||c|c|c|c|c|c|c|c}
    
    \hline
    ($s$, $t$)& \multicolumn{4}{c|}{(`Trouser', `Pullover')} & \multicolumn{4}{c}{(`Pullover', `Trouser')} \\ \hline
    Component &\multicolumn{2}{c|}{(I)} & \multicolumn{2}{c|}{(II)} & \multicolumn{2}{c|}{ (I)} & \multicolumn{2}{c}{(II)} \\ \hline
    \diagbox{DNN}{Factor} & \makecell{Acc   } & \makecell{ Visualisation   } & \makecell{ Acc   } & \makecell{Visualisation   } & \makecell{  Acc  } & \makecell{ Visualisation   } & \makecell{ Acc  } & \makecell{ Visualisation  }   \\ \hline 
    $\text{FashionMNIST}_{\text{CNN}} $ 
    & 0.01 
    & \makecell{\includegraphics[scale=0.15]{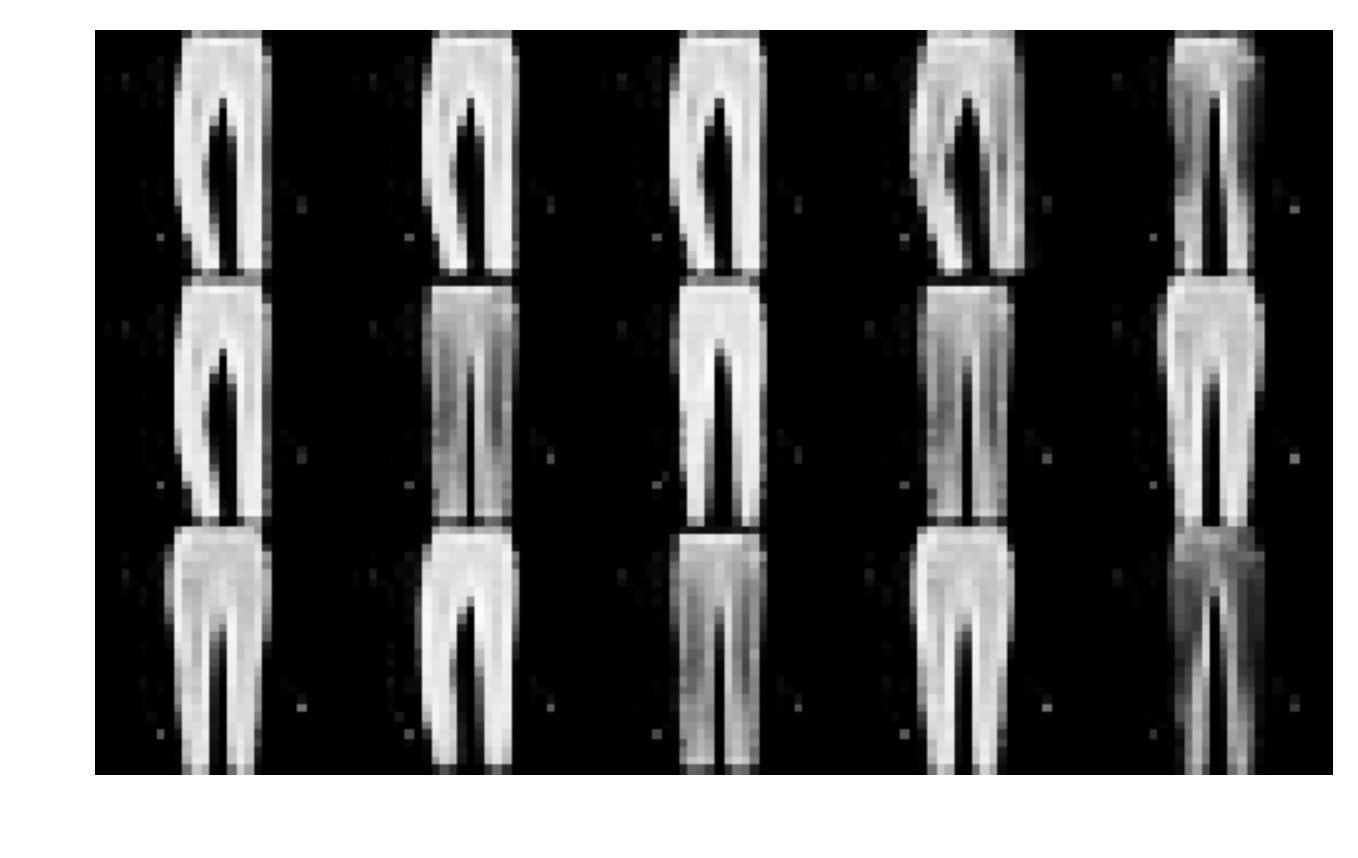}}
    & 1.0 
    & \makecell{\includegraphics[scale=0.15]{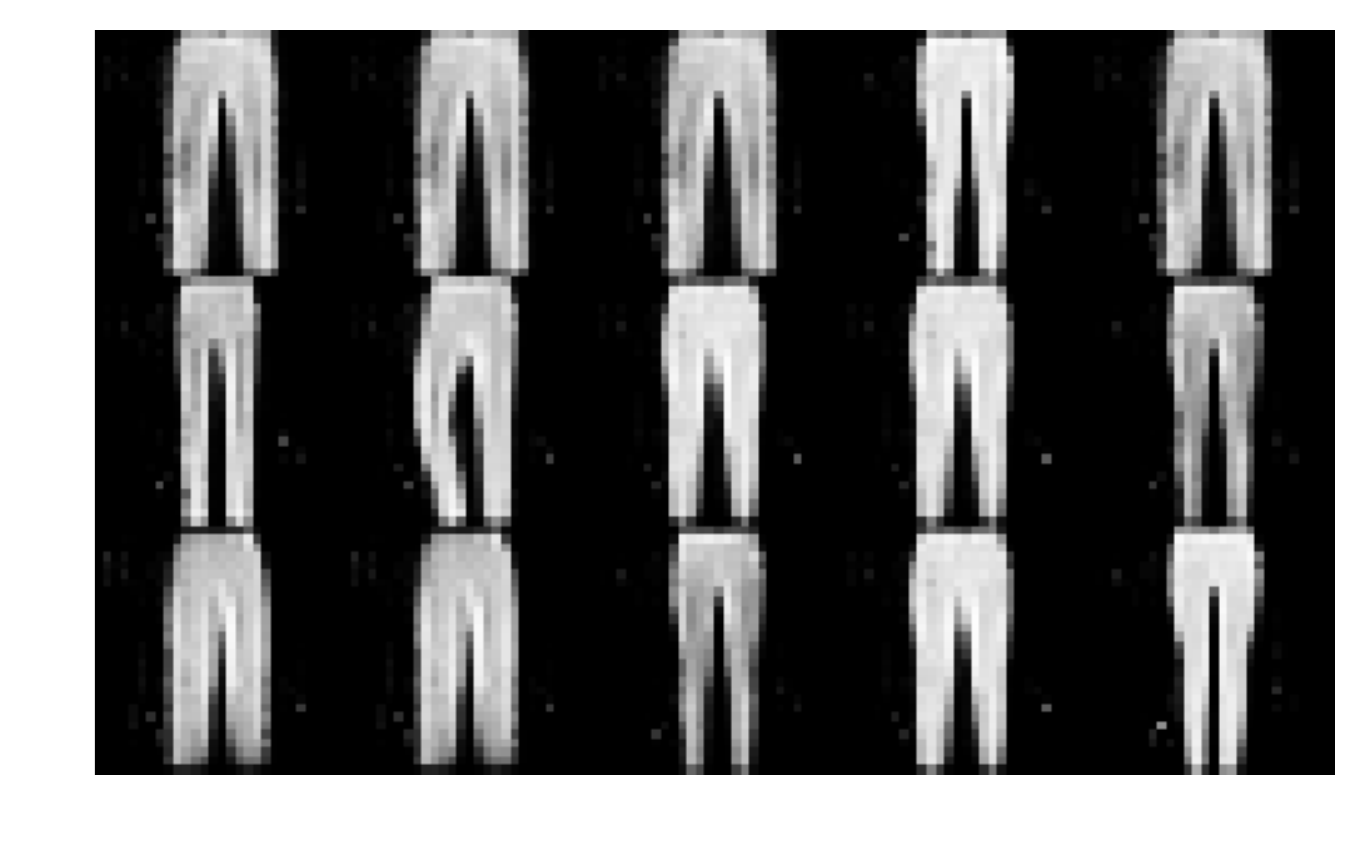}} 
    & 0.0 
    & \makecell{\includegraphics[scale=0.15]{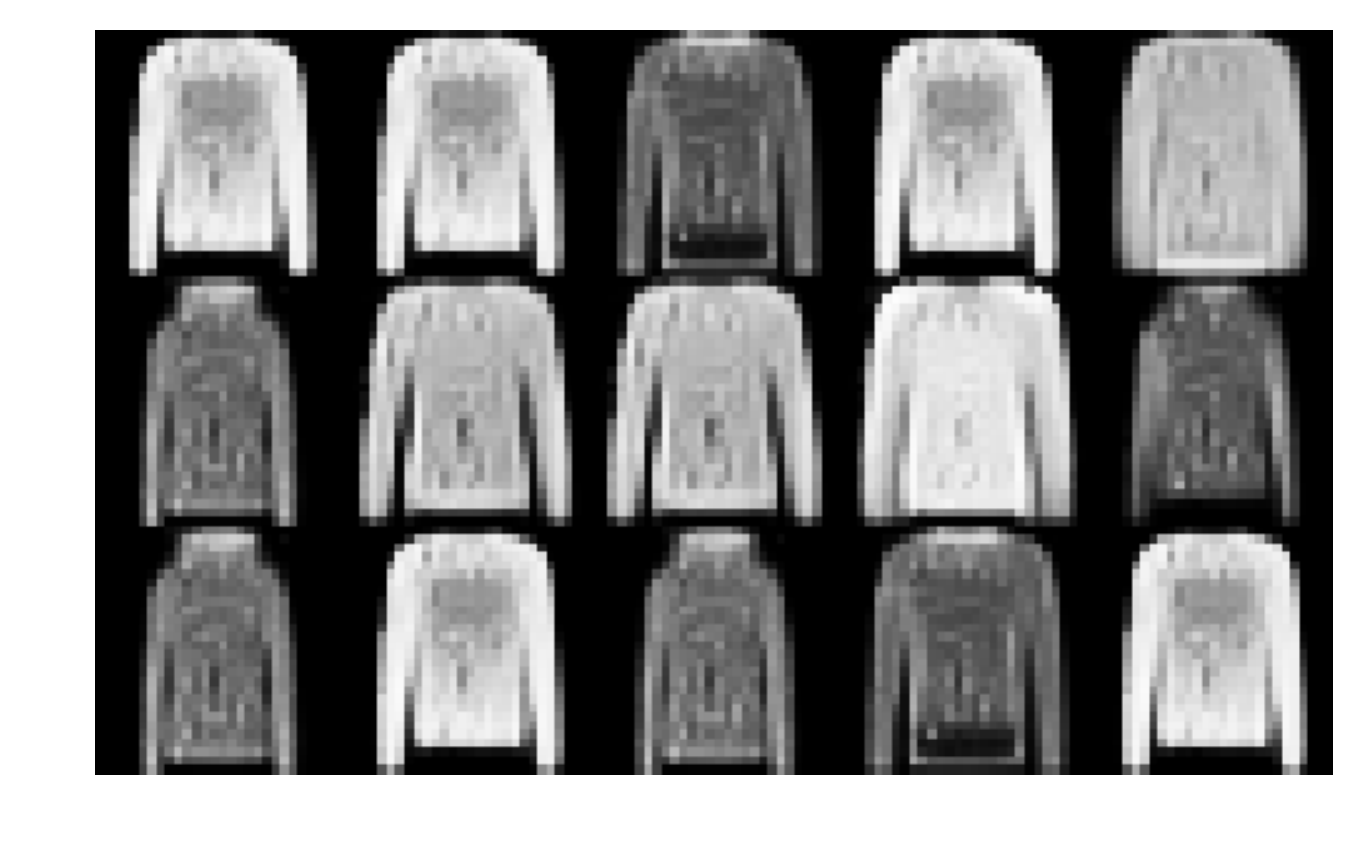}} 
    & 1.0 
    & \makecell{\includegraphics[scale=0.15]{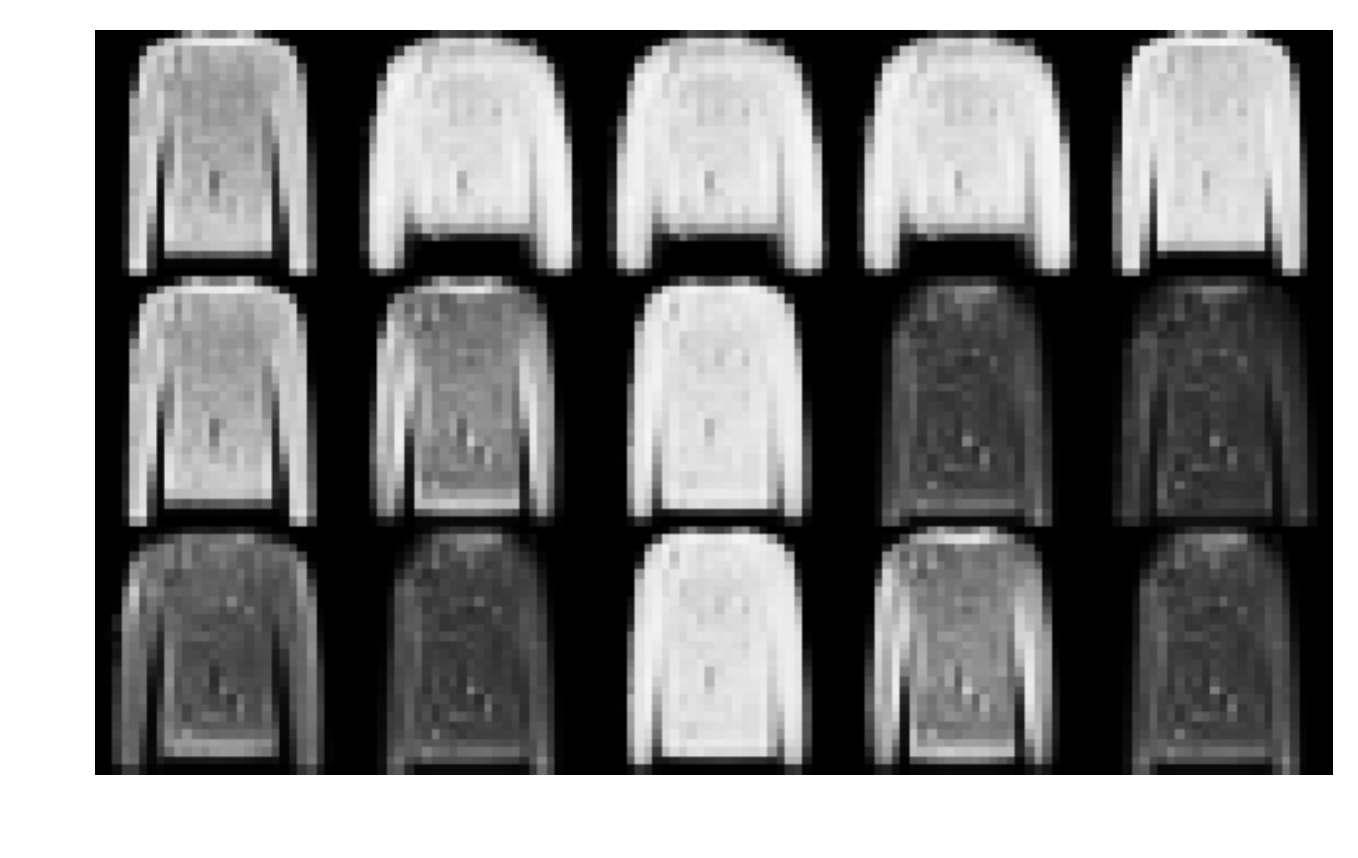}} \\ \hline
  $\text{FashionMNIST}_{\text{FCN}} $ 
    & 0.0 
    & \makecell{\includegraphics[scale=0.15]{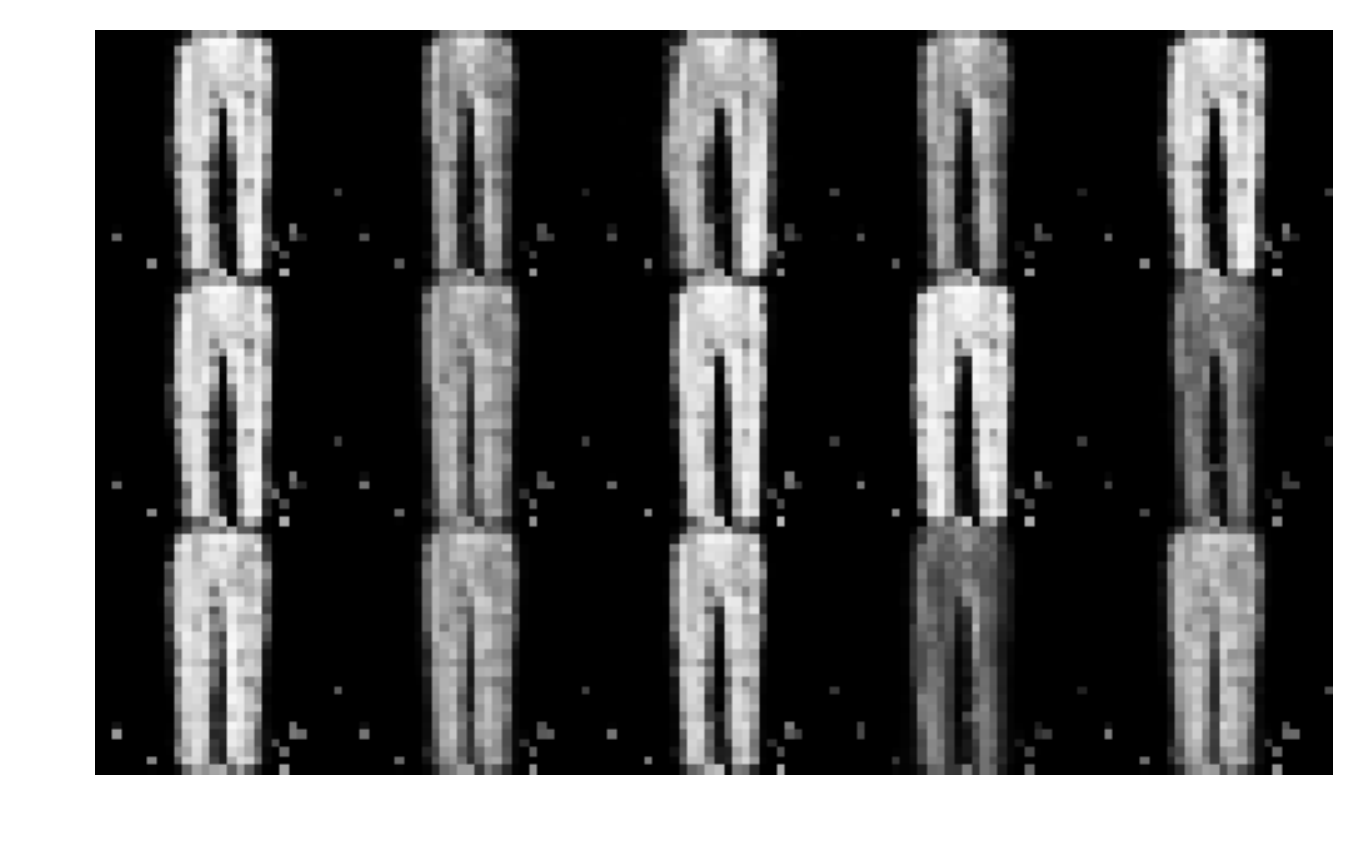}}
    & 1.0 
    & \makecell{\includegraphics[scale=0.15]{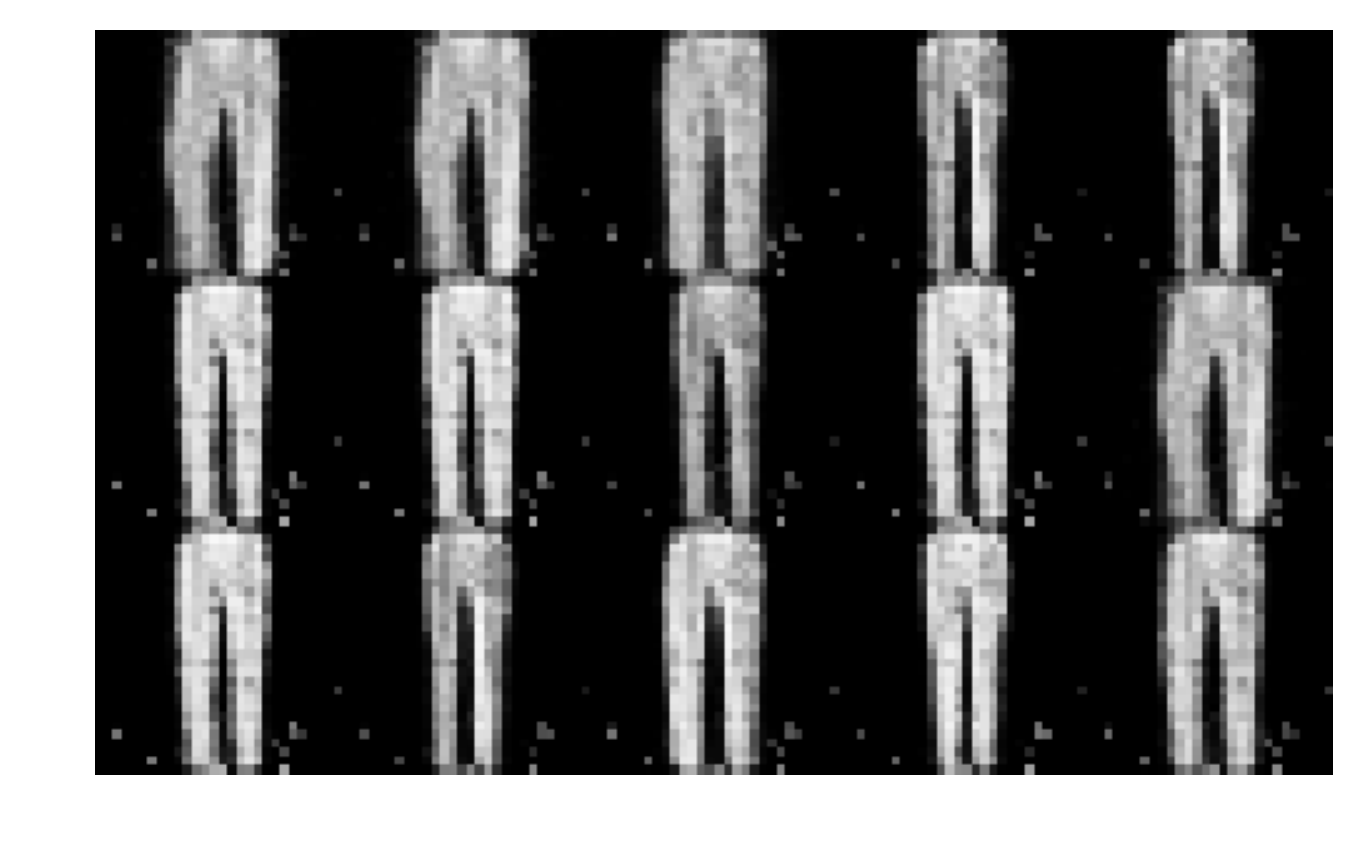}} 
    & 0.0 
    & \makecell{\includegraphics[scale=0.15]{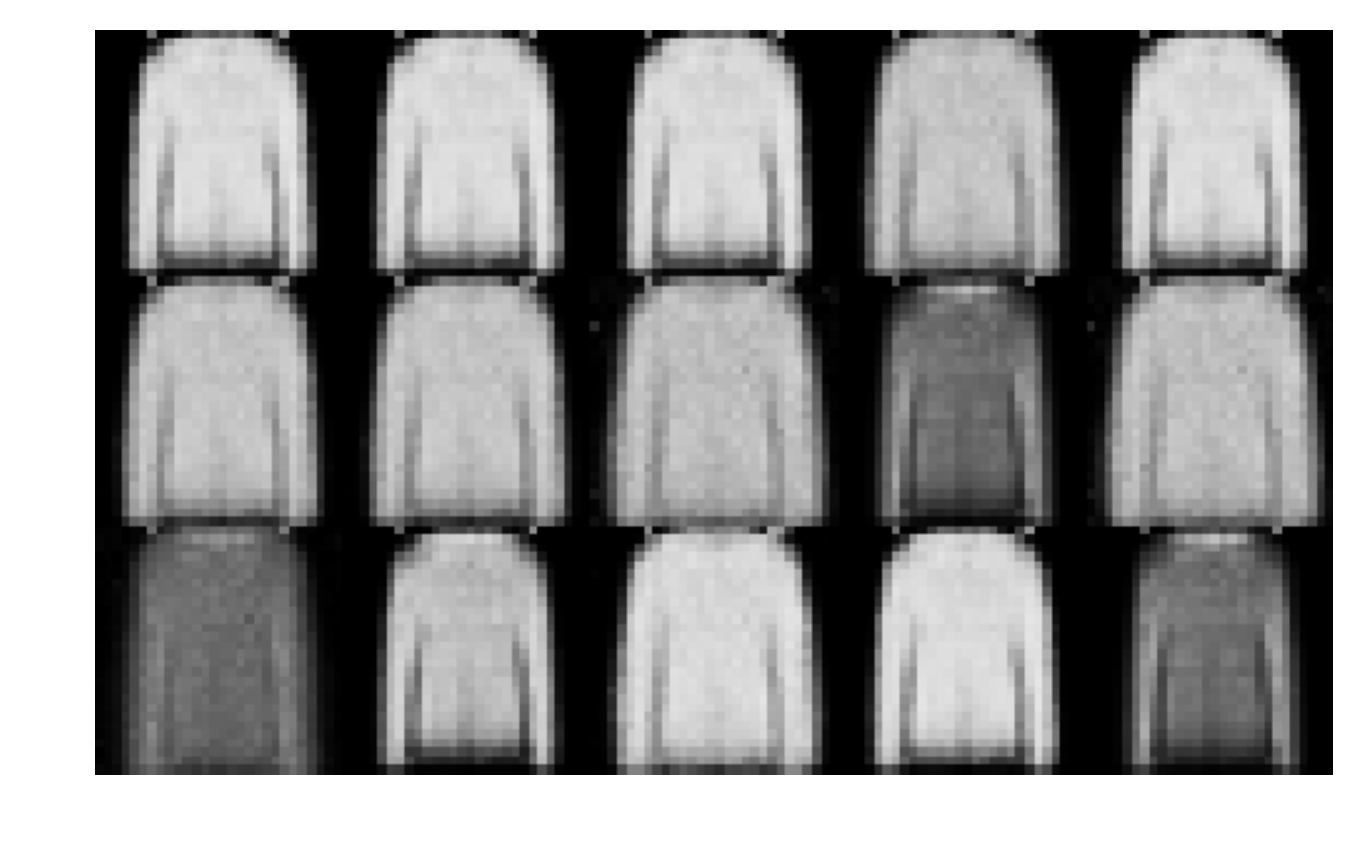}} 
    & 1.0 
    & \makecell{\includegraphics[scale=0.15]{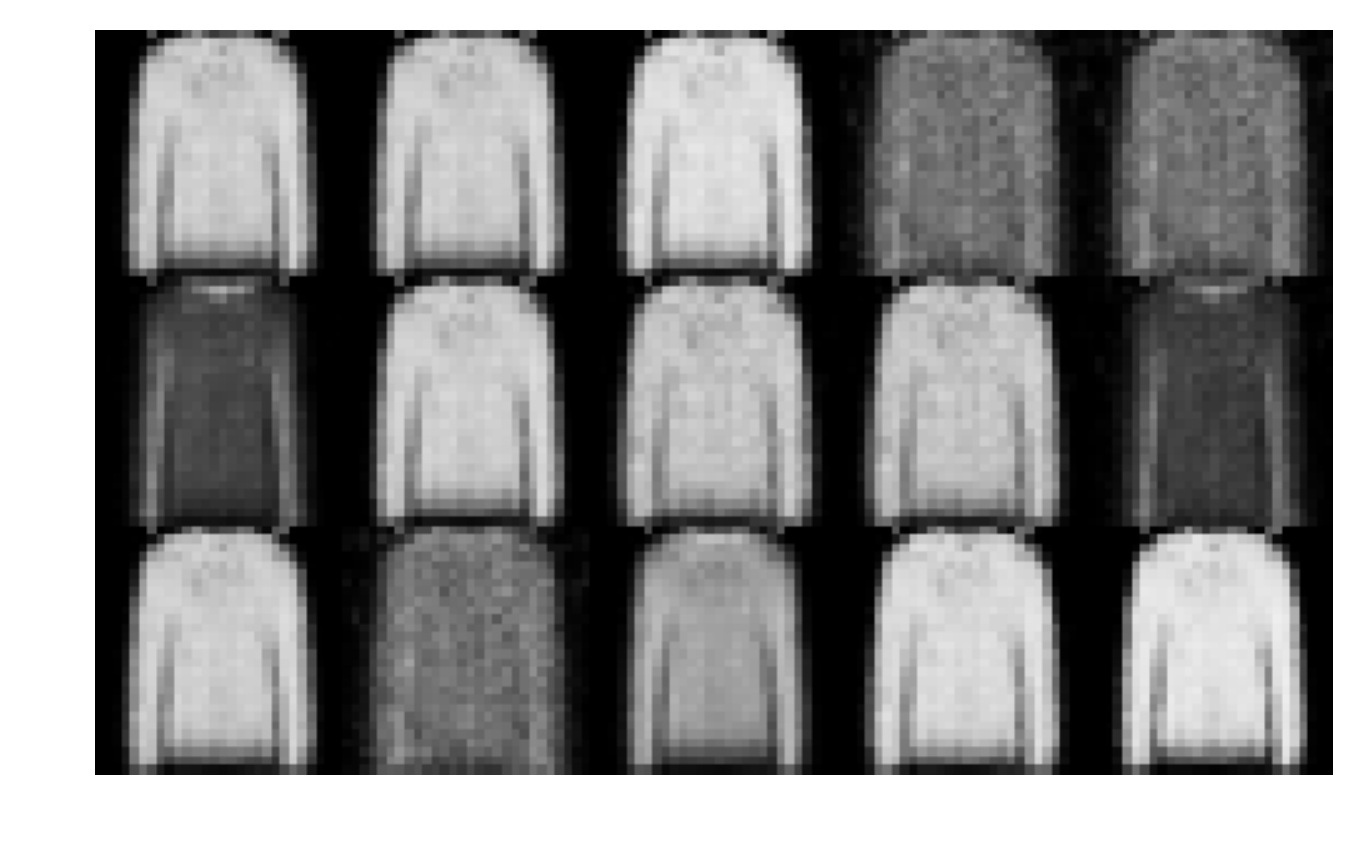}} \\ \hline
    \end{tabular}
    \label{tab:components12fashionmnsit}
\end{table*}{}

\begin{table*}[!htb]
\setlength\tabcolsep{1.5pt} 
    \renewcommand{\arraystretch}{0.5} 
\caption{Experimental results of investigating components (I) and (II) of DeepDIG for CIFAR10 dataset }
    \centering
        
    \begin{tabular}{c||c|c|c|c|c|c|c|c}
    
    \hline
    ($s$, $t$)& \multicolumn{4}{c|}{(`Automobile', `Bird')} & \multicolumn{4}{c}{(`Bird', `Automobile')} \\ \hline
    Component &\multicolumn{2}{c|}{(I)} & \multicolumn{2}{c|}{(II)} & \multicolumn{2}{c|}{ (I)} & \multicolumn{2}{c}{(II)} \\ \hline
    \diagbox{DNN}{Factor} & \makecell{Acc   } & \makecell{ Visualisation   } & \makecell{ Acc   } & \makecell{Visualisation   } & \makecell{  Acc  } & \makecell{ Visualisation   } & \makecell{ Acc  } & \makecell{ Visualisation  }   \\ \hline 
 $\text{CIFAR10}_{\text{ResNet}} $
    & 0.00 
    & \makecell{\includegraphics[scale=0.15]{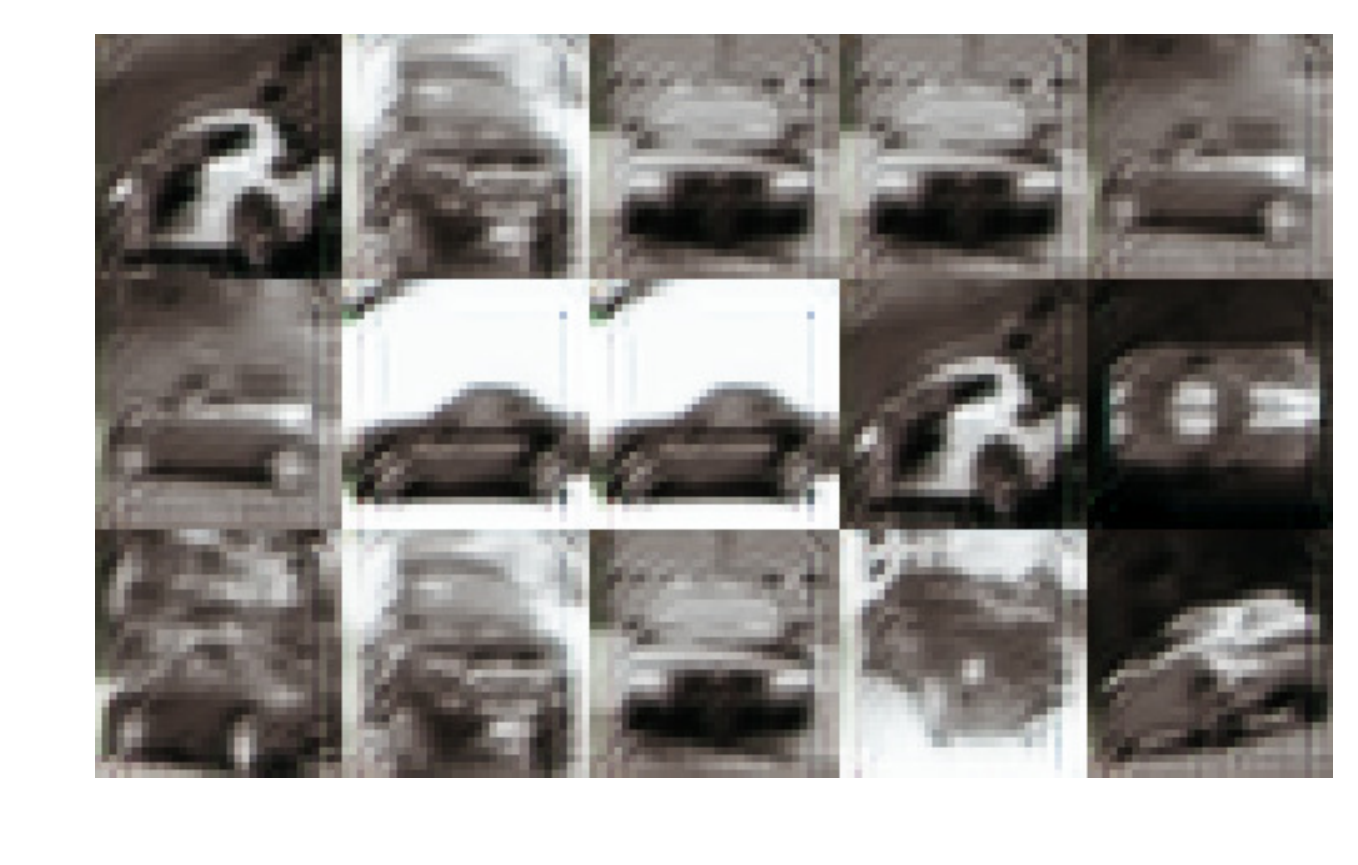}}
    & 0.99 
    & \makecell{\includegraphics[scale=0.15]{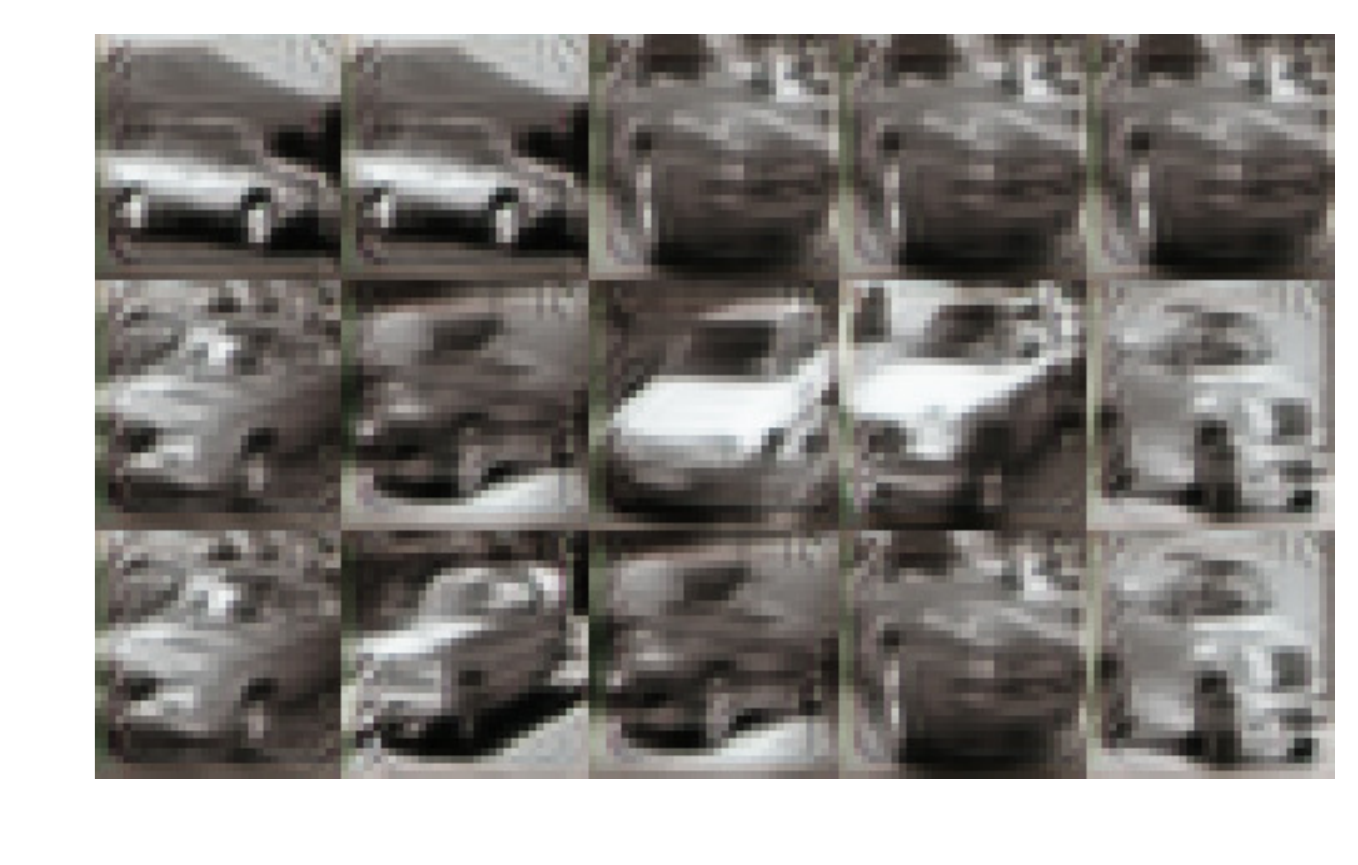}} 
    & 0.01 
    & \makecell{\includegraphics[scale=0.15]{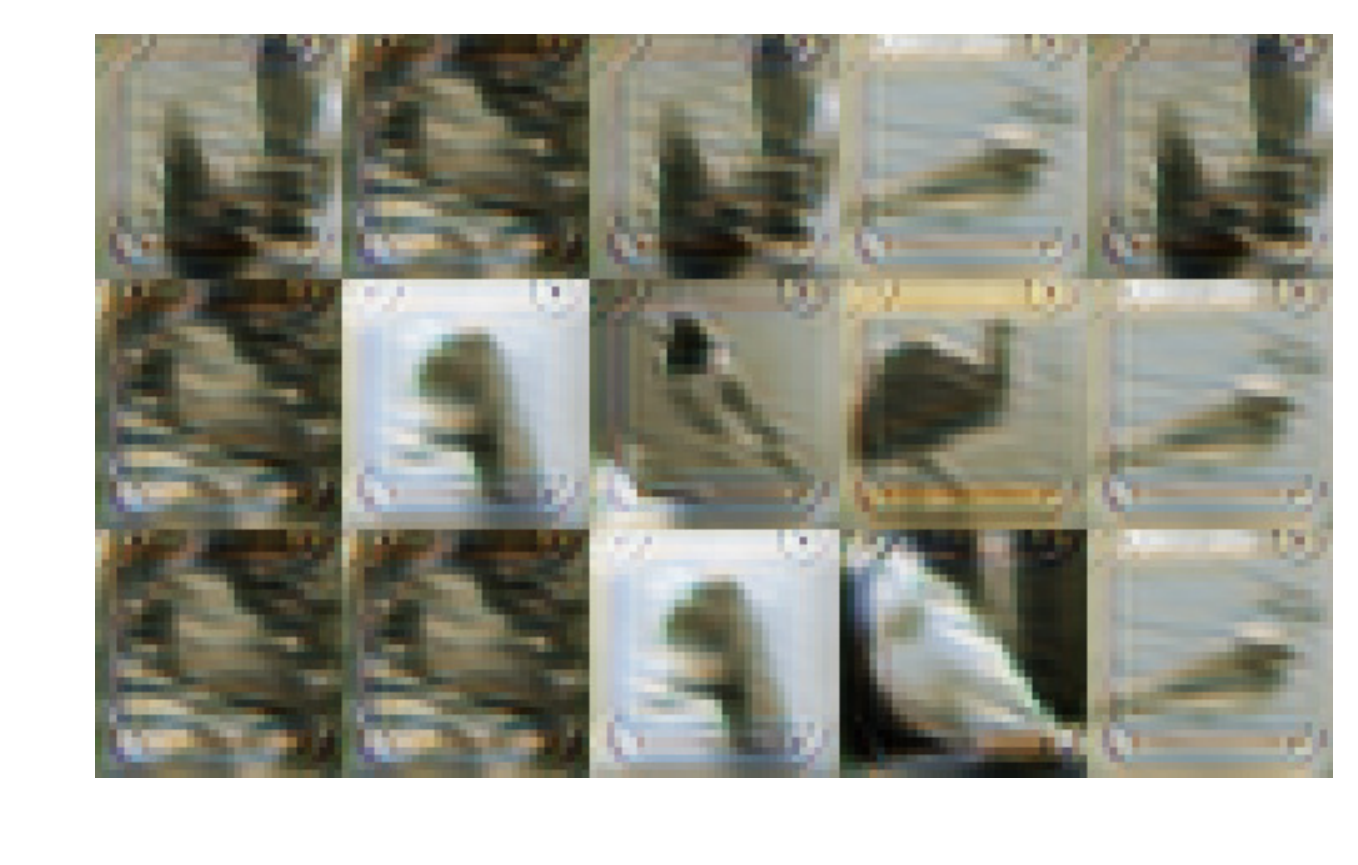}} 
    & 0.99 
    & \makecell{\includegraphics[scale=0.15]{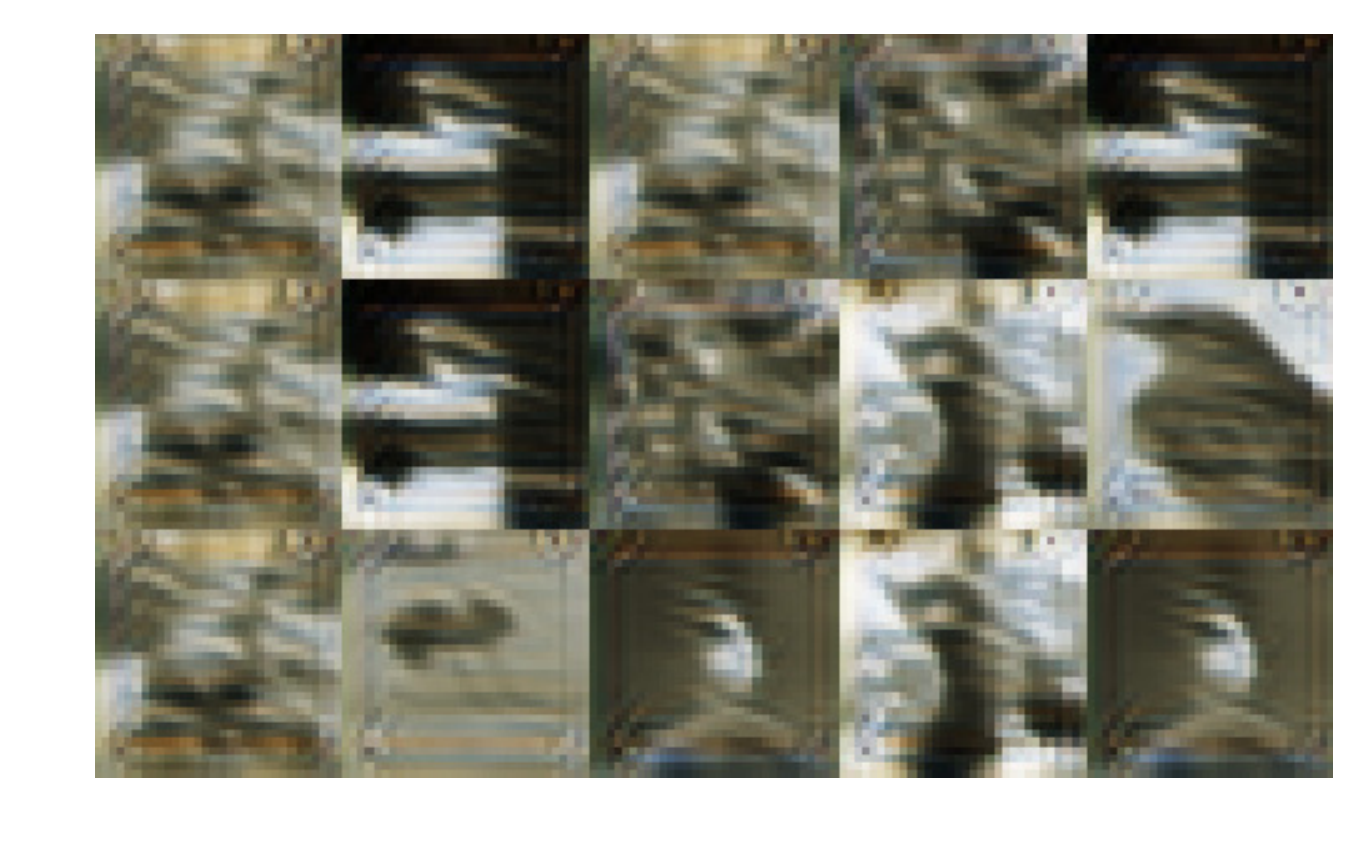}} \\ \hline
 $\text{CIFAR10}_{\text{GoogleNet}} $ 
    & 0.00 
    & \makecell{\includegraphics[scale=0.15]{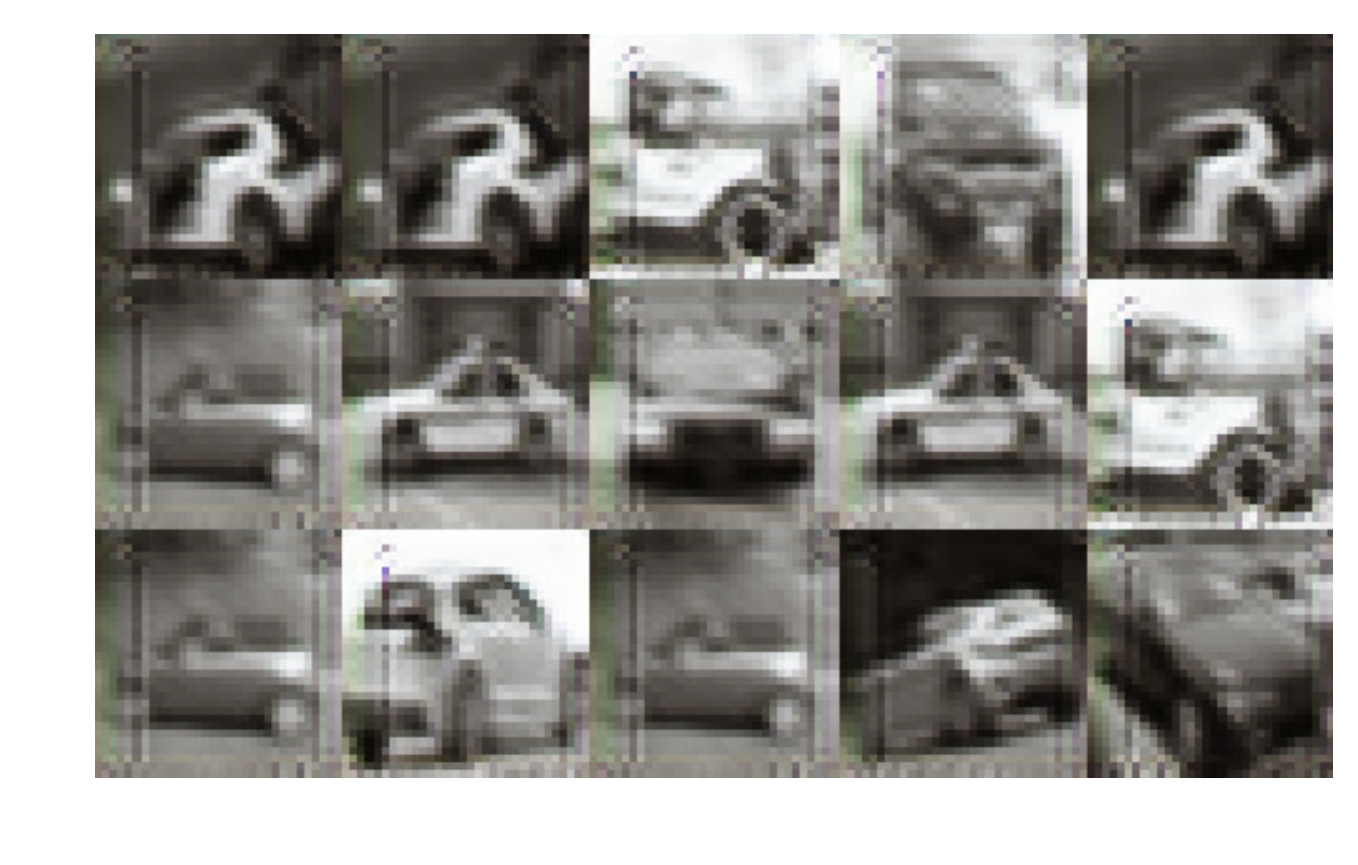}}
    & 0.99 
    & \makecell{\includegraphics[scale=0.15]{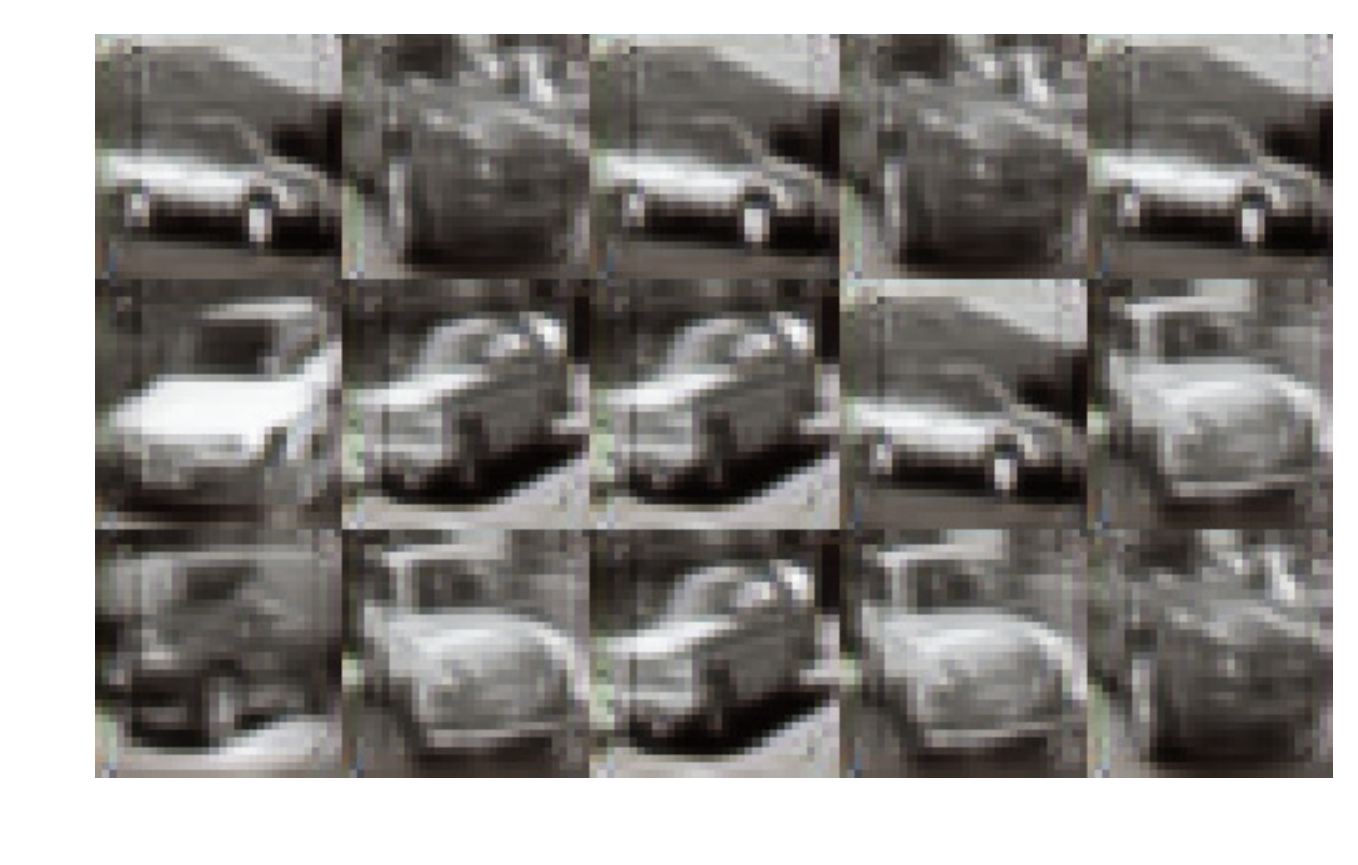}} 
    & 0.01 
    & \makecell{\includegraphics[scale=0.15]{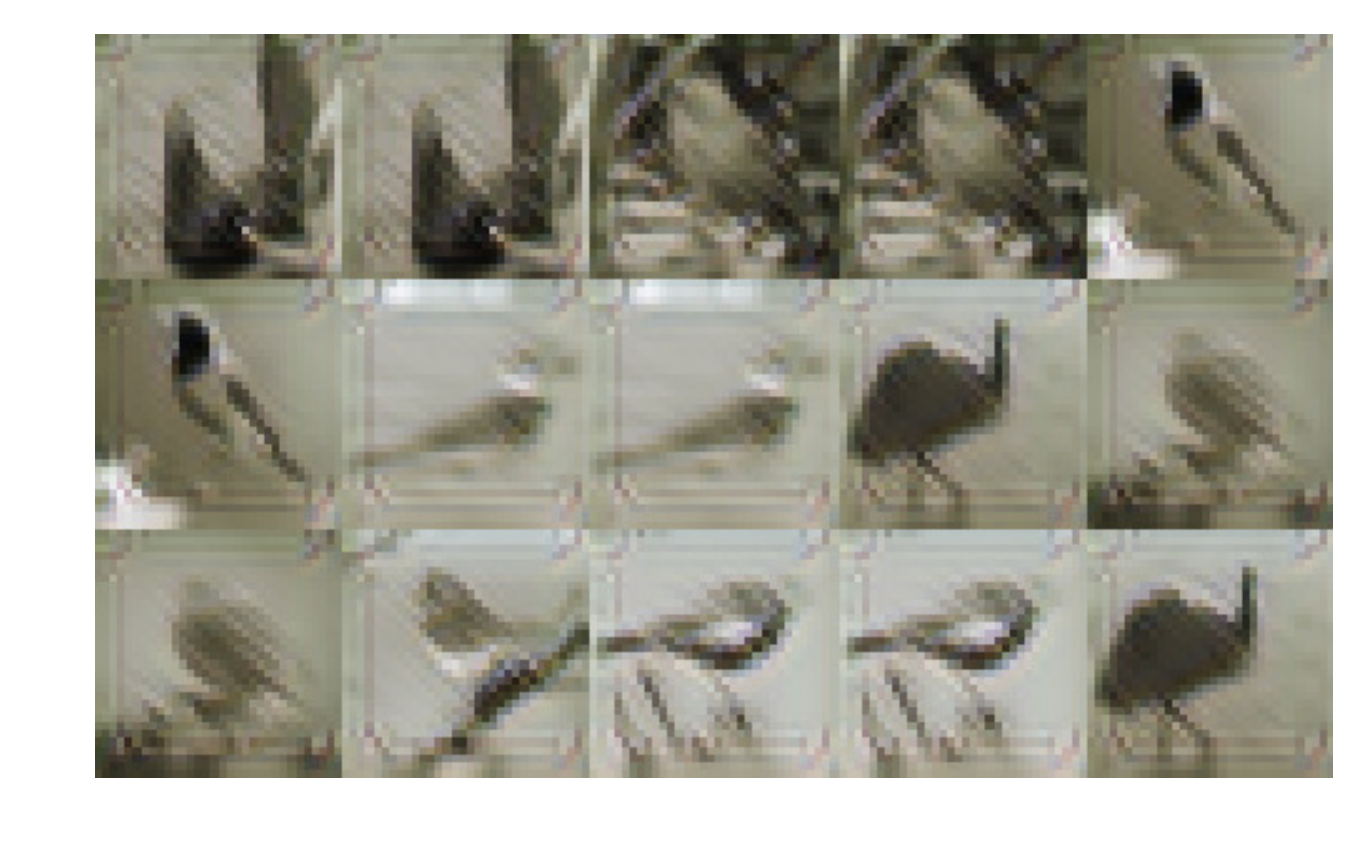}} 
    & 0.99 
    & \makecell{\includegraphics[scale=0.15]{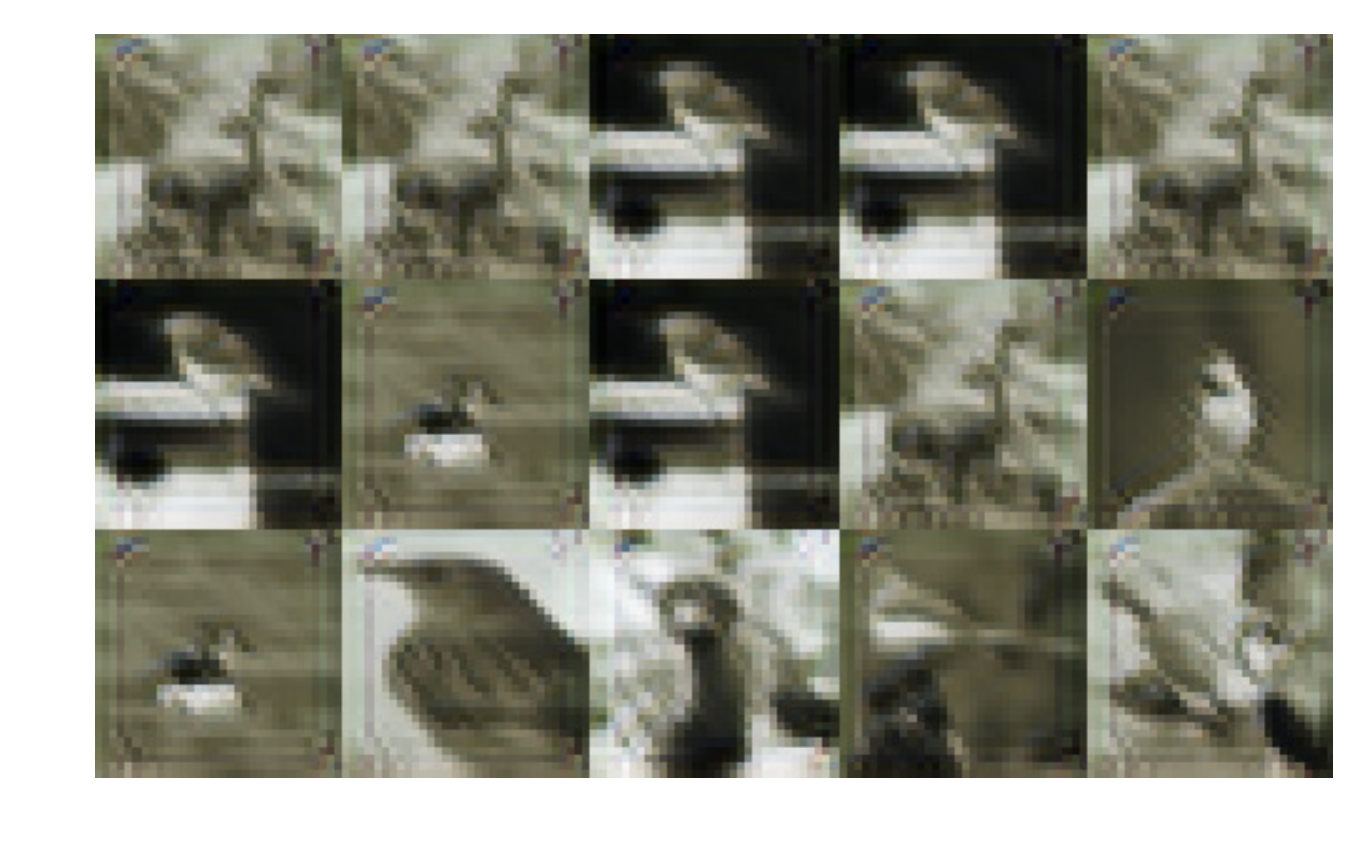}} \\ \hline
    \end{tabular}
    \label{tab:components12cifar10}
\end{table*}{}

\subsection{DeepDIG Component Analysis}
\label{sec:deepdig-component-results}
To recall from Section~\ref{sec:framework}, for a given source class $s$ and target class $t$, DeepDIG entails two important components including an adversarial example generation  from class $s$ to $t$ --See Figure~\ref{fig:deepdig} (I), another adversarial example generation model mapping adversarial examples found in component (I) back to the class $s$ region --See Figure~\ref{fig:deepdig} (II). Since adversarial example generation methods in components (I) and (II) plays an essential role in generating borderline samples, it is necessary to evaluate and analyze their performance. To this end, we run DeepDIG against models described in Table~\ref{tab:dnns}. For DNNs of MNIST, FashionMNIST, and CIFAR10  we investigate the class pairs (`1', `2'), (`Trouser', `Pullover'), and  (`Automobile', `Bird'), whose results are shown in Tables~\ref{tab:components12mnsit},~\ref{tab:components12fashionmnsit}, and \ref{tab:components12cifar10},  respectively. We evaluate the components (I) and (II) in terms of two factors, namely the accuracy of their adversarial example generation and the quality of the generated examples. Hence, results for each DNN include the accuracy score as well as the visualization of some of the generated samples (chosen randomly). Note that since component (I) generates adversarial examples that are miss-classified by a DNN, the smaller value for accuracy means more success. Component (II), however,  performs the \emph{reverse} adversarial example generation (i.e., mapping component (I)'s adversarial examples back to their correct classification region) and thus in this component higher value for accuracy means more success. Based on the results presented in Tables~\ref{tab:components12mnsit},~\ref{tab:components12fashionmnsit}, and \ref{tab:components12cifar10}, we make the following observation.

\begin{itemize}
    \item For all DNNs, both components (I) and (II) are shown to be capable of generating samples with very high accuracy\footnote{We found similar results for the f1 score.}.
    \item Overall, for all of DNNs the generated examples have high quality. The colors have been lost in generated examples for  $\text{CIFAR10}_\text{ResNet}$ and $\text{CIFAR10}_\text{GoogleNet}$. However, generated samples are visibly automobiles and birds. 
\end{itemize}{}

We can conclude that components (I) and (II) of DeepDIG perform as expected and are reliable modules for borderline instance generation.

\subsection{Baseline Comparison}
\label{sec:baselines}

To further evaluate the performance of DeepDIG for borderline instance generation, we compare it against several baseline methods described as follows.

\begin{itemize}
    \item \textbf{Random Pair Borderline Search (RPBS).} One may wonder that Algorithm~\ref{alg:middle} can be applied directly to any two samples as long as they are at the opposite sides of the decision boundary. Hence, in this baseline, we randomly pair up training samples from classes $s$ and $t$ and then apply Algorithm~\ref{alg:middle}. More specifically, the input to Algorithm~\ref{alg:middle} is $\{(x_i,x_j) | C(x_i)=s, C(x_j)=t\}$ where $x_i$ and $x_j$ belong to the training set and are randomly paired up. The authors in~\cite{yousefzadeh2019investigating} used a similar method to study the decision boundary of DNNs.
    \item \textbf{Embedding-nearest Pair Borderline Search (EPBS).} In this baseline method, instead of randomly pairing up samples at the opposite sides of the decision boundary between two classes, we pair a sample classified as $s$ with its \emph{nearest} sample at the opposite side of the boundary that is classified as $t$. The distance between the two samples is calculated in the embedding space. More formally,  the input to Algorithm~\ref{alg:middle} is $\{(x_i,x_j) | C(x_i)=s, C(x_j)=t, x_j = min_{x_t} || \mathcal{F}(x_i) - \mathcal{F}(x_t)||_2^2\}$ where, recalling from Section~\ref{sec:problem},  $\mathcal{F}$ denotes the embedding space learned by a DNN. The reason for including this baseline is that by considering a better-guided trajectory between two samples,  EPBS can hopefully generate borderline samples more effectively than RPBS.   
\end{itemize}{}

Tables~\ref{tab:baselinemnsit}, \ref{tab:baselinefashionmnist}, and \ref{tab:baselinecifar10} show the results for MNIST, FashionMNIST, and CIFAR10, respectively. Based on the results presented in these tables, we compare DeepDIG with baseline methods in terms of three important factors explained in the following. We should emphasize that an effective method is expected to succeed in all three factors.
 
\begin{table*}[!htb]
\setlength\tabcolsep{1.6pt} 
    \renewcommand{\arraystretch}{1.0} 
\caption{Comparing DeepDIG  with baseline methods (MNIST dataset) }
    \centering
       
    \begin{tabular}{c||c|c|c|c|c|c}
    \hline
    DNN & \multicolumn{3}{c|}{$\text{MNIST}_{\text{CNN}}$} & \multicolumn{3}{c}{$\text{MNIST}_{\text{FCN}}$} \\
    \hline
    \diagbox{Method}{Factor}  & $\overline{|f_s(x) - f_t(x)|}$ & Visualisation &  \makecell{ Success\\ Rate}  &   $\overline{|f_s(x) - f_t(x)|}$ & Visualisation &  \makecell{ Success\\ Rate} \\ \hline \hline
  DeepDIG 
  & $4.41 \times 10^{-5}$ 
  & \makecell{\includegraphics[scale=0.12]{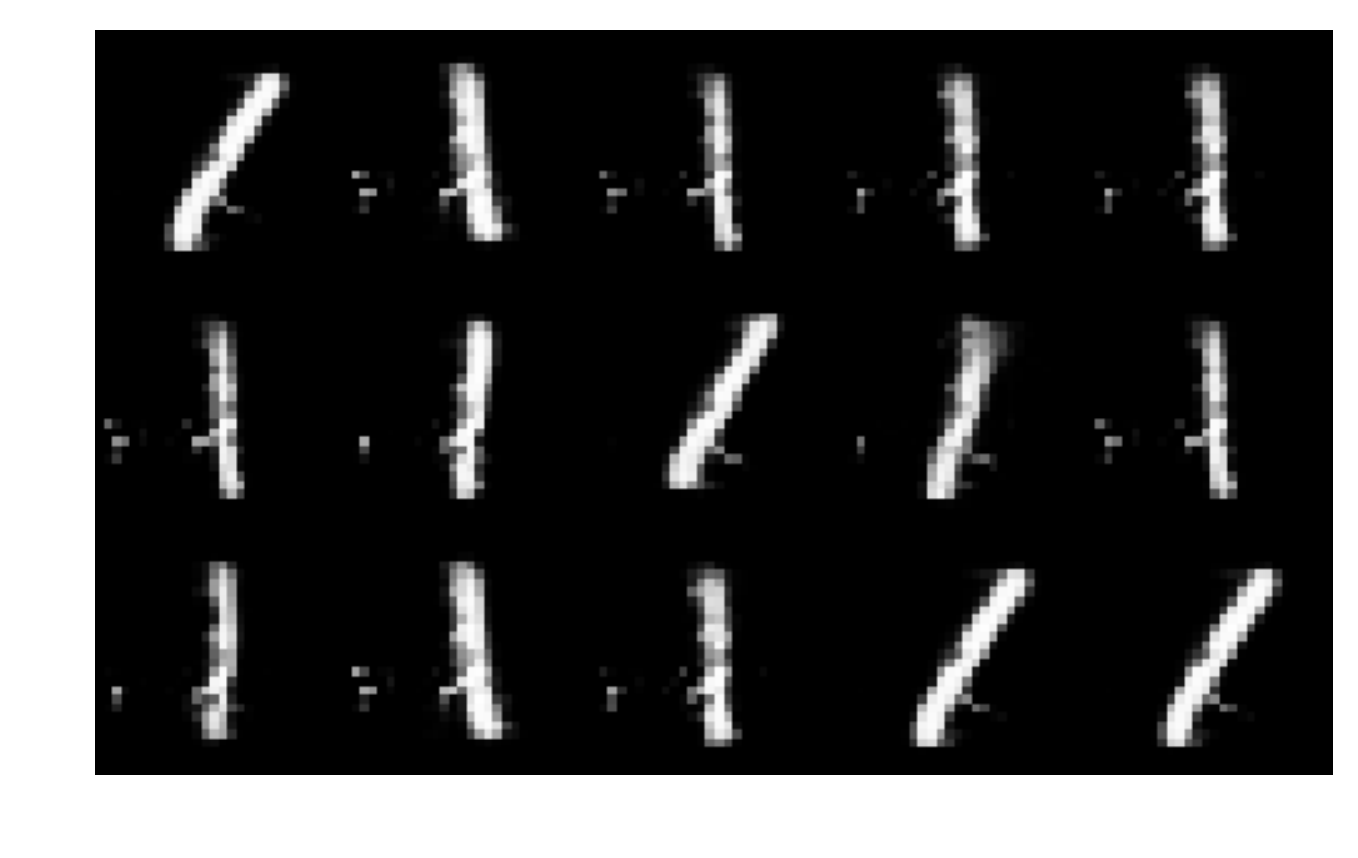} 
  \includegraphics[scale=0.12]{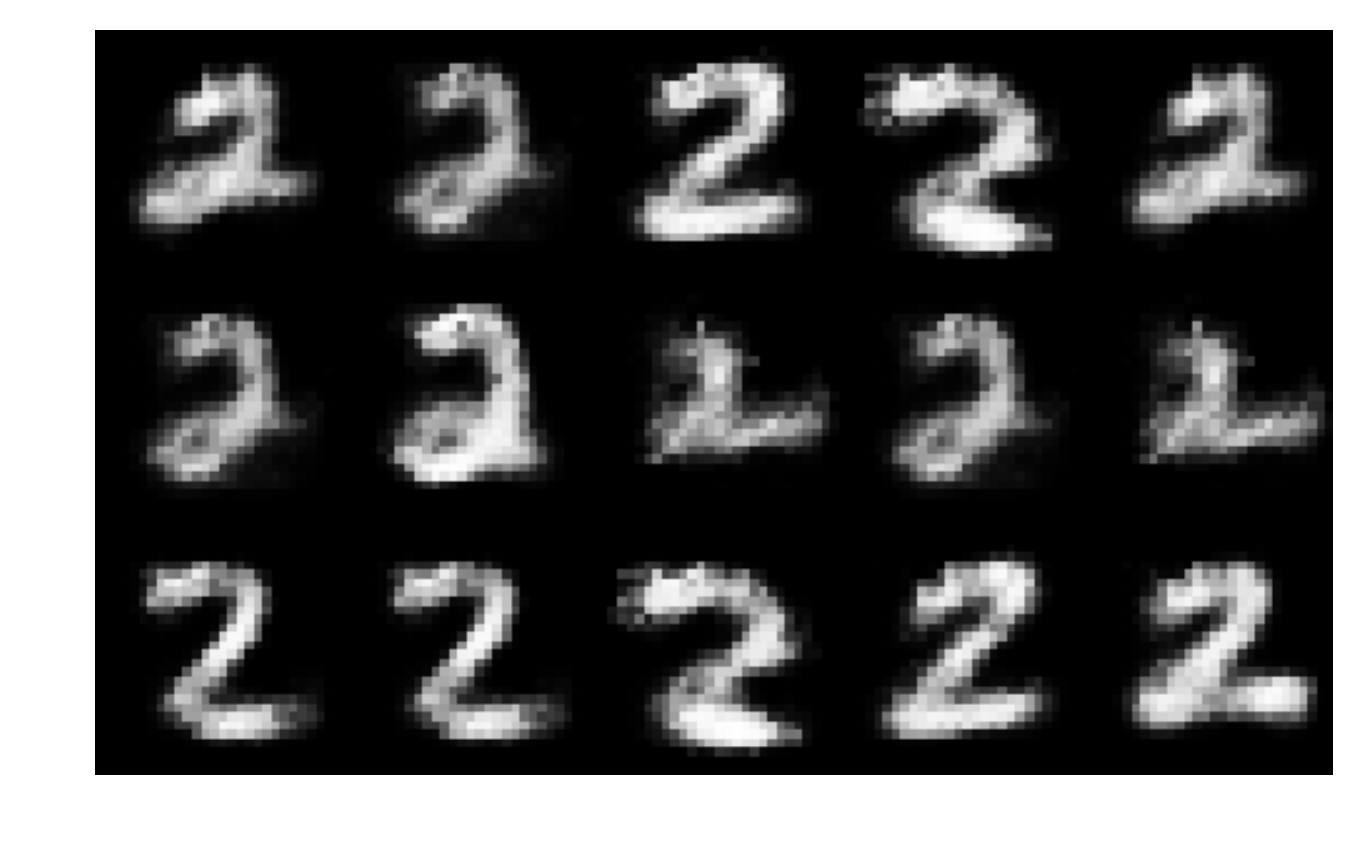}} 
  & 98.66
  &   $4.43 \times 10^{-5}$
  & \makecell{\includegraphics[scale=0.12]{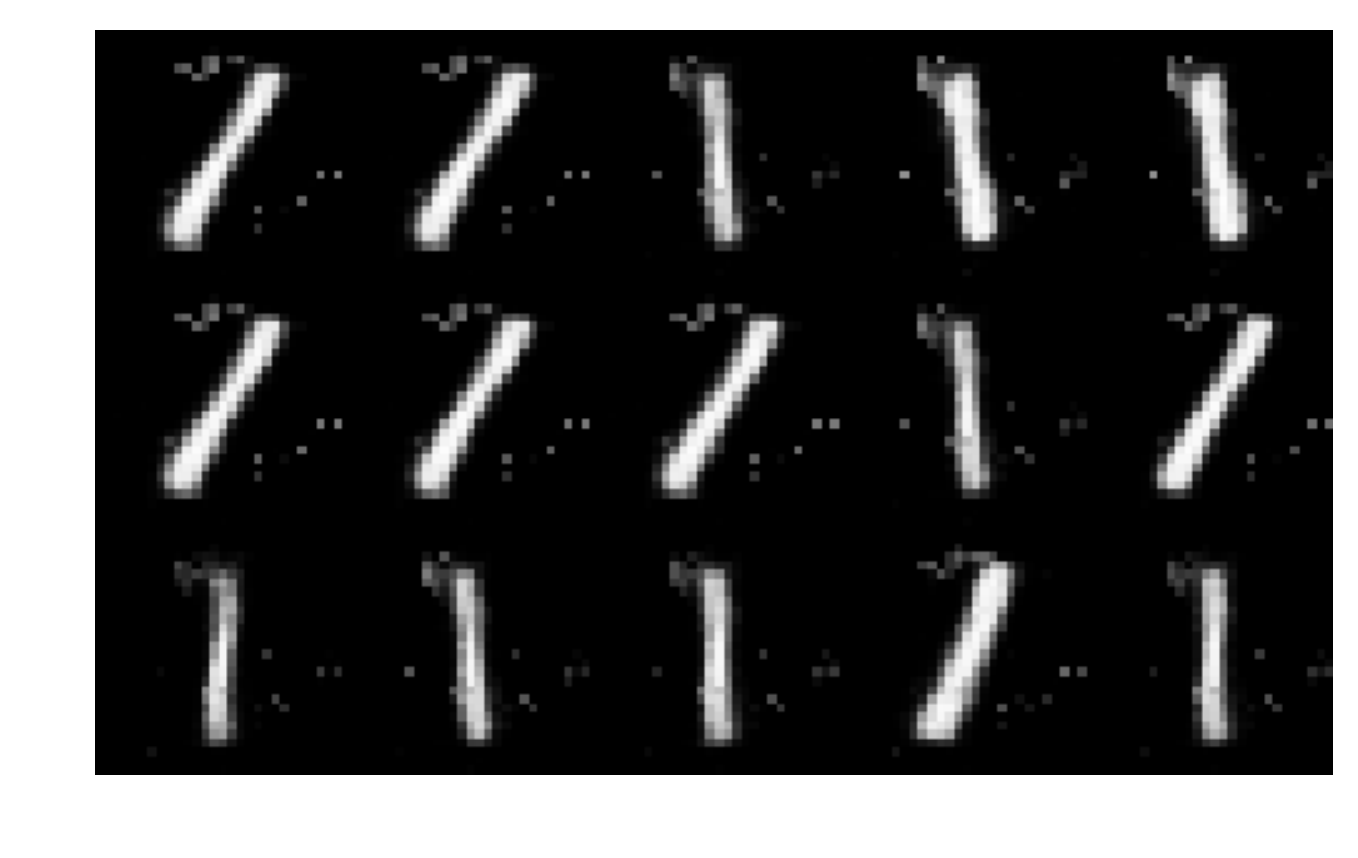} 
  \includegraphics[scale=0.12]{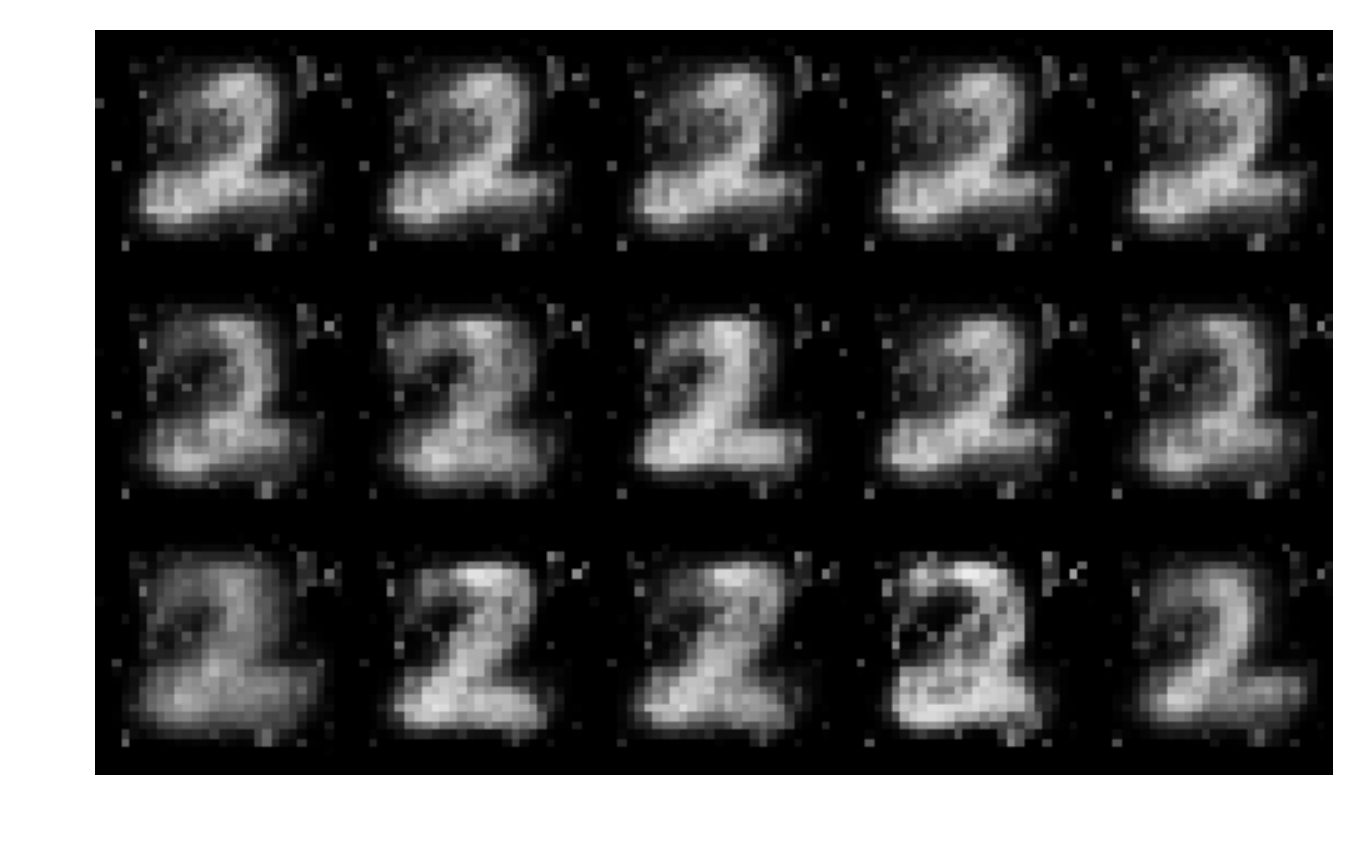}}
  & 99.19  \\ \hline
  
  RPBS 
  & $4.44 \times 10^{-5}$ 
  &\makecell{\includegraphics[scale=0.12]{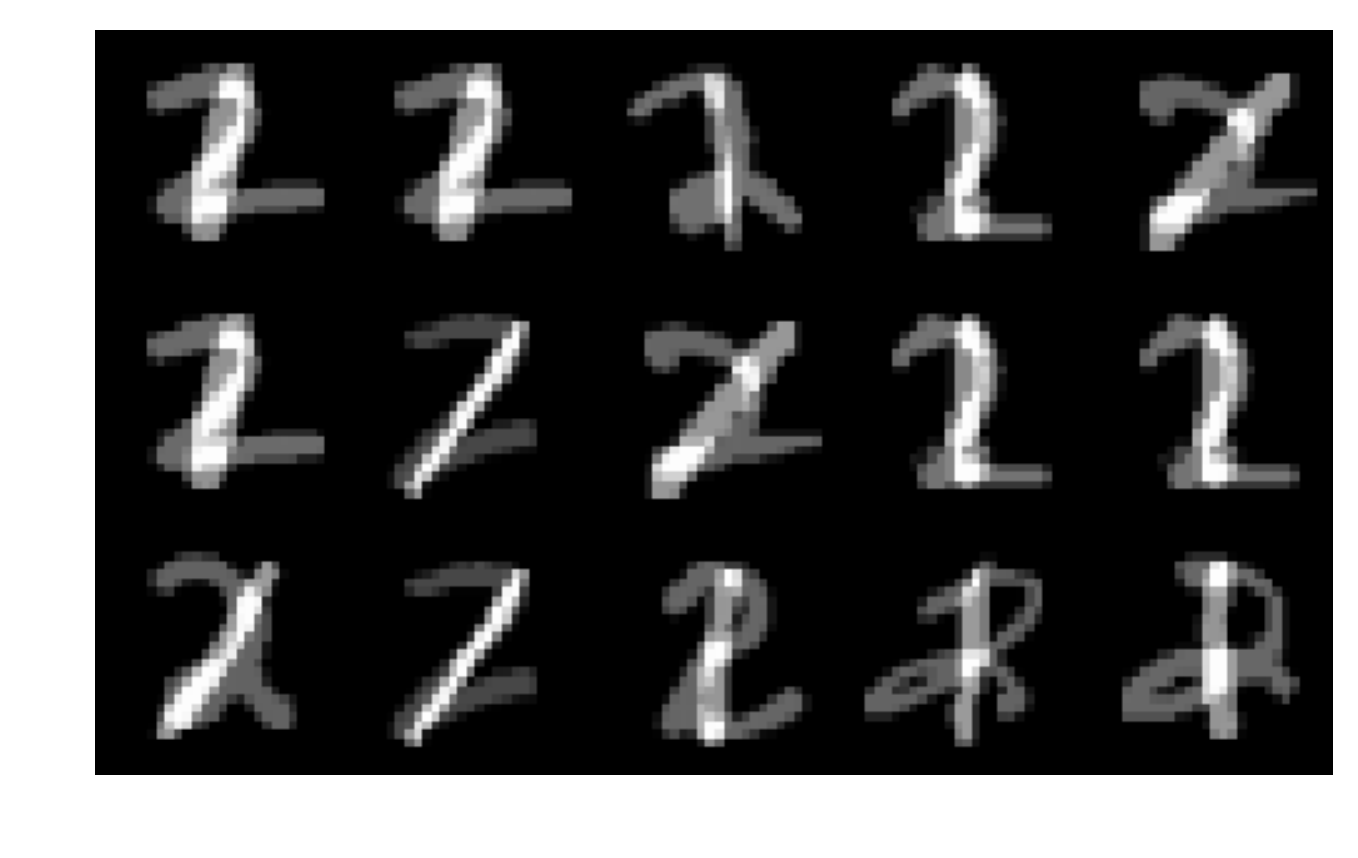}} 
  & 79.33 
  & $4.4 \times 10^{-4}$
  & \makecell{\includegraphics[scale=0.12]{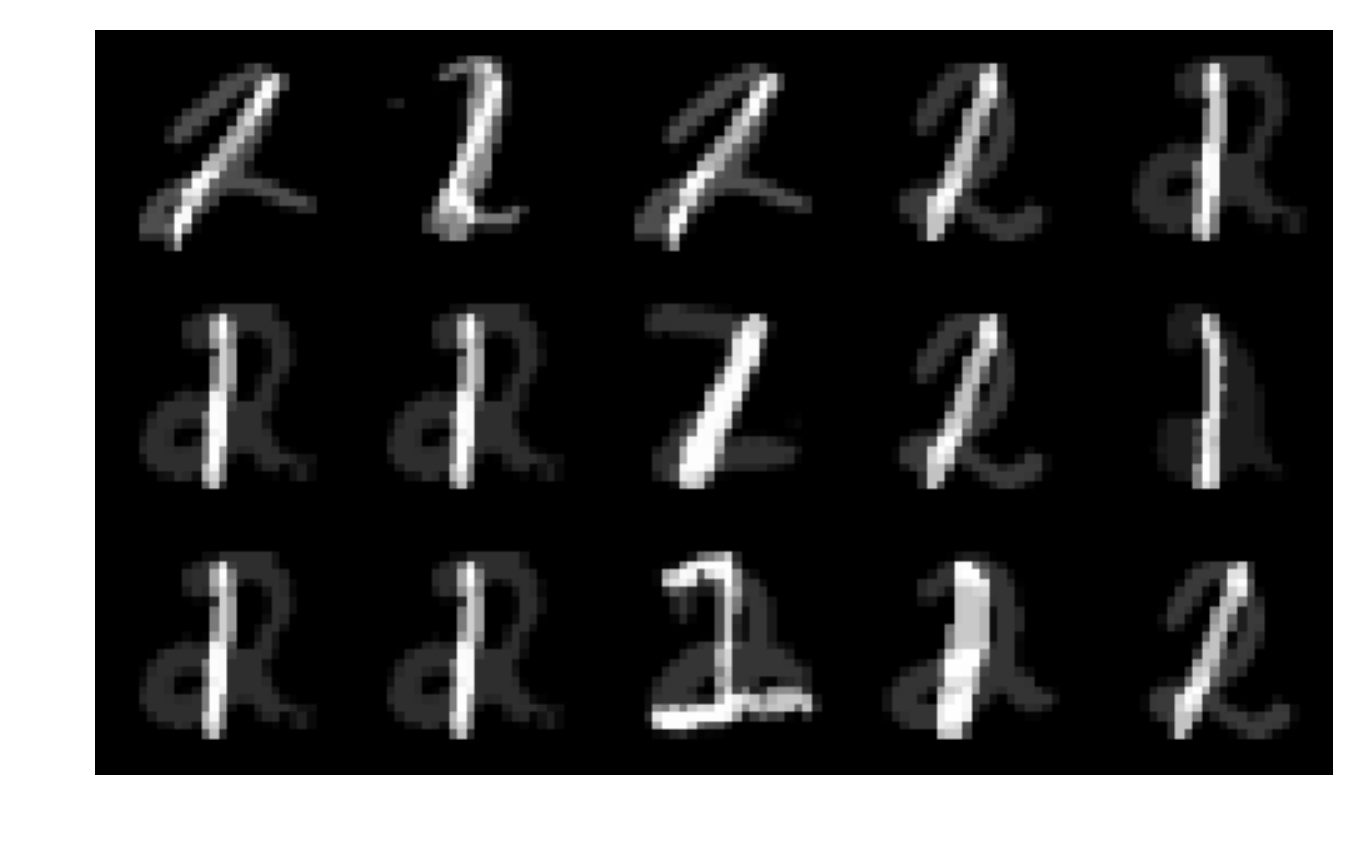}}
  & 48.48 \\ \hline
  
  EPBS 
  & $4.5 \times 10^{-4}$
  & \makecell{\includegraphics[scale=0.12]{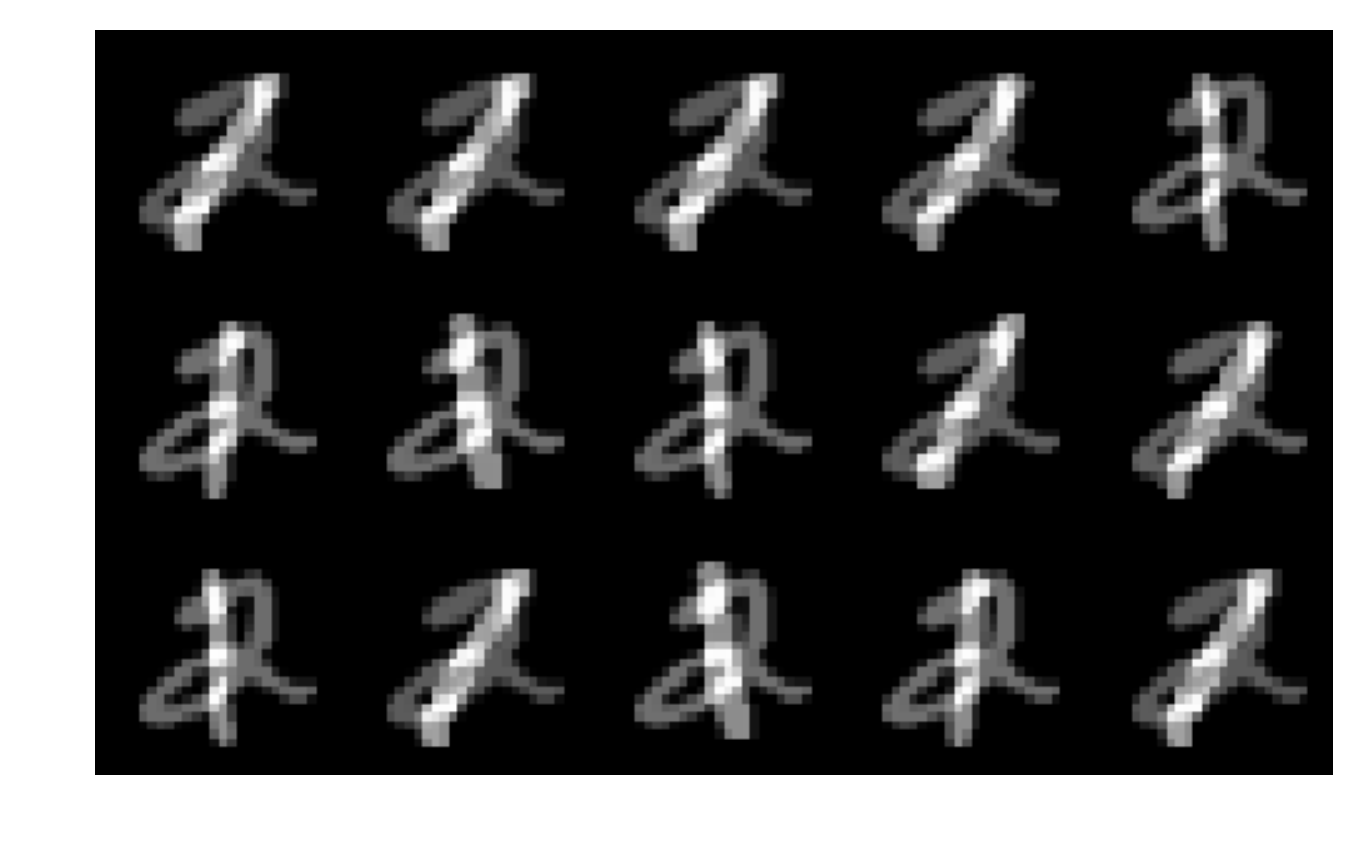} } 
  & 86.48 
  & $4.40 \times 10^{-5}$ 
  & \makecell{\includegraphics[scale=0.12]{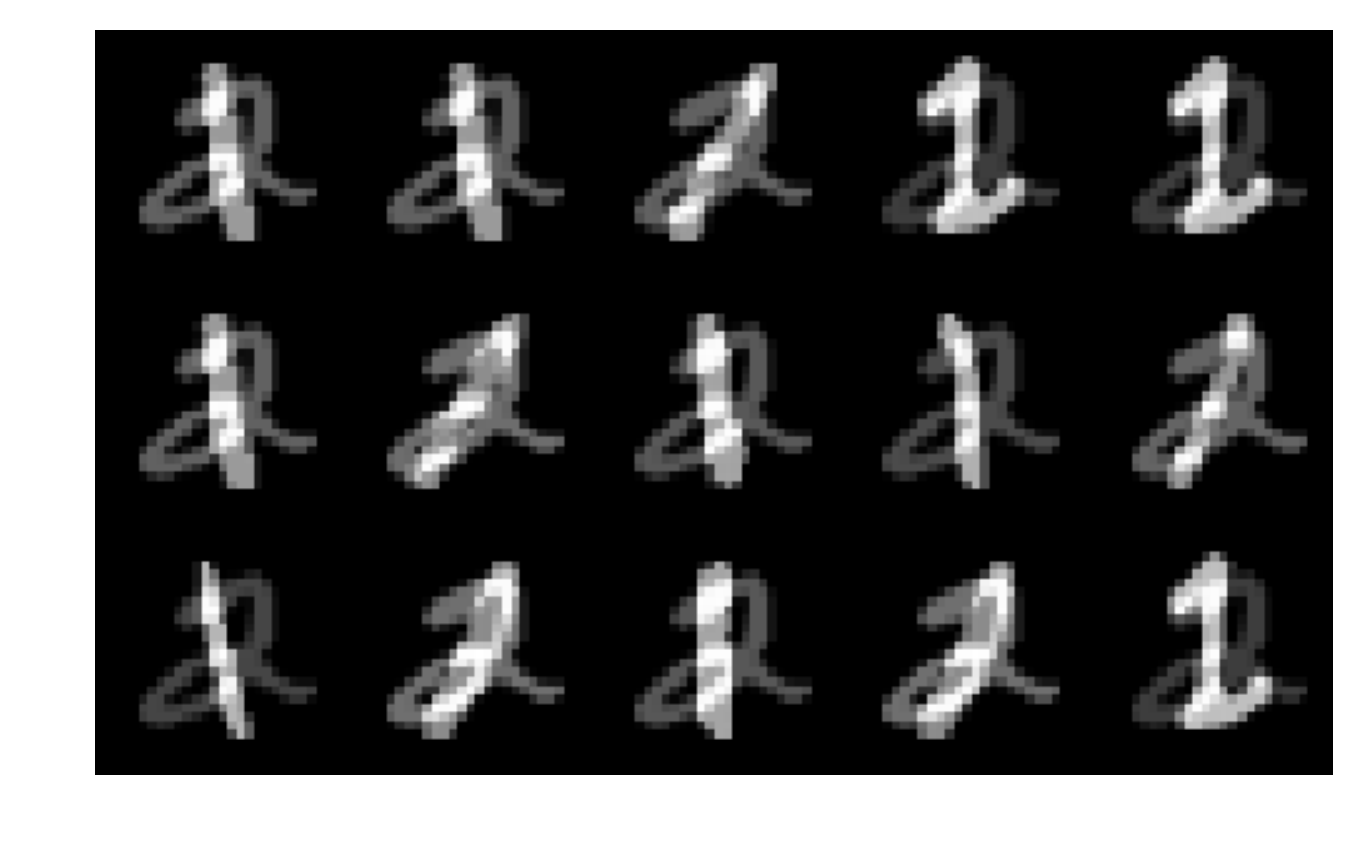}} 
  & 22.50
\\ \hline

    \end{tabular}
    \label{tab:baselinemnsit}
\end{table*}{}

\begin{table*}[!htb]
\setlength\tabcolsep{1.6pt} 
    \renewcommand{\arraystretch}{1.0} 
\caption{Comparing DeepDIG  with baseline methods (FashionMNIST dataset) }
    \centering
       
    \begin{tabular}{c||c|c|c|c|c|c}
    \hline
    DNN & \multicolumn{3}{c|}{$\text{FashionMNIST}_{\text{CNN}}$} & \multicolumn{3}{c}{$\text{FashionMNIST}_{\text{FCN}}$} \\
    \hline
    \diagbox{Method}{Factor}  & $\overline{|f_s(x) - f_t(x)|}$ & Visualisation &  \makecell{ Success\\ Rate} &   $\overline{|f_s(x) - f_t(x)|}$ & Visualisation &  \makecell{ Success\\ Rate} \\ \hline \hline
  DeepDIG 
  
  & $4.38 \times 10^{-5}$ 
  & \makecell{\includegraphics[scale=0.12]{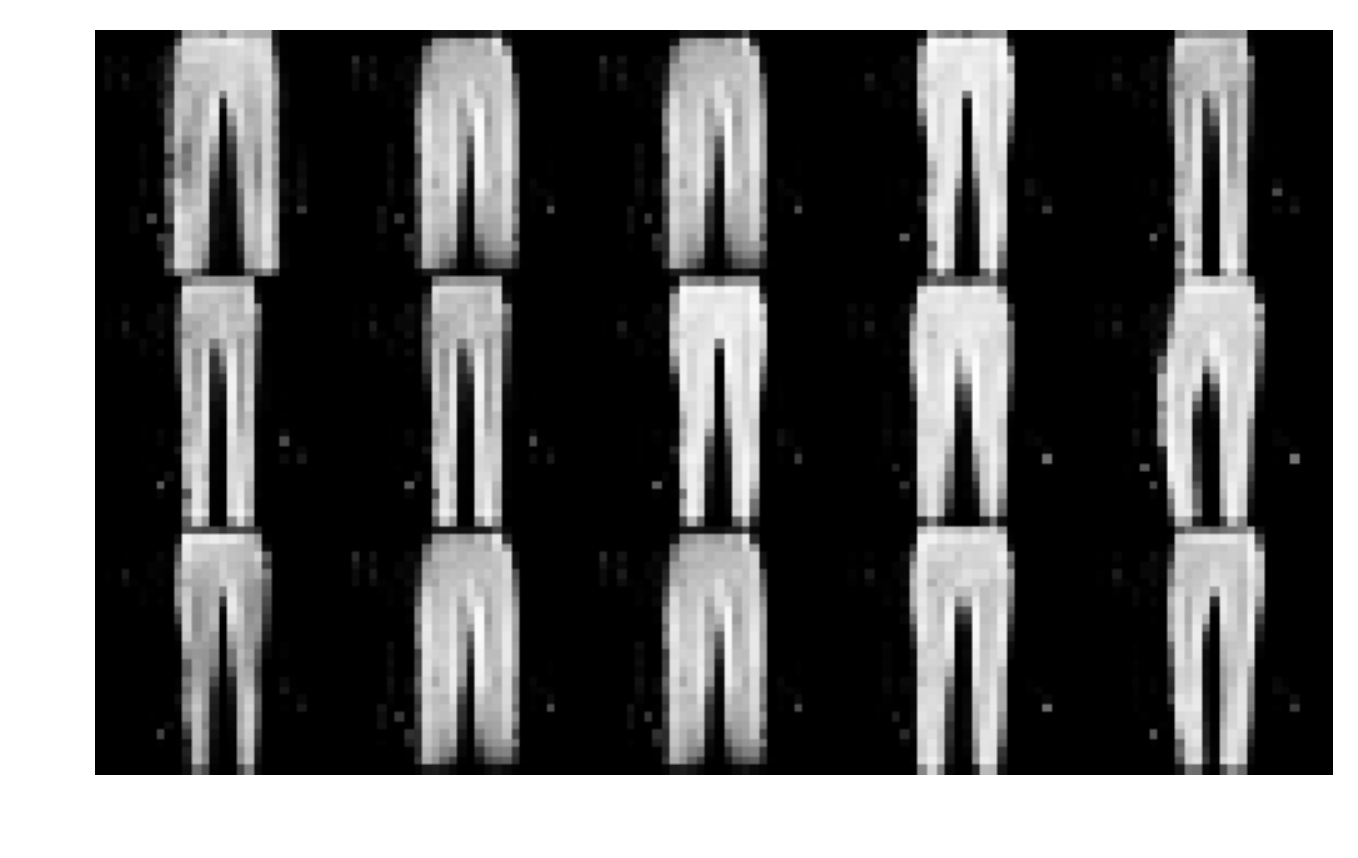} 
  \includegraphics[scale=0.12]{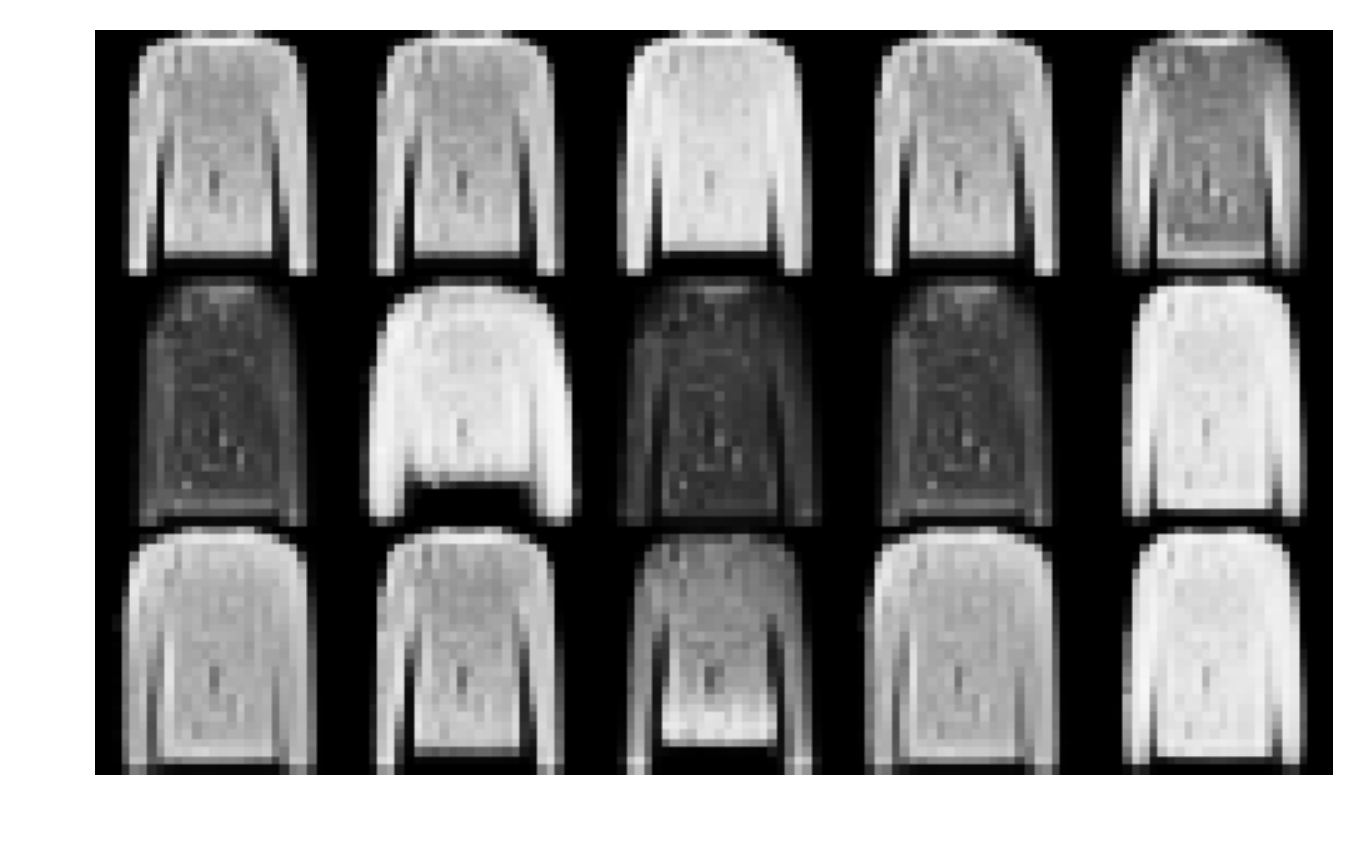}} 
  & 98.93 
  & $4.44 \times 10^{-5}$
  & \makecell{\includegraphics[scale=0.12]{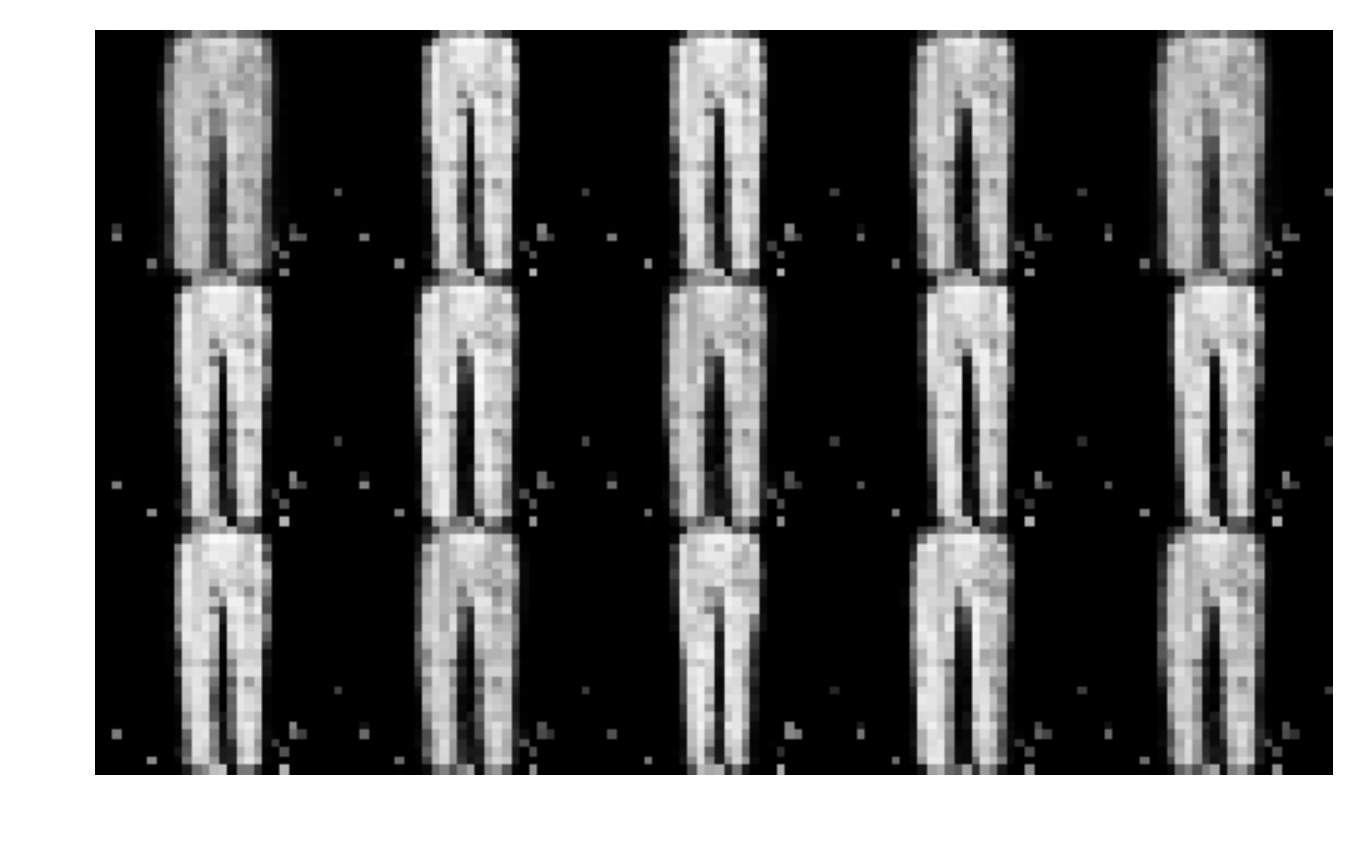} 
  \includegraphics[scale=0.12]{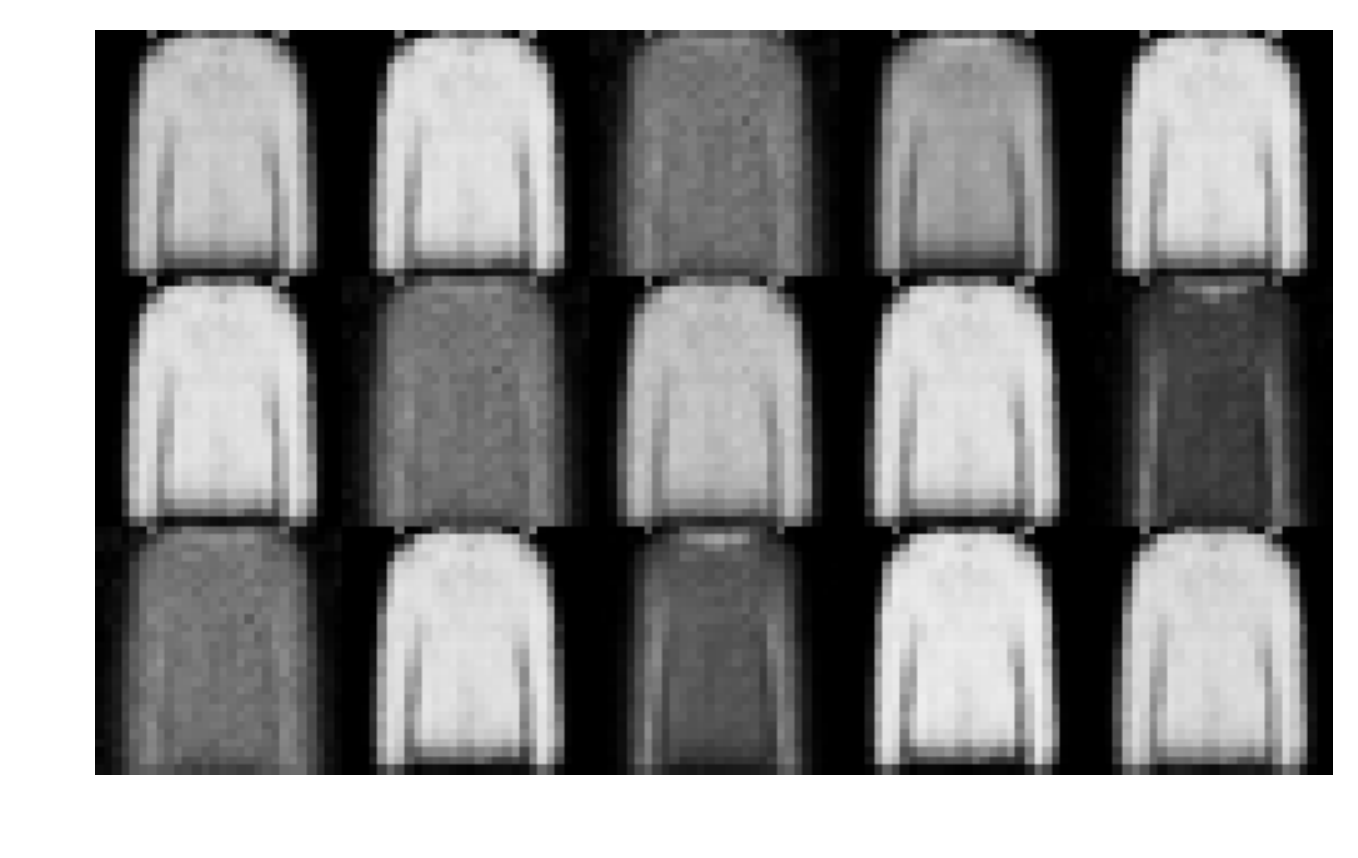}} 
  & 93.18
  \\ \hline
  
  RPBS 
  & $4.41 \times 10^{-5}$
  & \makecell{\includegraphics[scale=0.12]{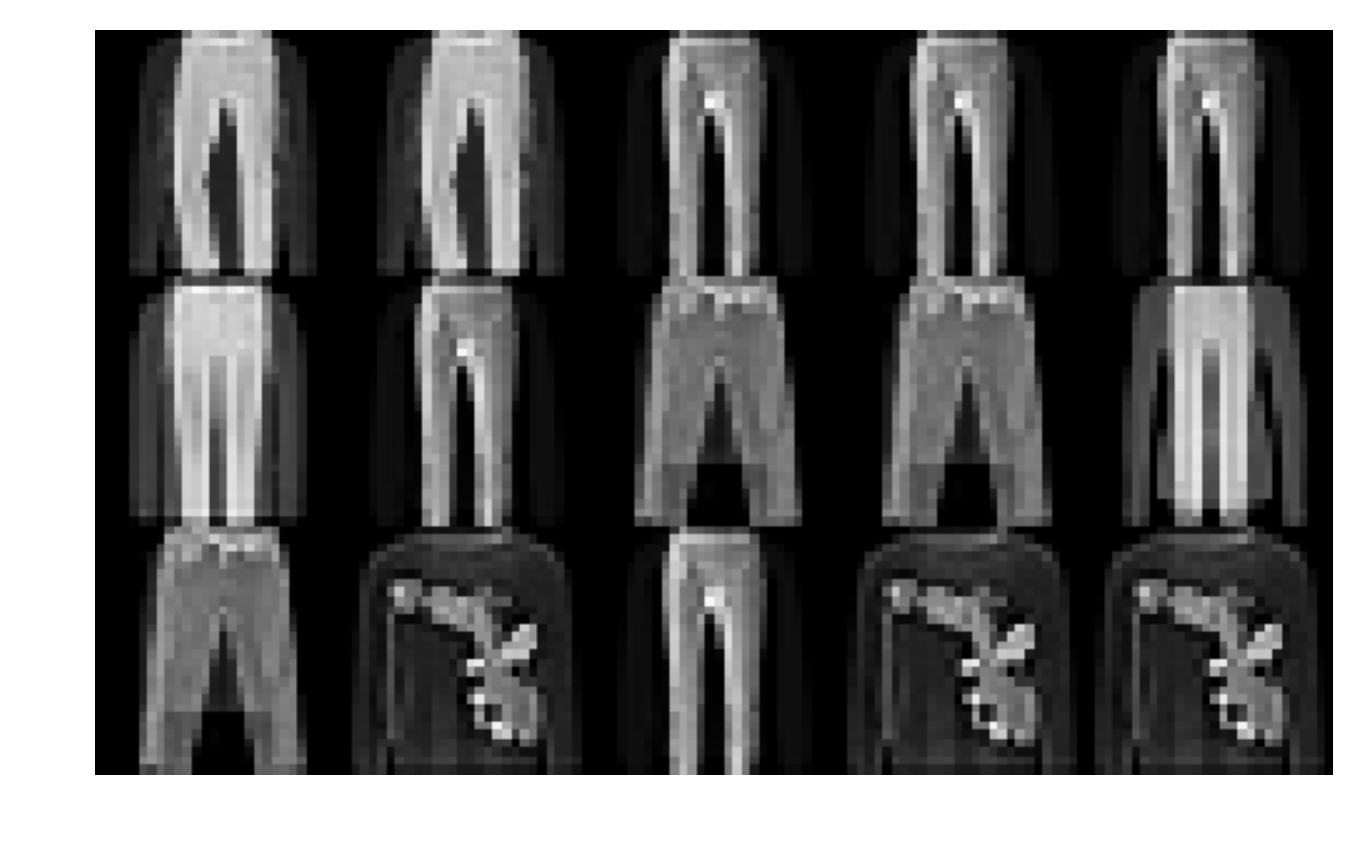}} 
  & 15.98
  & $4.47 \times 10^{-5}$
  & \makecell{\includegraphics[scale=0.12]{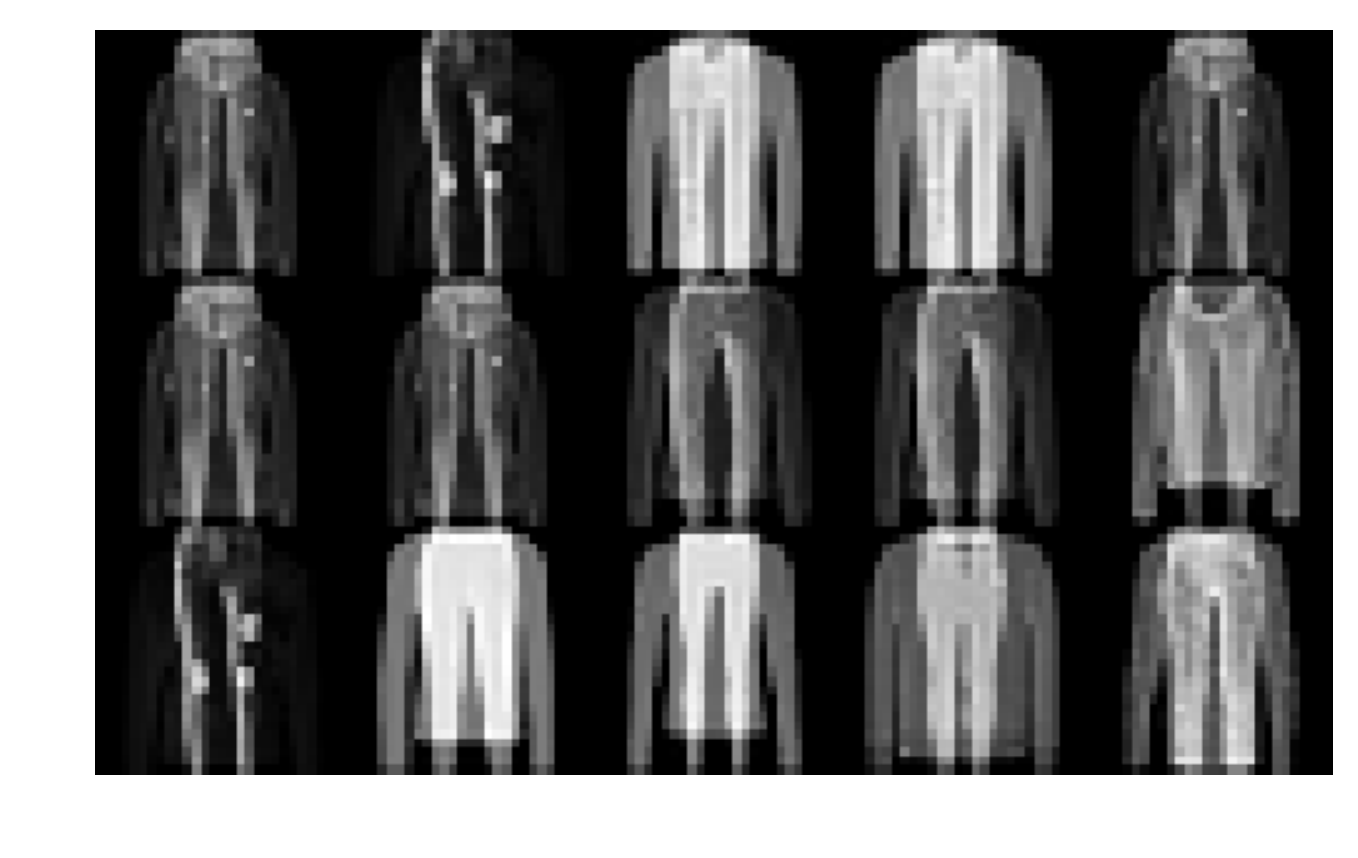}} 
  & 30.64 \\ \hline
 
 EPBS 
 & $4.42 \times 10^{-5}$ 
 & \makecell{\includegraphics[scale=0.12]{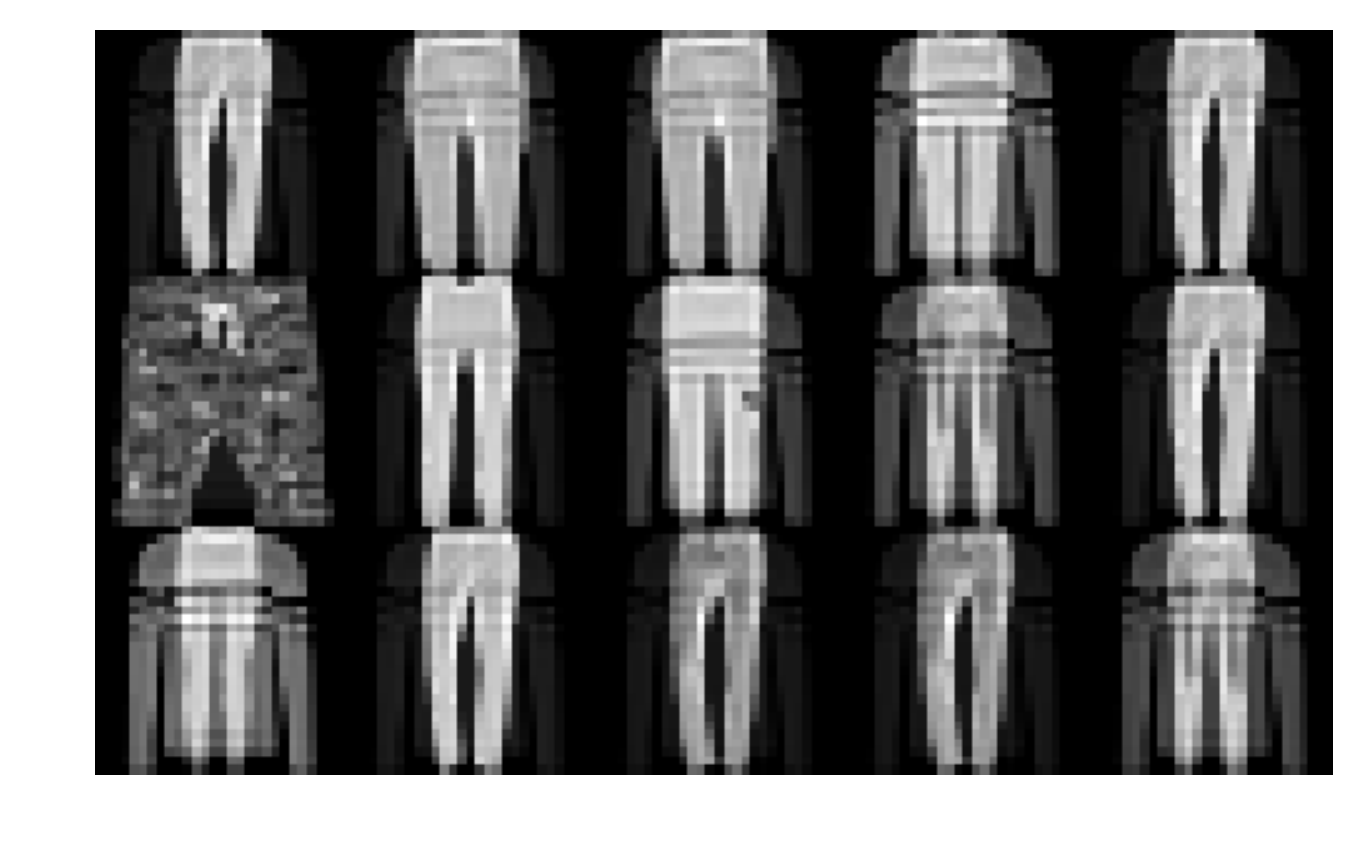}} 
 & 33.48
 & $4.39 \times 10^{-5}$
 & \makecell{\includegraphics[scale=0.12]{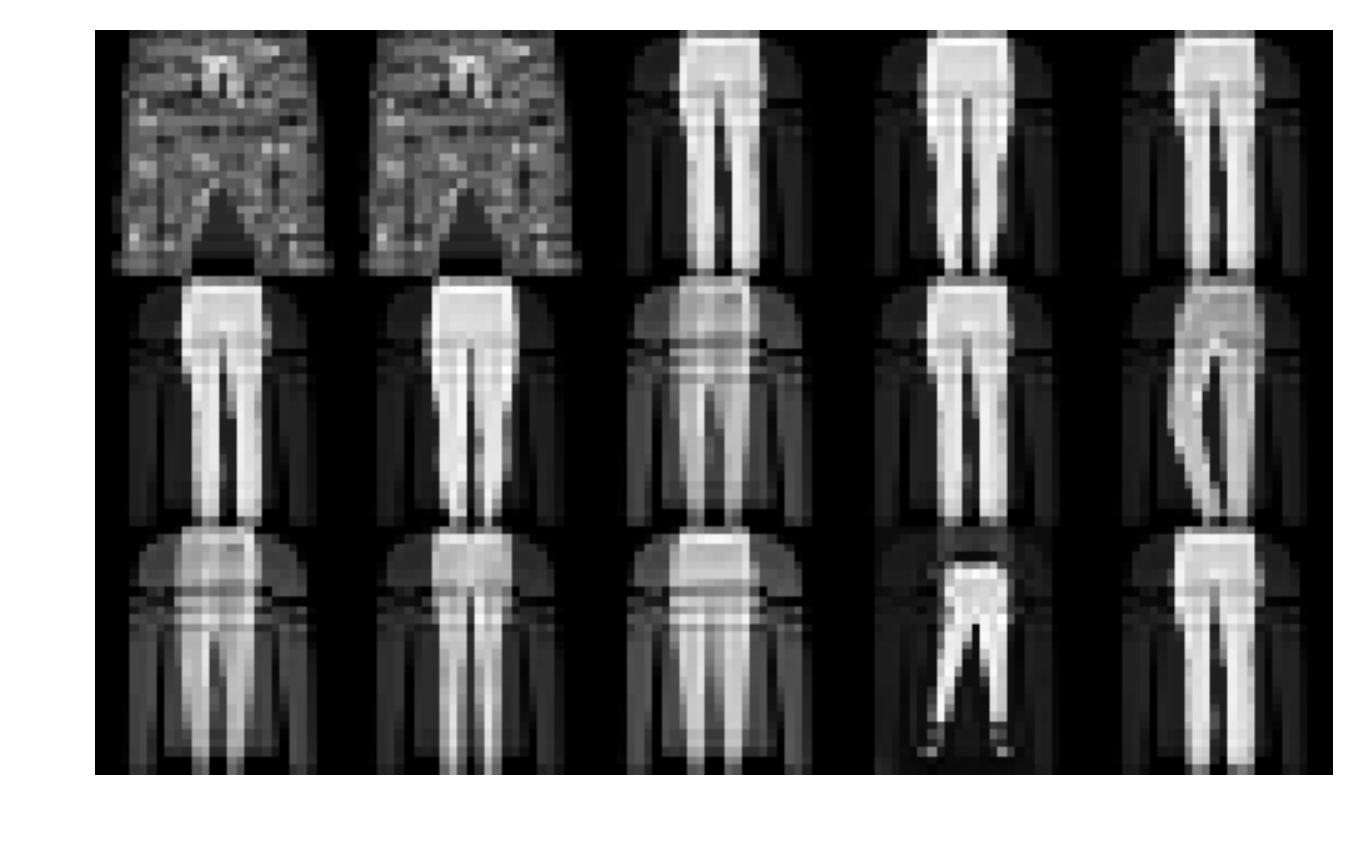}}  
 & 08.40 \\ \hline

    \end{tabular}
    \label{tab:baselinefashionmnist}
\end{table*}{}

\begin{table*}[!htb]
\setlength\tabcolsep{1.6pt} 
    \renewcommand{\arraystretch}{1.0} 
\caption{Comparing DeepDIG  with baseline methods (CIFAR10 dataset) }
    \centering
       
    \begin{tabular}{c||c|c|c|c|c|c}
    \hline
    DNN & \multicolumn{3}{c|}{$\text{CIFAR10}_{\text{ResNet}}$} & \multicolumn{3}{c}{$\text{CIFAR10}_{\text{GoogleNet}}$} \\
    \hline
    \diagbox{Method}{Factor}  & $\overline{|f_s(x) - f_t(x)|}$ & Visualisation &  \makecell{ Success\\ Rate} &   $\overline{|f_s(x) - f_t(x)|}$ & Visualisation &  \makecell{ Success\\ Rate} \\ \hline \hline
  
  DeepDIG 
   
  & $4.36 \times 10^{-5}$ 
  & \makecell{\includegraphics[scale=0.12]{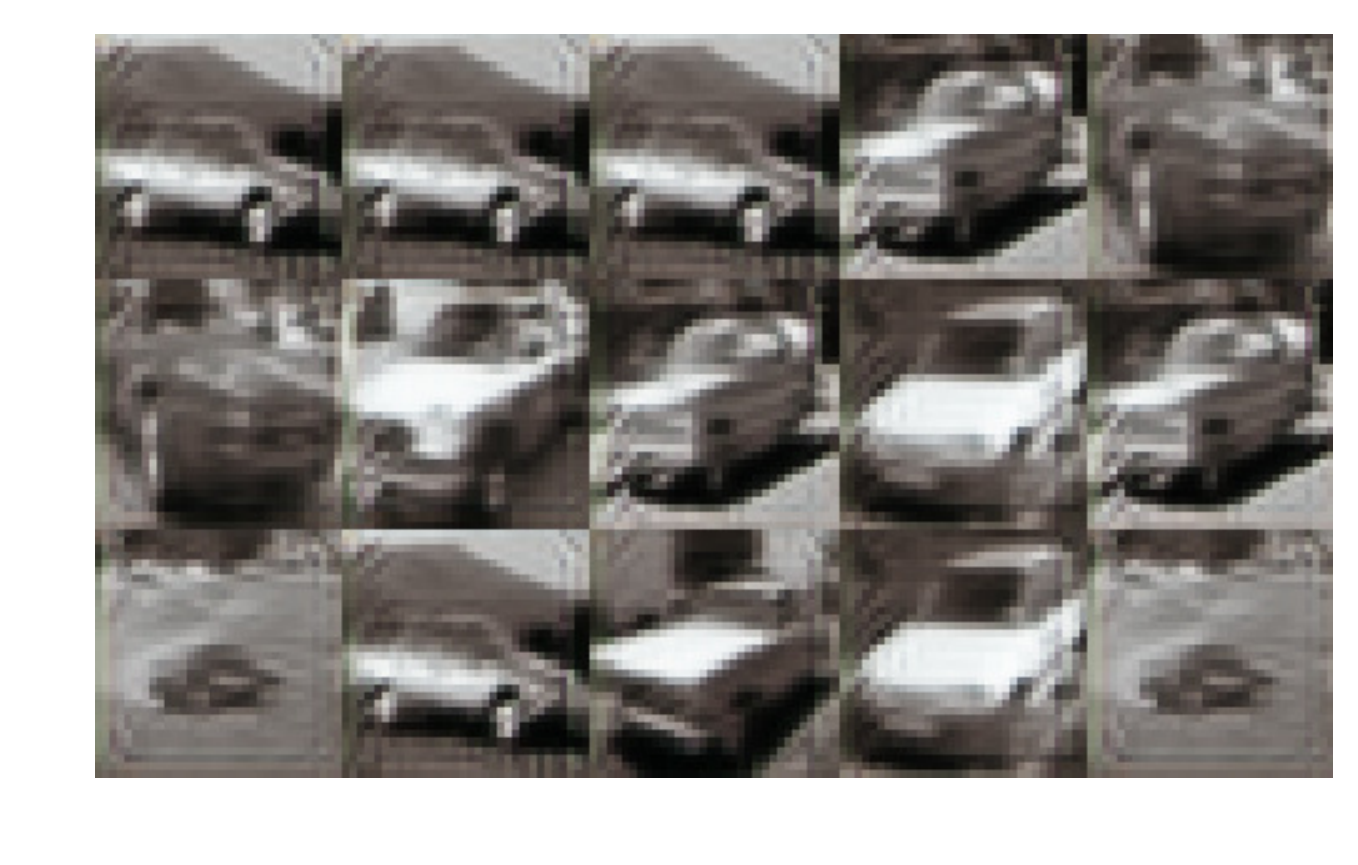} 
  \includegraphics[scale=0.12]{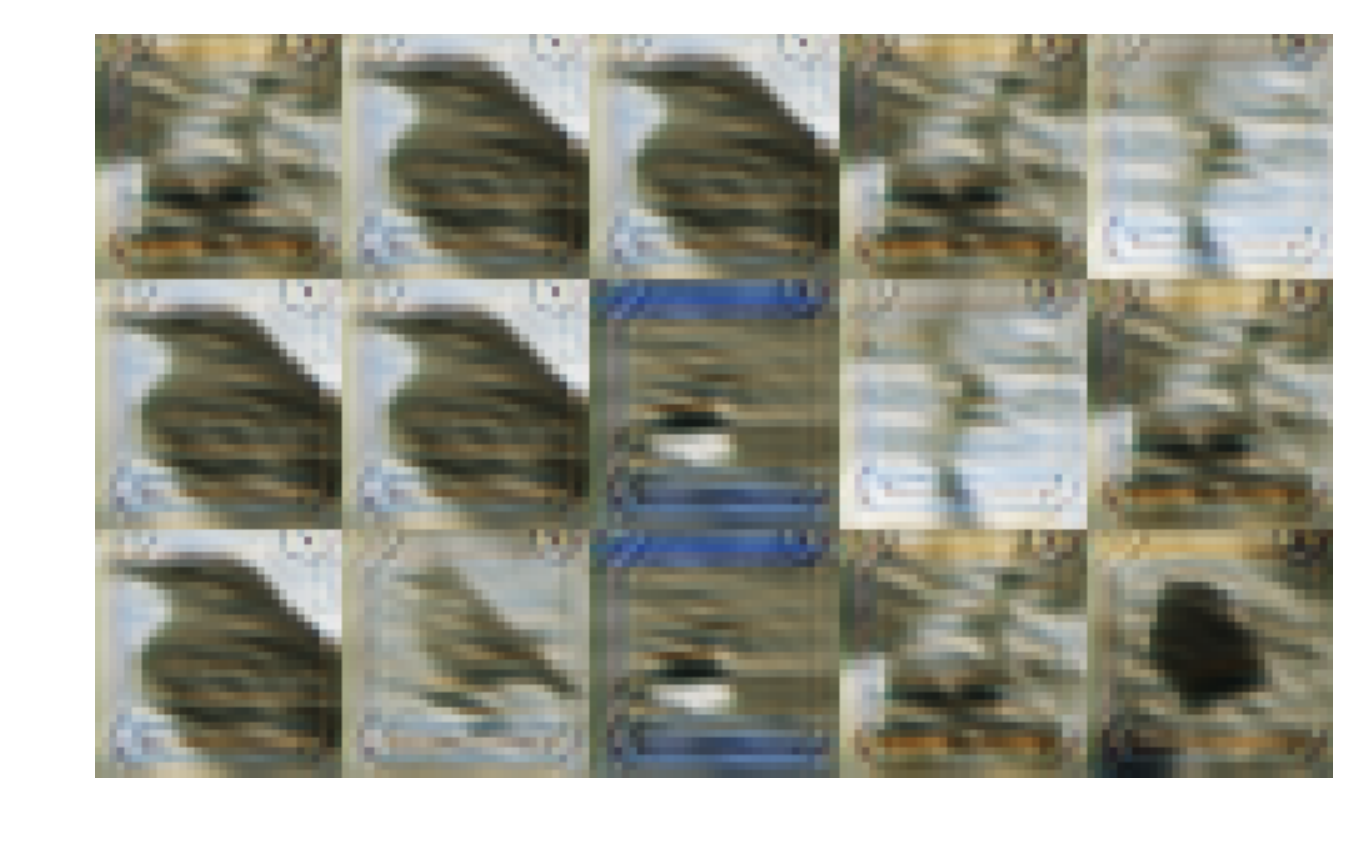}}
  & 70.60
  & $4.44 \times 10^{-5}$ 
  & \makecell{\includegraphics[scale=0.12]{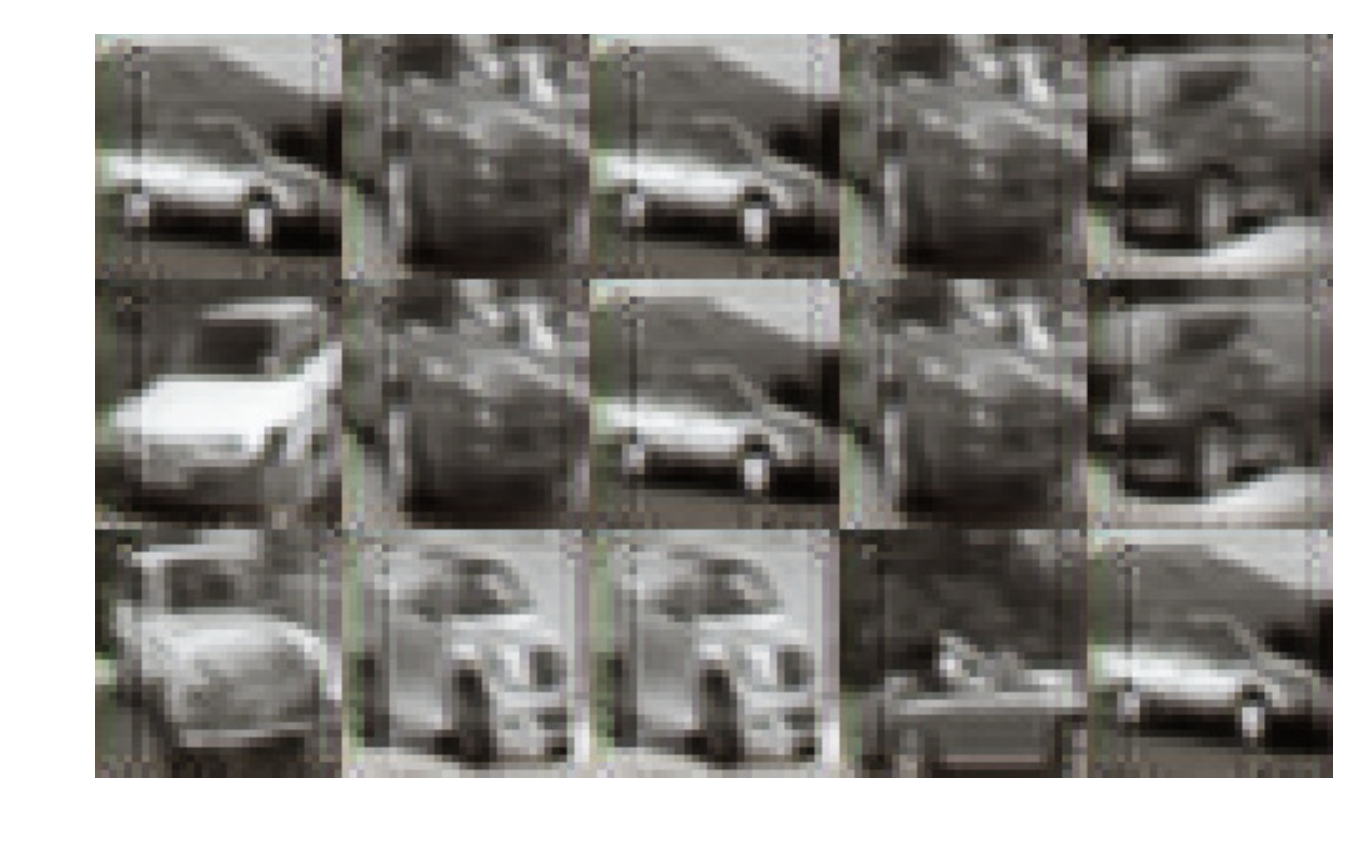} 
  \includegraphics[scale=0.12]{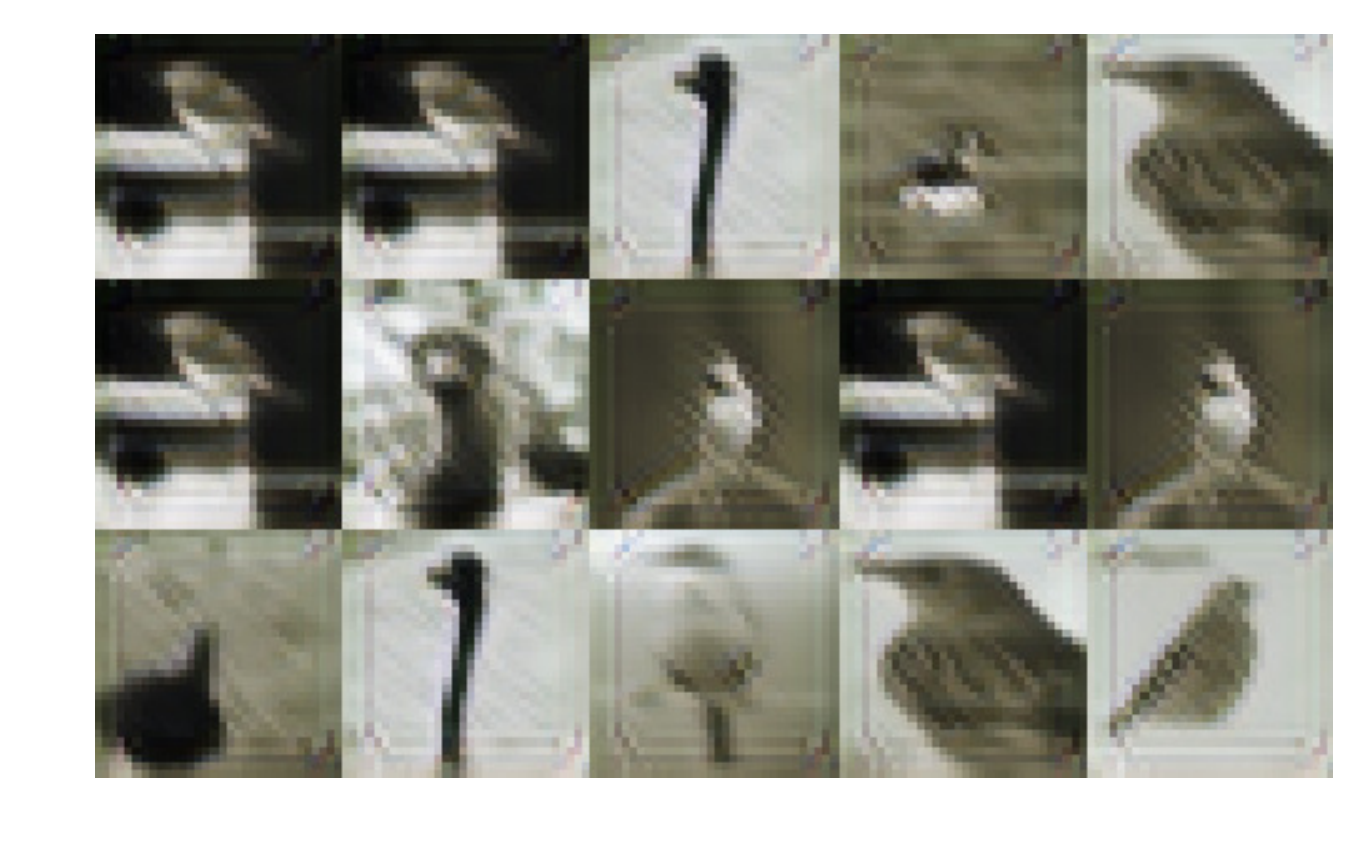}} 
  & 73.65 \\ \hline
  
  RPBS 
  
  & $4.46 \times 10^{-5}$ 
  & \makecell{\includegraphics[scale=0.12]{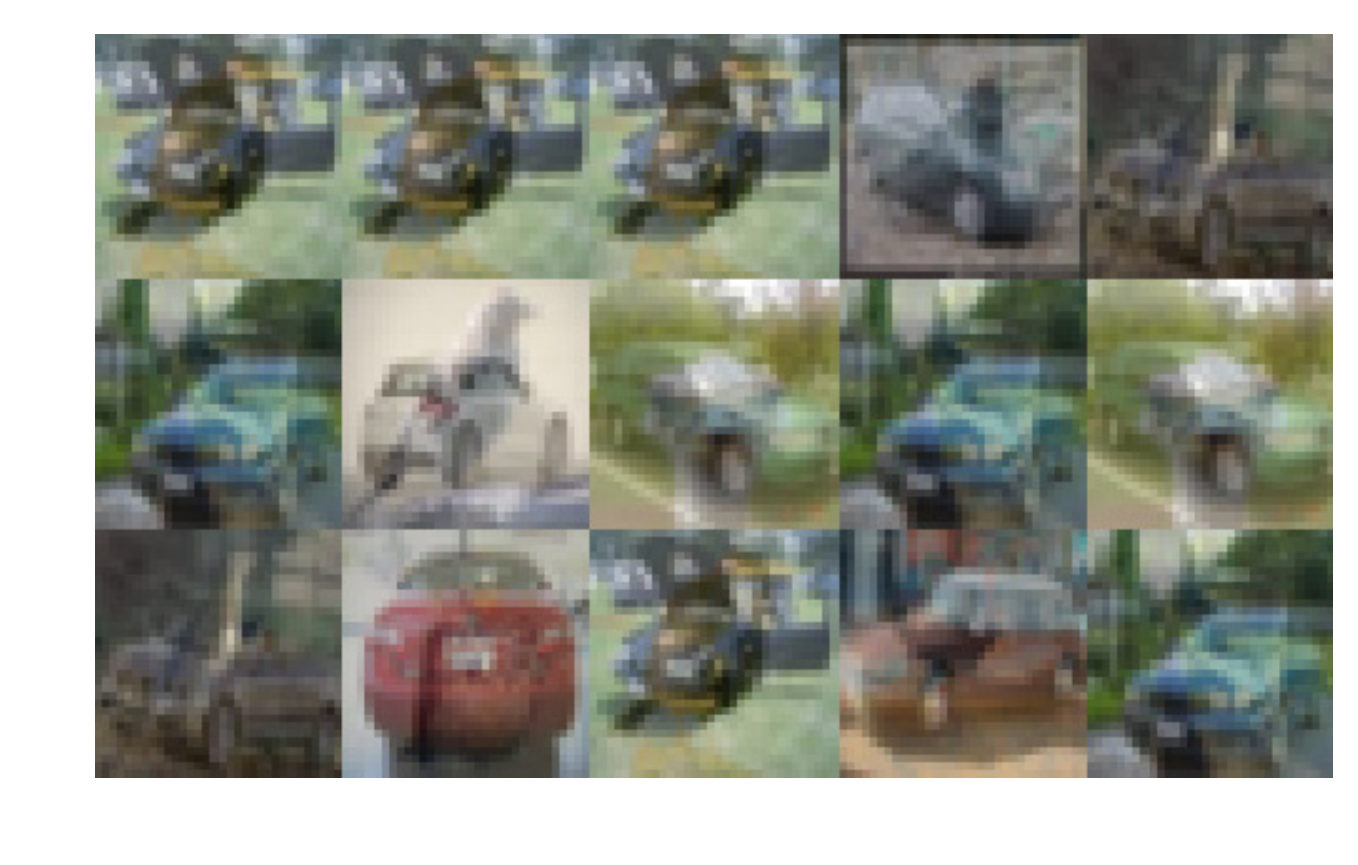}} 
  & 12.06 
  & $4.32 \times 10^{-5}$ 
  & \makecell{\includegraphics[scale=0.12]{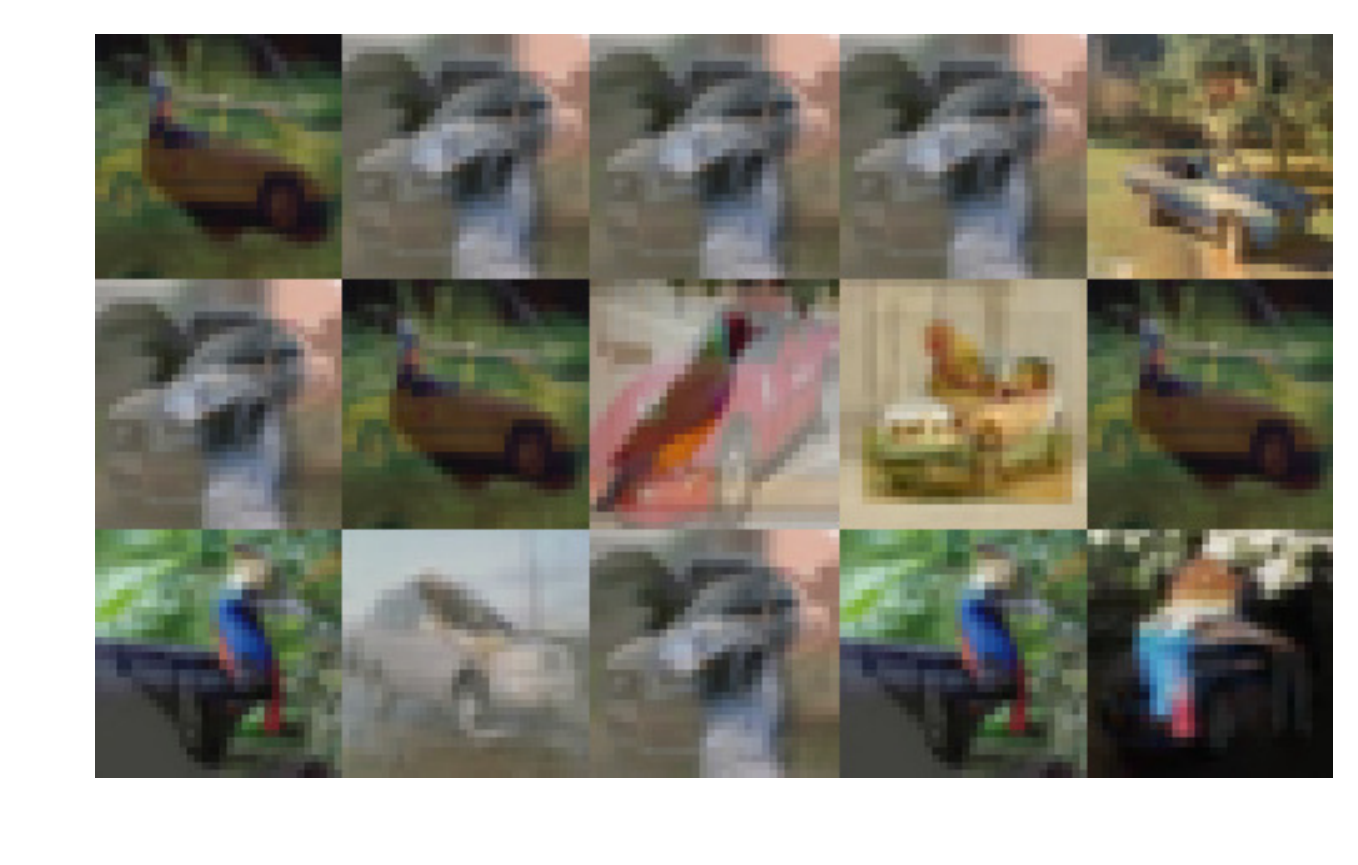}}  
  & 13.96\\ \hline
  
 EPBS 
 & $4.40 \times 10^{-5}$ 
 & \makecell{\includegraphics[scale=0.12]{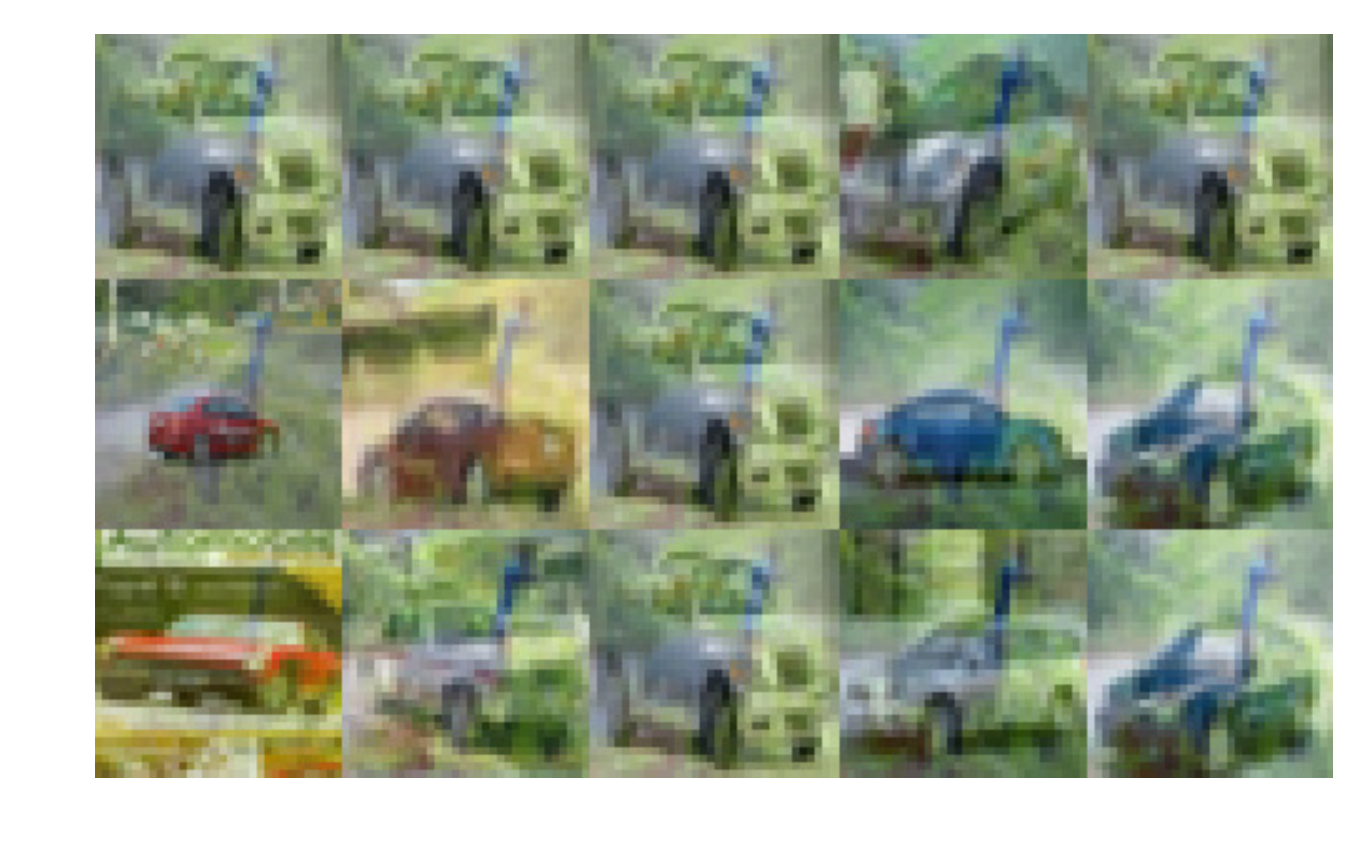}} 
 & 13.62 
 & $4.03 \times 10^{-4}$ 
 & \makecell{\includegraphics[scale=0.12]{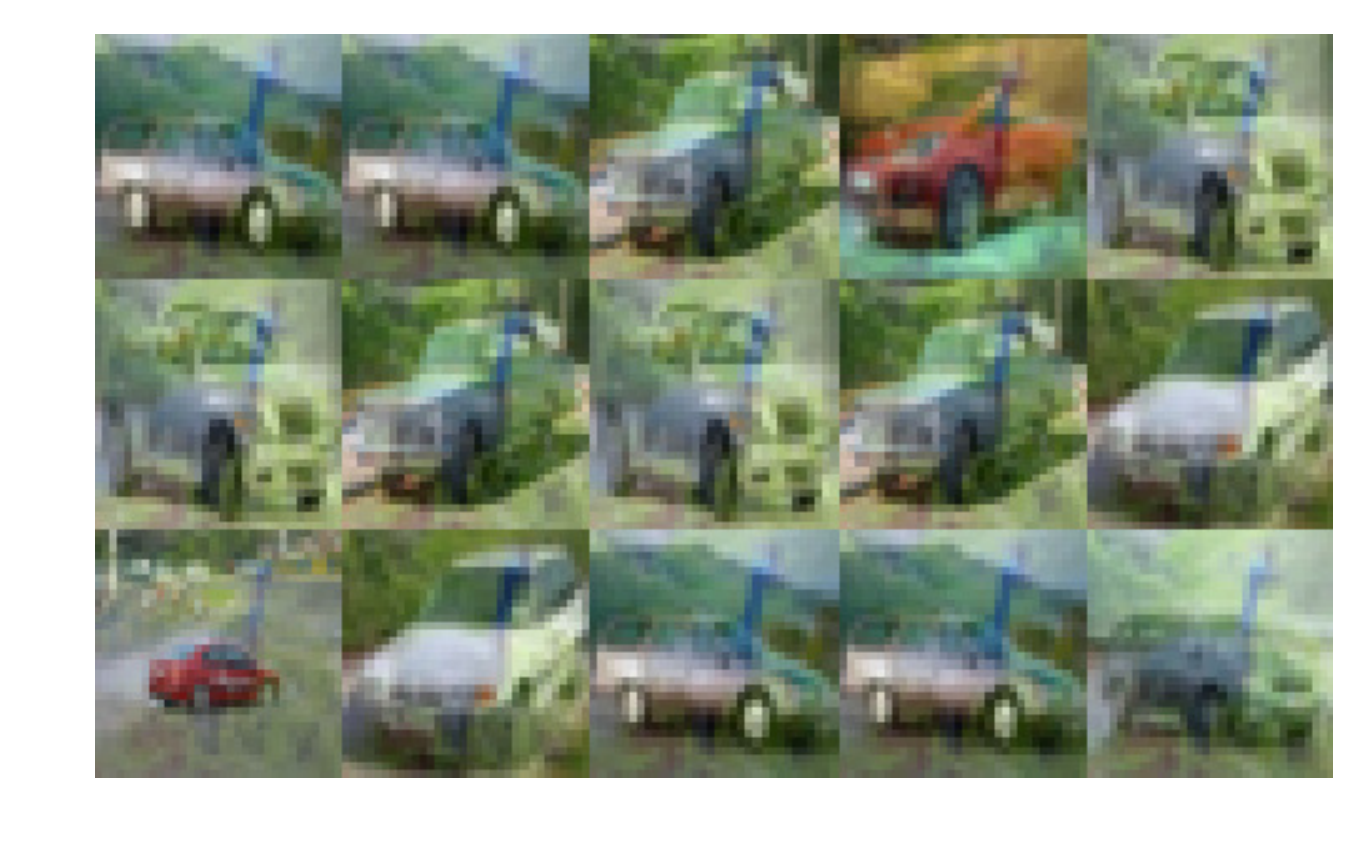}} 
 & 38.47 \\ \hline

    \end{tabular}
    \label{tab:baselinecifar10}
\end{table*}{}
\begin{enumerate}

    \item As for the first factor, we measure the average absolute difference in a DNN's prediction probabilities of classes $s$ and $t$ for borderline samples. This factor has been shown as $\overline{|f_s(x) - f_t(x)|}$ in Tables~\ref{tab:baselinemnsit}, \ref{tab:baselinefashionmnist}, and \ref{tab:baselinecifar10}. This factor is in line with the definition of the decision boundary (refer to Section~\ref{sec:problem}) and to ensure the criterion (a) discussed in Section~\ref{sec:framework}.  We can observe that all methods including DeepDIG succeed in discovering borderline instances whose difference in prediction probabilities for classes $s$ and $t$  is significantly small on average. In other words, a DNN is `confused' to categorically classify generated borderline instances.  
    
    \item  For the second factor, we inspect the quality of the generated borderline samples. We expect a good borderline generation method to generate borderline instances that are visibly similar to real samples. This is in line with the criterion (b) explained in Section~\ref{sec:framework}. Note that unlike baselines methods, DeepDIG approaches the decision boundary between two classes $s$ and $t$ in a two-way fashion i.e., once from $s$ to $t$ and once from $t$ to $s$ as described at the end of Section~\ref{sec:framework}; accordingly, we have included two sets of visualized images. As can be observed in Tables~\ref{tab:baselinemnsit}, \ref{tab:baselinefashionmnist}, and \ref{tab:baselinecifar10}, DeepDIG can generate borderline instances that are indistinguishable from real samples specially for MNIST and FashionMNIST datasets. However, RPBS and EPBS fail to generate similar-to-real samples. A closer look reveals that RPBS and EPBS generate a `sloppy' combination of the samples of classes $s$ and $t$ e.g., for FashionMNIST  the generated borderline instances contain both  \emph{trouser} and a \emph{pullover}. 
   
    \item Finally, for the last factor, we expect a method to generate borderline instances with a high \emph{success rate}. To quantify this, we measure the ratio of the number of sample pairs fed to Algorithm~\ref{alg:middle} to the number of successfully generated borderline instances i.e., those returned in line~\ref{alg:success}.  Remember that the baseline methods RPBS and EPBS still utilize Algorithm~\ref{alg:middle} to generate borderline methods. As shown in Tables~\ref{tab:baselinemnsit}, \ref{tab:baselinefashionmnist}, and \ref{tab:baselinecifar10}, DeepDIG significantly outperforms the baseline methods respect to the success rate.    
\end{enumerate}{}
Based on the observations above, we can infer that DeepDIG is an effective method capable of generating borderline instances for different DNNs. Now let's pinpoint the reason for the success of DeepDIG in comparison to baseline approaches. To achieve a good classification performance, DNNs carve out connected decision regions for different classes~\cite{fawzi2018empirical} that are complicatedly intertwined with each other. With this in mind, to generate a borderline sample, RPBS and EPBS form a \emph{simple} trajectory between two samples in the two decision regions $r_s$ and $r_t$ and then perform the greedy binary search (i.e., Algorithm~\ref{alg:middle}). However, given the complexity of the decision regions and their intertwined nature, the binary search along this trajectory is likely to end up in a region other than $r_s$ and $r_t$ i.e., ``Fail" in line~\ref{alg:fail} of Algorithm~\ref{alg:middle}. In contrast, DeepDIG is equipped with two effective components (I) and (II) that make the end-points of its established trajectory very close to the decision boundary between decision regions $r_s$ and $r_t$. Hence, the binary search for DeepDIG is less prone to end up in a different region other than $r_s$ and $r_t$ and thus higher success rate for DeepDIG is ensued.


\subsection{Inter-model Decision Boundary Characterization}
\label{sec:charac-dnns}

\begin{figure*}[!htb]
    \centering
    \begin{subfigure}[t]{0.5\textwidth}
        \centering
        \includegraphics[scale=0.26]{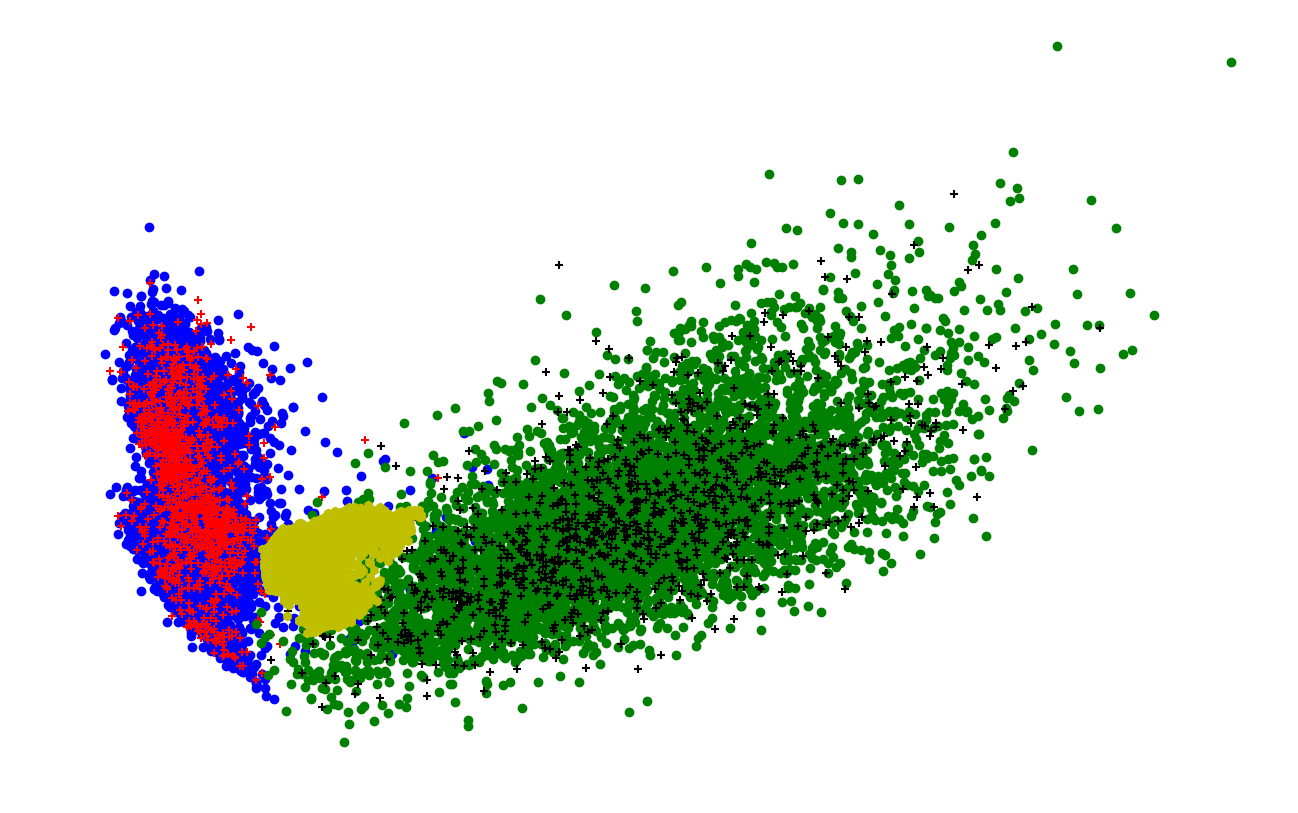}
        \caption{$\text{MNIST}_{\text{FCN}}$}
        \label{fig:pcamnistfcn}
    \end{subfigure}%
    ~ 
    \begin{subfigure}[t]{0.5\textwidth}
        \centering
        \includegraphics[scale=0.26]{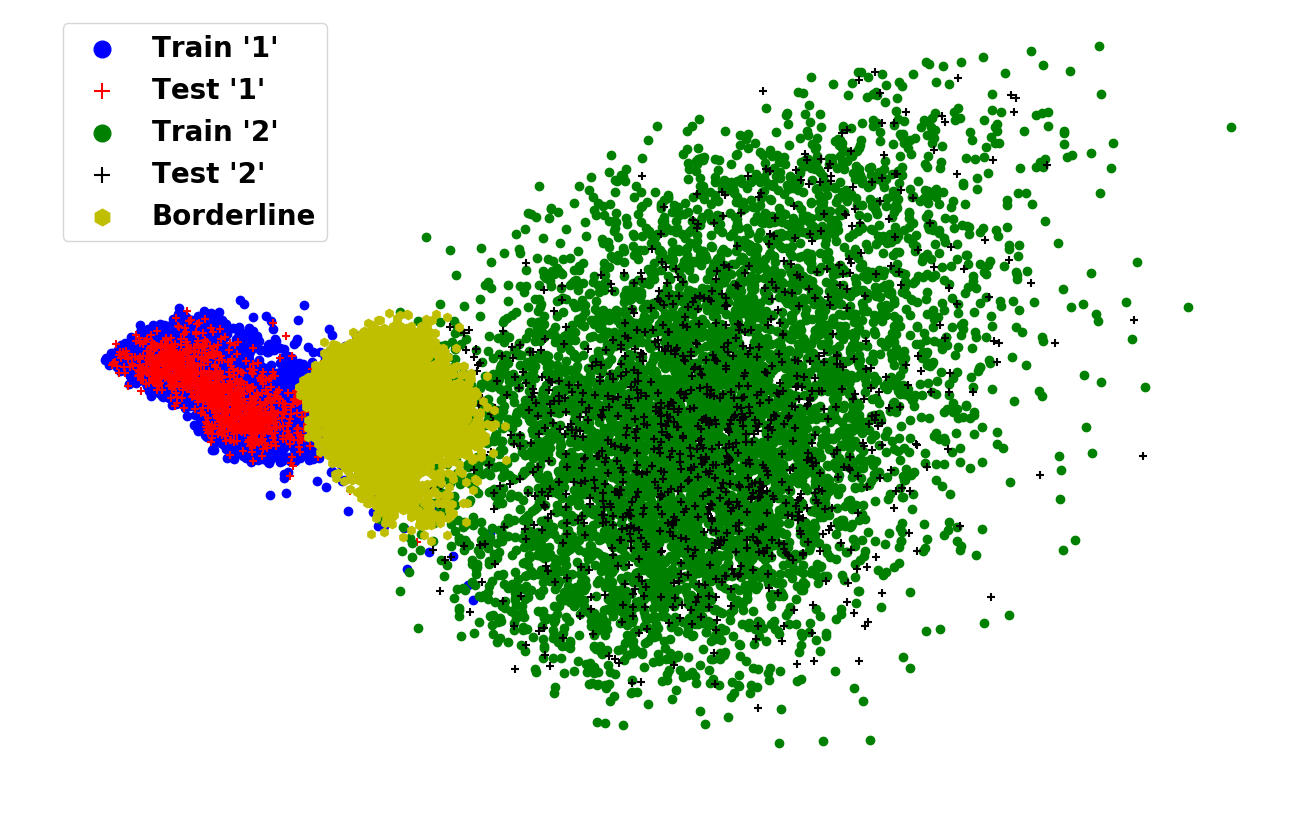}
        \caption{$\text{MNIST}_{\text{CNN}}$}
        \label{fig:pcamnistcnn}
    \end{subfigure}
    \caption{Projection of embeddings of training and test samples as well as borderline instances onto a 2D space (MNIST)  }
    \label{fig:pcamnist}
\end{figure*}

\begin{figure*}[!htb]
    \centering
    \begin{subfigure}[t]{0.54\textwidth}
        \centering
        \includegraphics[scale=0.26]{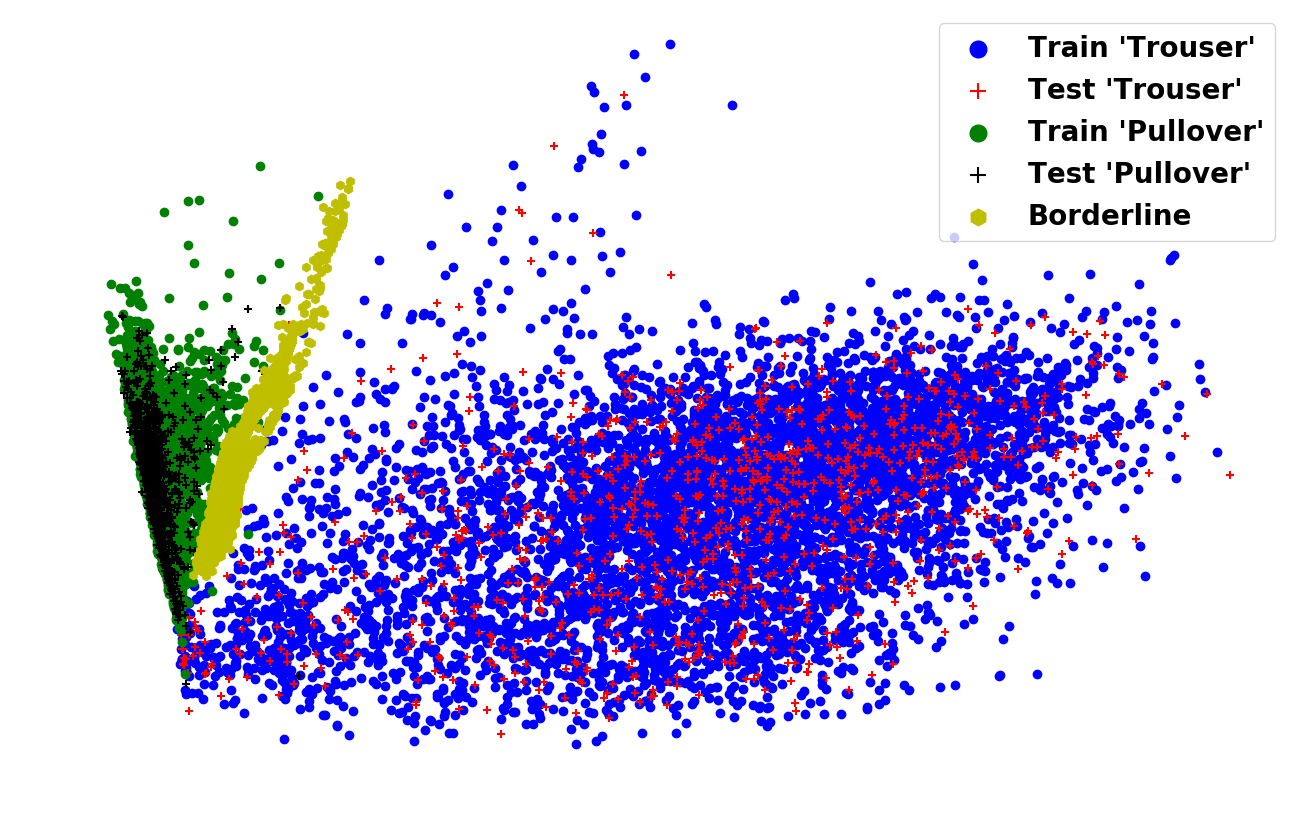}
        \caption{$\text{FashionMNIST}_{\text{FCN}}$}
        \label{fig:pcafashionmnistfcn}
    \end{subfigure}%
    ~ 
    \begin{subfigure}[t]{0.46\textwidth}
        \centering
        \includegraphics[scale=0.26]{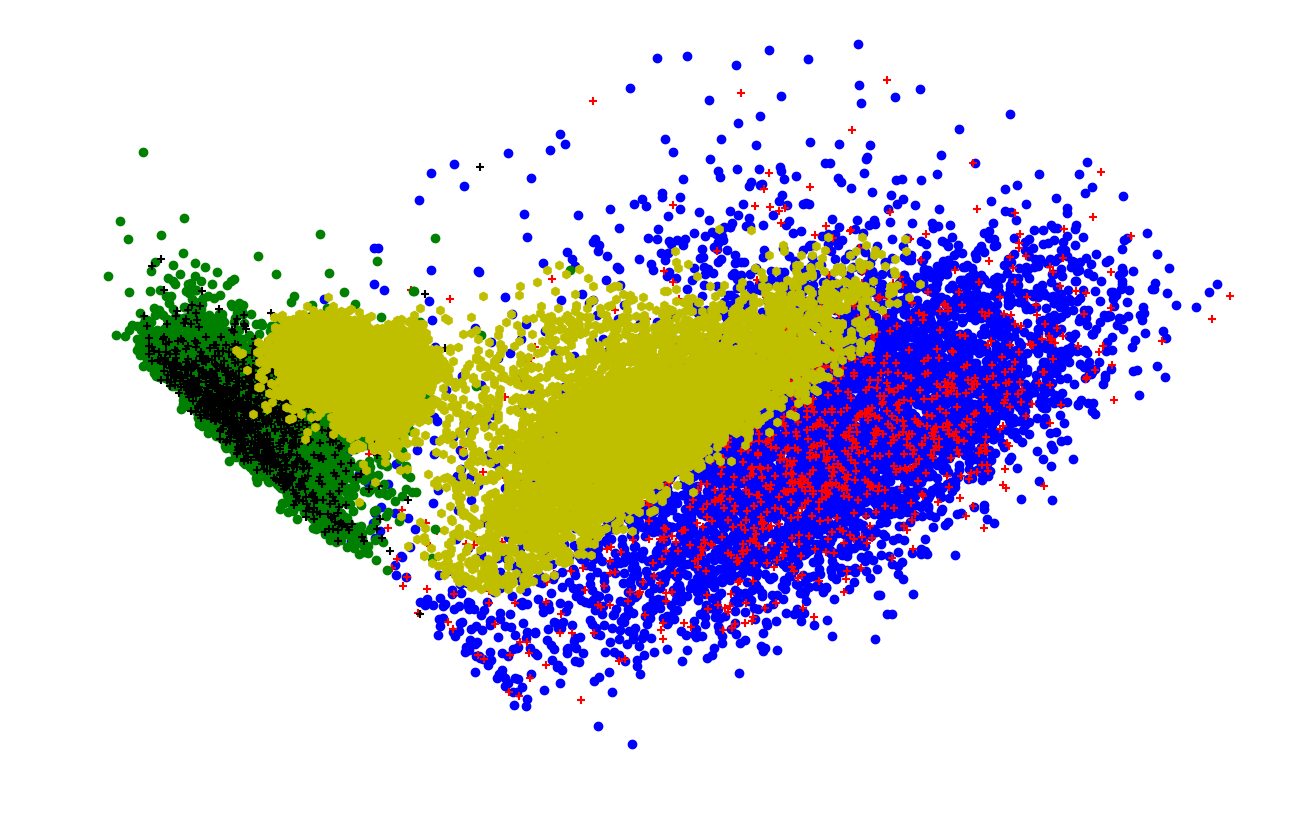}
        \caption{$\text{FashionMNIST}_{\text{CNN}}$}
        \label{fig:pcafashionmnistcnn}
    \end{subfigure}
    \caption{Projection of embeddings of training and test samples as well as borderline instances  onto a 2D space (FashionMNIST)  }
    \label{fig:pcafashionmnist}
\end{figure*}

\begin{figure*}[!htb]
    \centering
    \begin{subfigure}[t]{0.53\textwidth}
        \centering
        \includegraphics[scale=0.26]{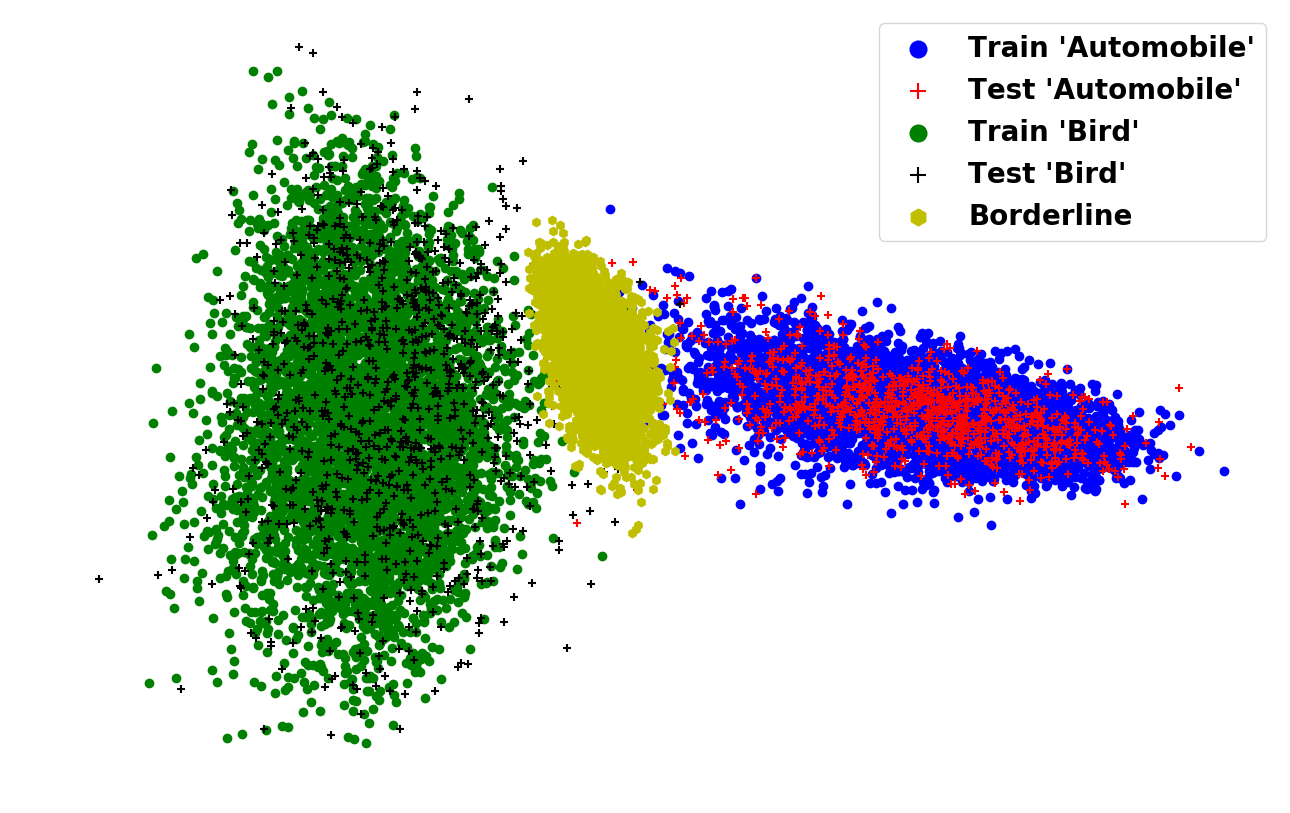}
        \caption{$\text{CIFAR10}_{\text{ResNet}}$}
        \label{fig:pcacifar10resnet}
    \end{subfigure}%
    ~ 
    \begin{subfigure}[t]{0.47\textwidth}
        \centering
        \includegraphics[scale=0.26]{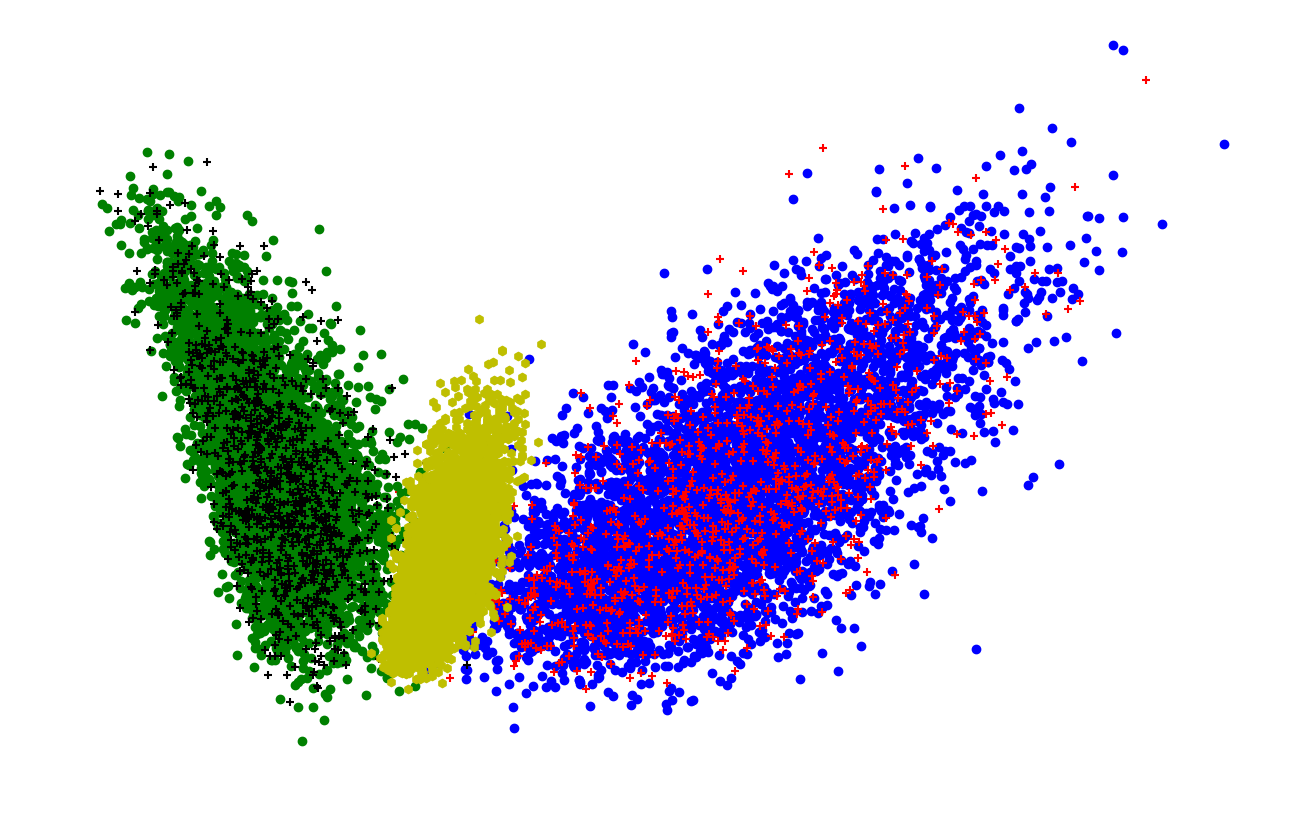}
        \caption{$\text{CIFAR10}_{\text{GoogleNet}}$}
        \label{fig:pcacifar10googlenet}
    \end{subfigure}
    \caption{Projection of embeddings of training and test samples as well as borderline instances onto a 2D space (CIFAR10)  }
    \label{fig:pcacifar10}
\end{figure*}

Thus far, we have investigated the components of DeepDIG and made sure of their working. We also compared DeepDIG with baseline methods and showed that it outperforms them. Now it is time to utilize DeepDIG to characterize the decision boundary of DNNs. In Section~\ref{sec:characteristics}, we developed several measures to characterize the decision boundary between the two classes. We utilize borderline instances generated by DeepDIG and compute those measures for the pre-trained models in Table~\ref{tab:dnns}. The results are shown in Table~\ref{tab:inter-model}. To further illustrate how borderline samples are spatially located in the embedding space learned by a DNN, we visualize them along with training and test samples. Figures~\ref{fig:pcamnist}, \ref{fig:pcafashionmnist}, and \ref{fig:pcacifar10} show the visualizations for DNNs trained on MNIST, FashionMNIST, and CIFAR10, respectively. To generate these figures, we project the embeddings learned by a DNN to a 2D space using PCA (Principal Component Analysis)~\cite{shlens2014tutorial}. We fit the PCA on the embeddings of the standard train samples and use it in the inference mode to project the embedding of test instances as well as borderline instances\footnote{For PCA we use scikit-learn package with \texttt{n\_components=2} and the other parameter settings as defaults.}. Note that as explained before, for the decision boundary between two classes $s$ and $t$, DeepDIG generates two sets of borderline instances, namely ones that are approached from $s$ and consequently their labels are $s$ and ones that are approached from class $t$ and their labels are $t$ -- See Tables~\ref{tab:baselinemnsit}, \ref{tab:baselinefashionmnist}, and \ref{tab:baselinecifar10} for some visualizations of these two sets. However, in projections presented in Figures~\ref{fig:pcamnist}, \ref{fig:pcafashionmnist}, and \ref{fig:pcacifar10}, we show both sets of borderline instances as `borderline' to signify their near decision boundary property rather than their labels. Based on the results presented in Table~\ref{tab:inter-model} as well as  Figures~\ref{fig:pcamnist}, \ref{fig:pcafashionmnist}, and \ref{fig:pcacifar10}, we make the following observations regarding the characteristics of decision boundaries of investigated DNNs. 

\begin{table}[!htb]\small
\setlength\tabcolsep{0.9pt} 
    \renewcommand{\arraystretch}{1.2} 
\caption{Results of inter-model decision boundary characterization}
    \centering
       
    \begin{tabu}{l|c|c|c|c|c}
    \hline
     & IDC & $\text{EDC1}_{\text{Test}}$ & $\text{EDC1}_{\text{Borderline}}$ & $\text{EDC2}_{\text{Test}}$ & $\text{EDC2}_{\text{Borderline}}$  \\ \hline
    \hline
    $\text{MNIST}_{\text{CNN}}$ & 0.037& 0.94 & 0.25 & 99.18 & 39.73 \\ \hline 
    $\text{MNIST}_{\text{FCN}}$ & 0.018& 0.92 & 0.29 & 99.72 & 58.36 \\ \tabucline[1.5pt]{---}
    $\text{FashionMNIST}_{\text{CNN}}$ & 0.035 & 0.77 & 0.18 & 99.40 & 37.09 \\ \hline 
    $\text{FashionMNIST}_{\text{FCN}}$ & 0.017 & 0.91 & 0.26 & 99.01 & 83.34 \\ \tabucline[1.5pt]{---}
    $\text{CIFAR10}_{\text{ResNet}}$ & 0.035 & 0.69 & 0.12 & 99.45 & 52.75 \\ \hline 
    $\text{CIFAR10}_{\text{GoogleNet}}$ & 0.038 & 0.55 & 0.06 & 99.65 & 38.51 
    \end{tabu}
    \label{tab:inter-model}
\end{table}{}

\begin{itemize}
    \item In Section~\ref{sec:linear}, we defined measure EDC1 to determine whether generated borderline instances are near the decision boundary in the embedding space or not.  We can observe from Table~\ref{tab:inter-model} that borderline samples --compared to unseen test samples-- are very close to the separating hyperplane i.e., $\text{EDC1}_{\text{Borderline}}$ is significantly smaller than $\text{EDC1}_{\text{Test}}$ for all DNNs.  Note that EDC1 is normalized to be in the range $[0,1]$. This proves our hypothesis that borderline instances are indeed in the separating region between two classes in the embedding space and thus are near the decision boundary in the embedding space. Visualizations in Figures~\ref{fig:pcamnist}, \ref{fig:pcafashionmnist}, and \ref{fig:pcacifar10} further corroborate this hypothesis where we can easily observe that borderline samples are between the original samples of two classes. In particular, borderline samples occupy a different region of the embedding space than that of original samples (train and test sets). 
    \item IDC measure is developed to inform us about the complexity of the decision boundary in the input space while EDC2's purpose is the same except in the embedding space. We can observe that these two are not disjoint and there is a strong correlation between these two complexity measures. More specifically, the more complex the decision boundary in the input space is (i.e., a larger value for IDC), the more complex the decision boundary in the embedding space is (i.e., a smaller value for $\text{EDC2}_{\text{Borderline}}$) and vice versa. 
    \item We can observe that a linear model can obtain a perfect accuracy score on the test set ($\text{EDC2}_{\text{Test}}$ > 99\%). The reason is that test samples follow the same distribution with training data as they are surrounded by training data points --See Figures~\ref{fig:pcamnist}, \ref{fig:pcafashionmnist}, and \ref{fig:pcacifar10}. Borderline samples, in contrast, have a considerably smaller accuracy which is due to again their different distribution than original data. Hence, we can conclude that \emph{the linear separability capability  of samples in the embedding space learned by a DNN holds as long as samples come from the same distribution with training data.} 
    \item  CNN architectures are sophisticated methodologies specifically designed to capture salient patterns in images while FCNs consist of simple multi-layer perceptrons. Based on our results, it seems that the capability of extracting complex patterns has caused creating more complex decision boundaries for CNNs compared to FCNs. This has been shown in Table~\ref{tab:inter-model} where FCN models have resulted in carving out less complicated decision boundaries than CNNs for MNIST and FashionMNIST datasets. This is particularly evident for $\text{FashionMNIST}_\text{CNN}$ in Figure~\ref{fig:pcafashionmnistcnn} wherein borderline instances are complicatedly intertwined with real samples.
    \item $\text{CIFAR10}_{\text{ResNet}}$ forms a less complicated decision boundary than $\text{CIFAR10}_{\text{GoogleNet}}$. We speculate this is due to the highly complex structure of  GoogleNet~\cite{szegedy2015going} and an excessive number of parameters of this model --See Table~\ref{tab:dnns}.    
\end{itemize}{}

Based on the above observations, we make the following conclusion. Although many factors can influence how a DNN establishes a decision boundary e.g., non-linear activation function, regularization, etc, thanks to DeepDIG and further proposed characteristics we can shed light on a DNN and its behavior in a systematic and principled manner. This is particularly useful for the model selection task where it can help us to complement other selection criteria including the common criterion used in this task i.e., the performance on a held-out test set. For instance, while CNN models for MNIST and FashionMNIST (i.e., $\text{MNIST}_\text{CNN}$ and $\text{FashionMNIST}_\text{CNN}$, respectively) achieve slightly better performance on the test set than FCN models (i.e., $\text{MNIST}_\text{FCN}$ and $\text{FashionMNIST}_\text{FCN}$, respectively) --See Table~\ref{tab:dnns}-- one might opt to use FCNs due to their simpler decision boundaries. Hence, we believe DeepDIG can help a practitioner/researcher to make a more informed decision regarding developing a deep model.

\subsection{Intra-model Decision Boundary Characterization}
\label{sec:intra-class}

\begin{table*}[!htb]
\setlength\tabcolsep{0.8pt} 
    \renewcommand{\arraystretch}{1.2} 
    \centering
    \caption{Illustration of the generated borderline samples for all pair-wise classes in  $\text{MNIST}_{\text{CNN}}$ }
    \begin{tabular}{|c||c|c|c|c|c|c|c|c|c|c|}
    \hline
       \diagbox{s}{t}  & `0' & `1' & `2' & `3' & `4' & `5' & `6' & `7' & `8' & `9'  \\ \hline \hline

`0'&--&\makecell{\includegraphics[scale=0.12]{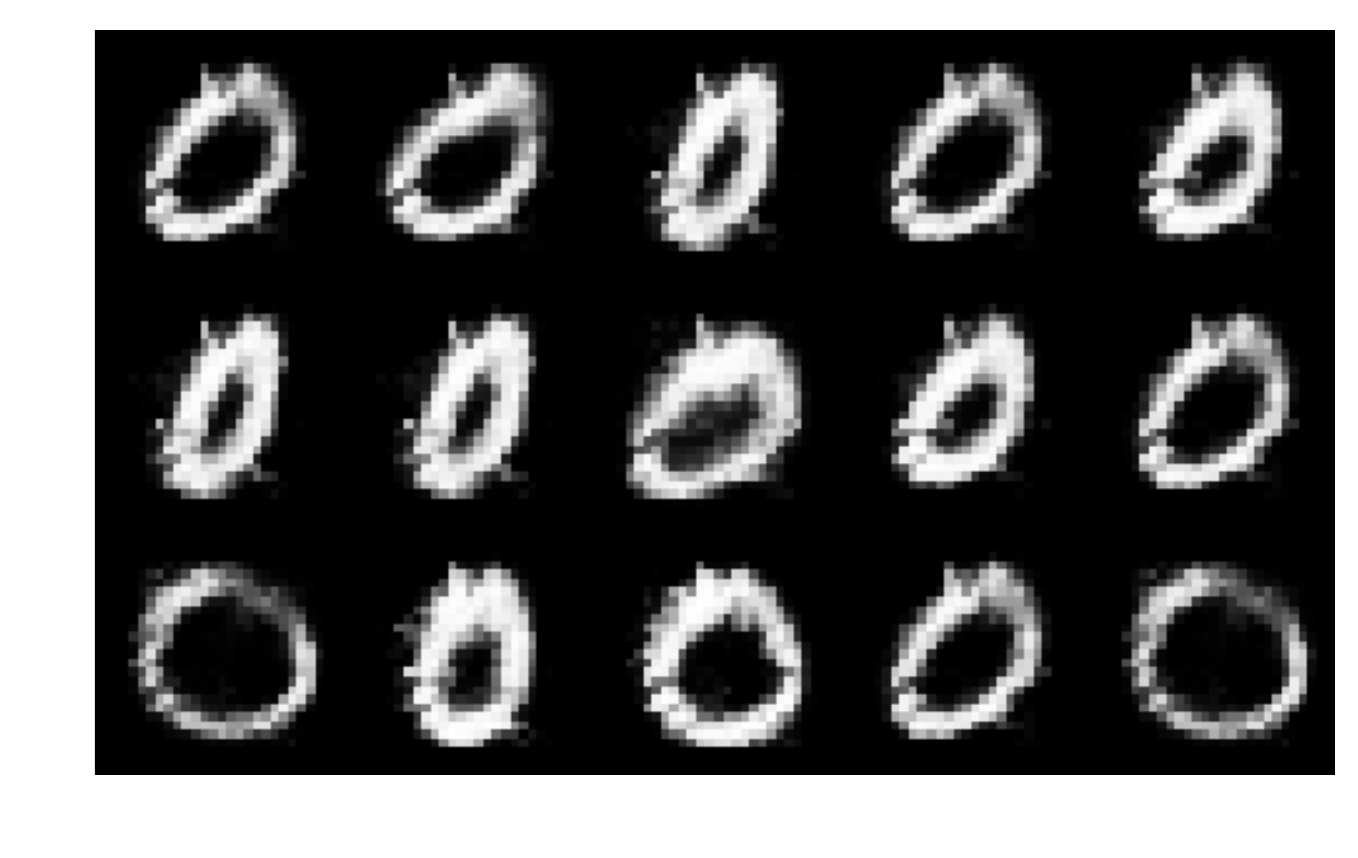}}&\makecell{\includegraphics[scale=0.12]{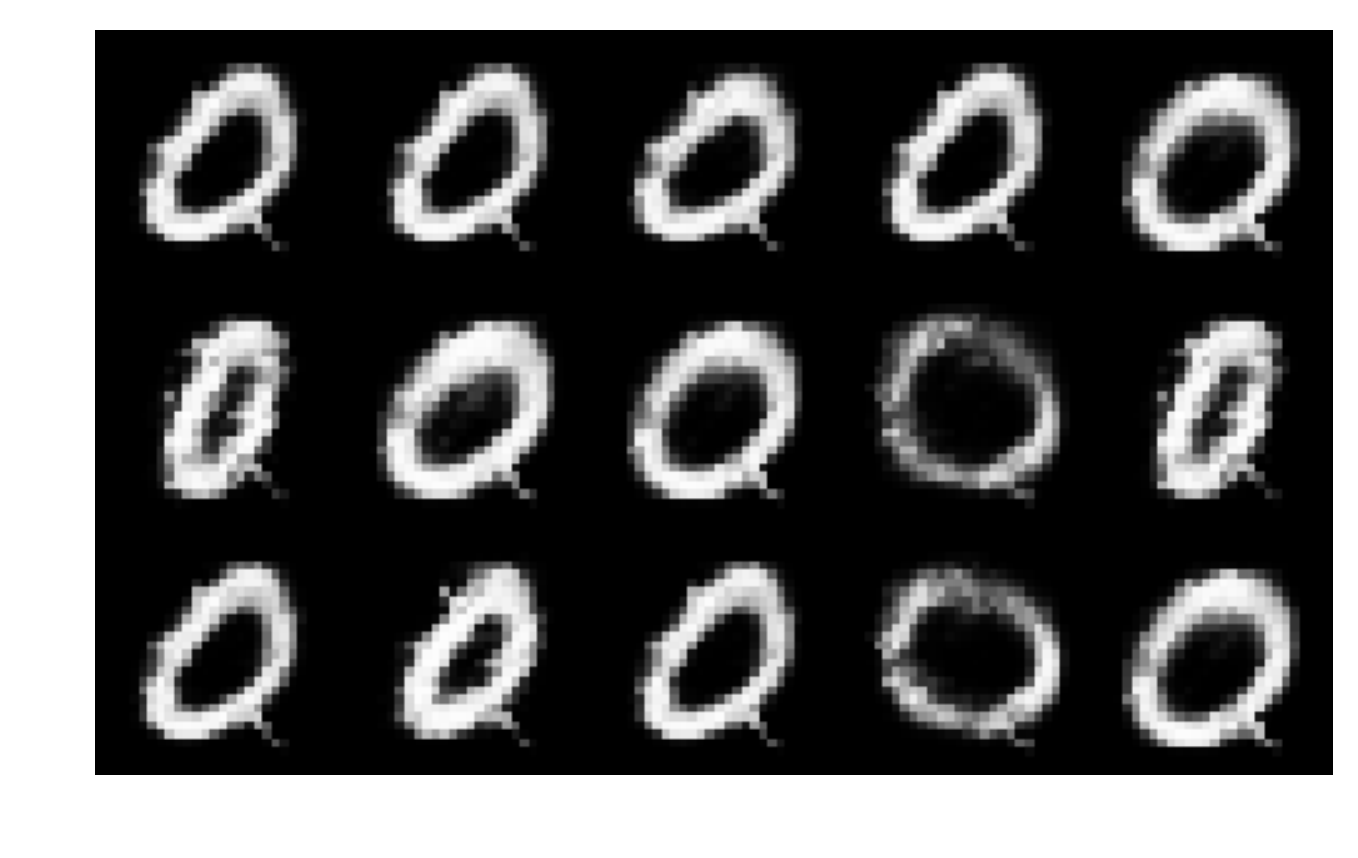}}&\makecell{\includegraphics[scale=0.12]{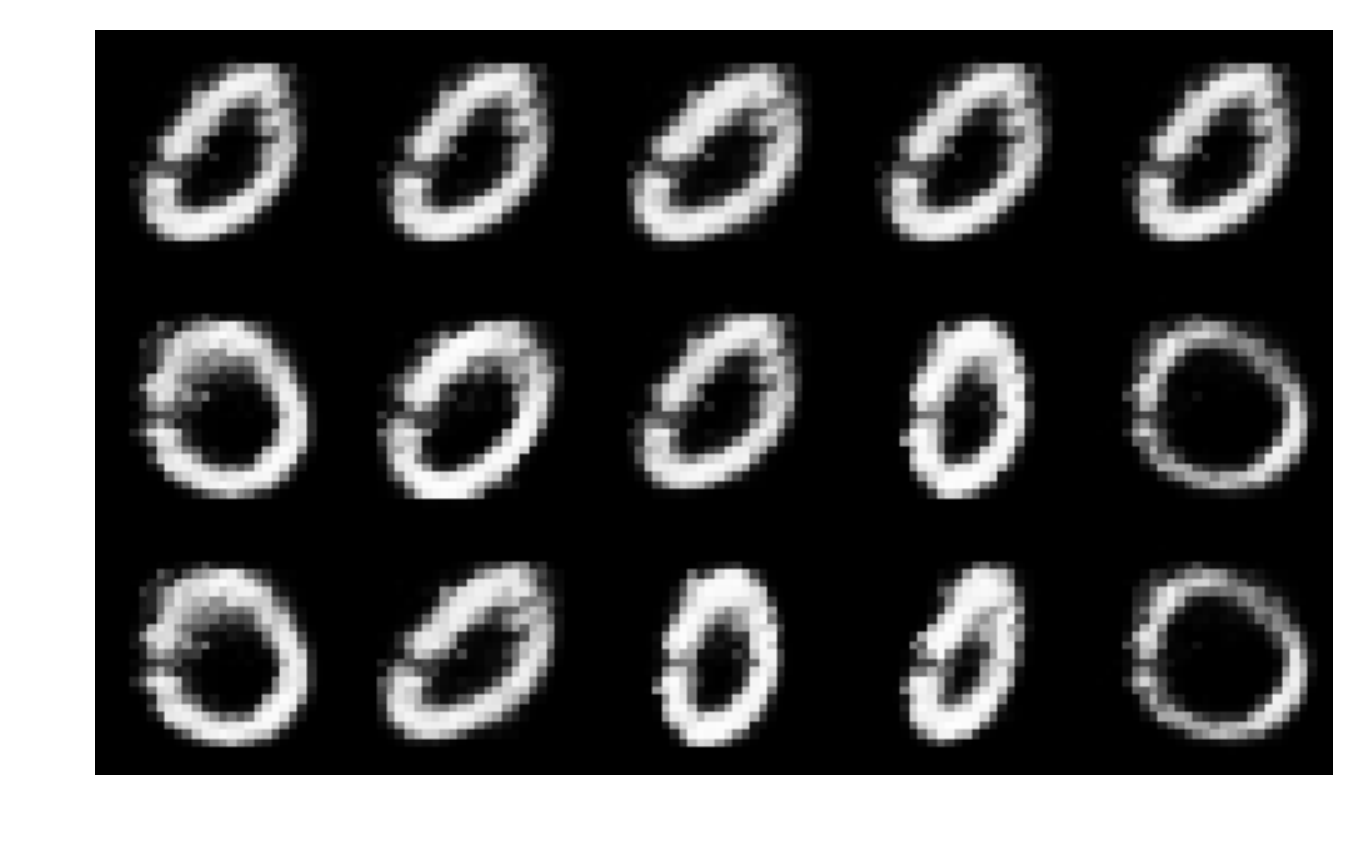}}&\makecell{\includegraphics[scale=0.12]{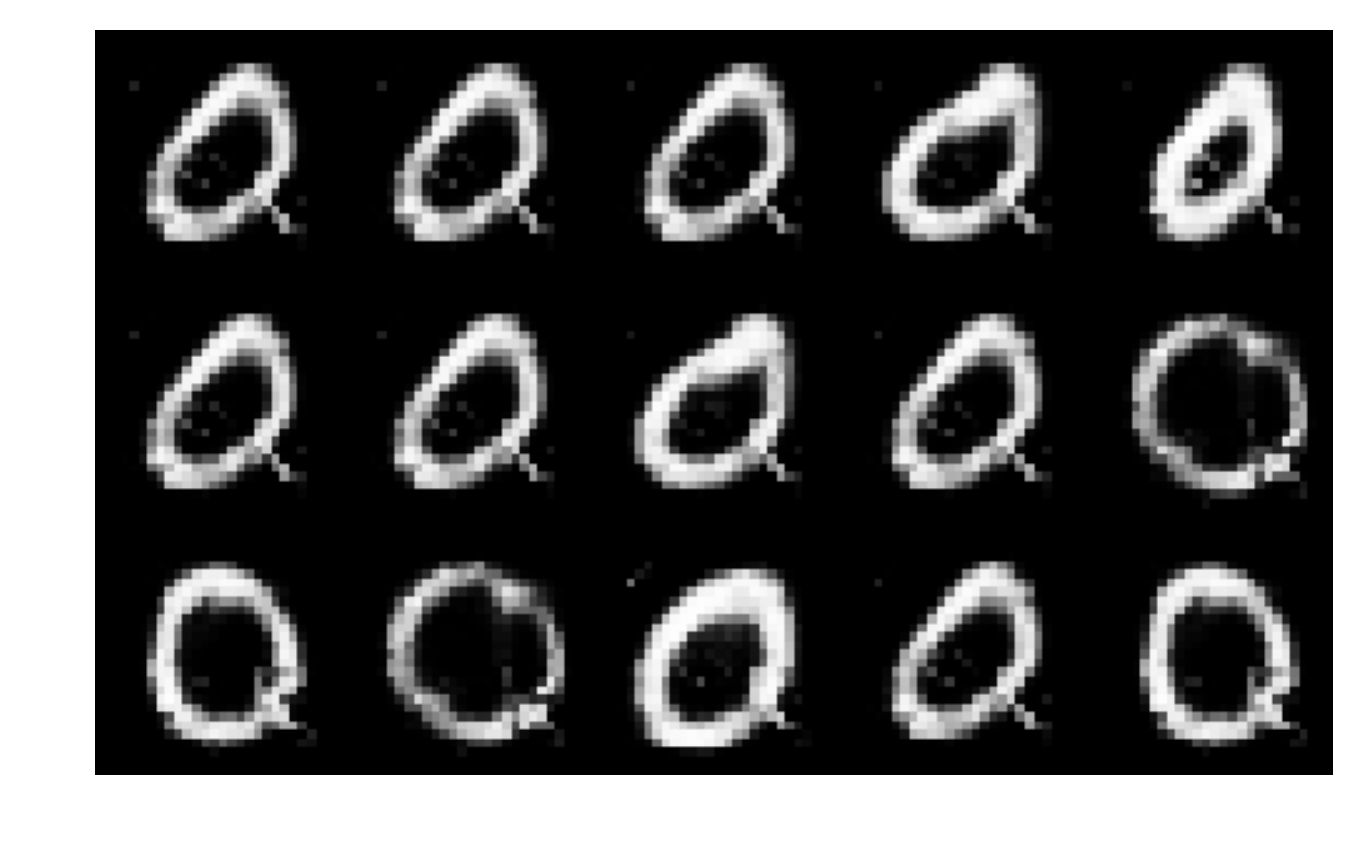}}&\makecell{\includegraphics[scale=0.12]{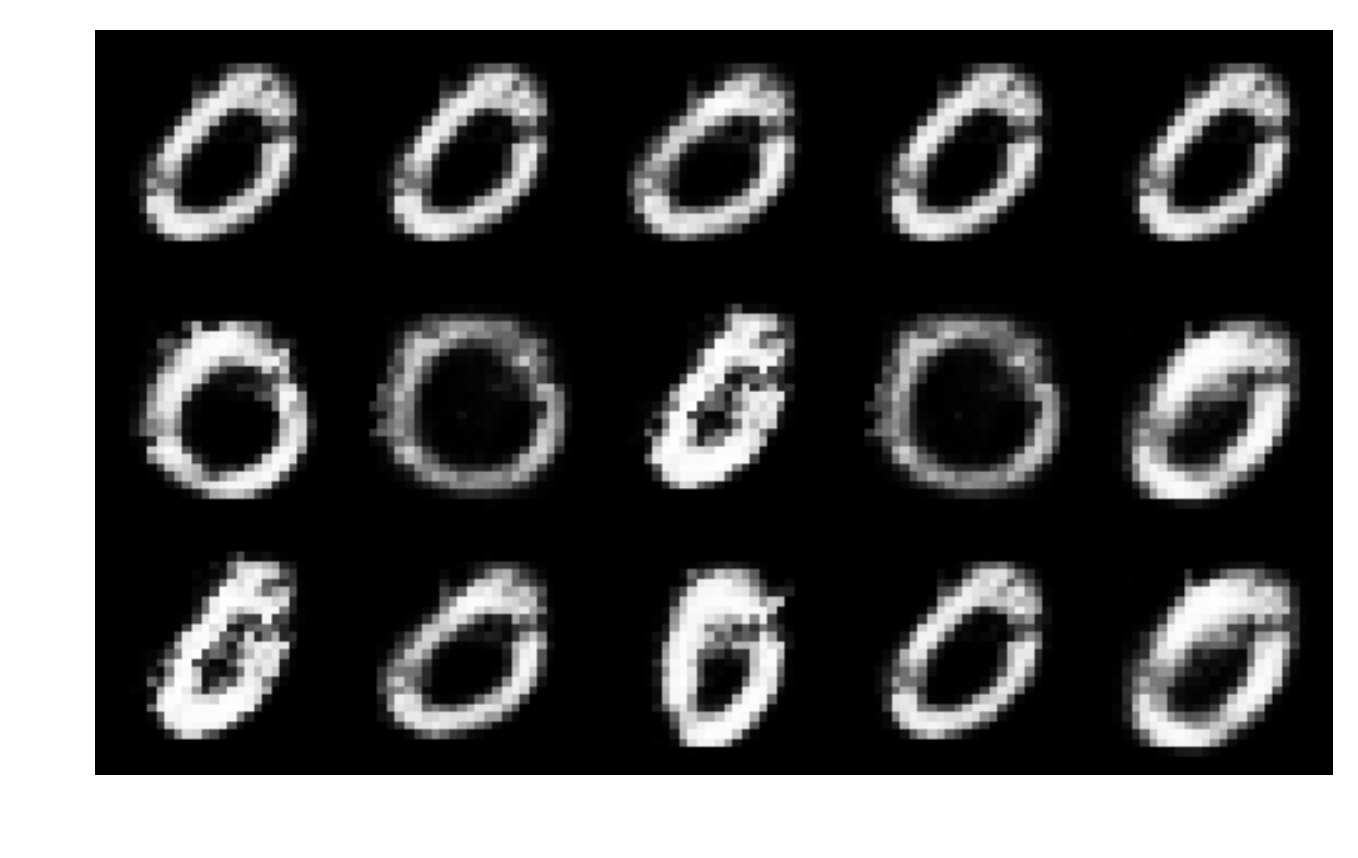}}&\makecell{\includegraphics[scale=0.12]{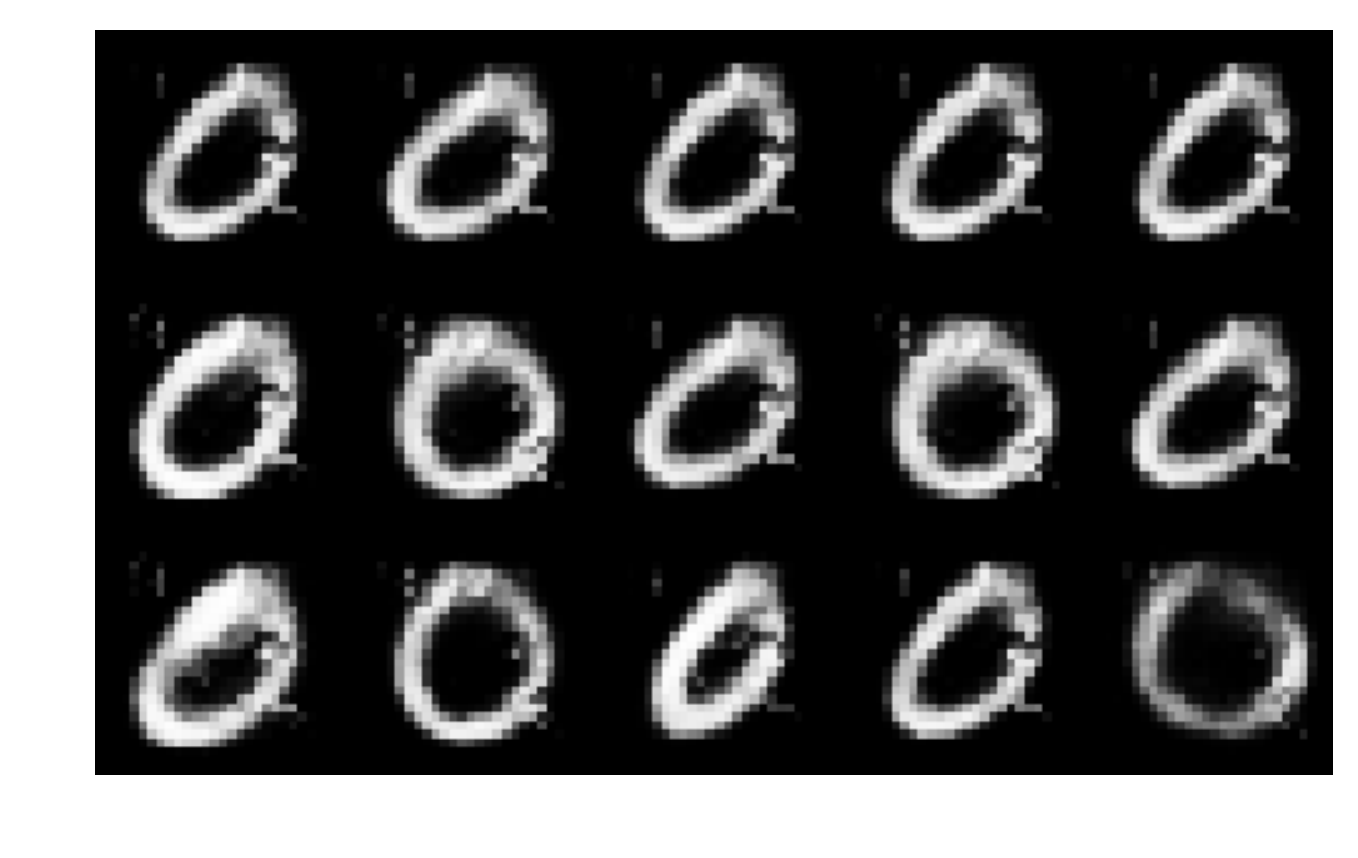}}&\makecell{\includegraphics[scale=0.12]{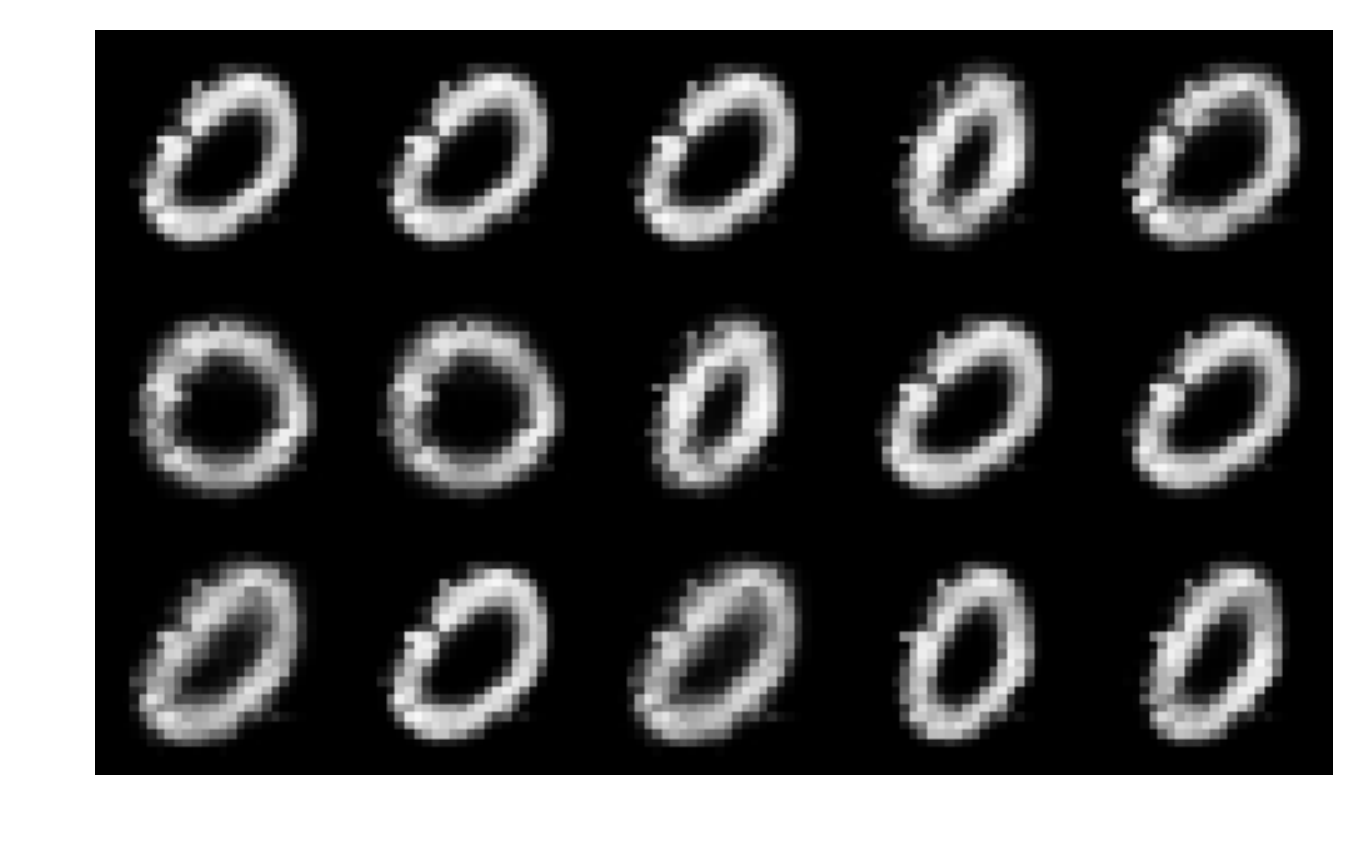}}&\makecell{\includegraphics[scale=0.12]{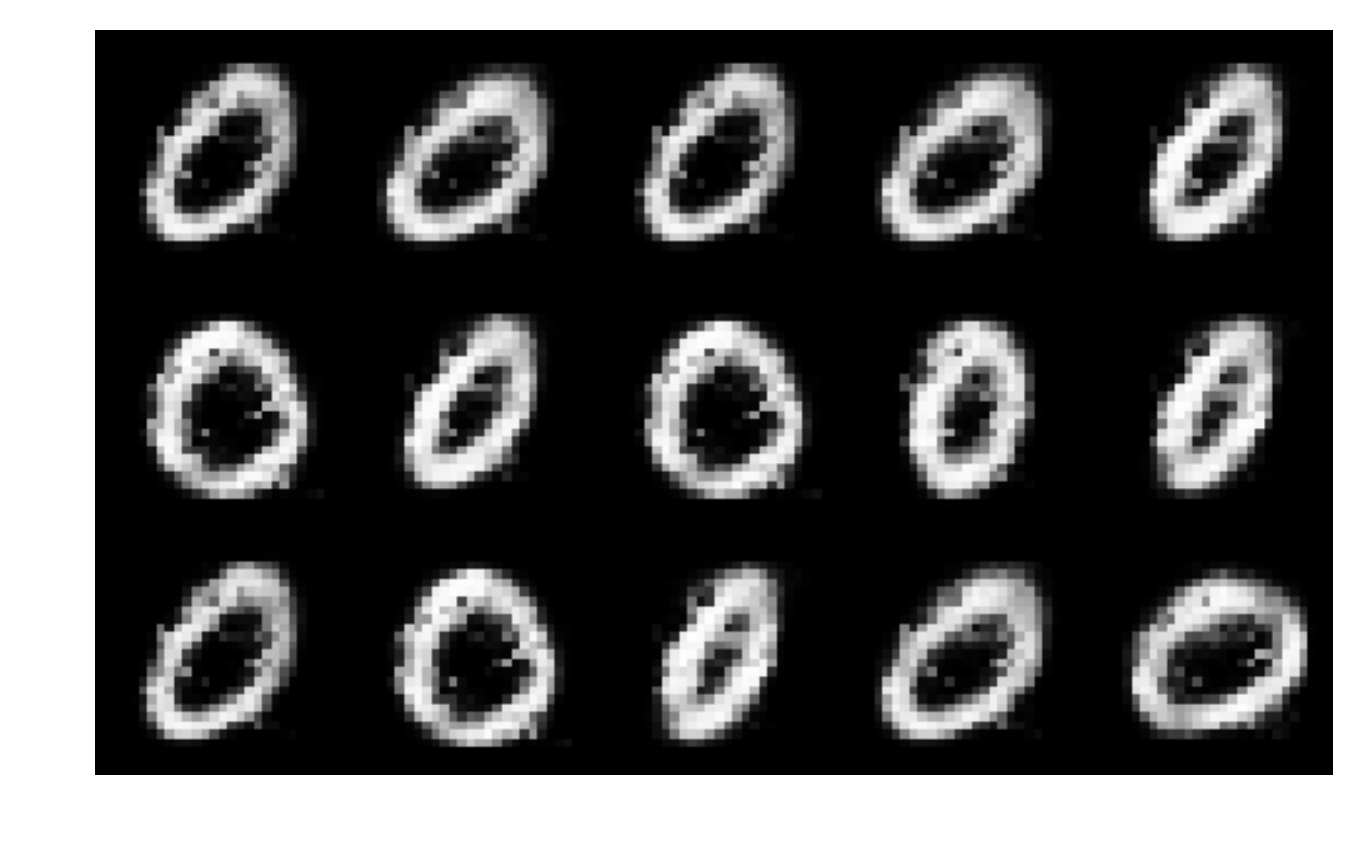}}&\makecell{\includegraphics[scale=0.12]{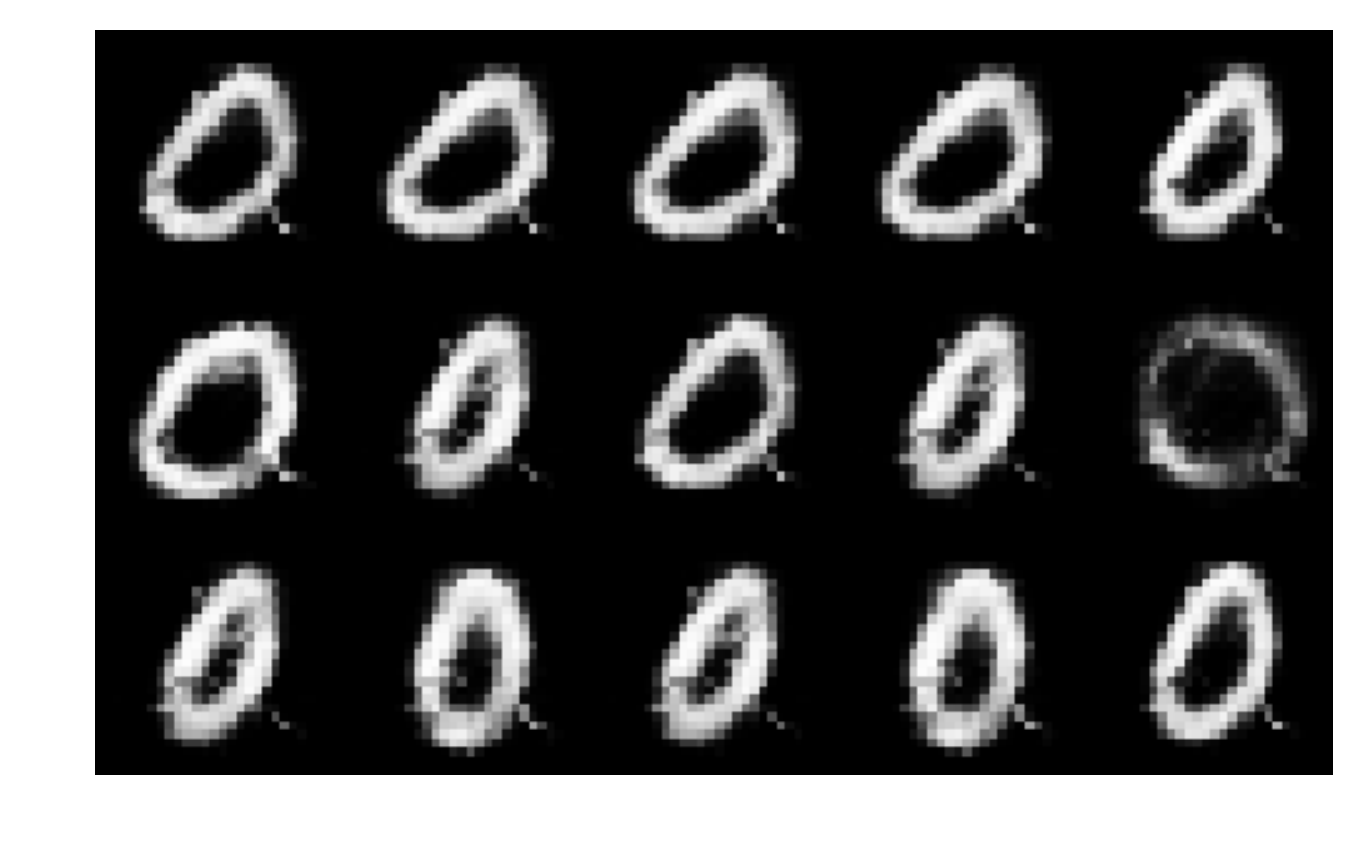}}\\ \hline
`1'&\makecell{\includegraphics[scale=0.12]{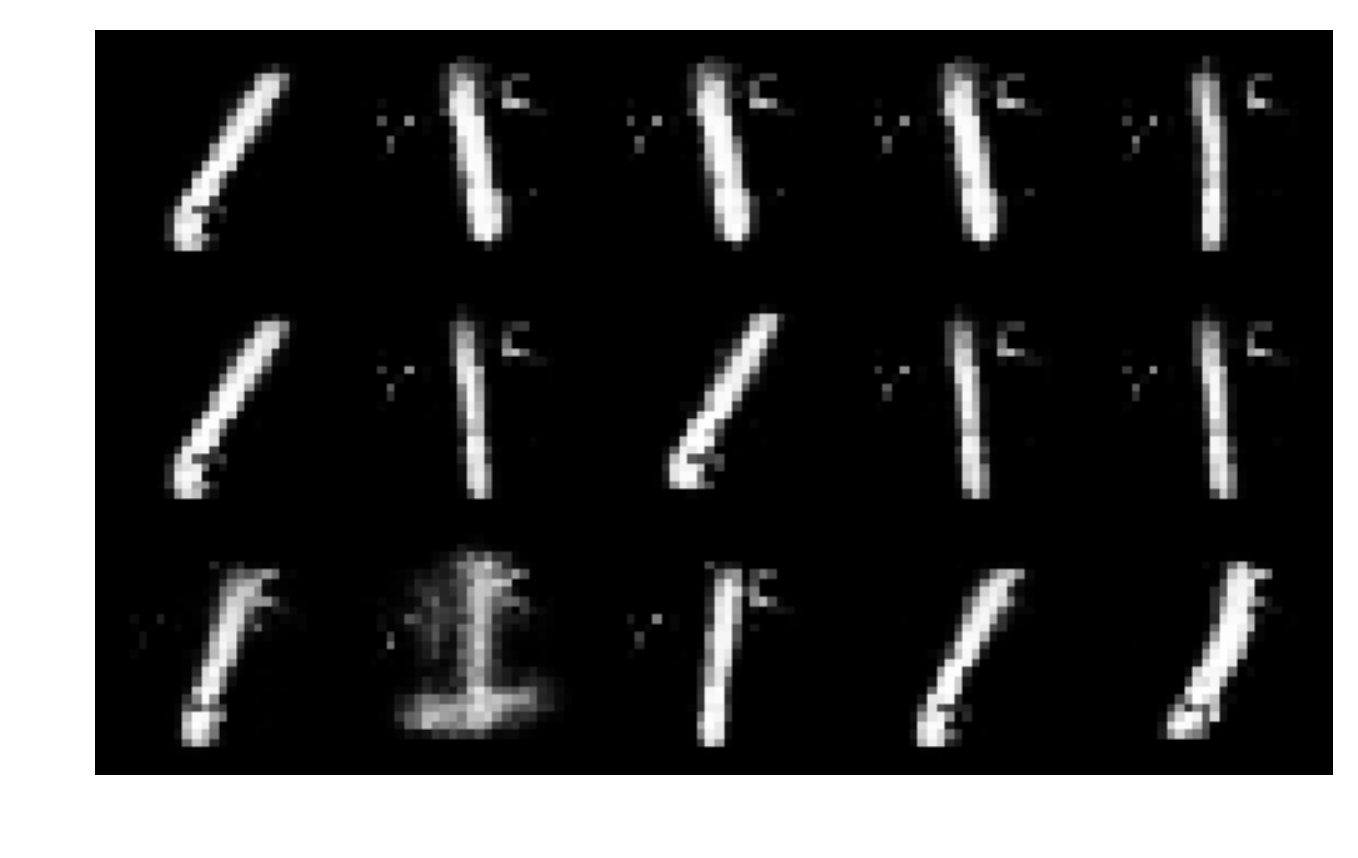}}&--&\makecell{\includegraphics[scale=0.12]{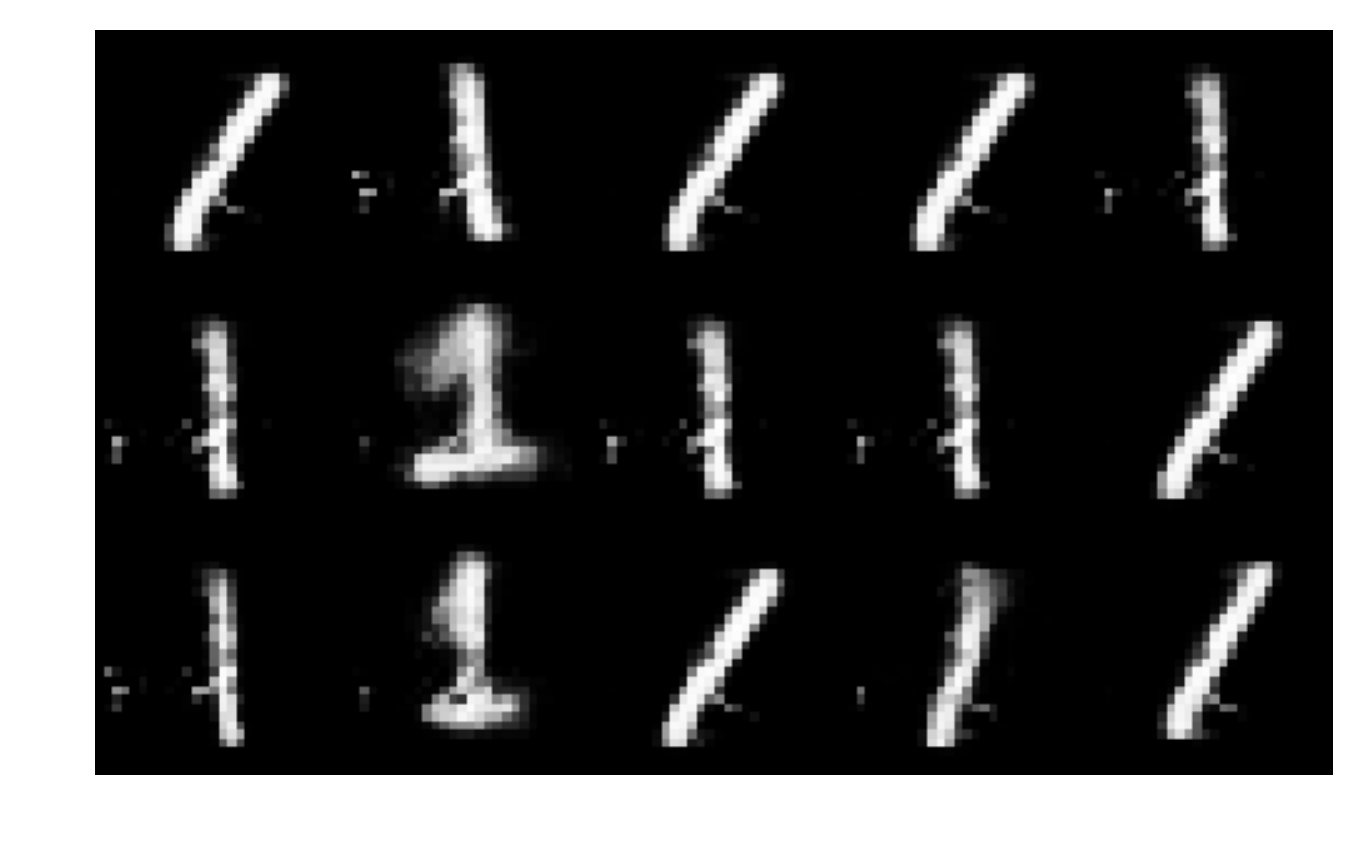}}&\makecell{\includegraphics[scale=0.12]{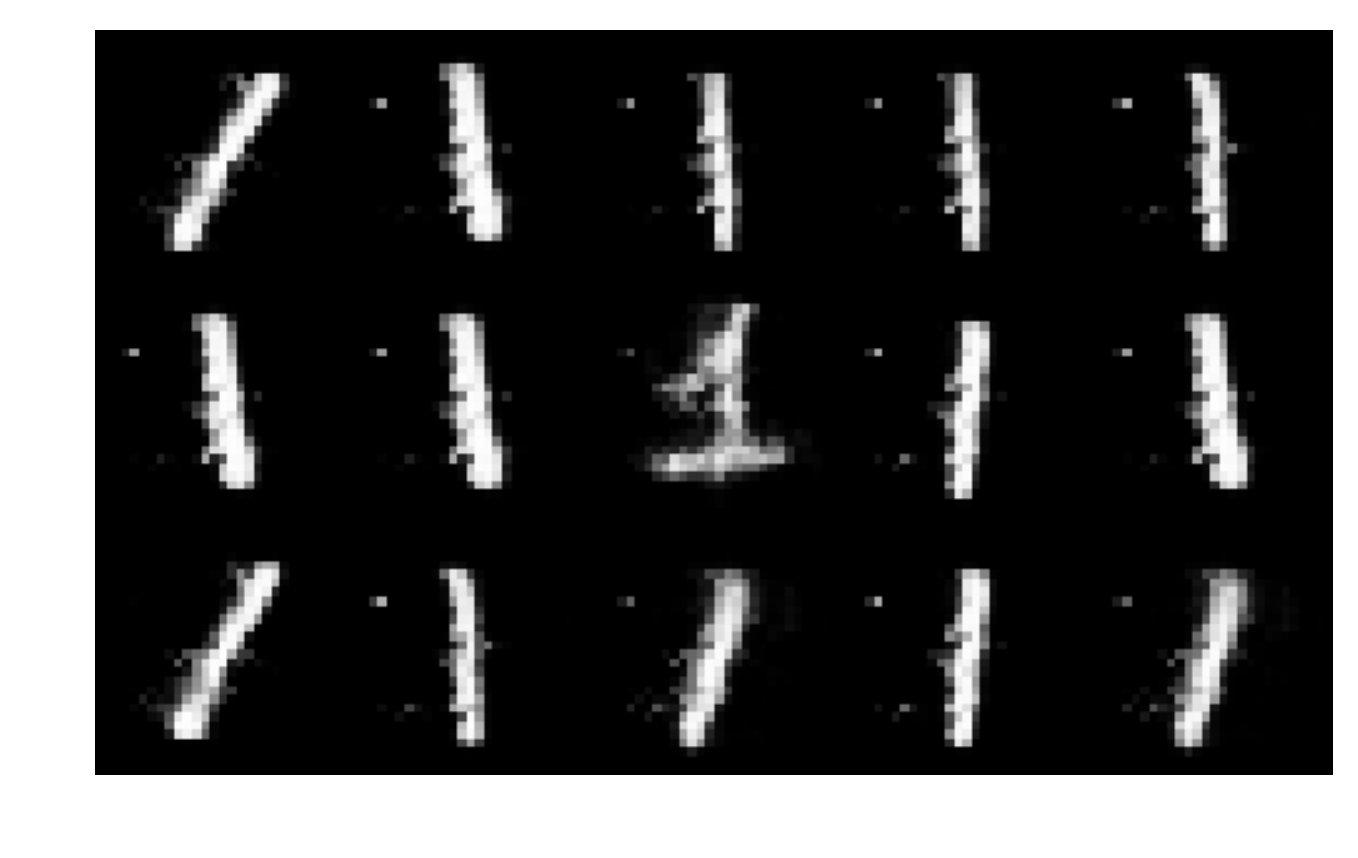}}&\makecell{\includegraphics[scale=0.12]{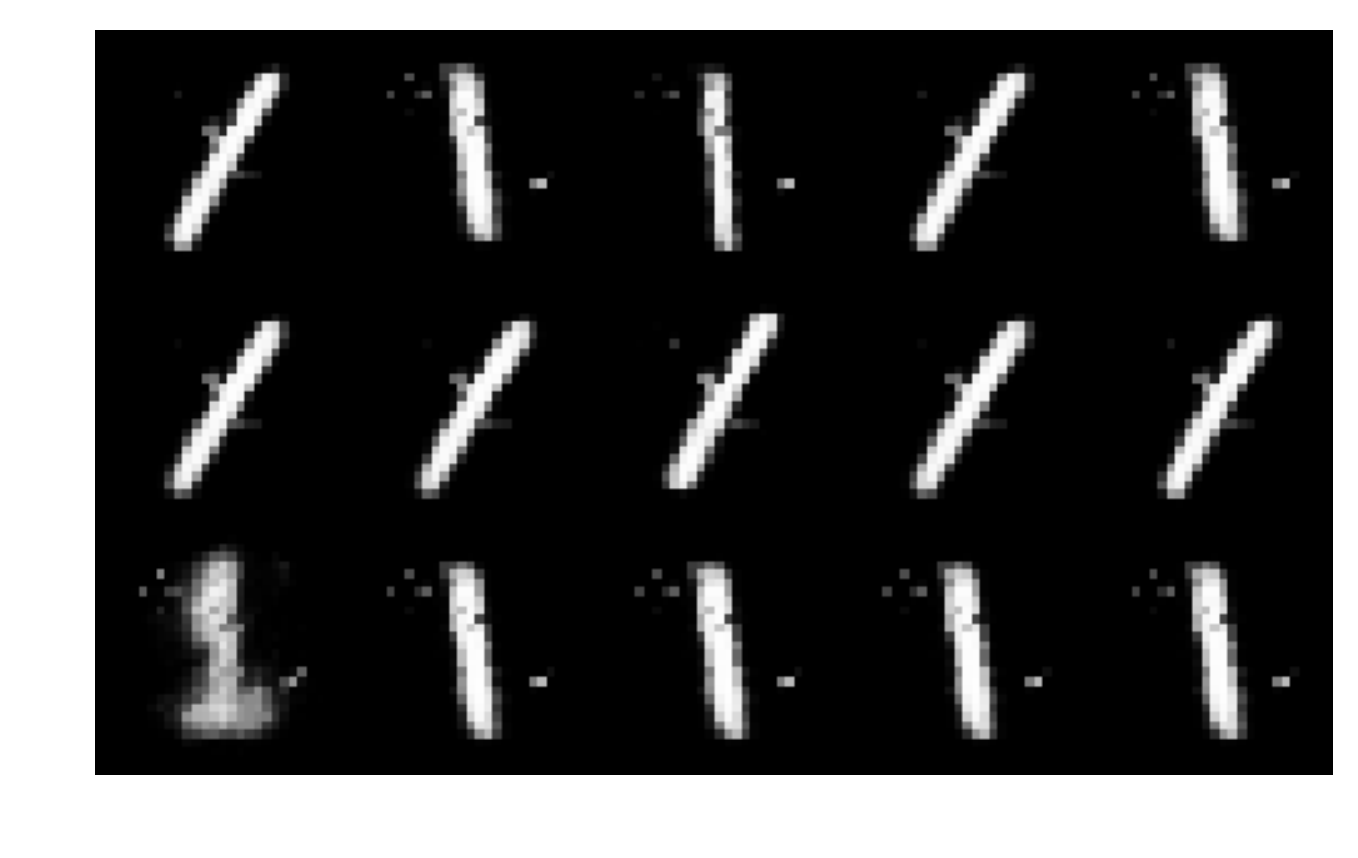}}&\makecell{\includegraphics[scale=0.12]{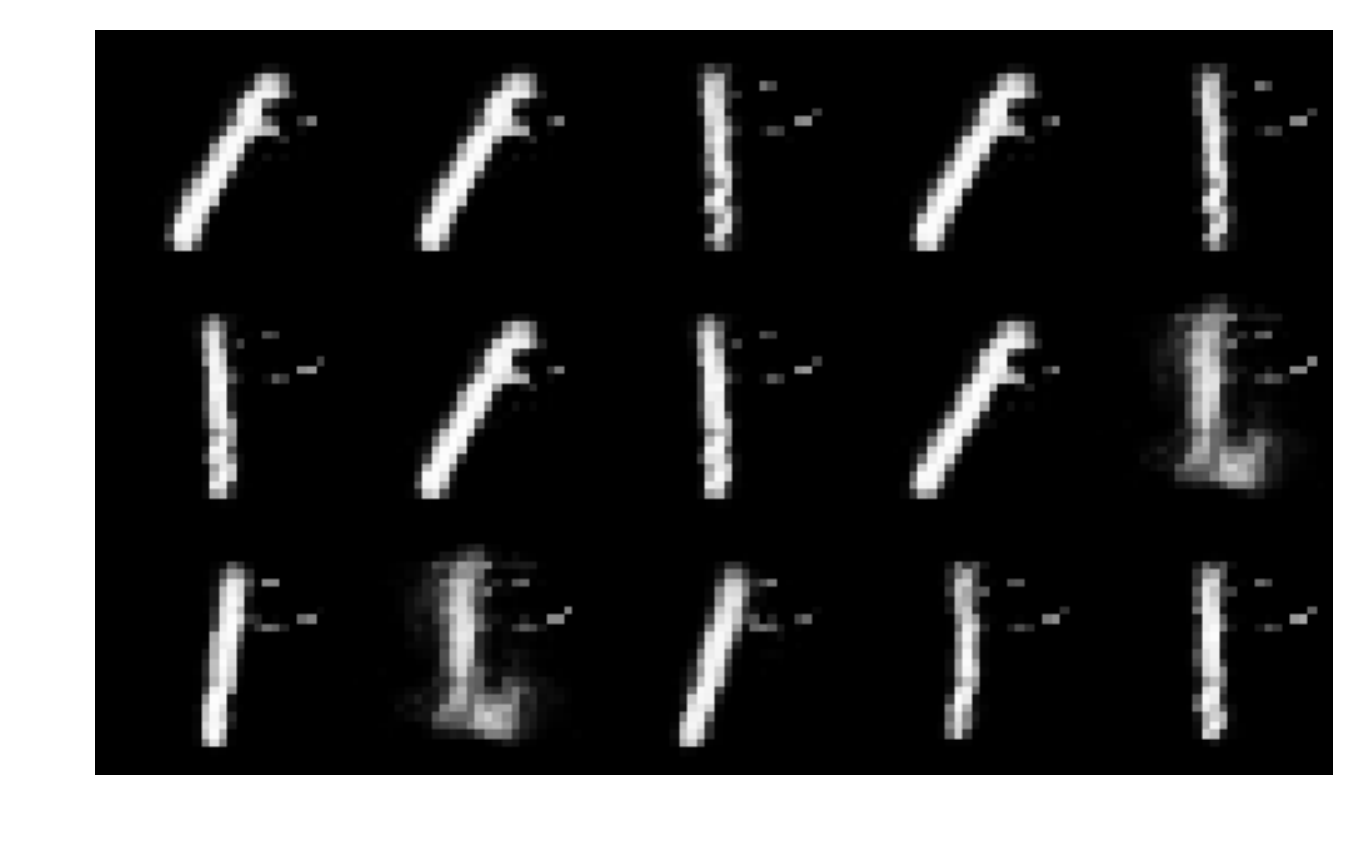}}&\makecell{\includegraphics[scale=0.12]{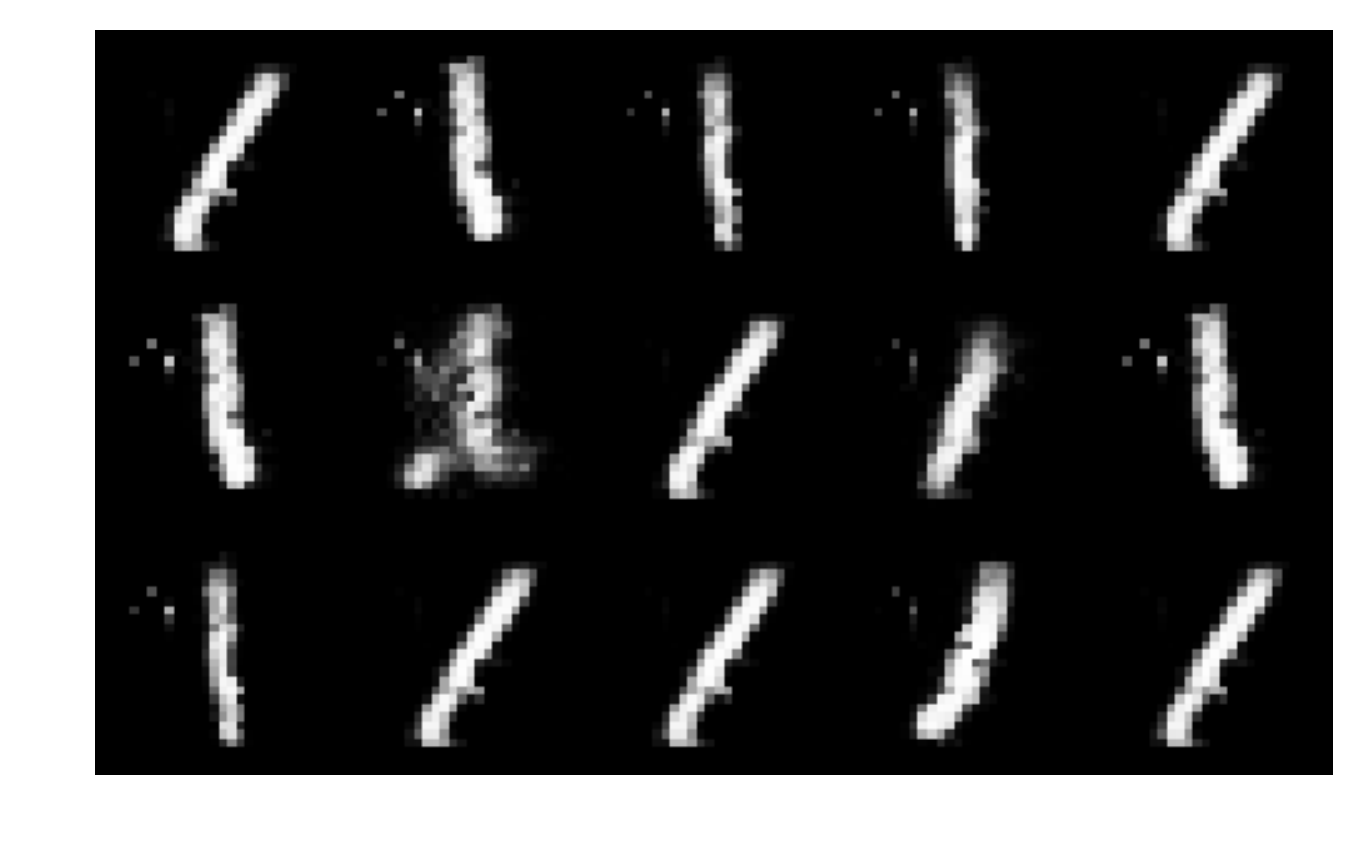}}&\makecell{\includegraphics[scale=0.12]{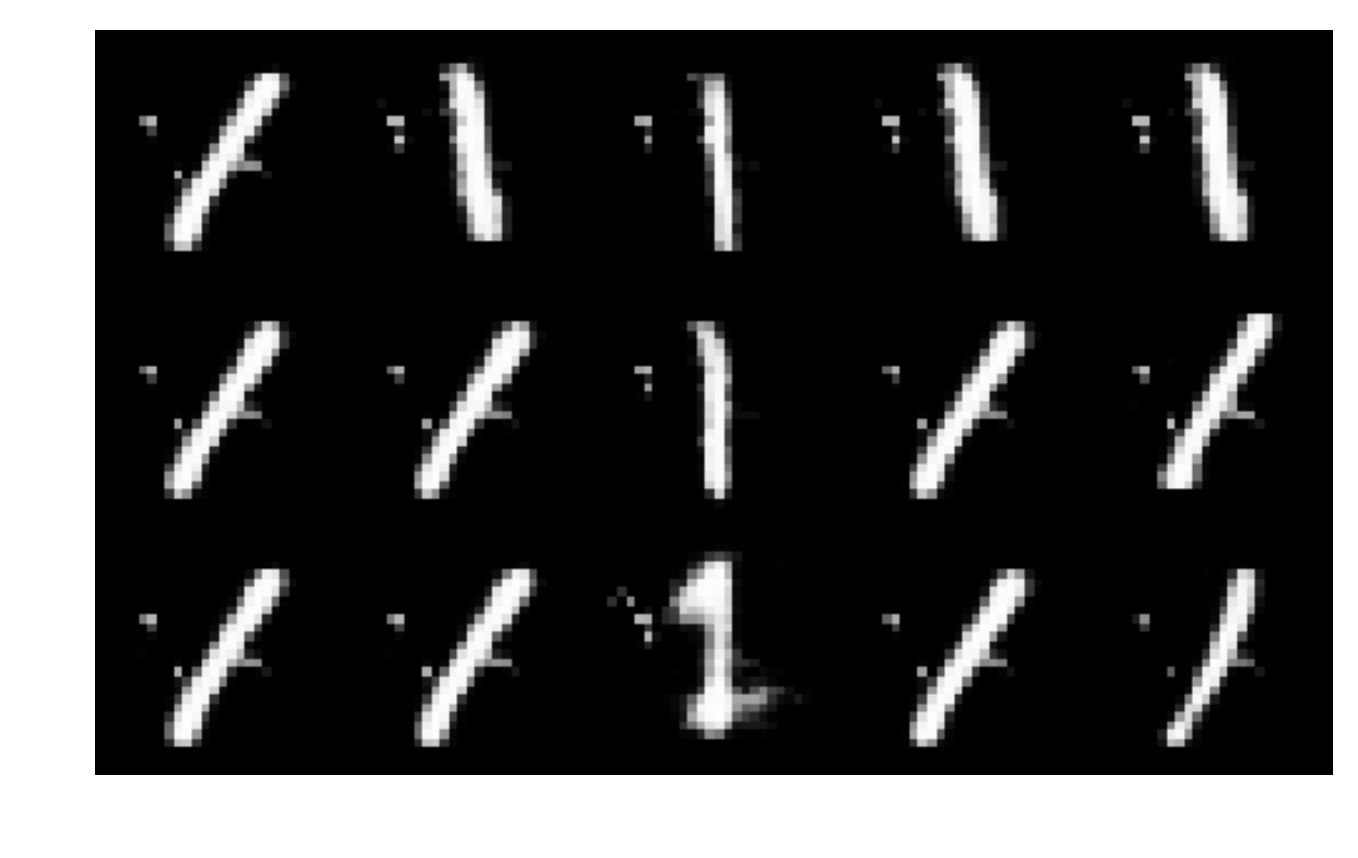}}&\makecell{\includegraphics[scale=0.12]{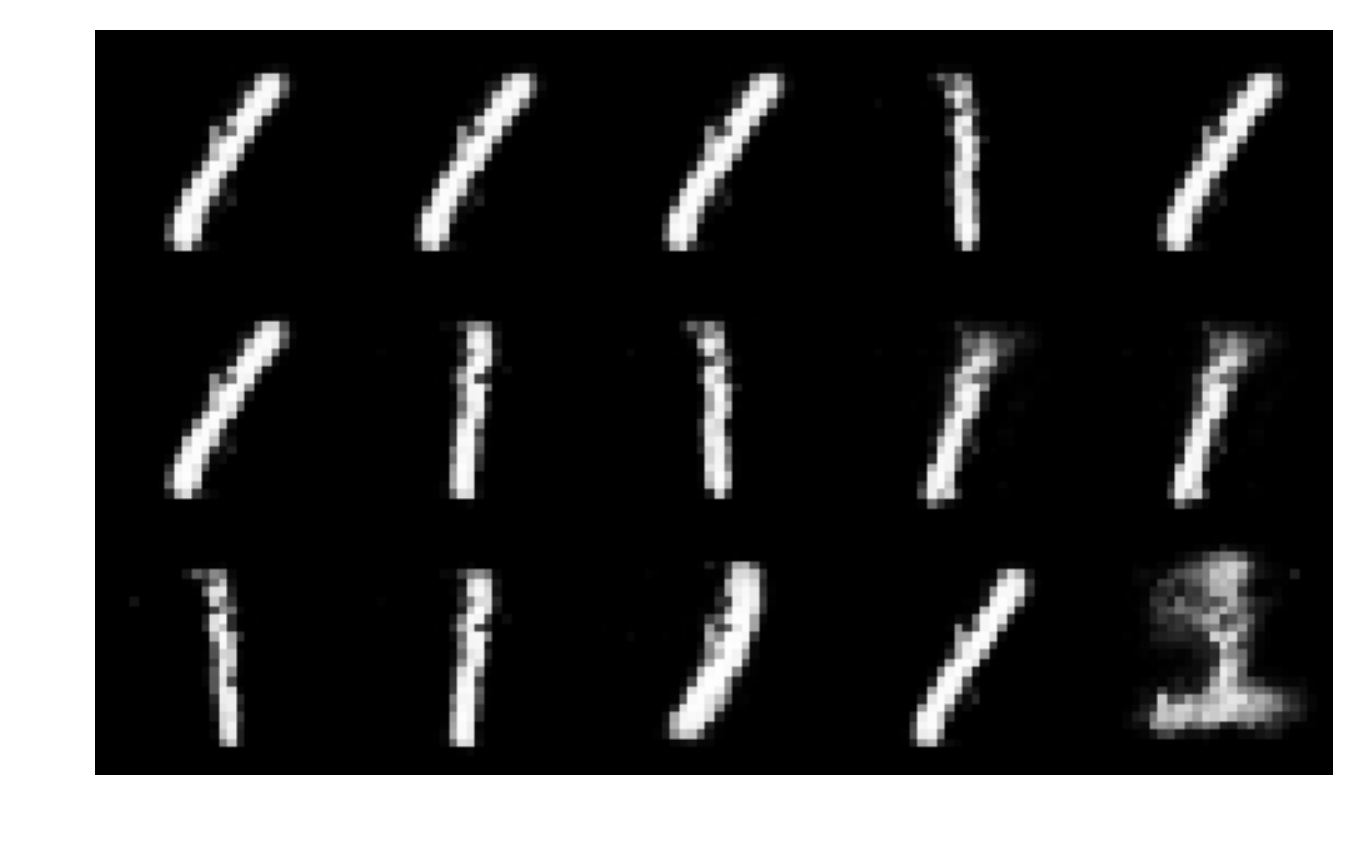}}&\makecell{\includegraphics[scale=0.12]{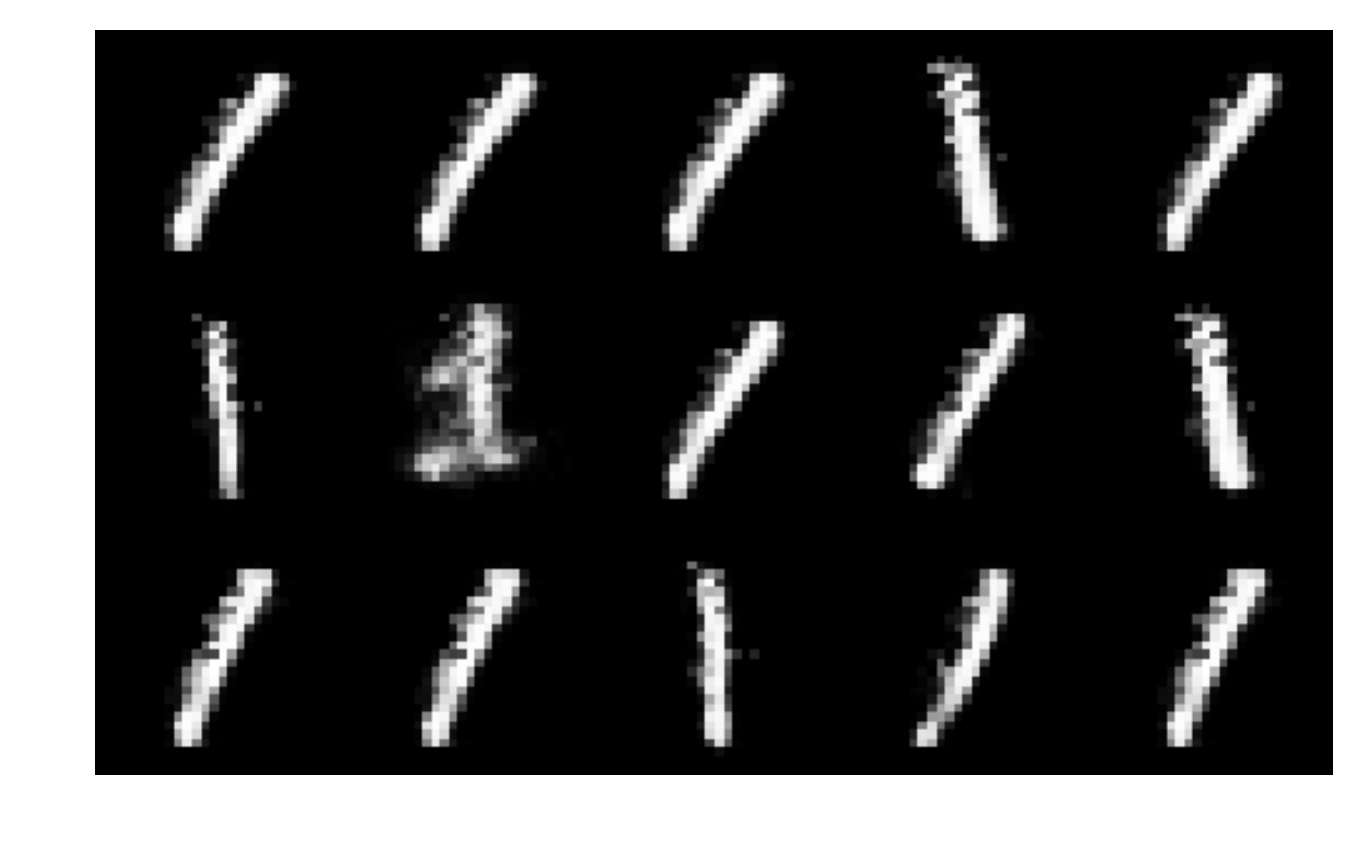}}\\ \hline
`2'&\makecell{\includegraphics[scale=0.12]{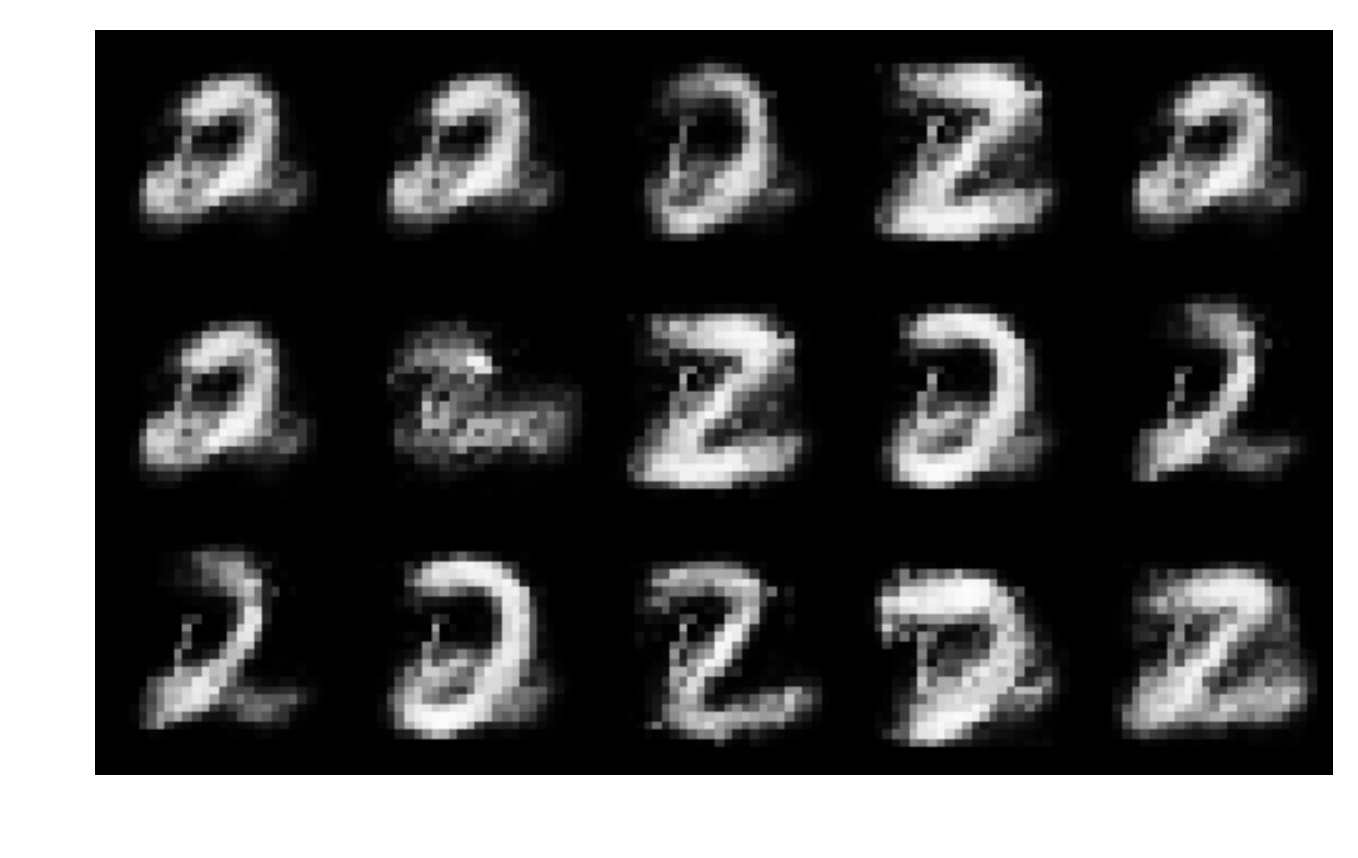}}&\makecell{\includegraphics[scale=0.12]{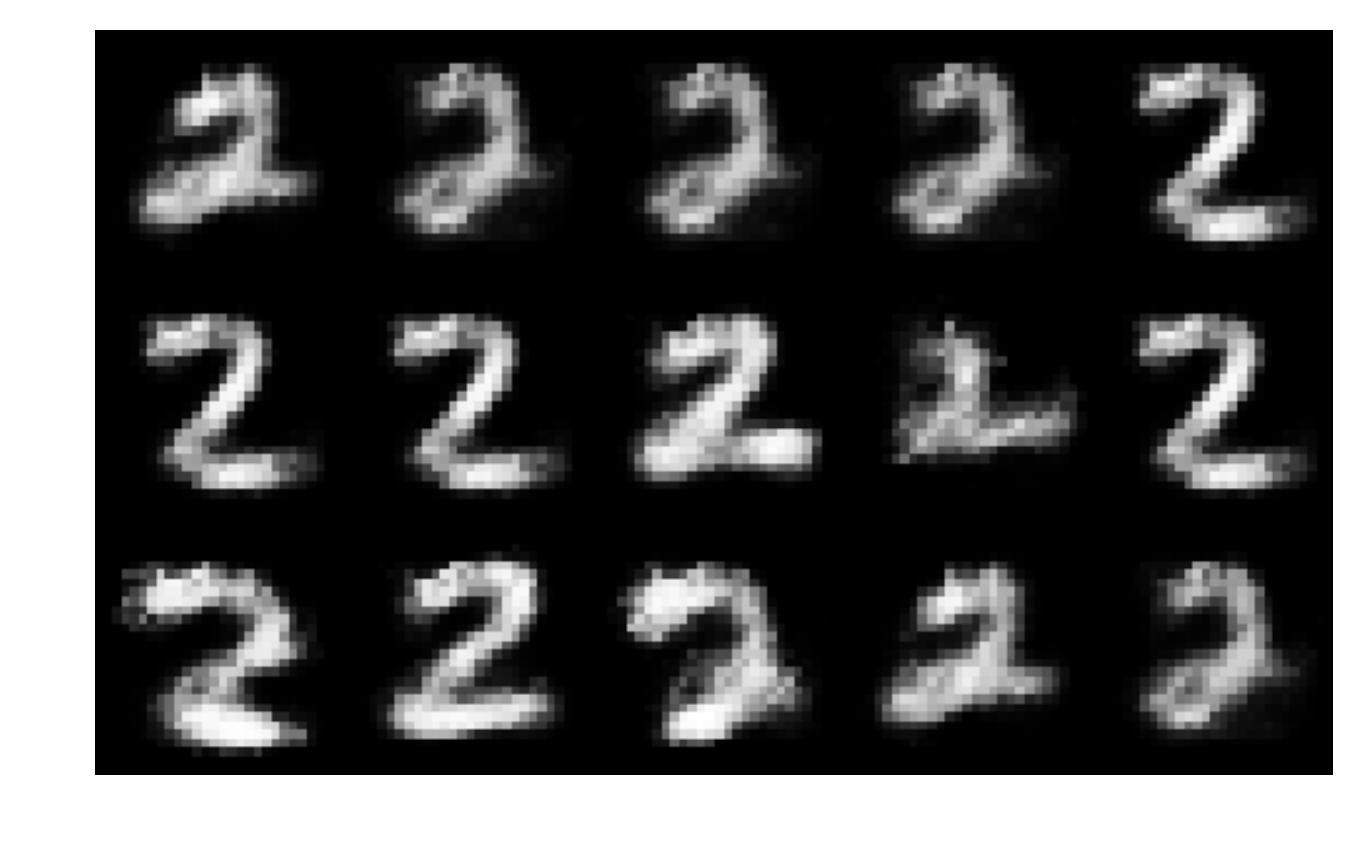}}&--&\makecell{\includegraphics[scale=0.12]{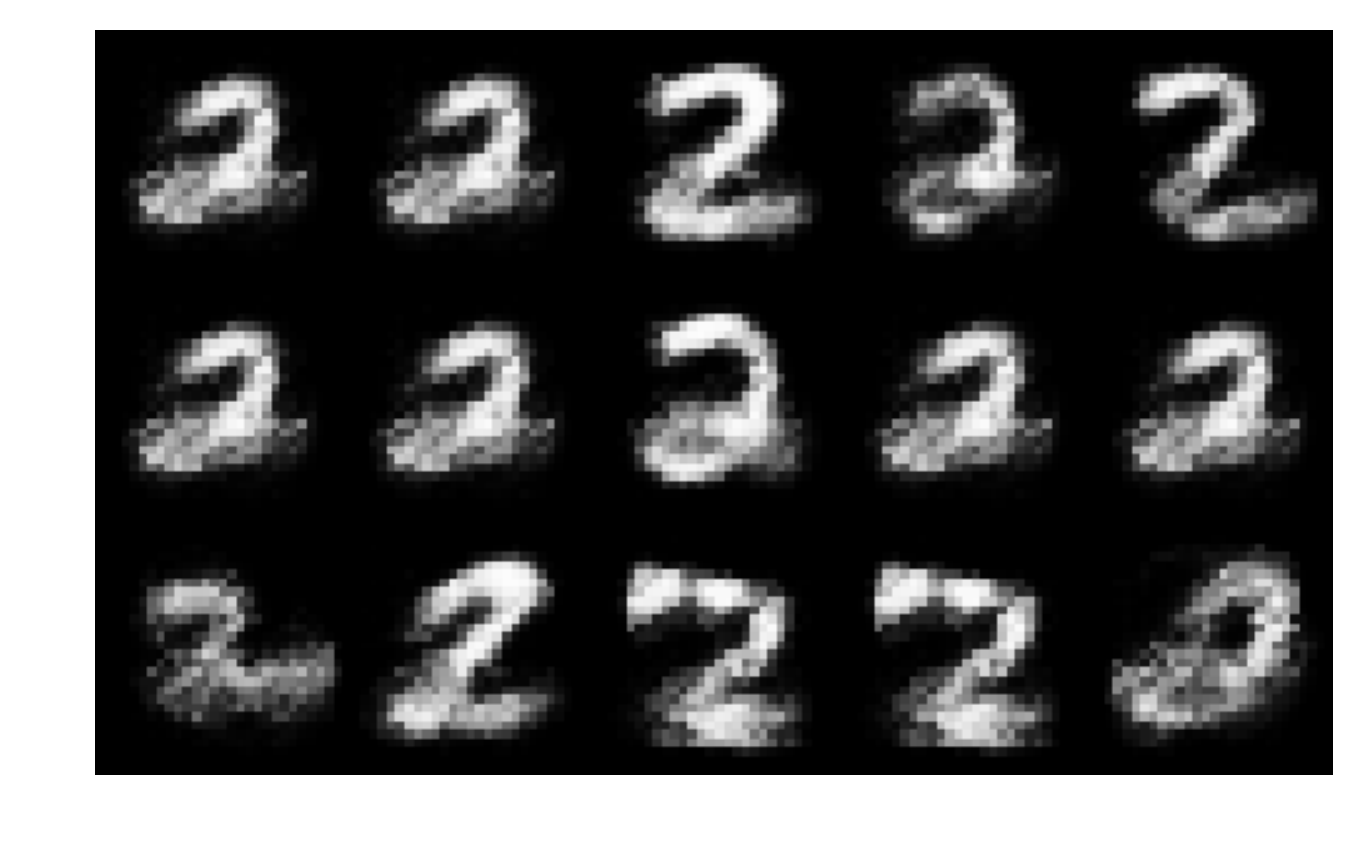}}&\makecell{\includegraphics[scale=0.12]{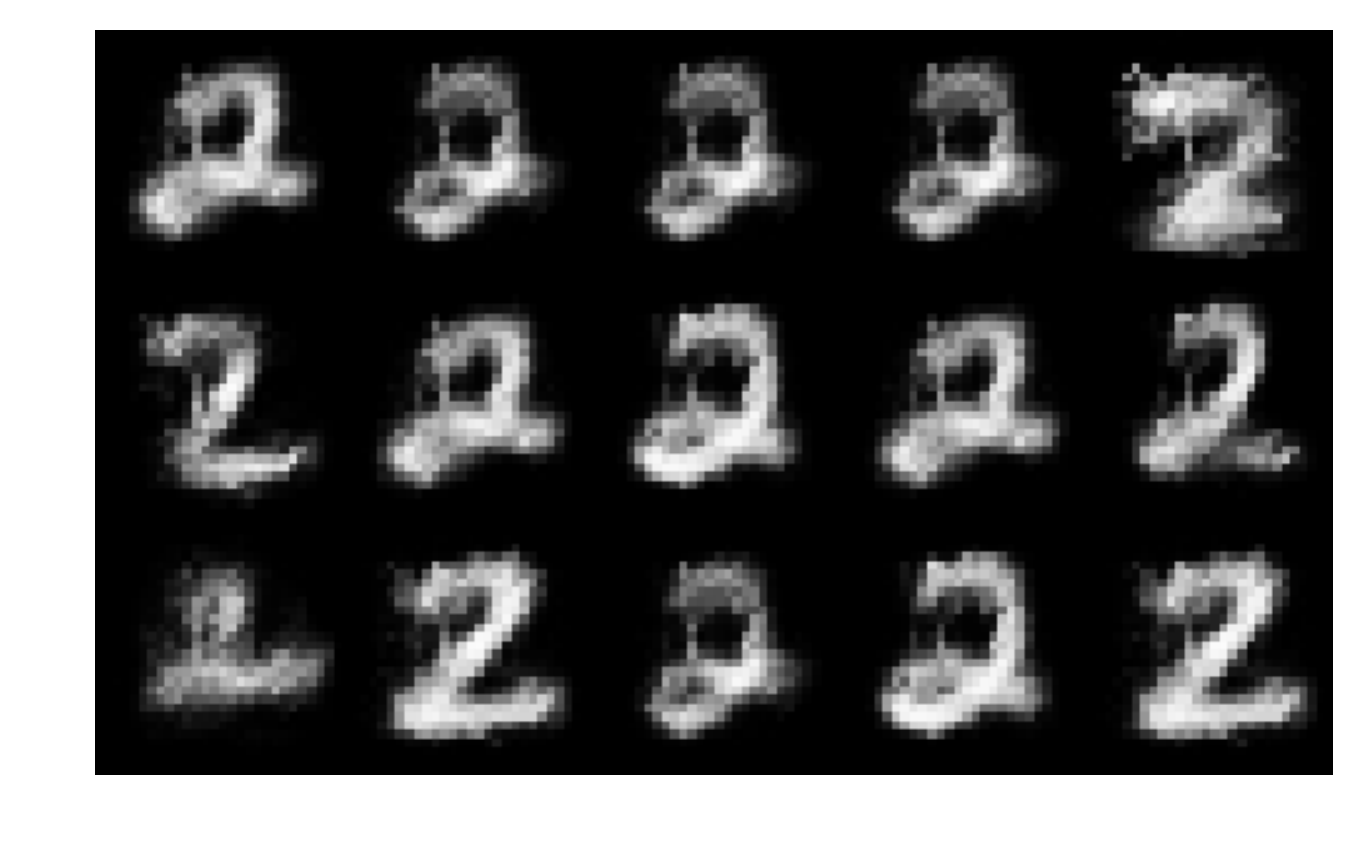}}&\makecell{\includegraphics[scale=0.12]{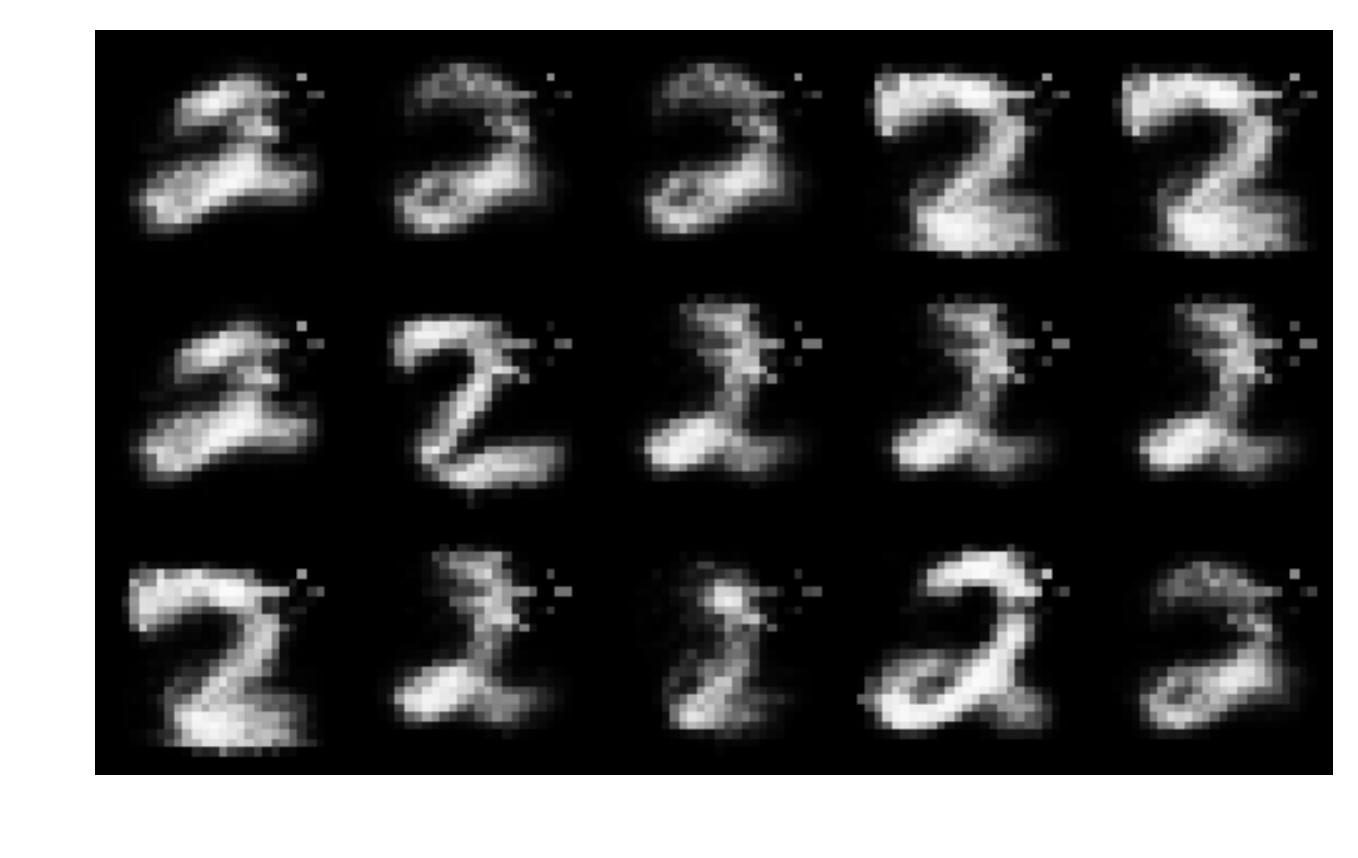}}&\makecell{\includegraphics[scale=0.12]{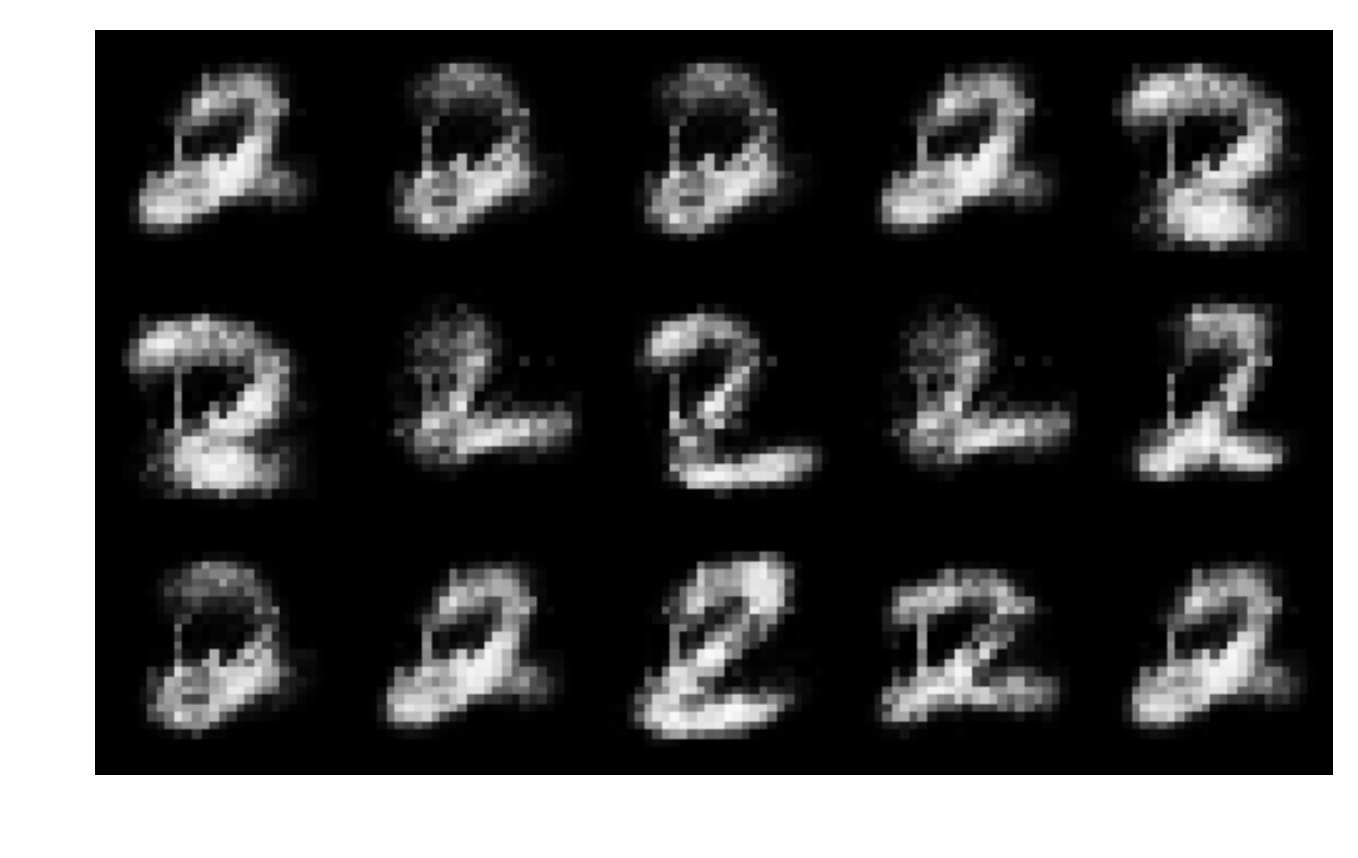}}&\makecell{\includegraphics[scale=0.12]{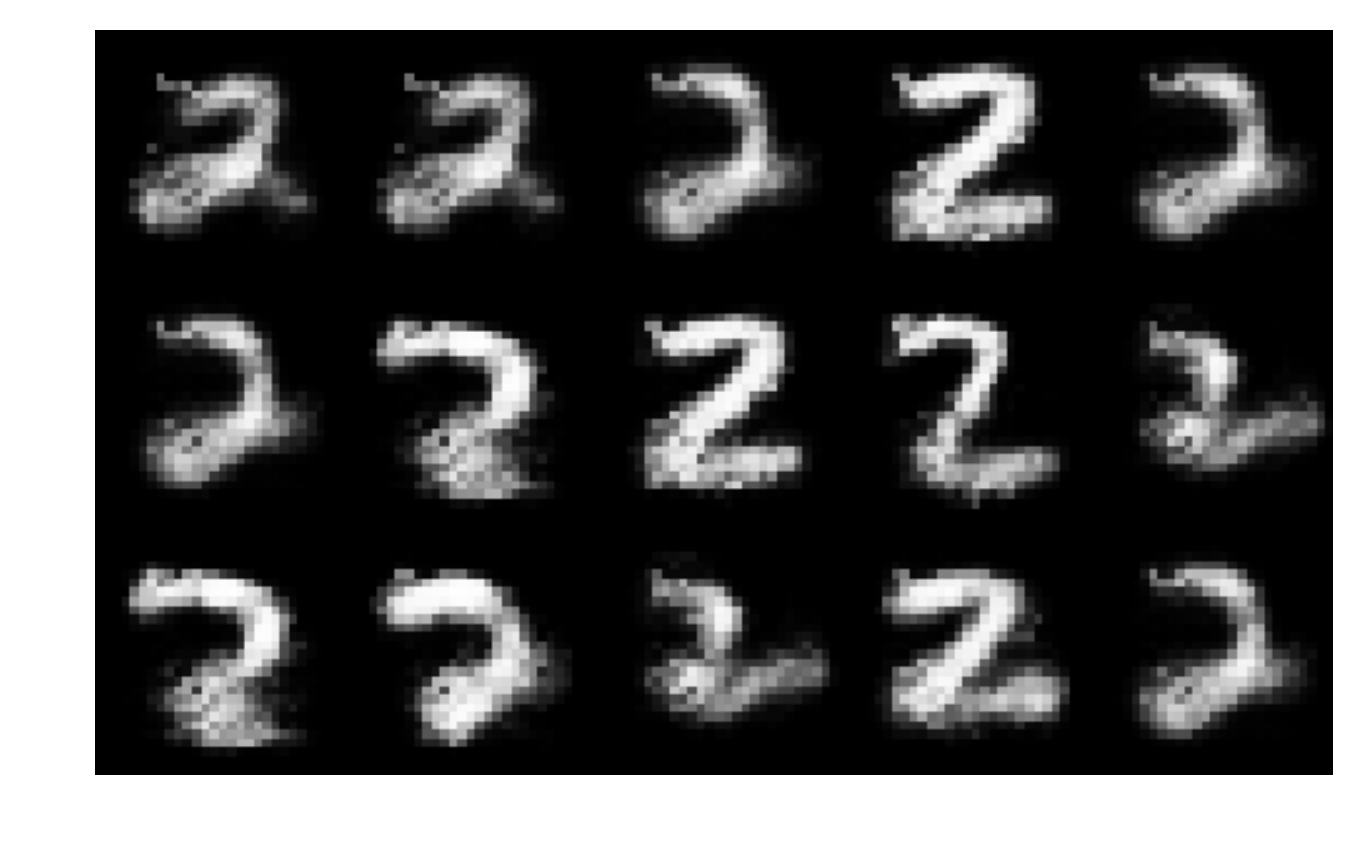}}&\makecell{\includegraphics[scale=0.12]{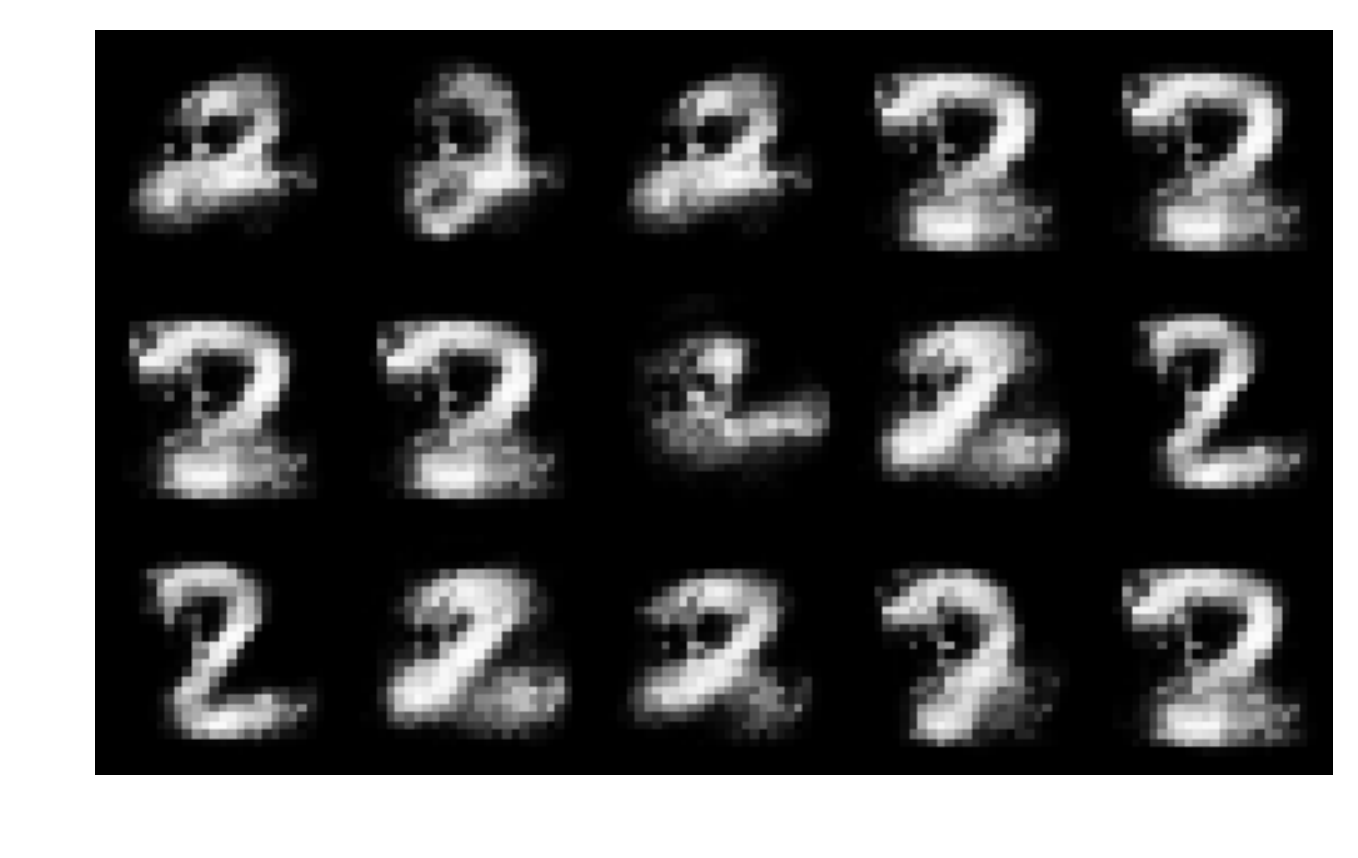}}&\makecell{\includegraphics[scale=0.12]{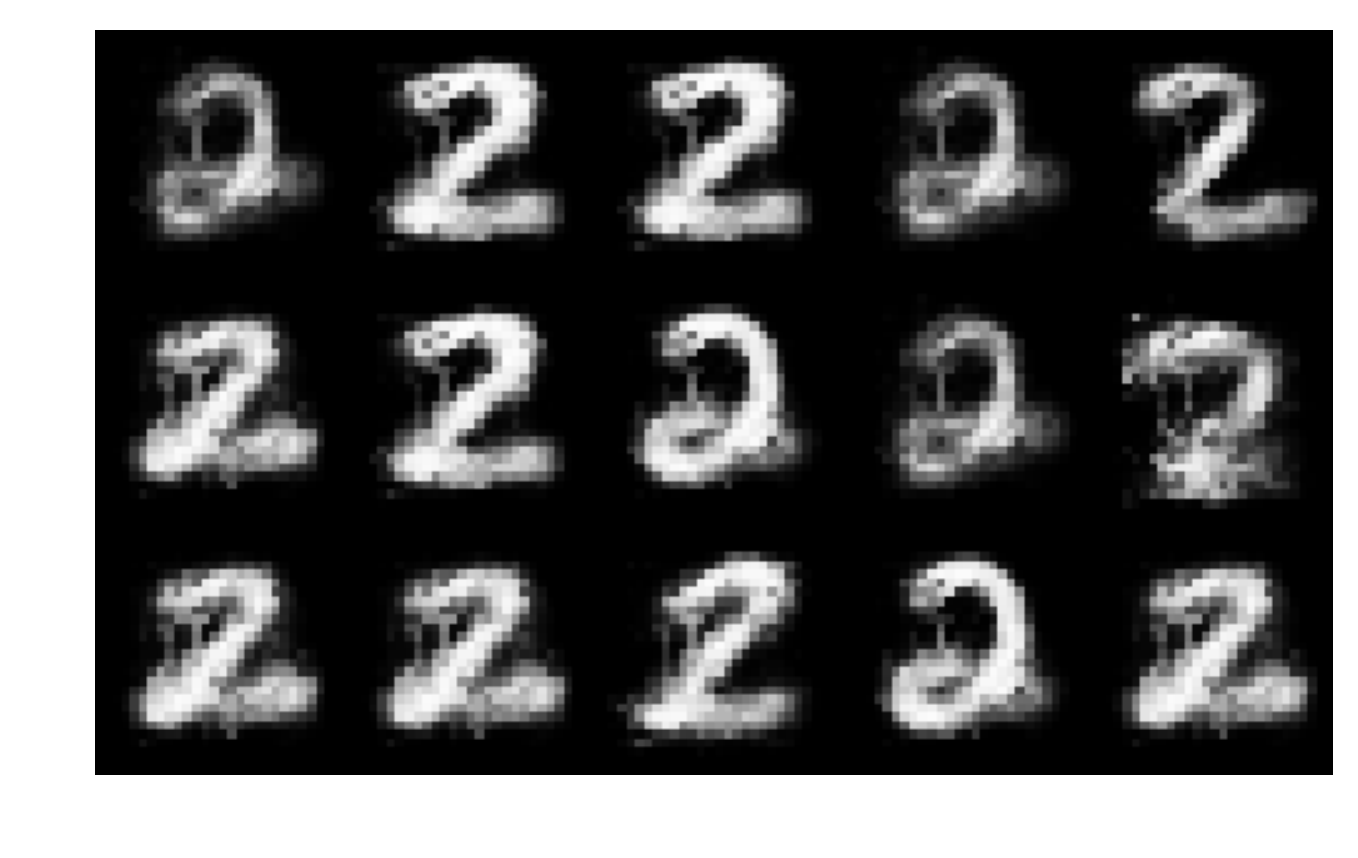}}\\ \hline
`3'&\makecell{\includegraphics[scale=0.12]{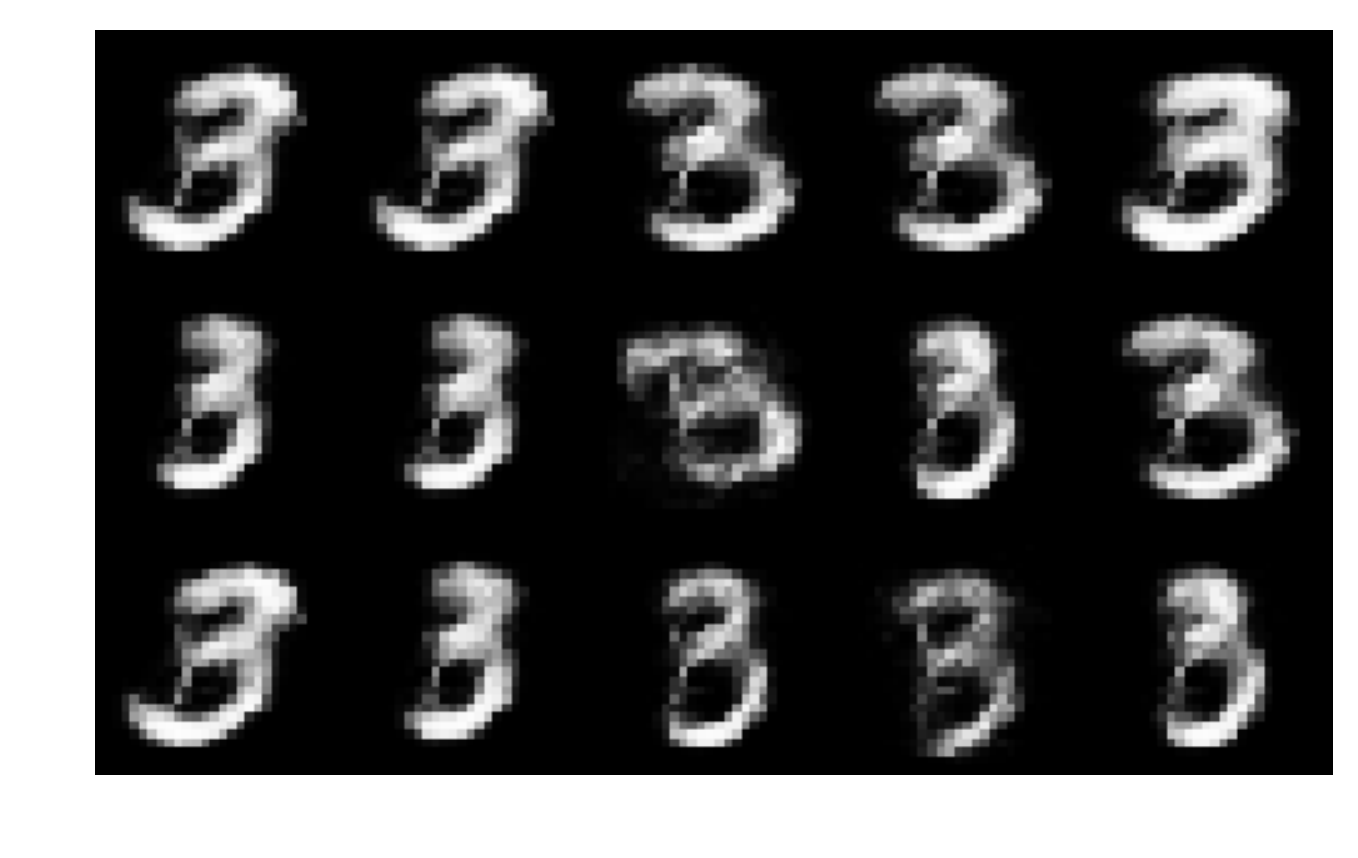}}&\makecell{\includegraphics[scale=0.12]{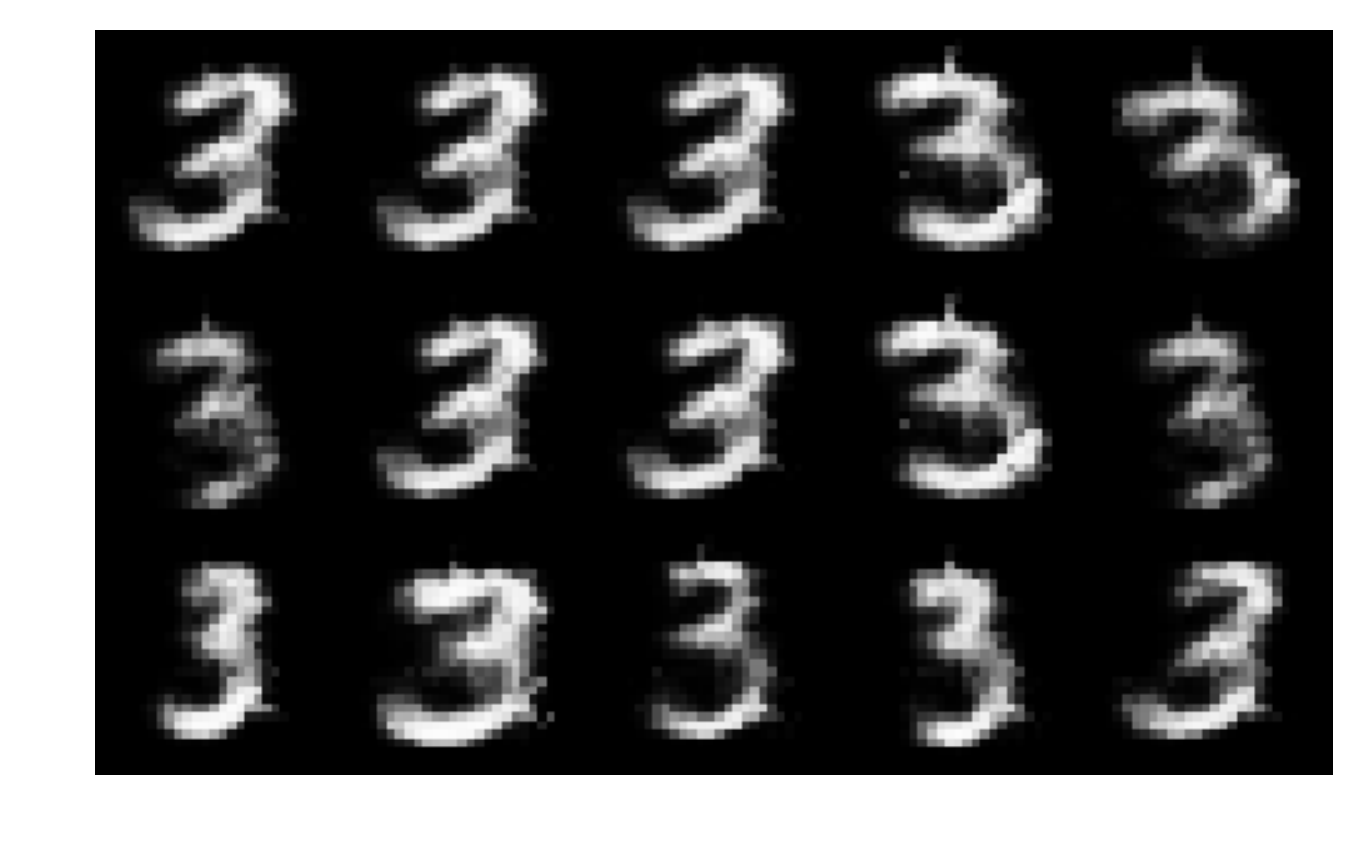}}&\makecell{\includegraphics[scale=0.12]{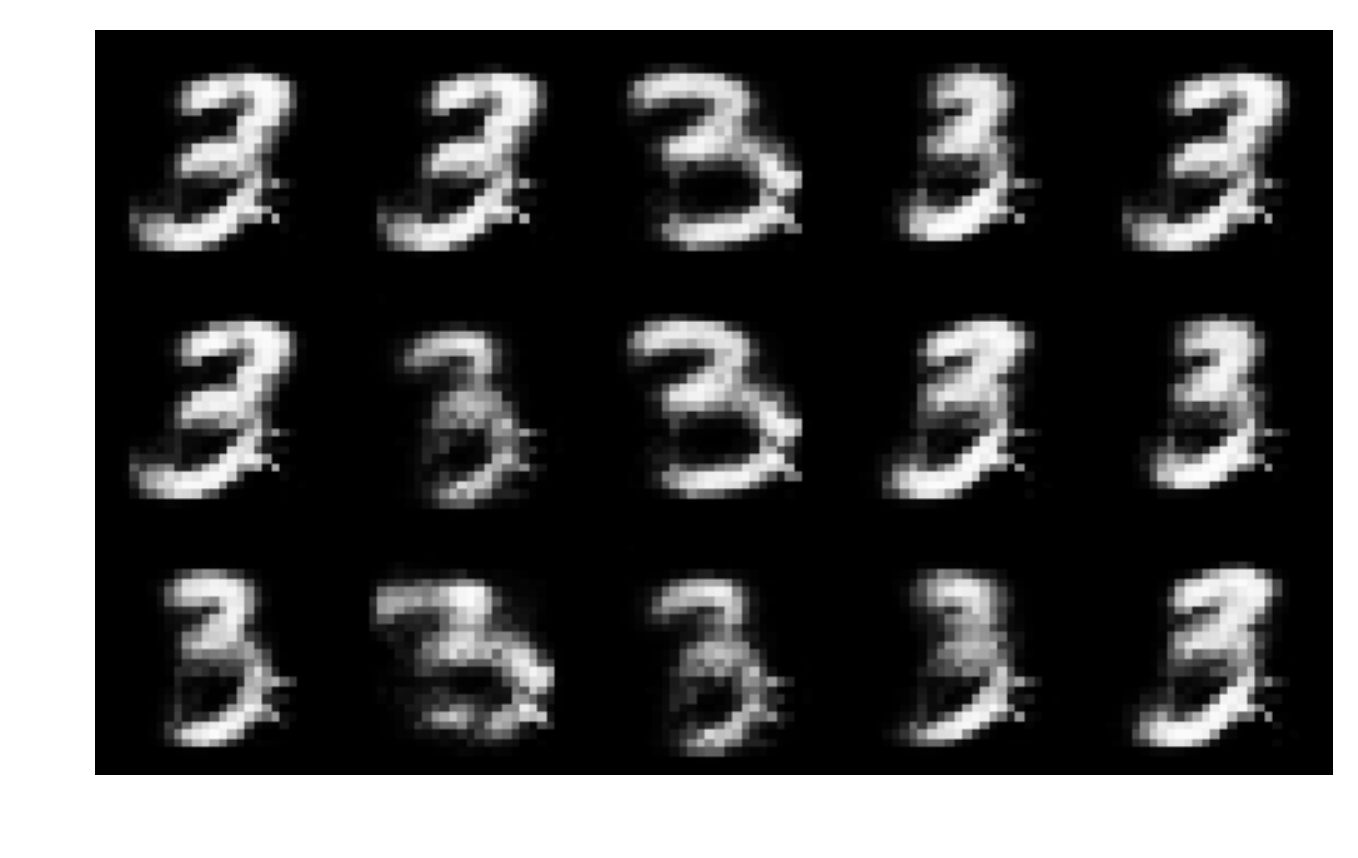}}&--&\makecell{\includegraphics[scale=0.12]{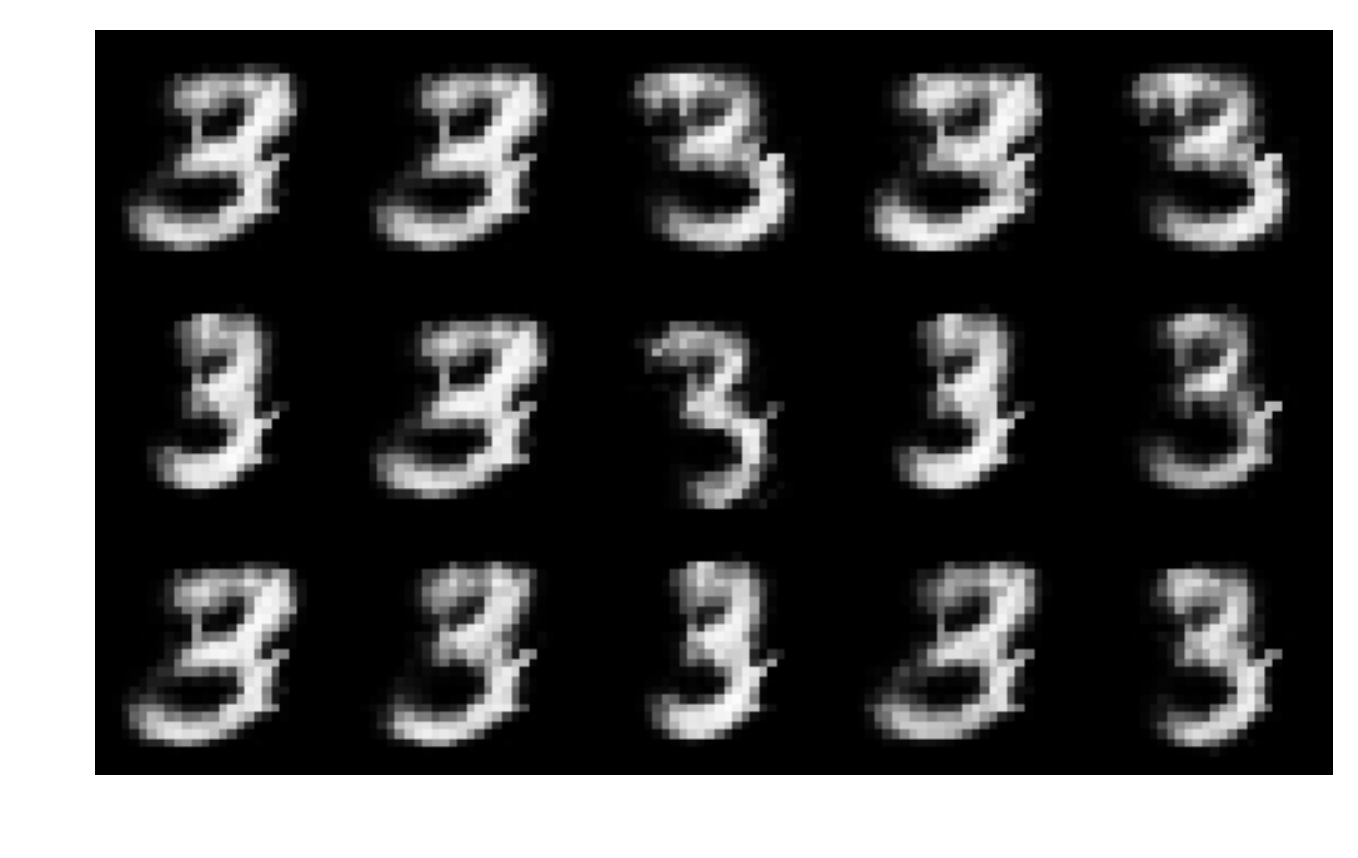}}&\makecell{\includegraphics[scale=0.12]{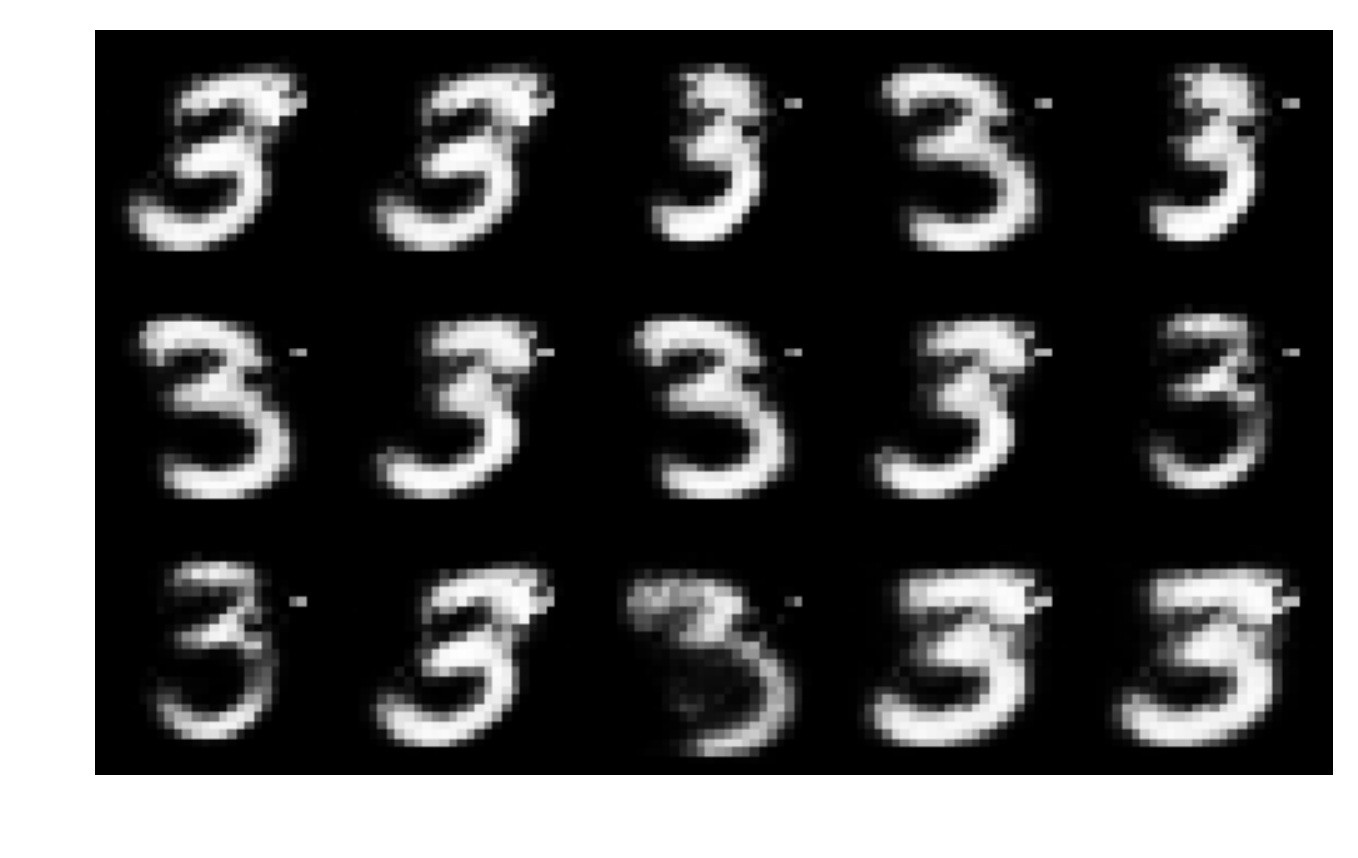}}&\makecell{\includegraphics[scale=0.12]{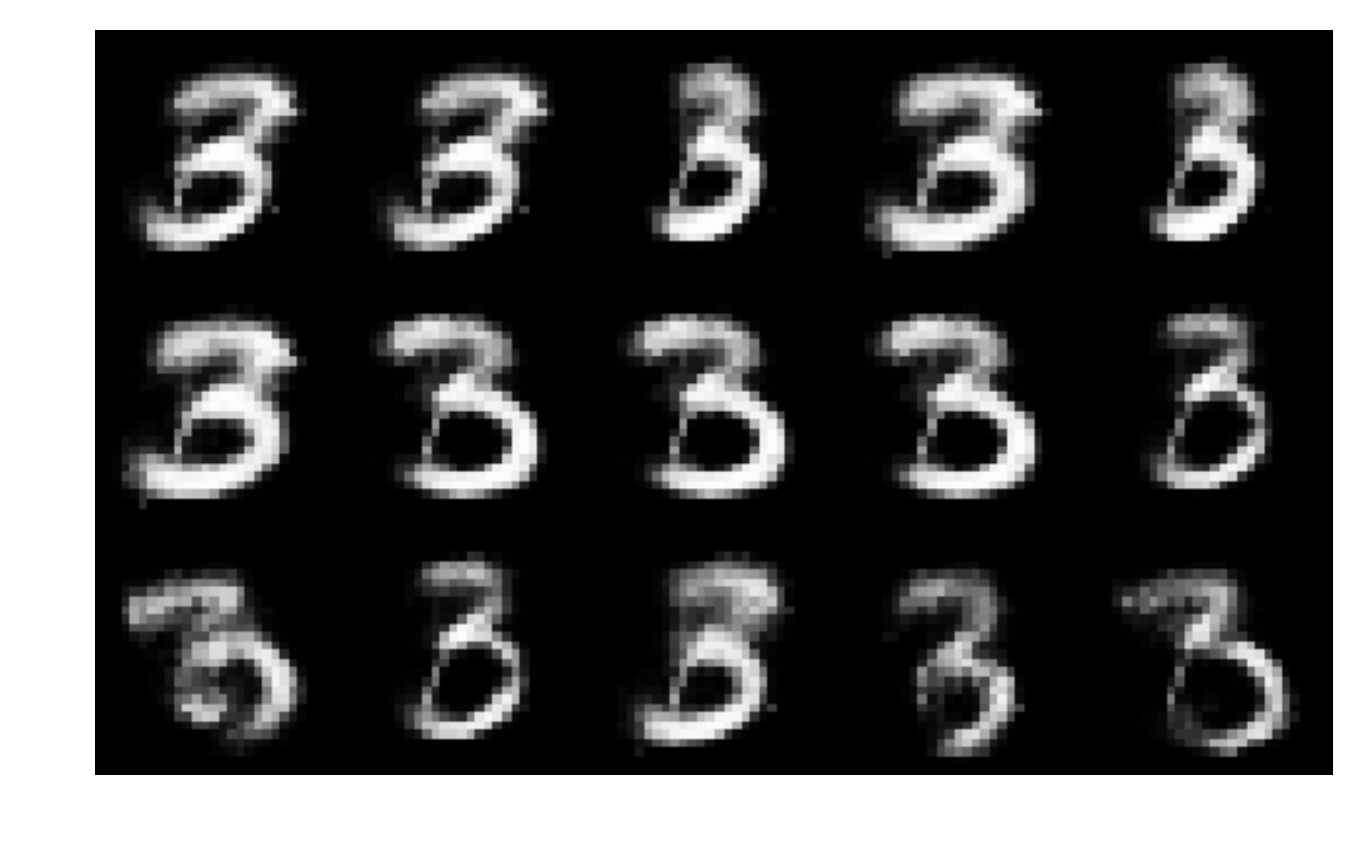}}&\makecell{\includegraphics[scale=0.12]{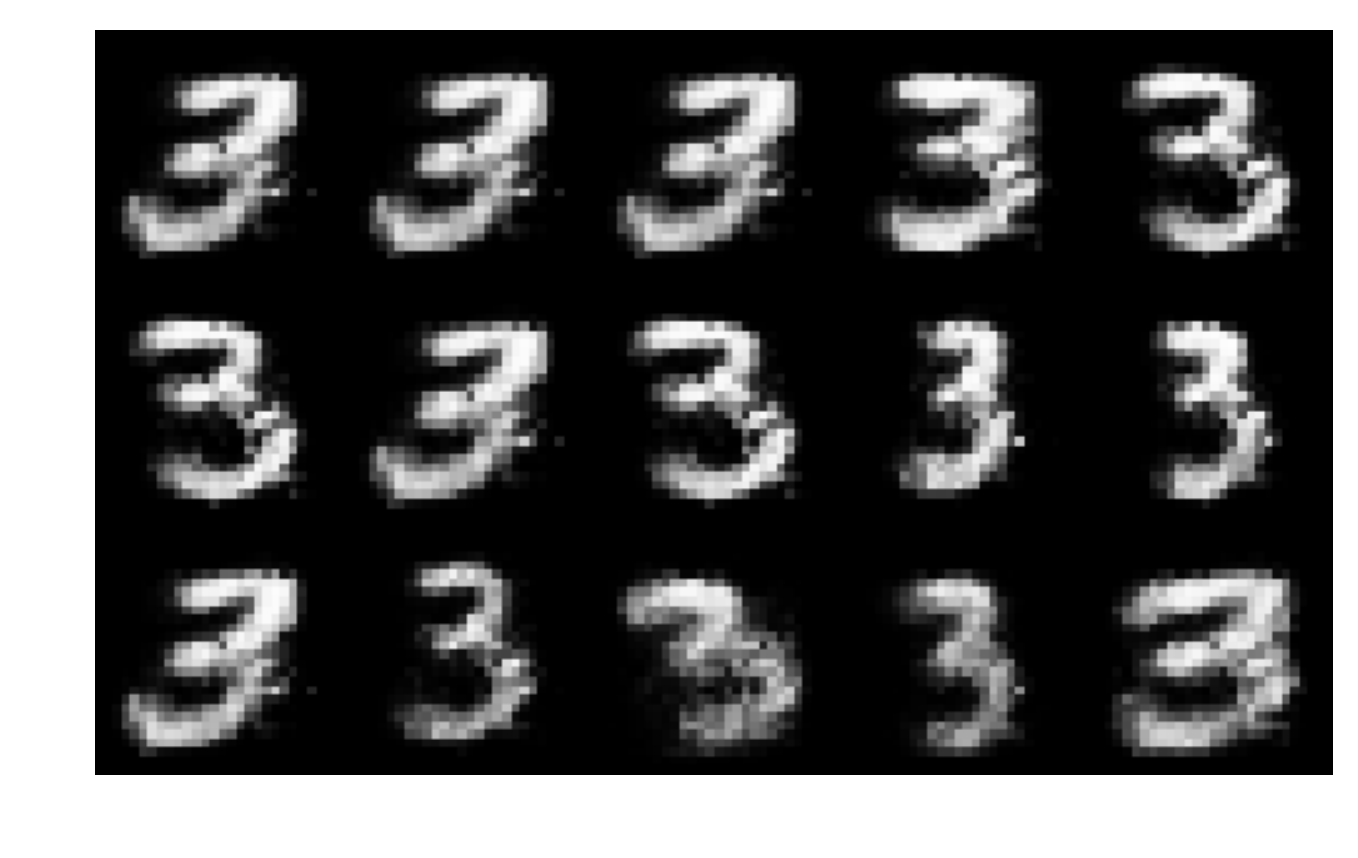}}&\makecell{\includegraphics[scale=0.12]{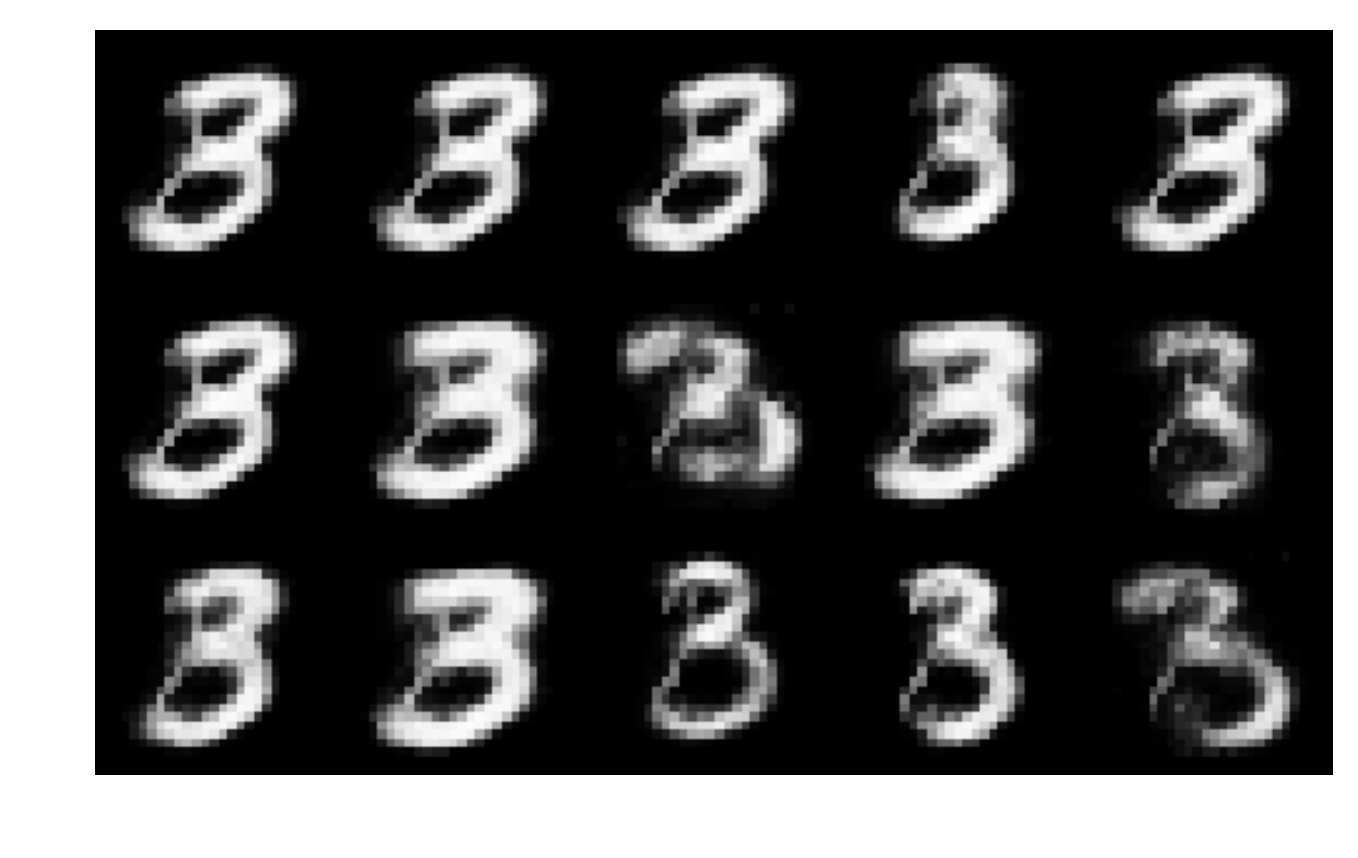}}&\makecell{\includegraphics[scale=0.12]{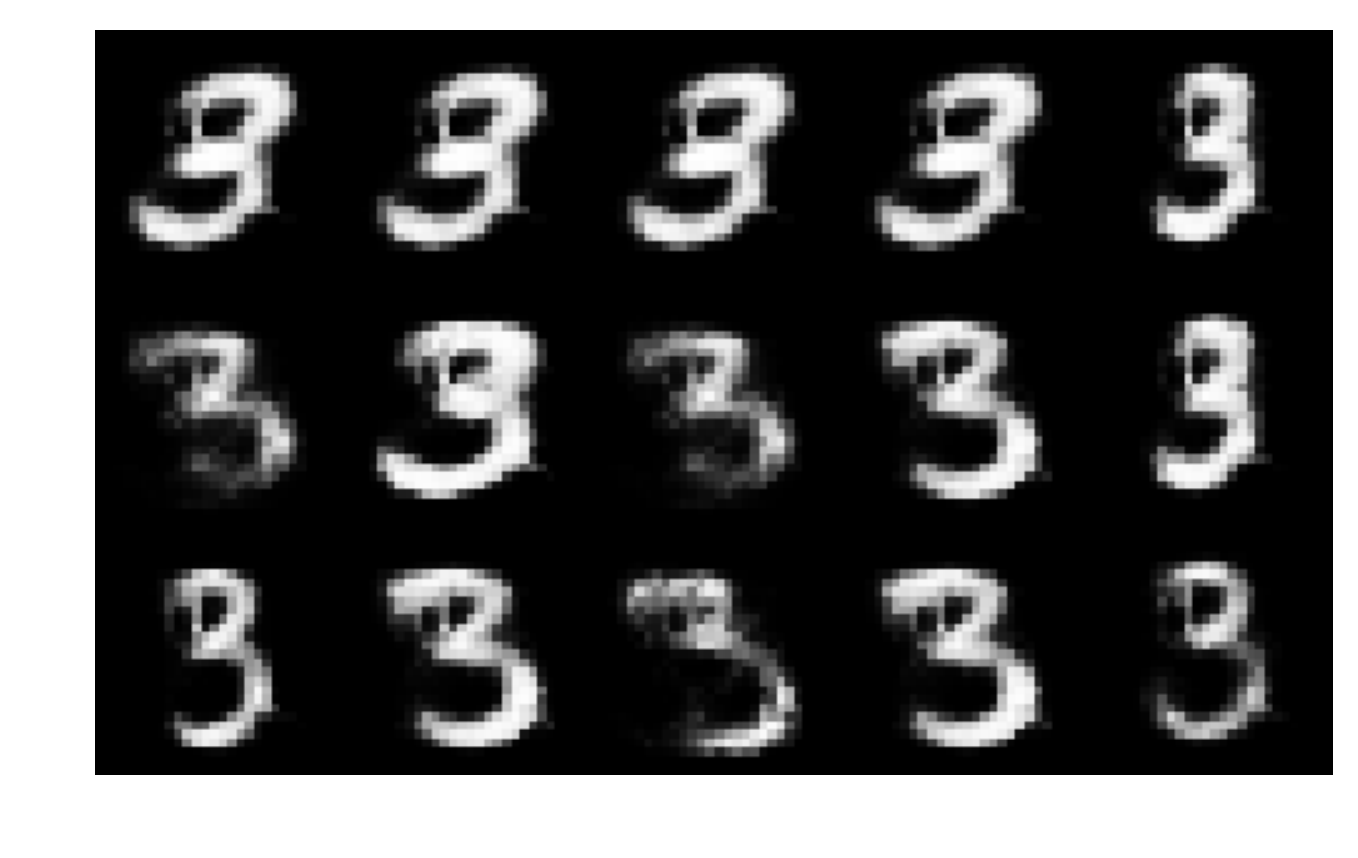}}\\ \hline
`4'&\makecell{\includegraphics[scale=0.12]{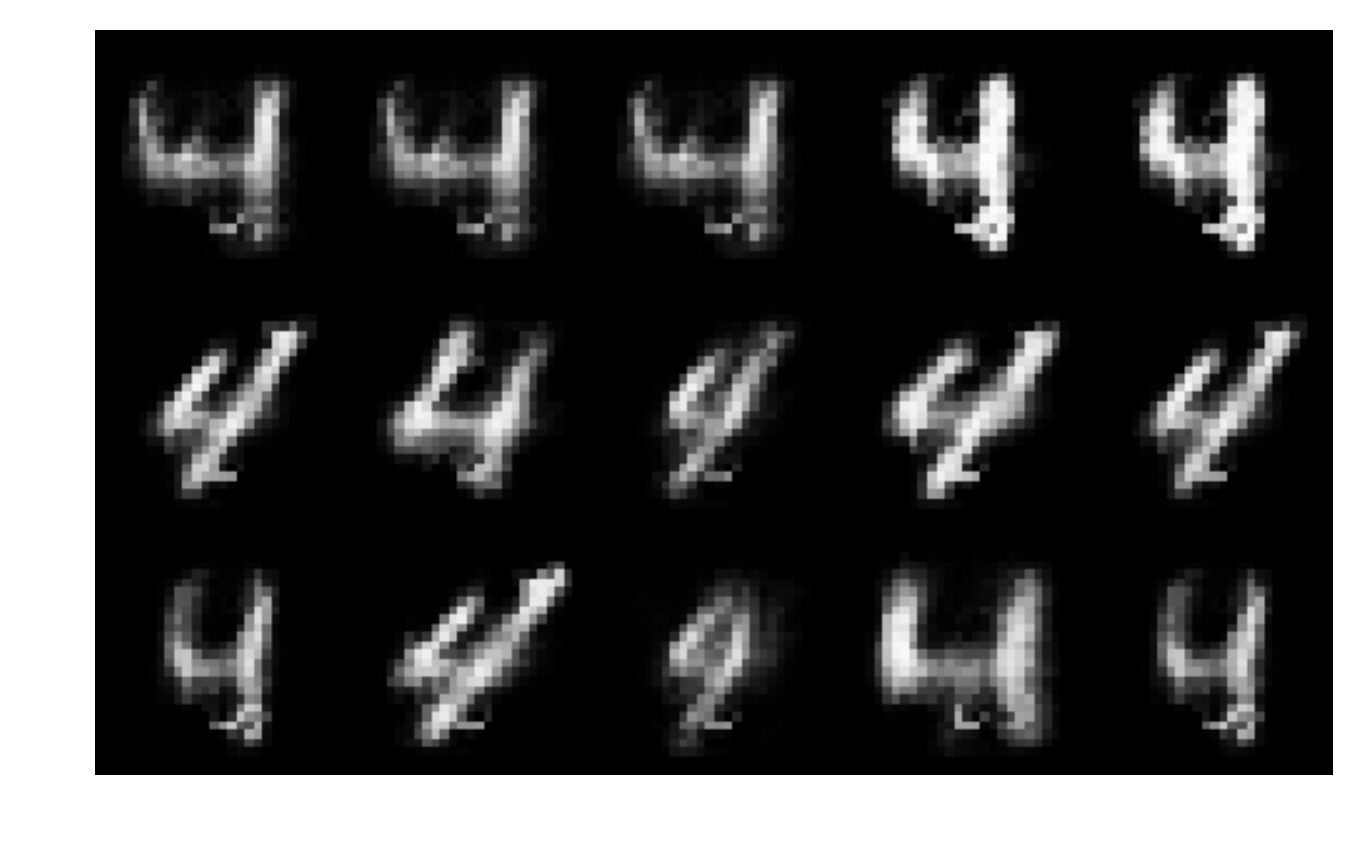}}&\makecell{\includegraphics[scale=0.12]{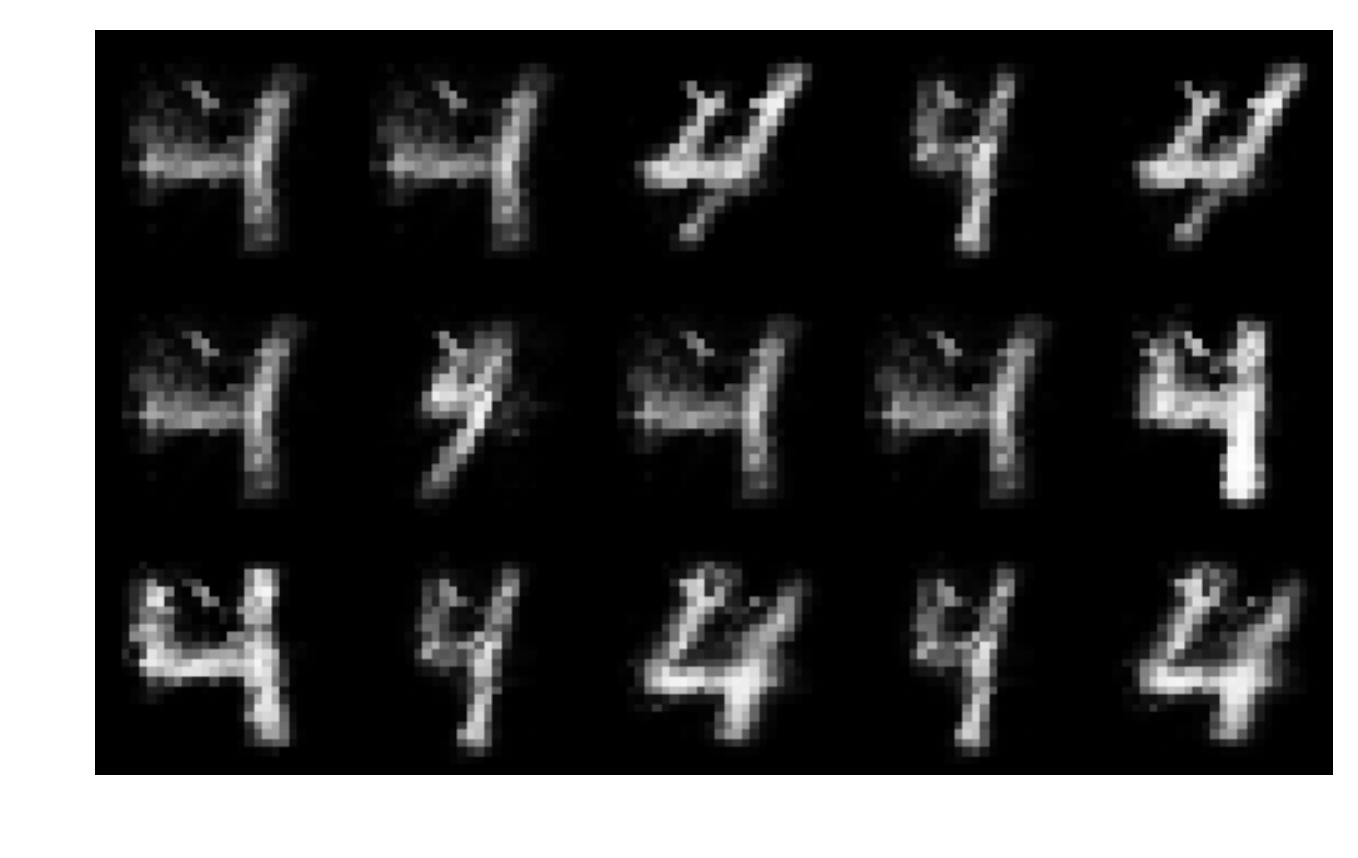}}&\makecell{\includegraphics[scale=0.12]{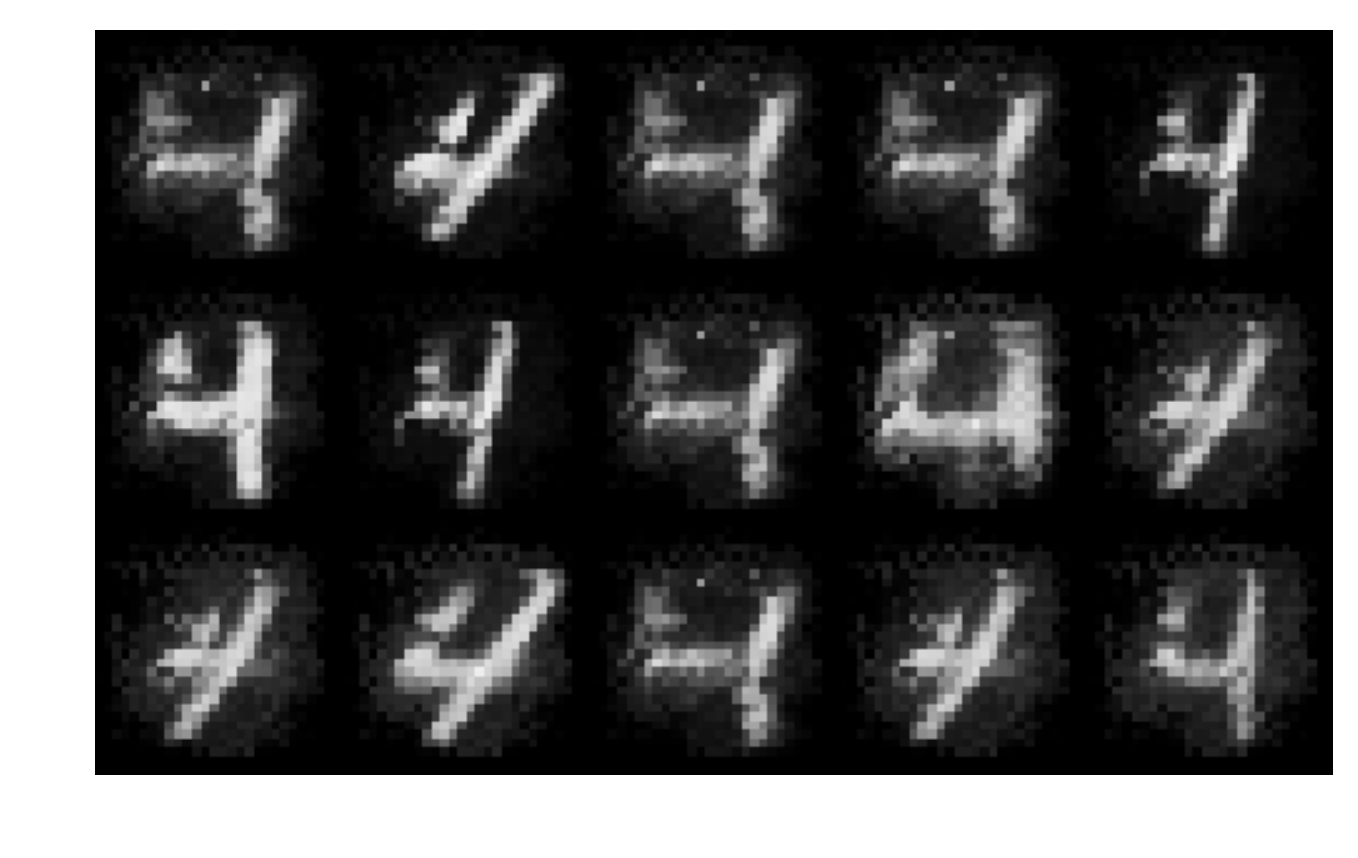}}&\makecell{\includegraphics[scale=0.12]{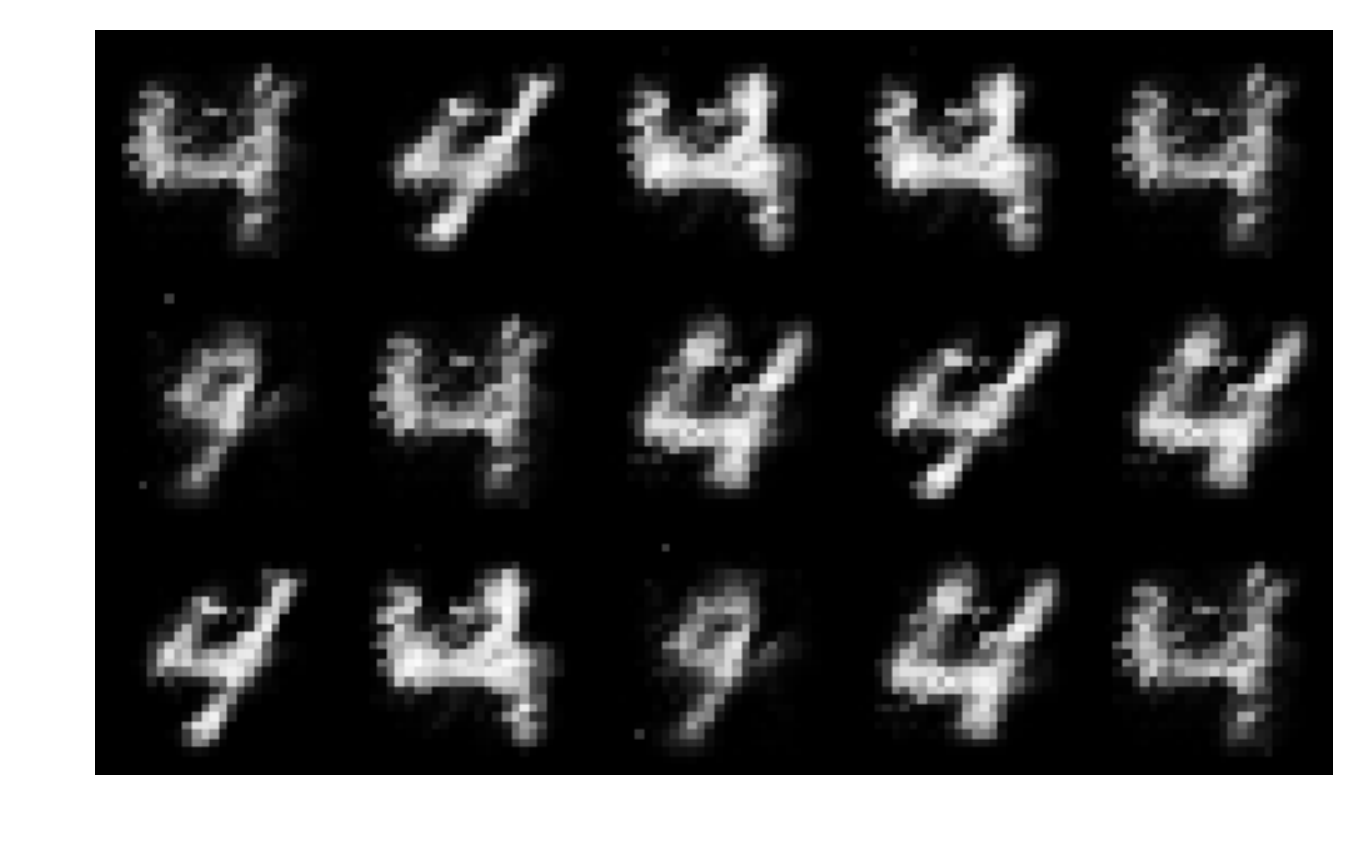}}&--&\makecell{\includegraphics[scale=0.12]{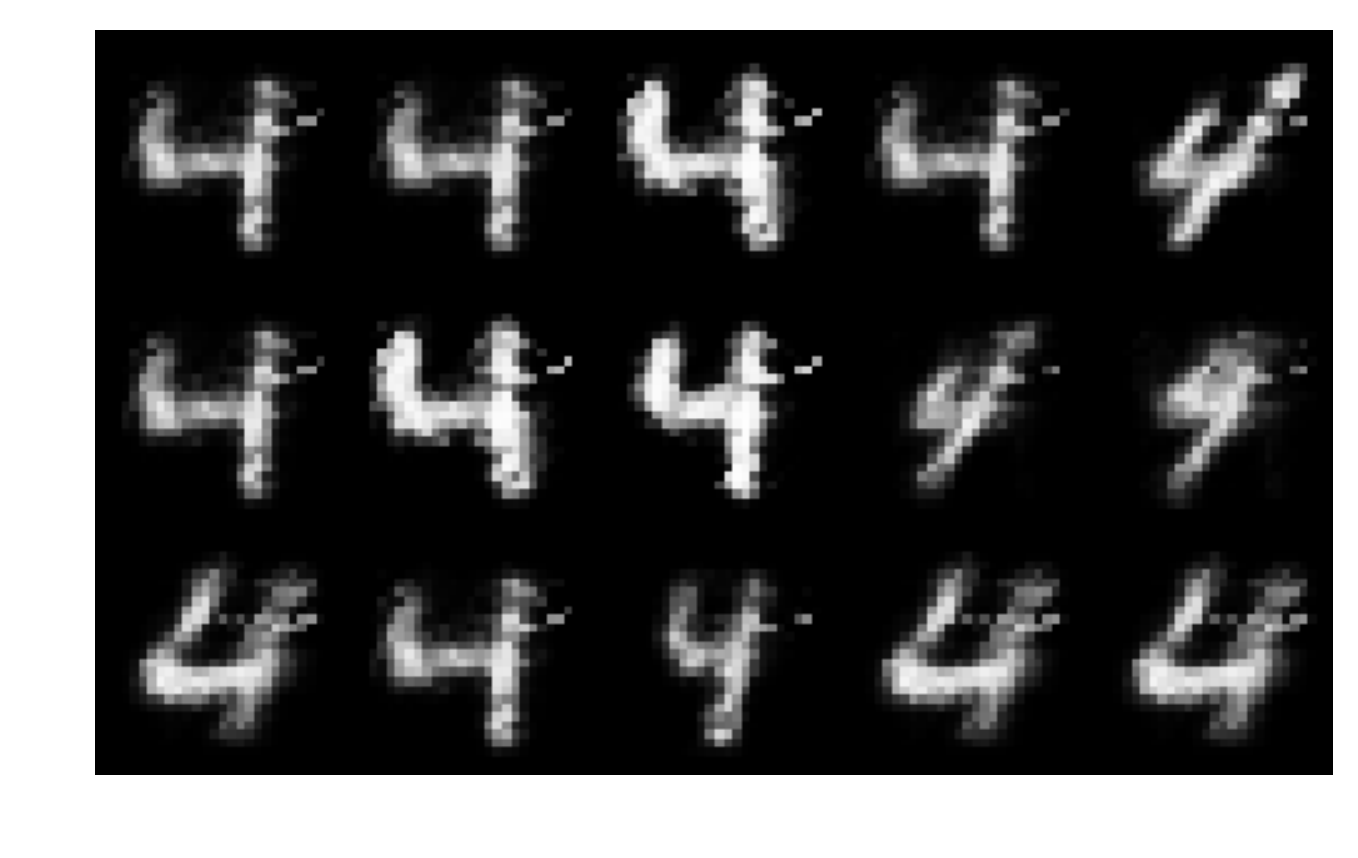}}&\makecell{\includegraphics[scale=0.12]{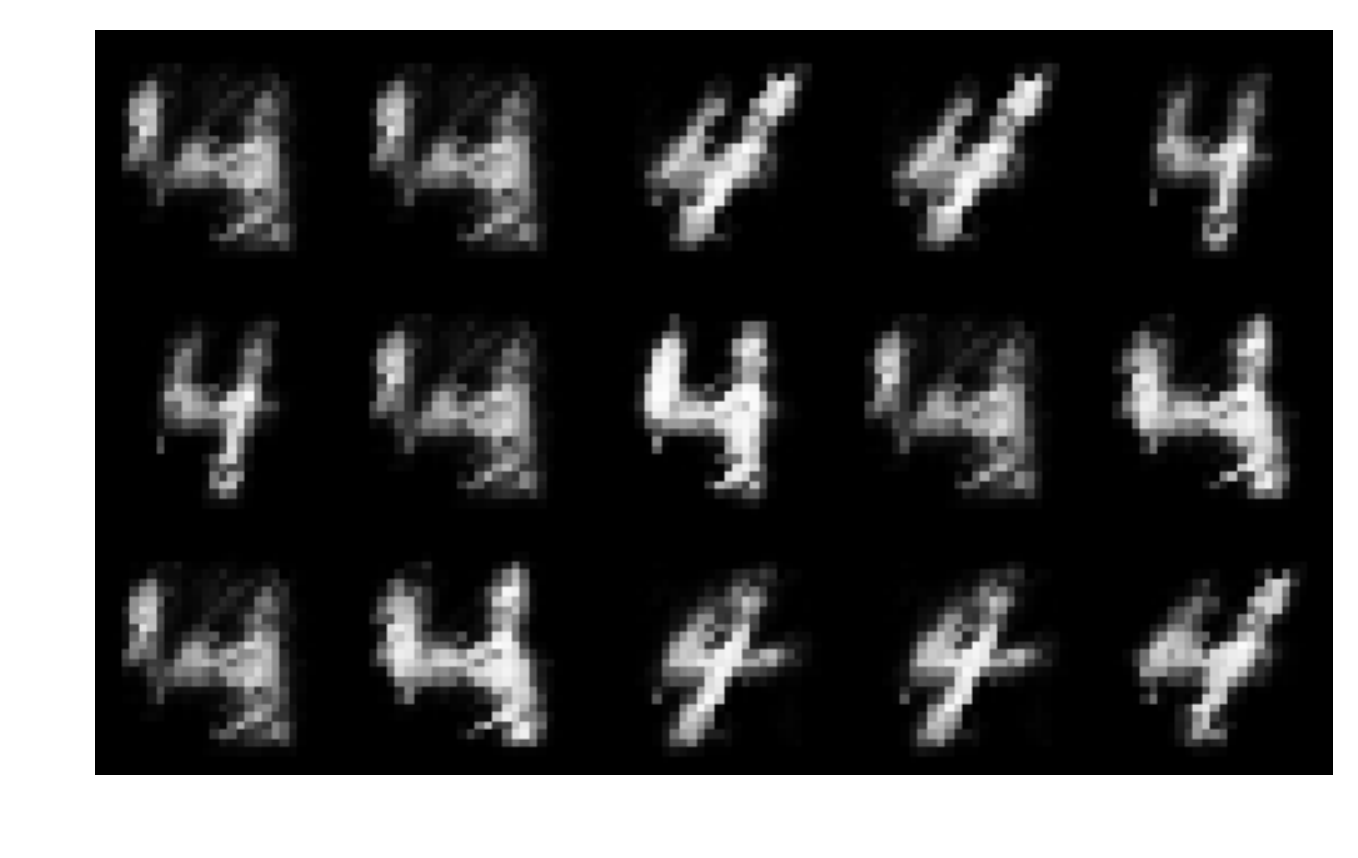}}&\makecell{\includegraphics[scale=0.12]{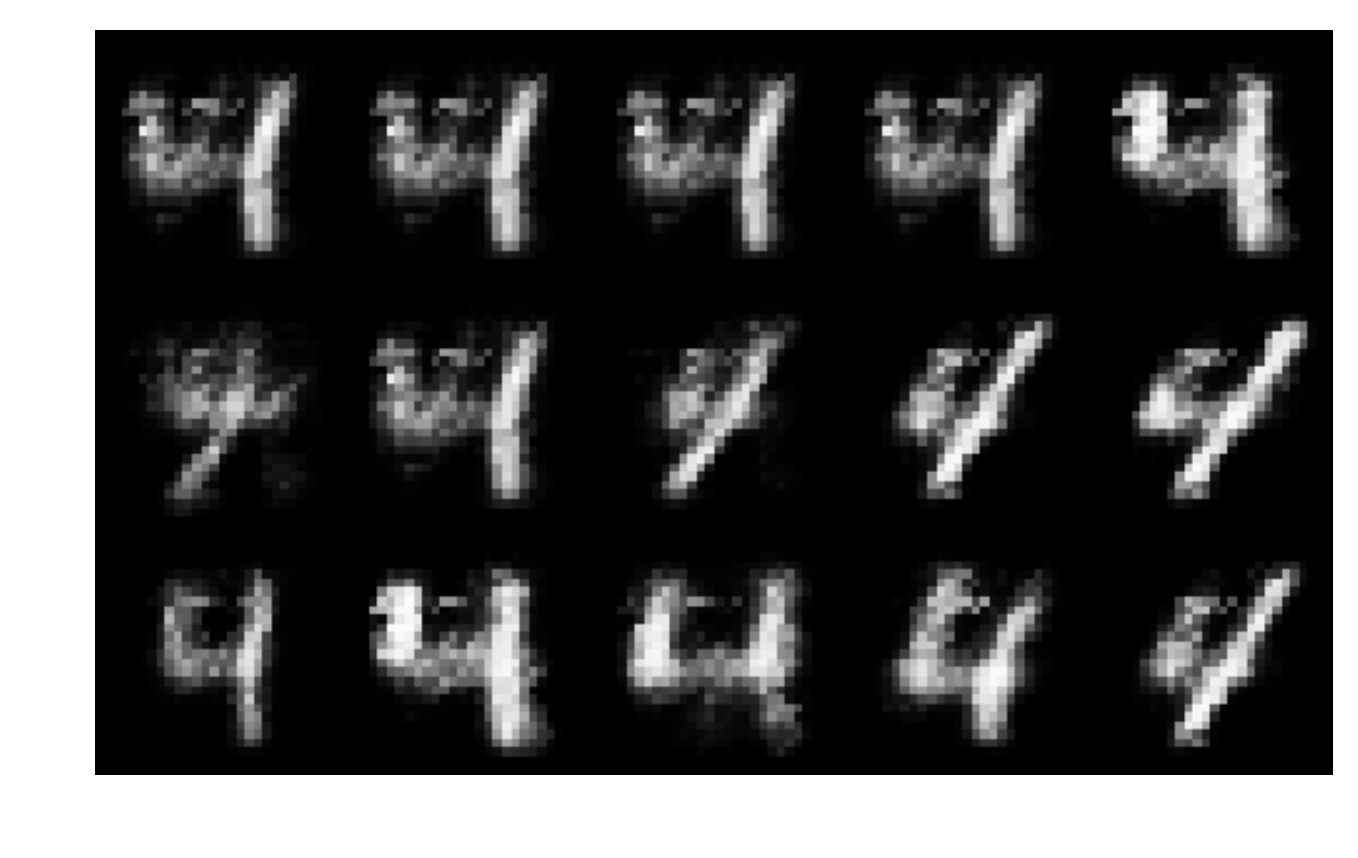}}&\makecell{\includegraphics[scale=0.12]{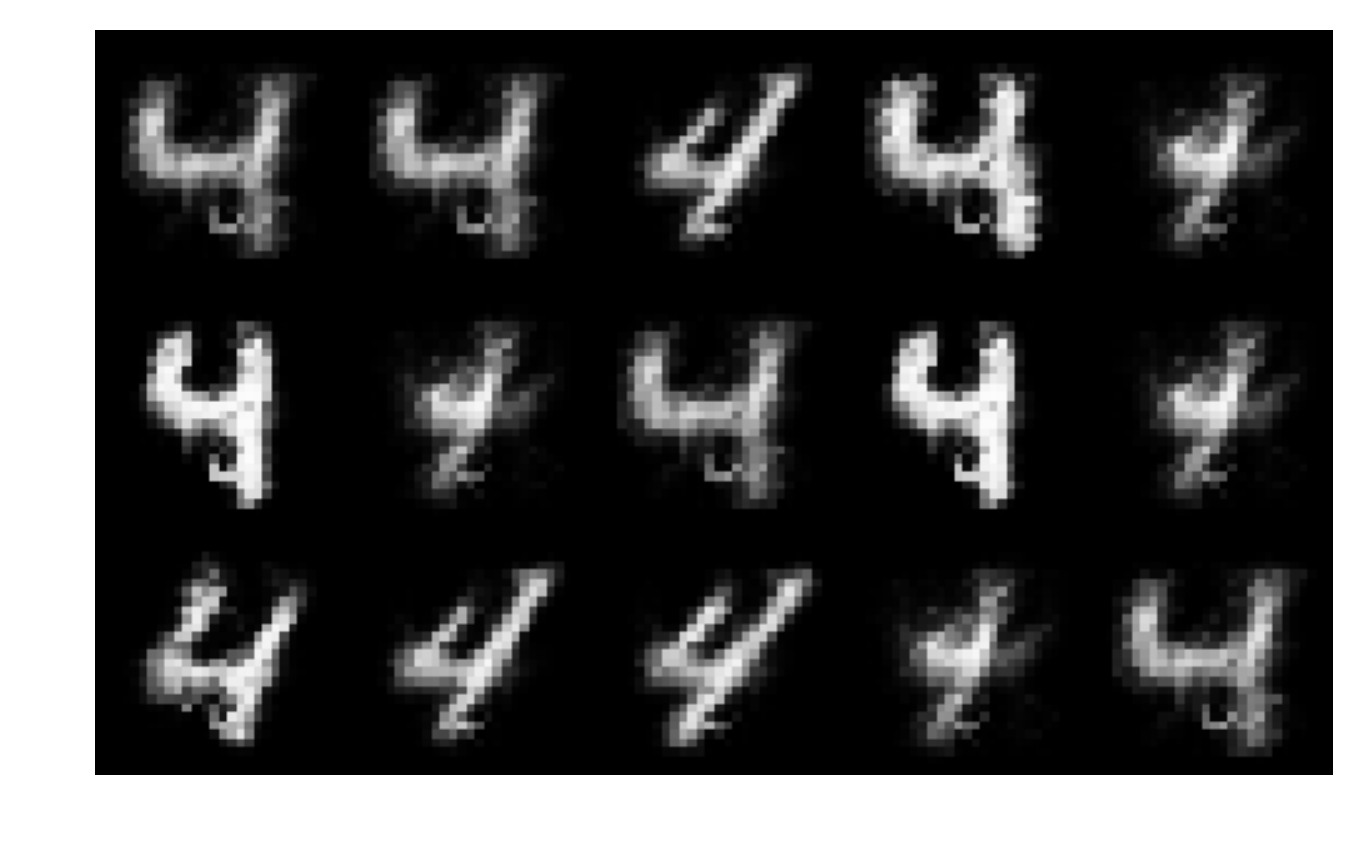}}&\makecell{\includegraphics[scale=0.12]{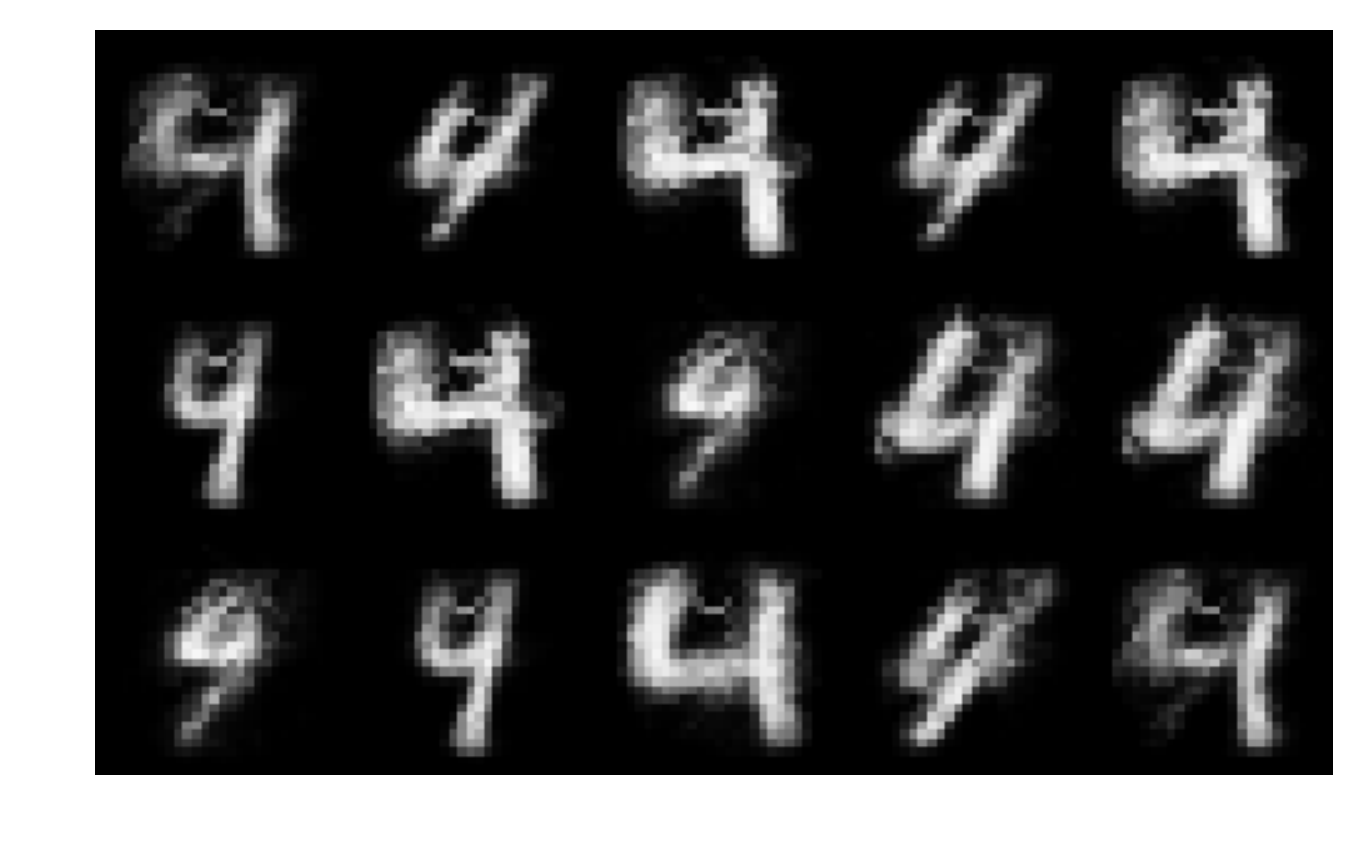}}\\ \hline
`5'&\makecell{\includegraphics[scale=0.12]{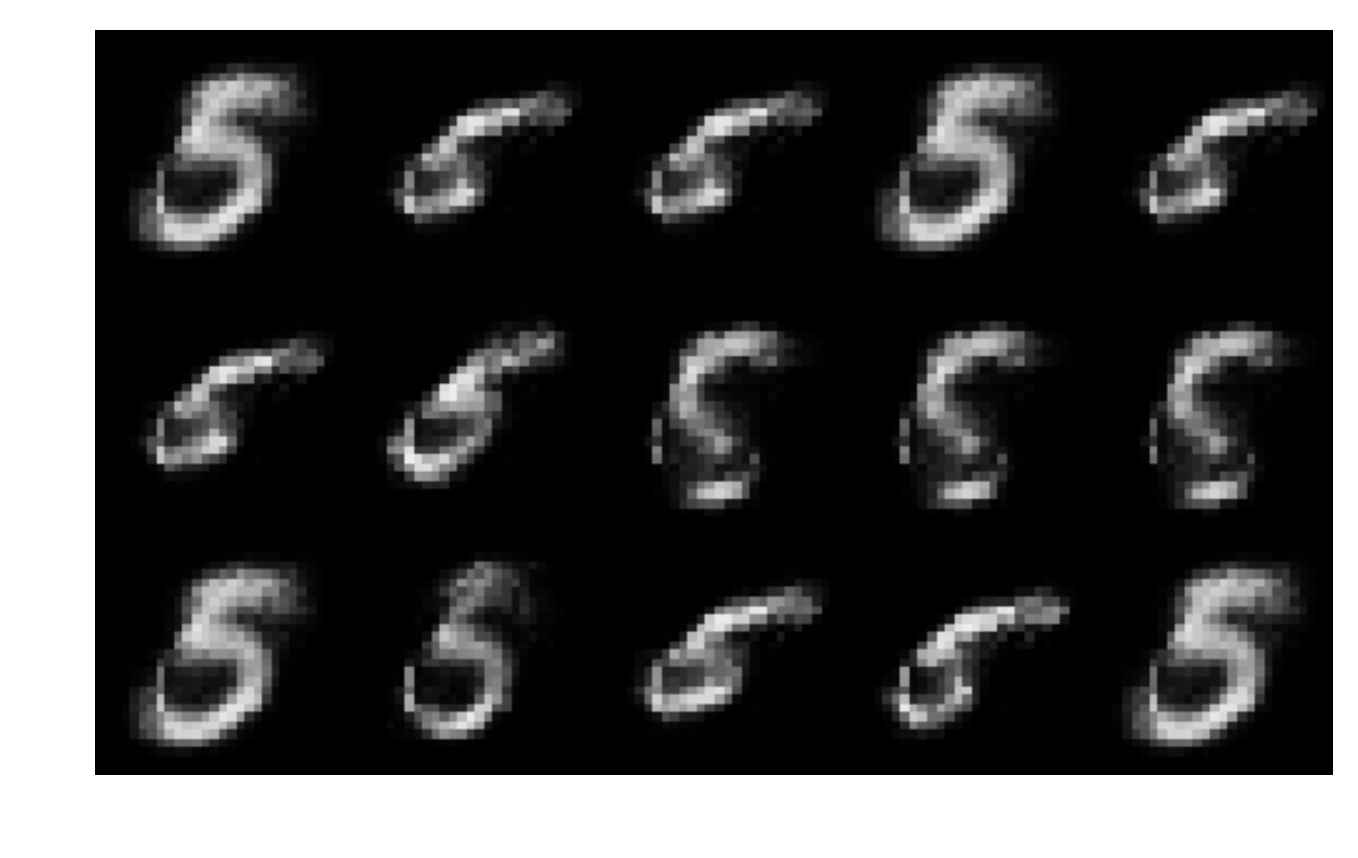}}&\makecell{\includegraphics[scale=0.12]{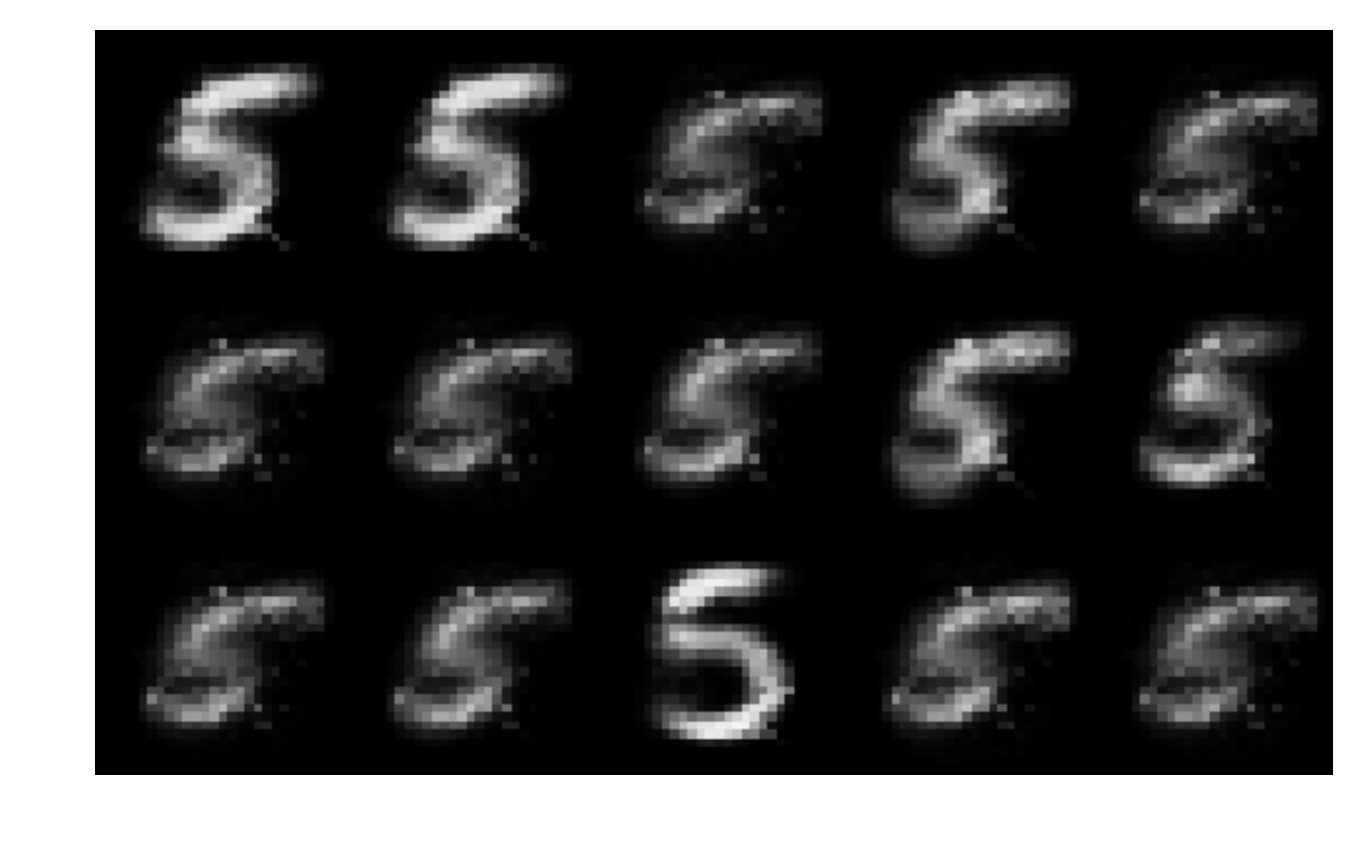}}&\makecell{\includegraphics[scale=0.12]{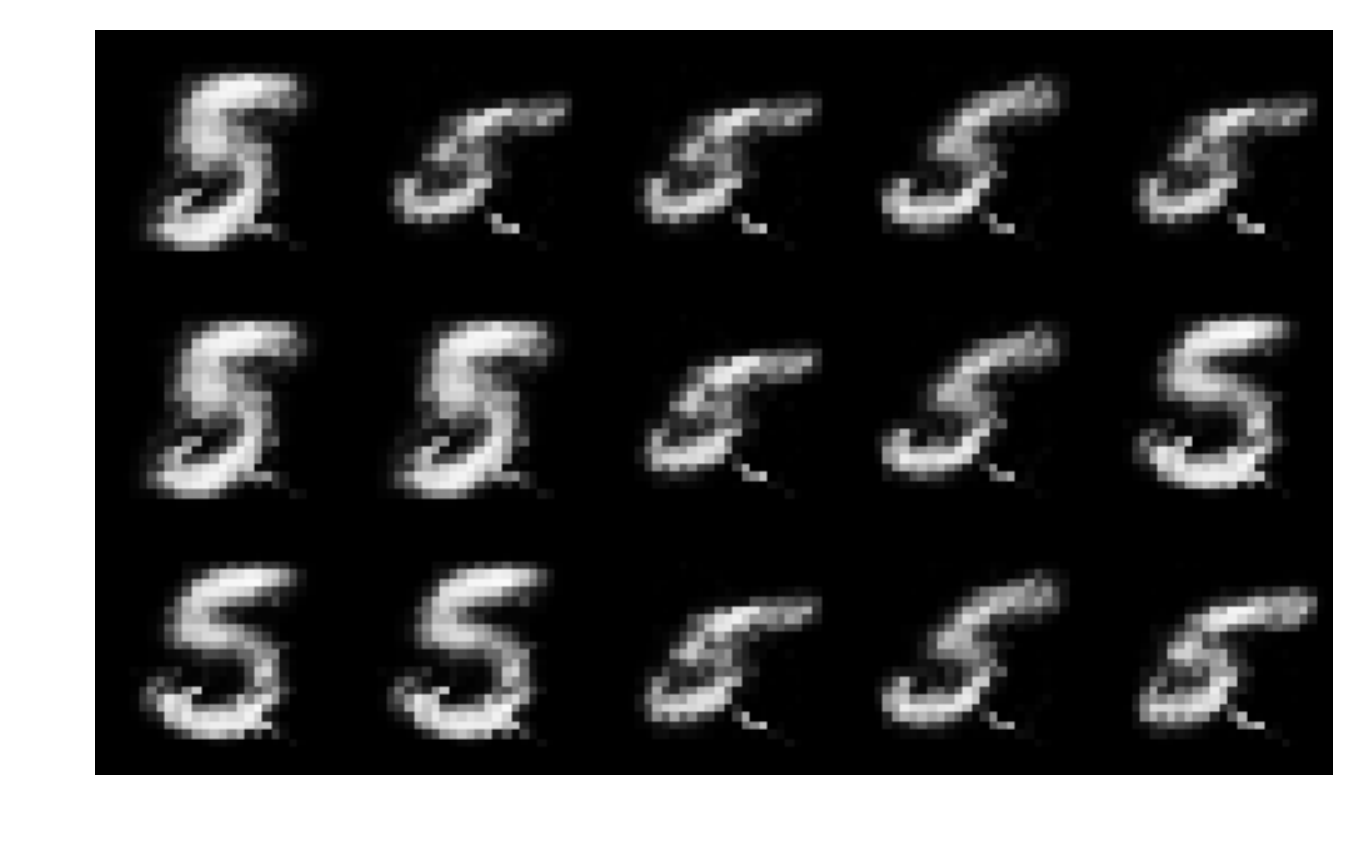}}&\makecell{\includegraphics[scale=0.12]{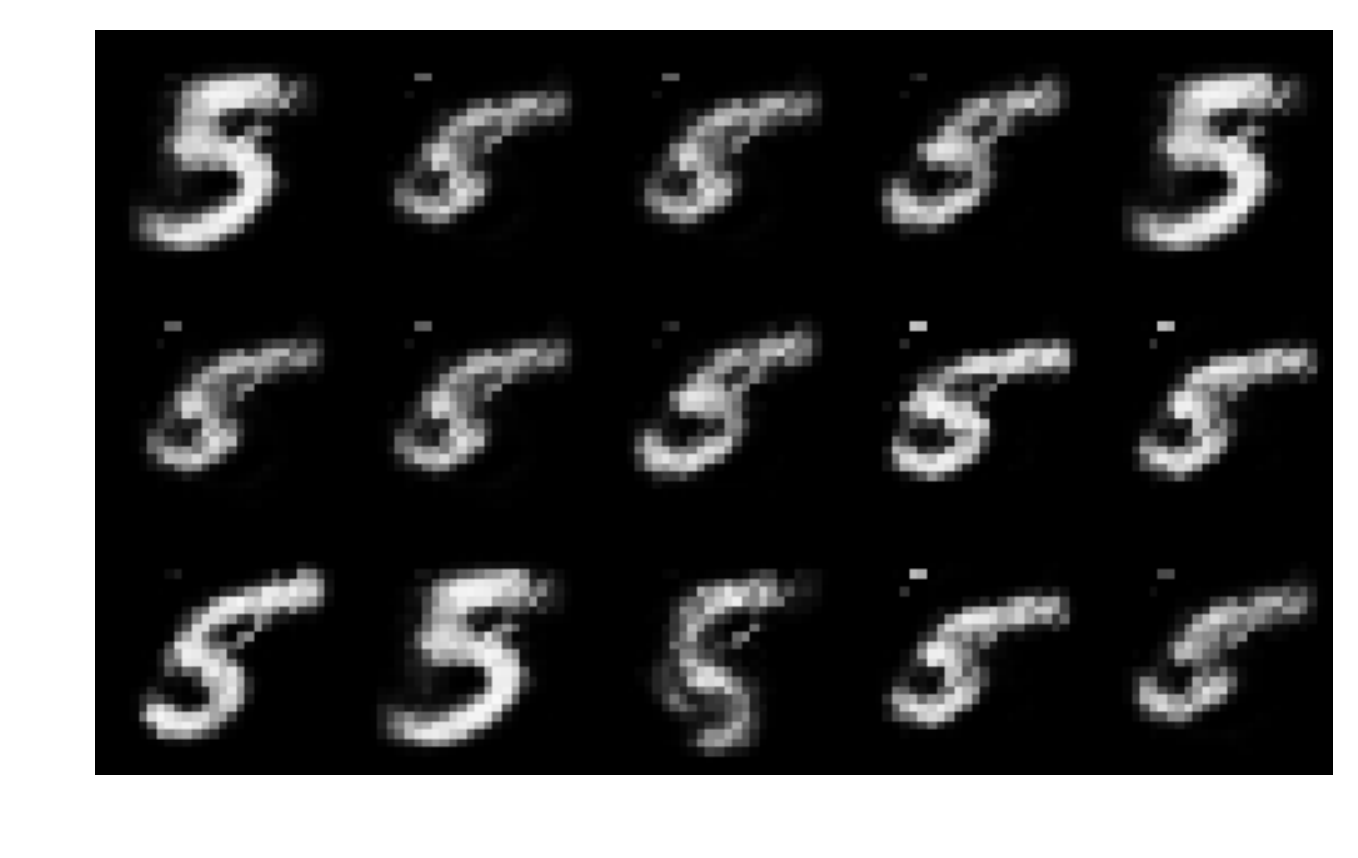}}&\makecell{\includegraphics[scale=0.12]{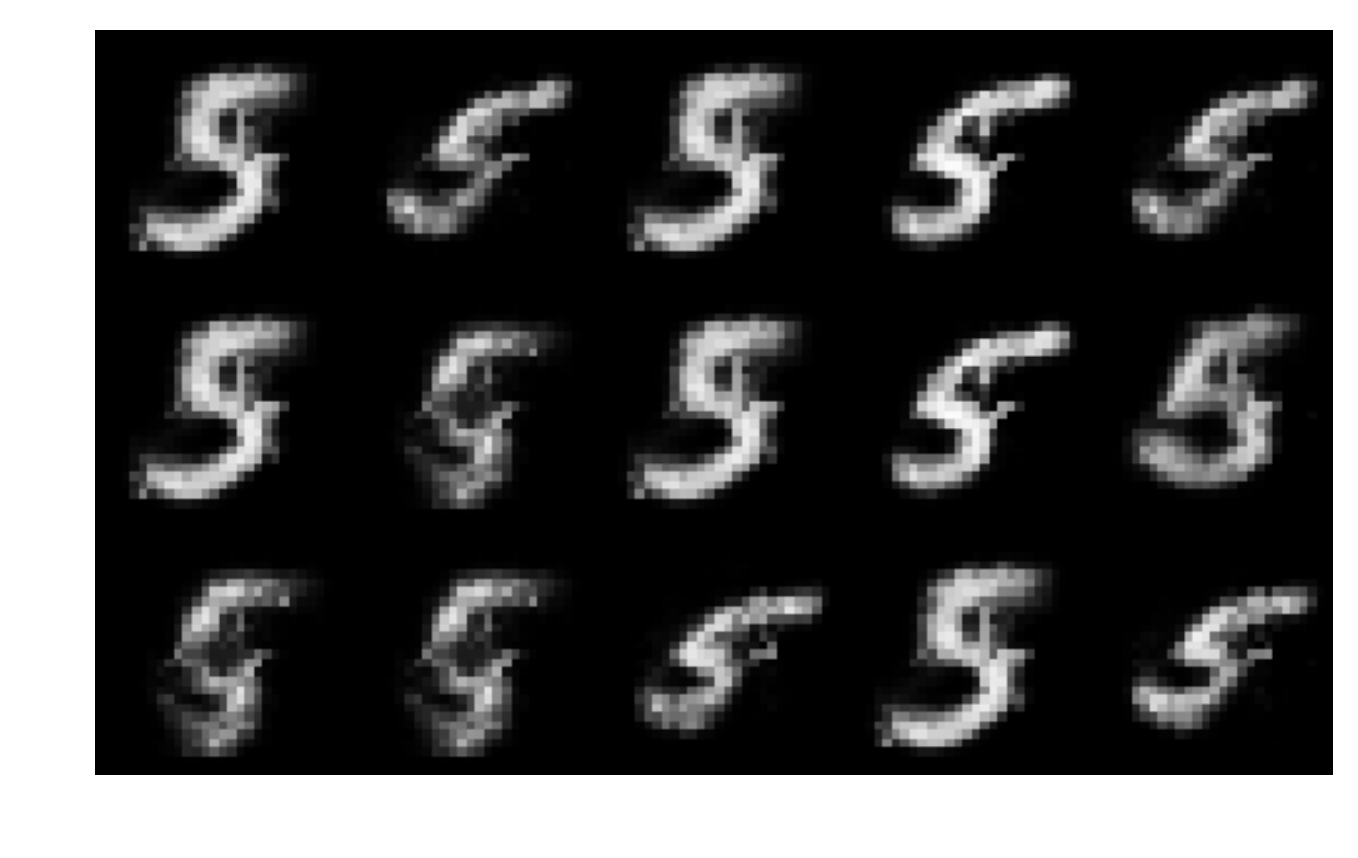}}&--&\makecell{\includegraphics[scale=0.12]{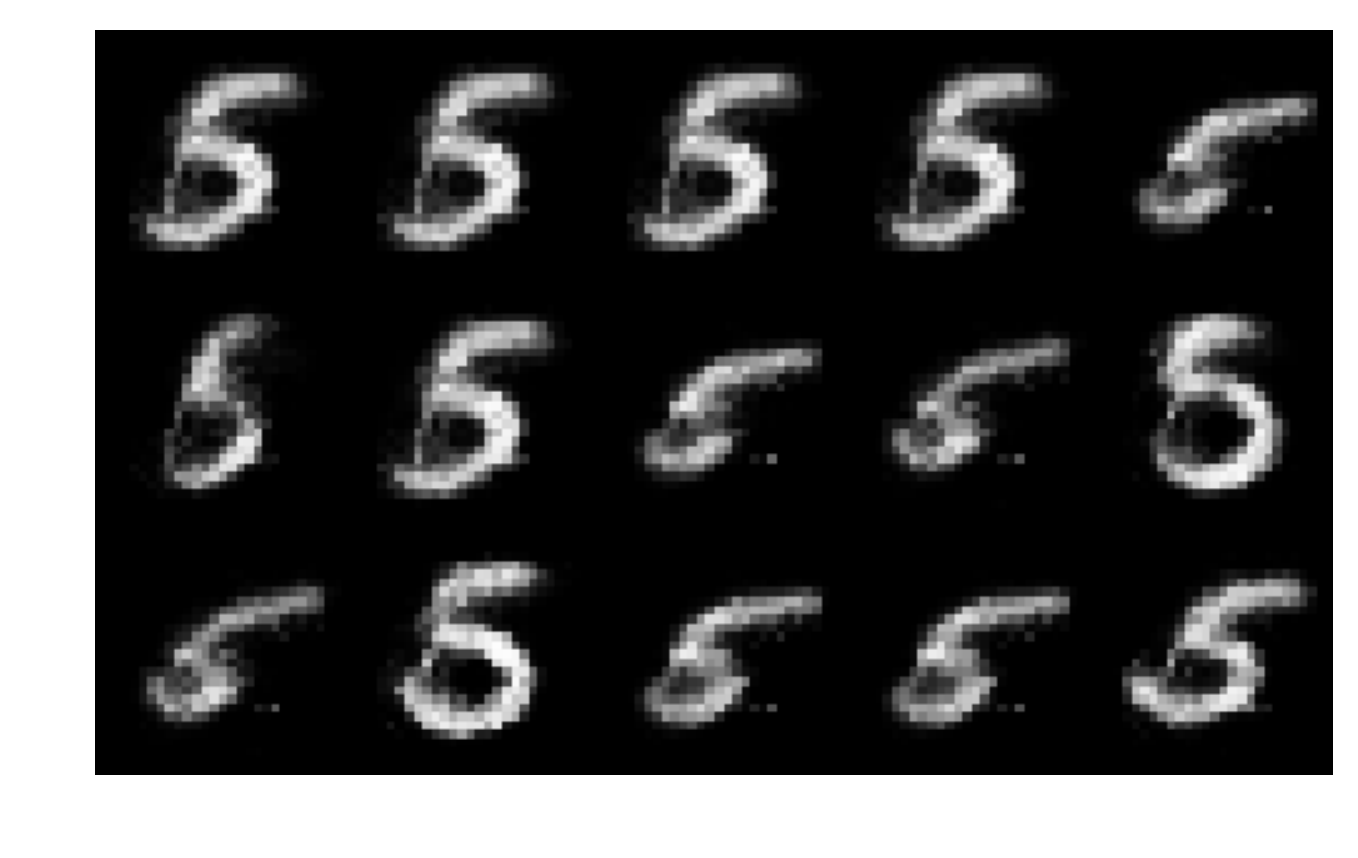}}&\makecell{\includegraphics[scale=0.12]{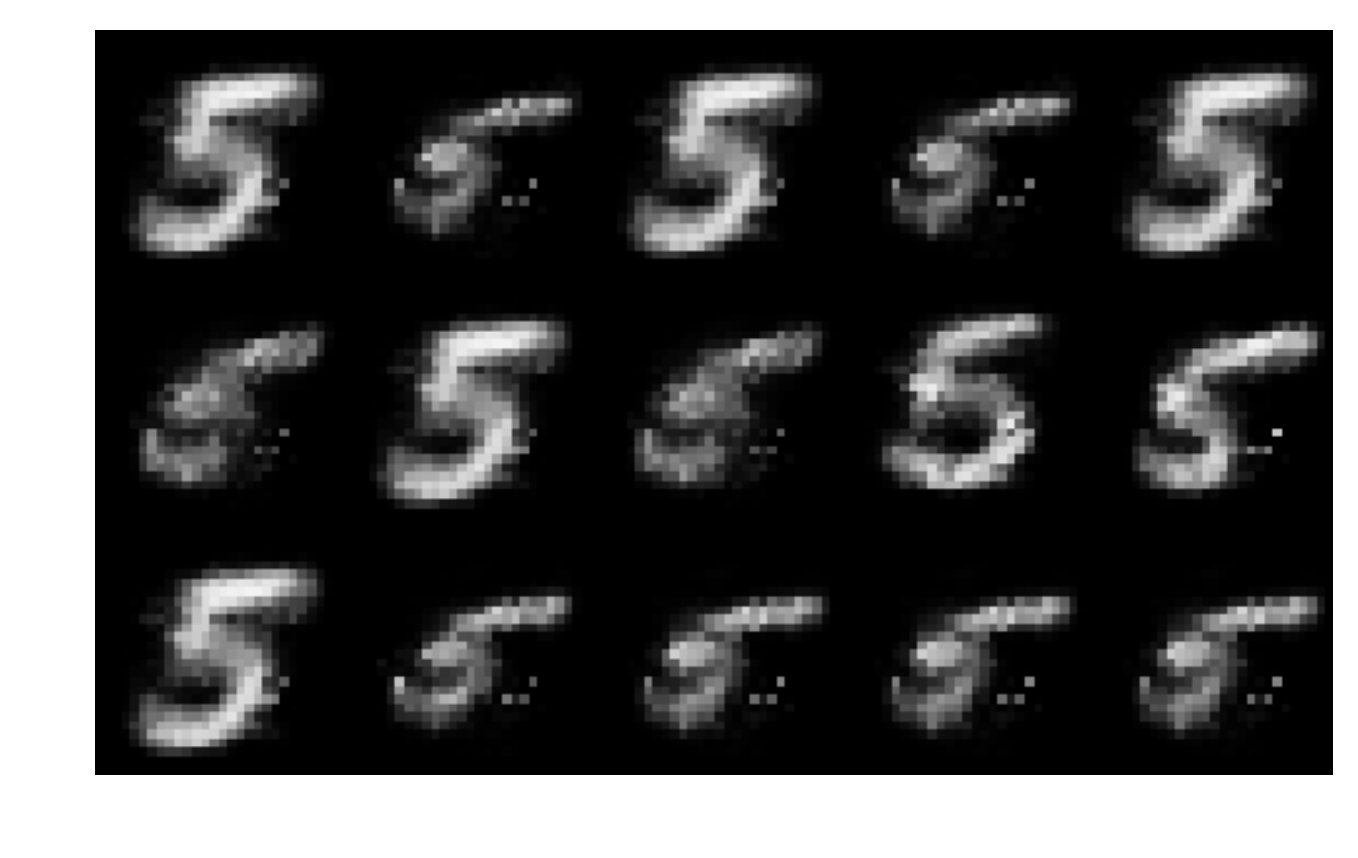}}&\makecell{\includegraphics[scale=0.12]{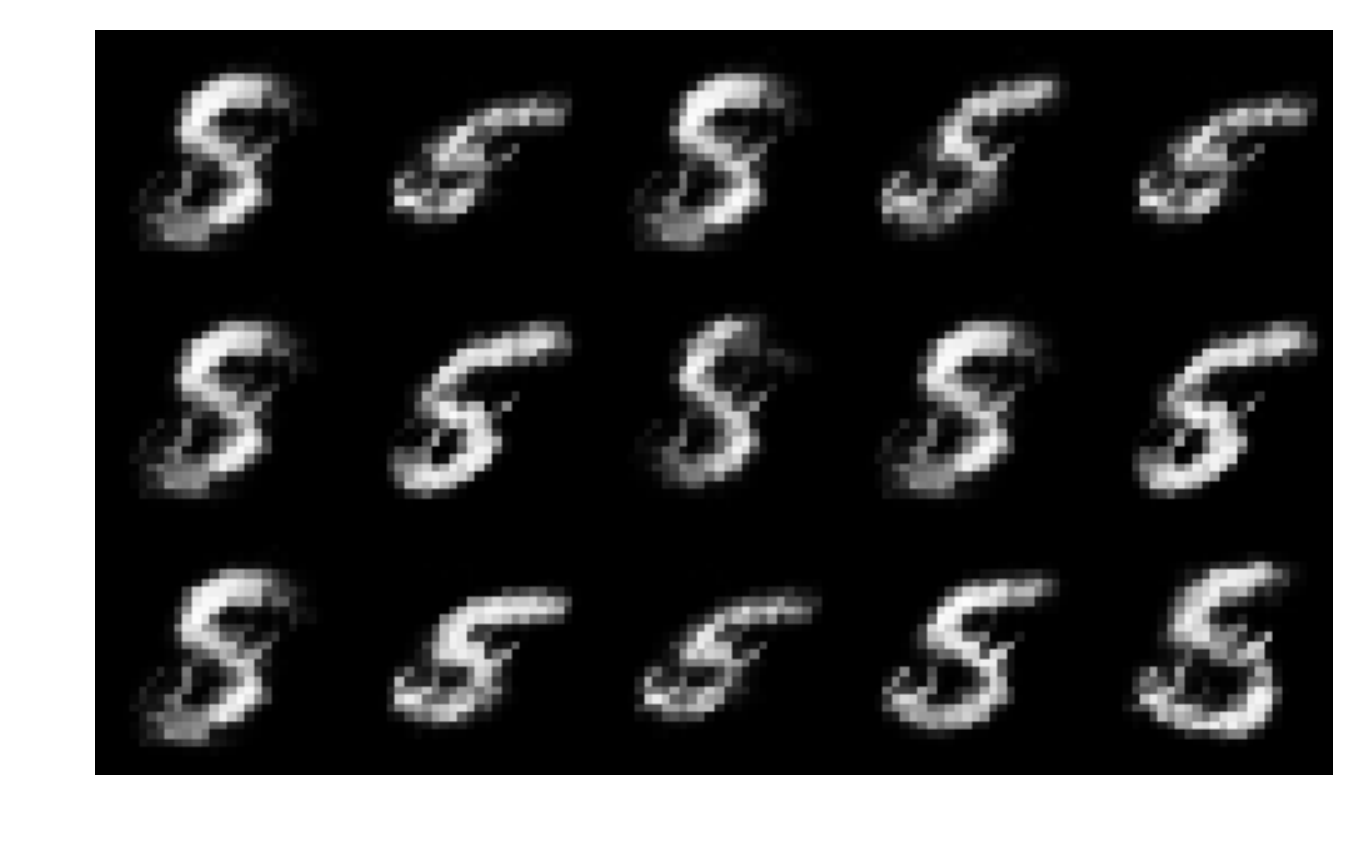}}&\makecell{\includegraphics[scale=0.12]{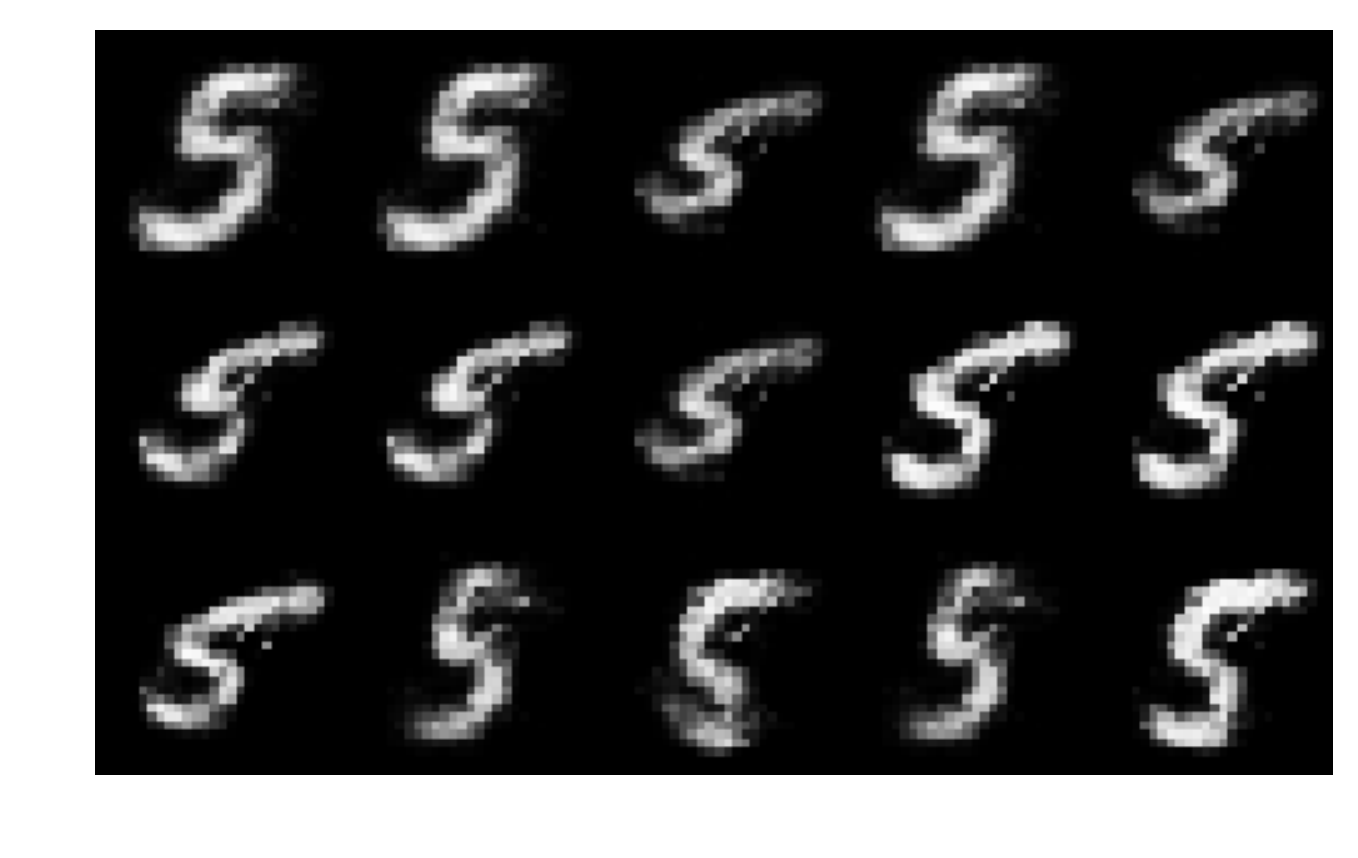}}\\ \hline
`6'&\makecell{\includegraphics[scale=0.12]{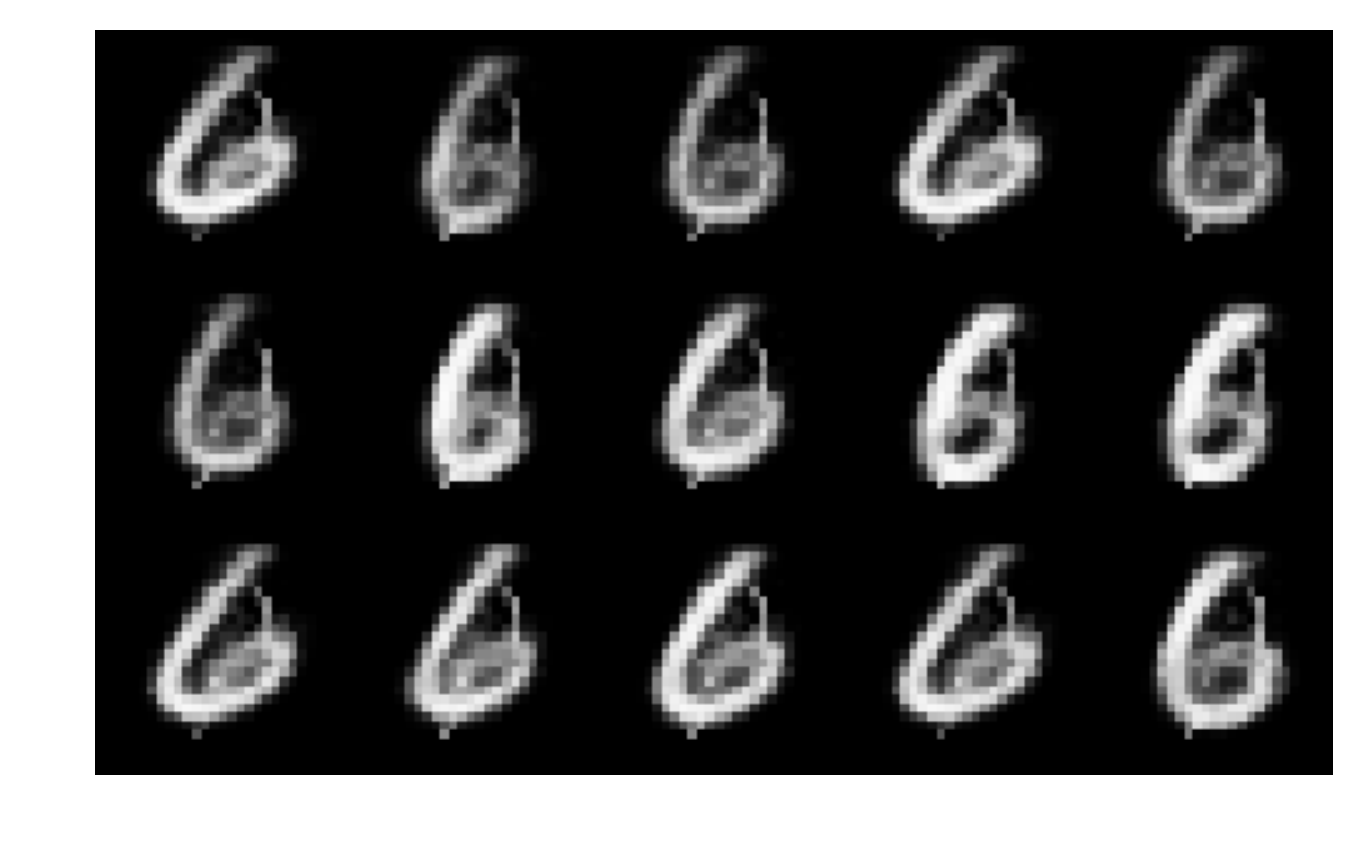}}&\makecell{\includegraphics[scale=0.12]{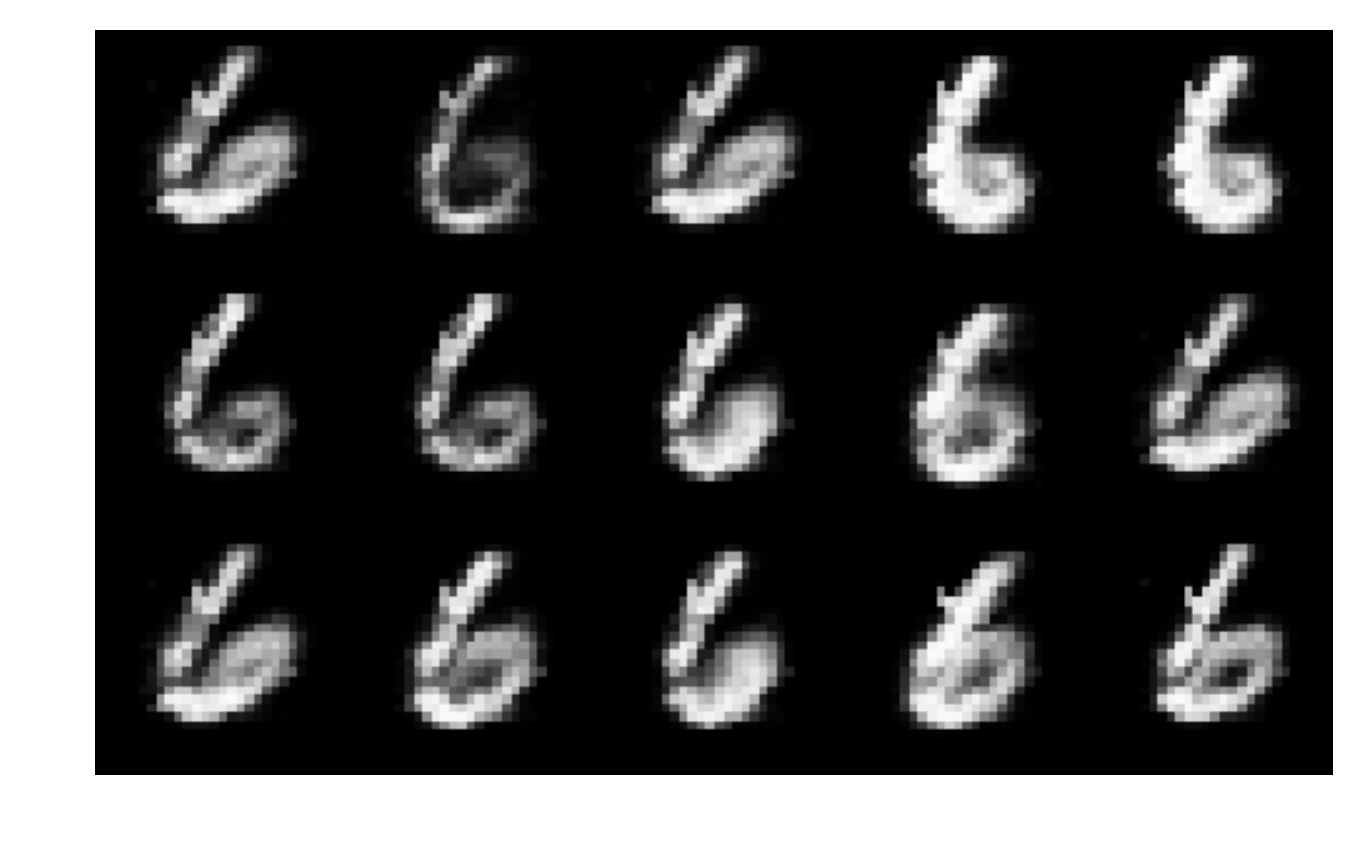}}&\makecell{\includegraphics[scale=0.12]{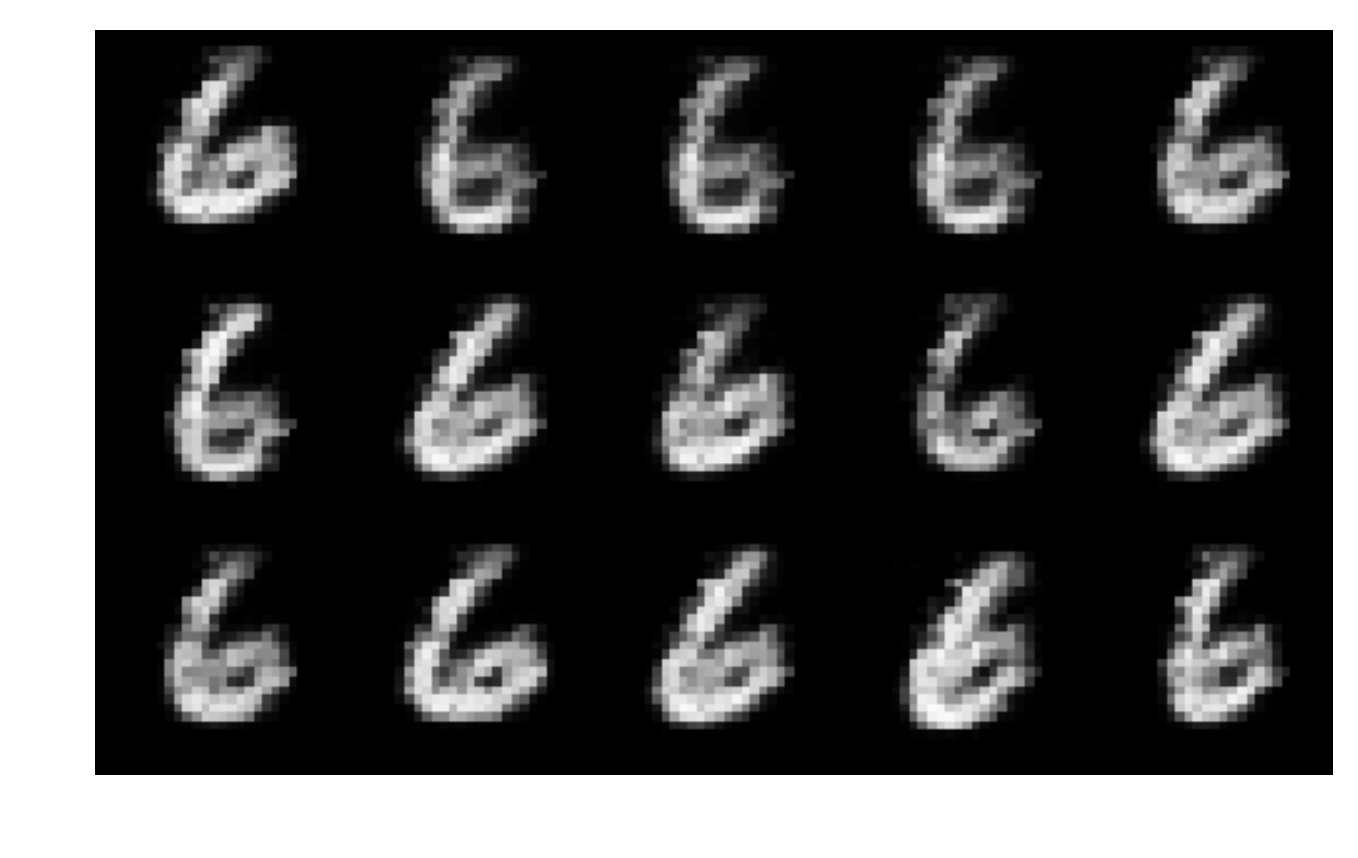}}&\makecell{\includegraphics[scale=0.12]{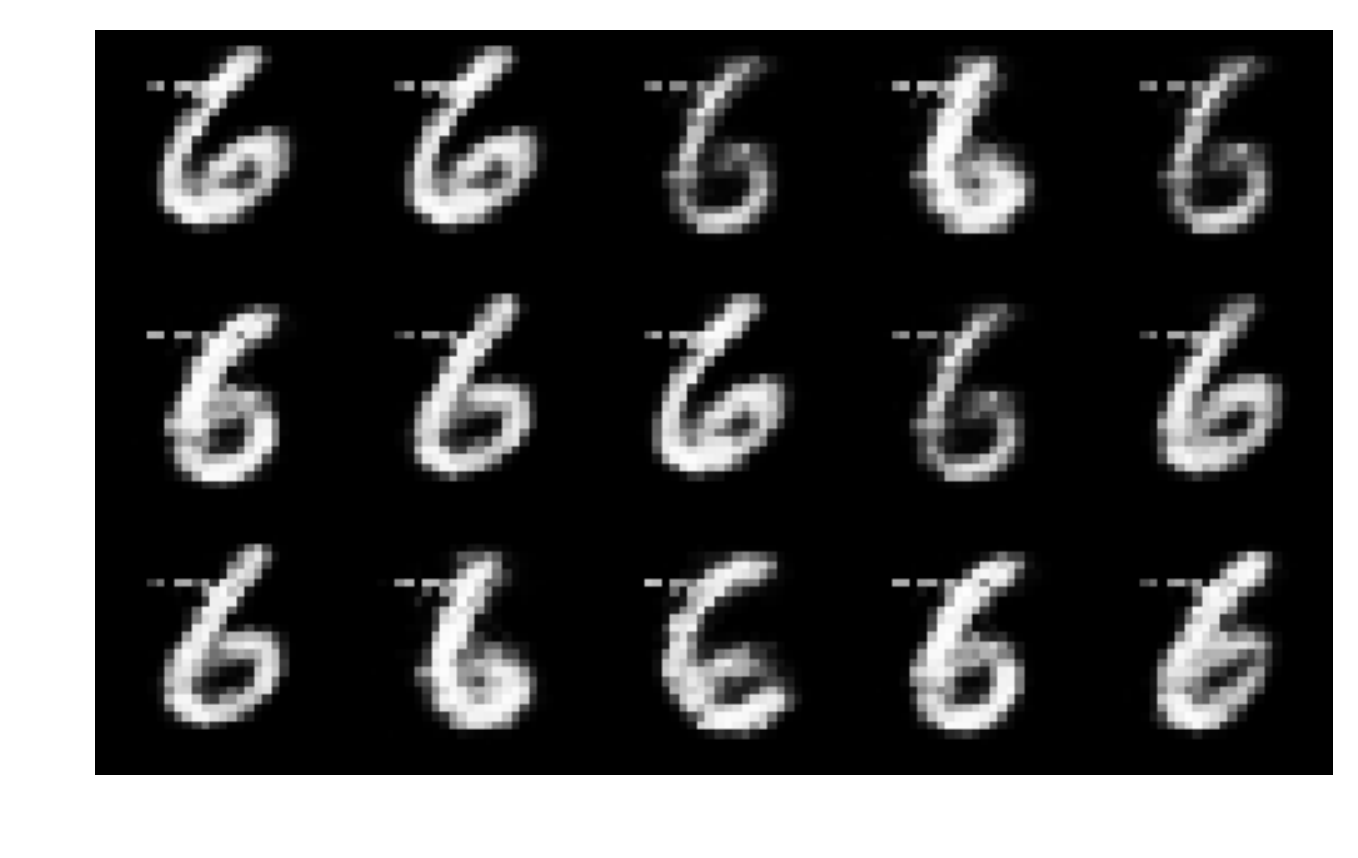}}&\makecell{\includegraphics[scale=0.12]{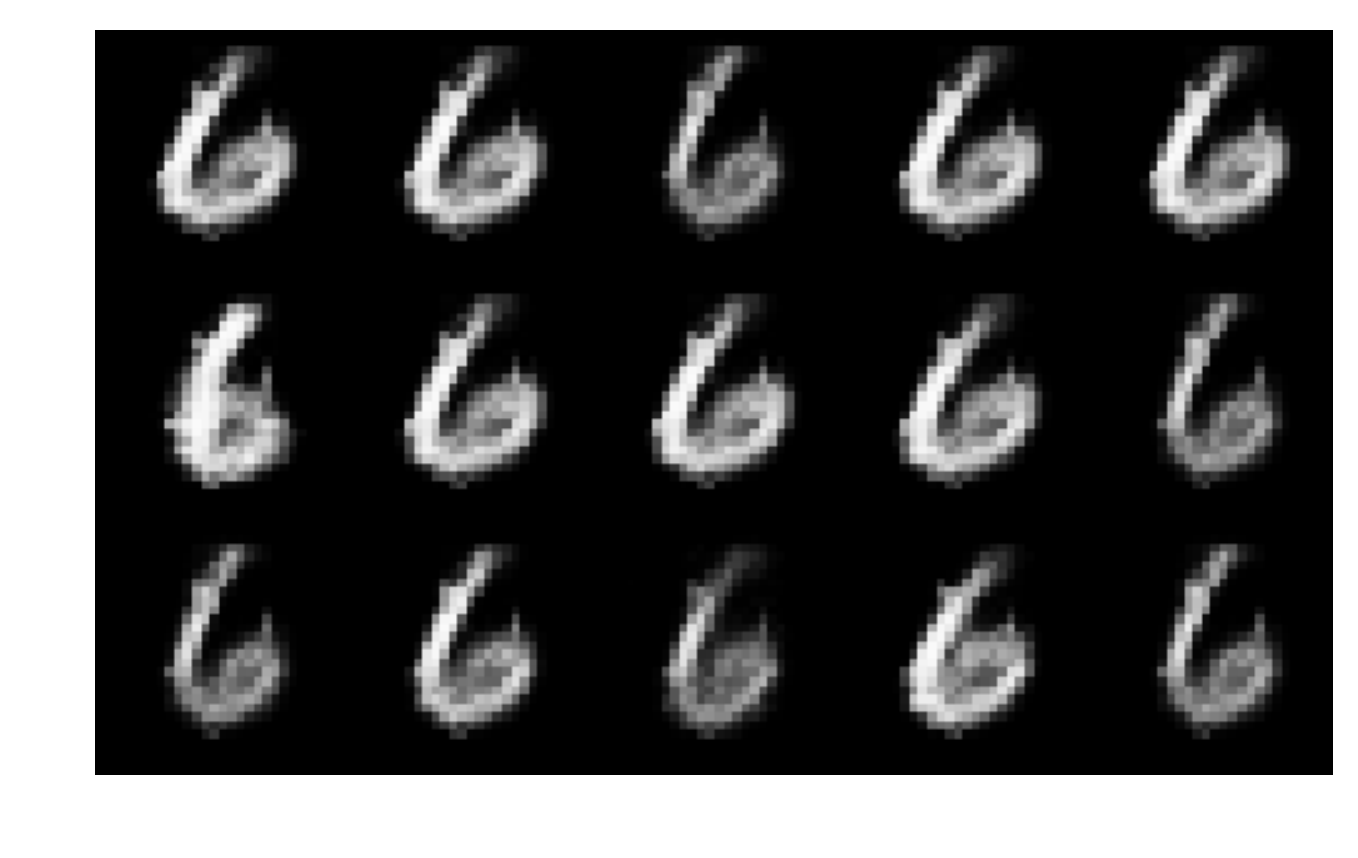}}&\makecell{\includegraphics[scale=0.12]{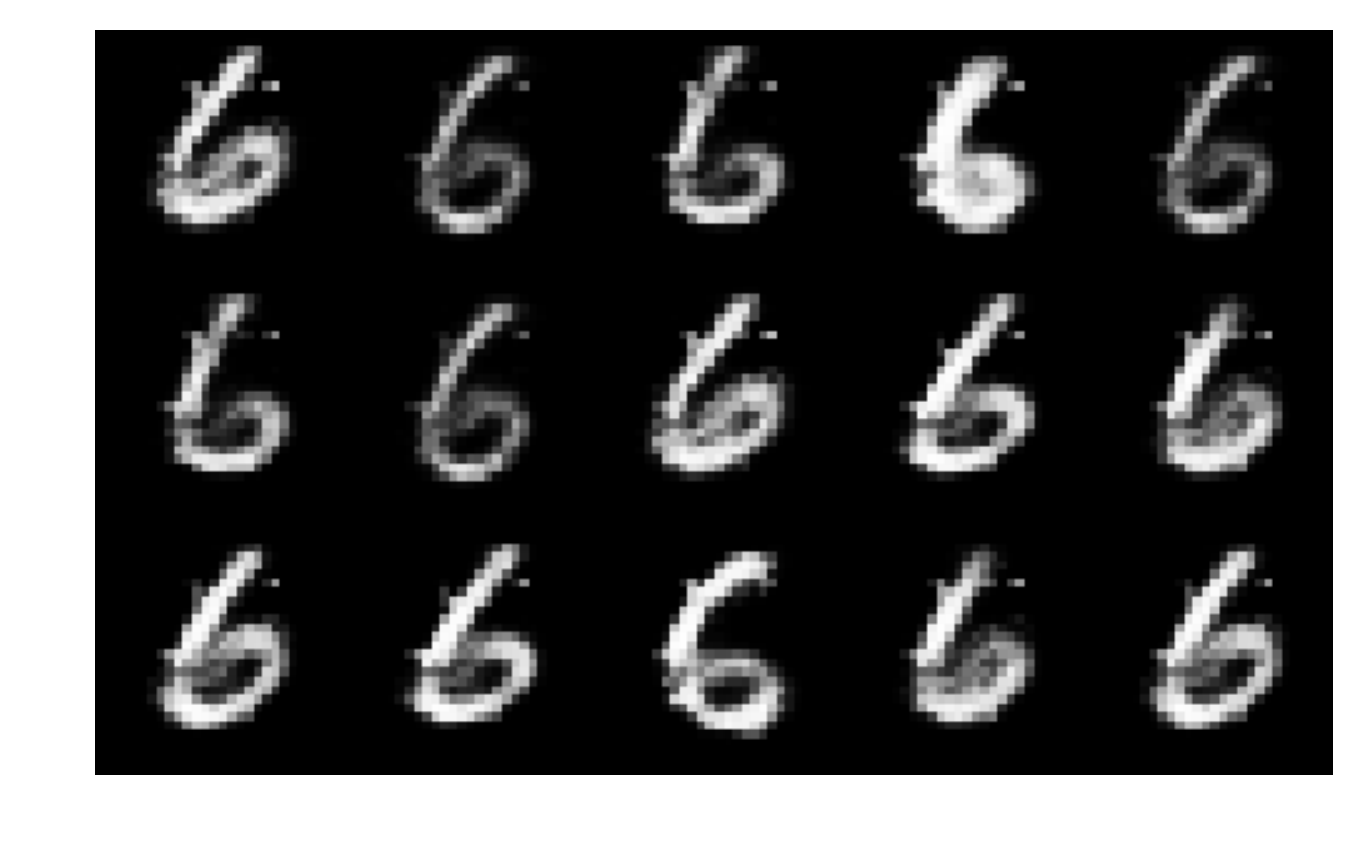}}&--&\makecell{\includegraphics[scale=0.12]{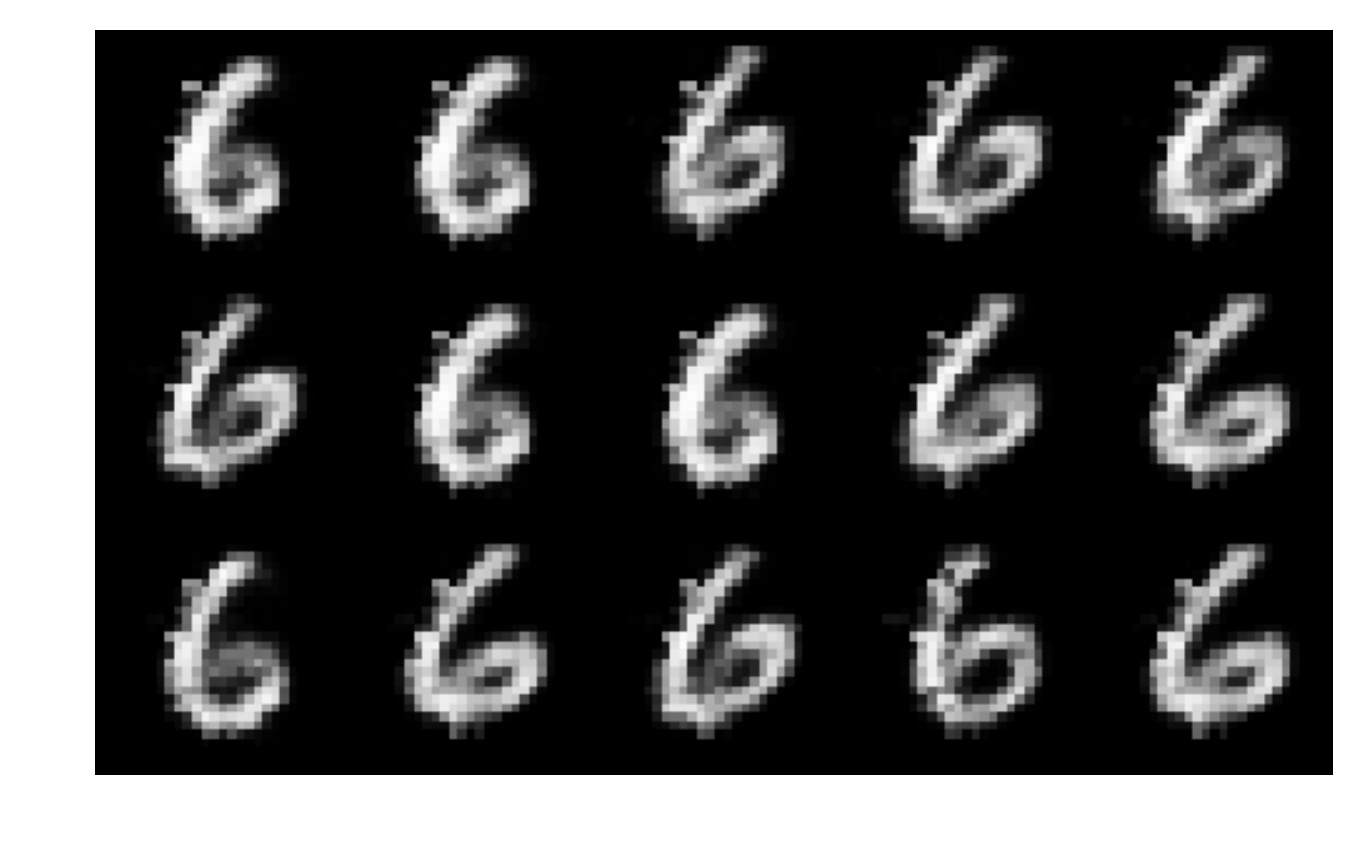}}&\makecell{\includegraphics[scale=0.12]{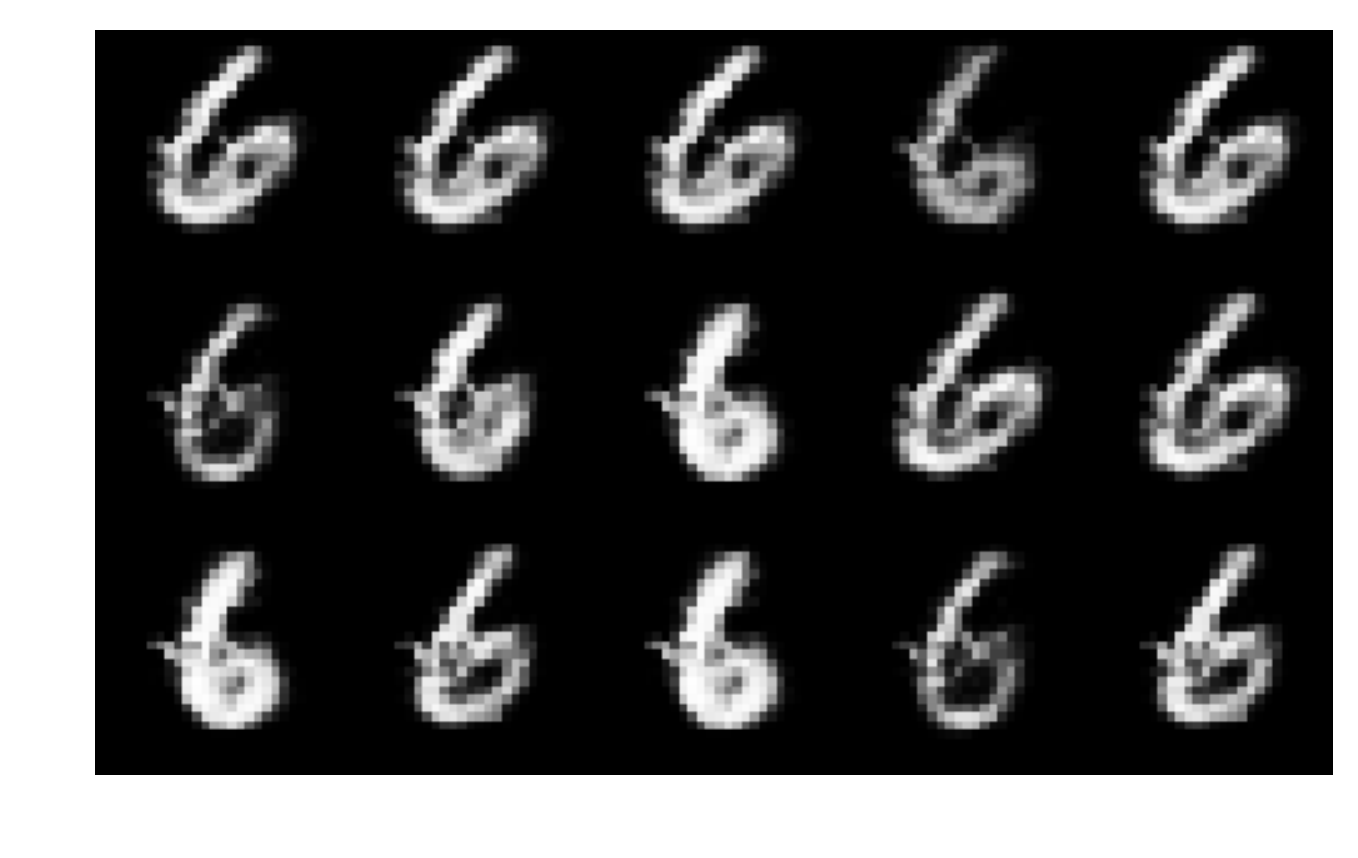}}&\makecell{\includegraphics[scale=0.12]{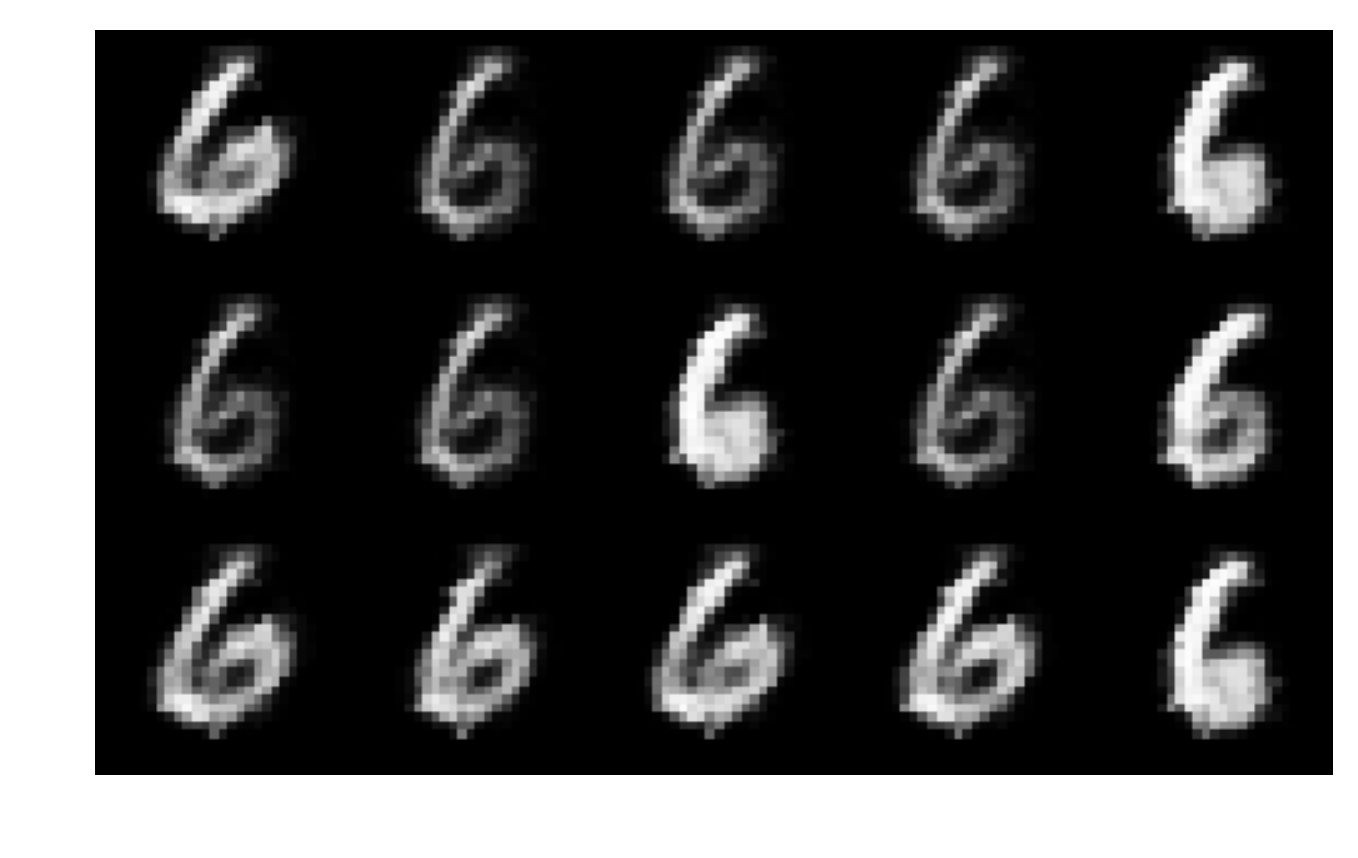}}\\ \hline
`7'&\makecell{\includegraphics[scale=0.12]{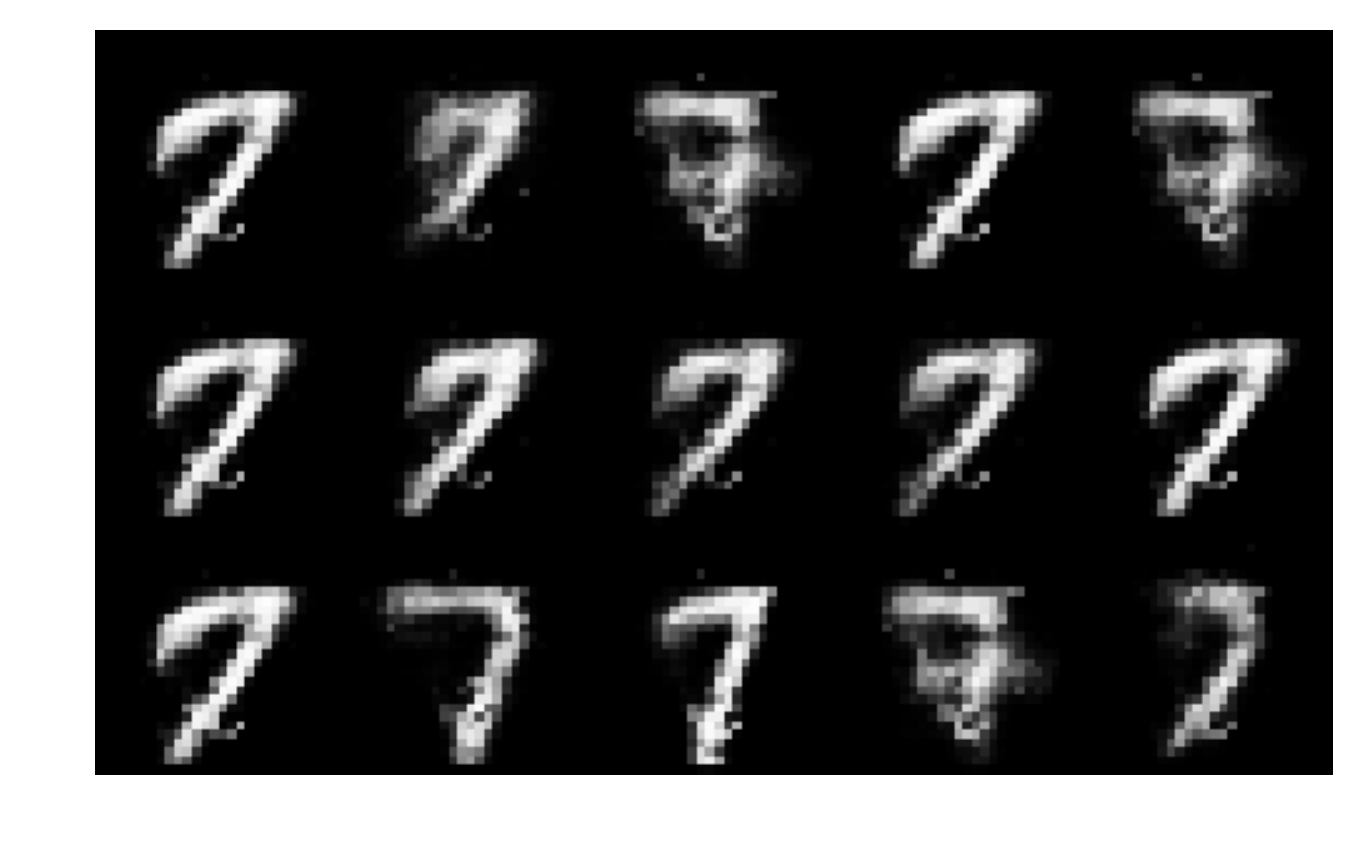}}&\makecell{\includegraphics[scale=0.12]{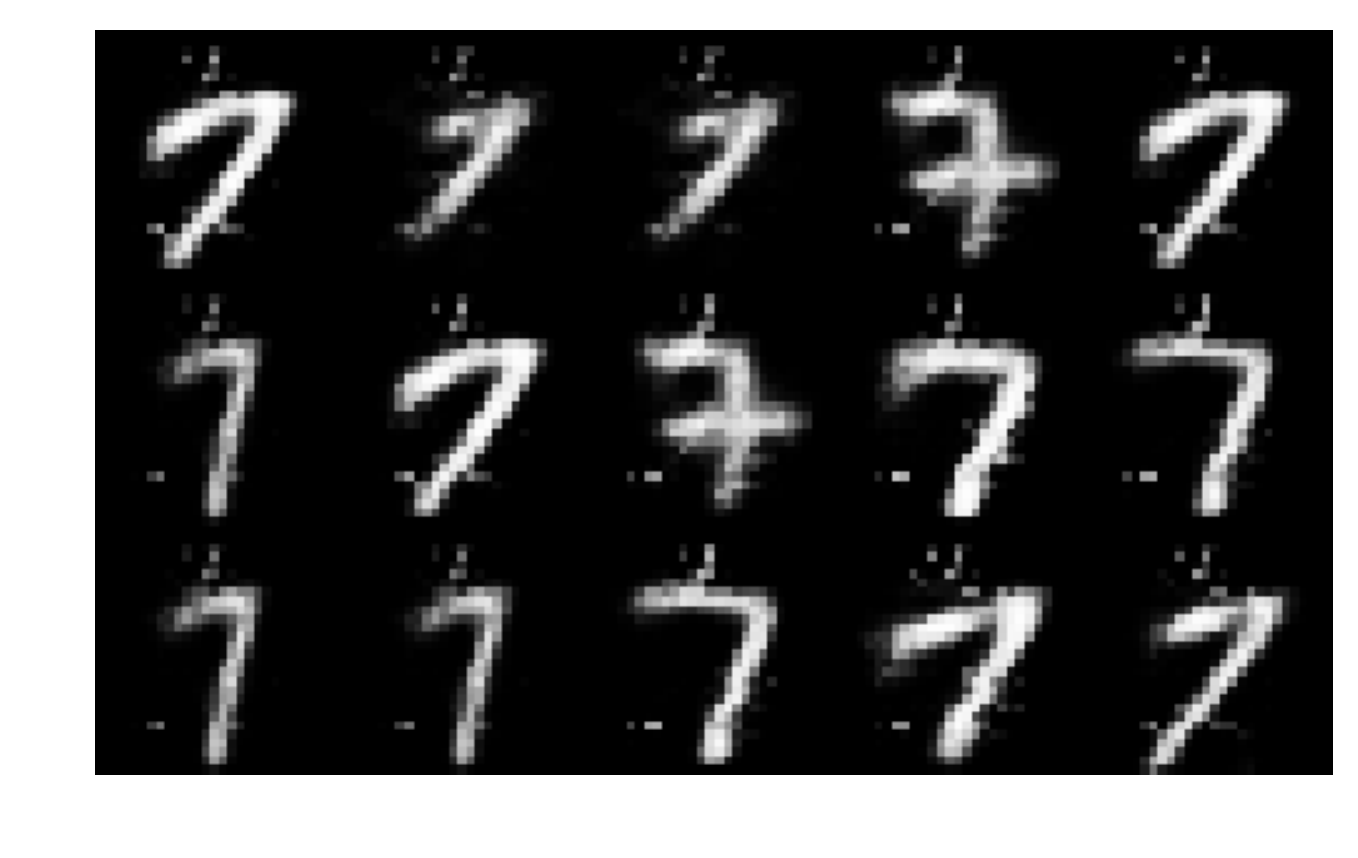}}&\makecell{\includegraphics[scale=0.12]{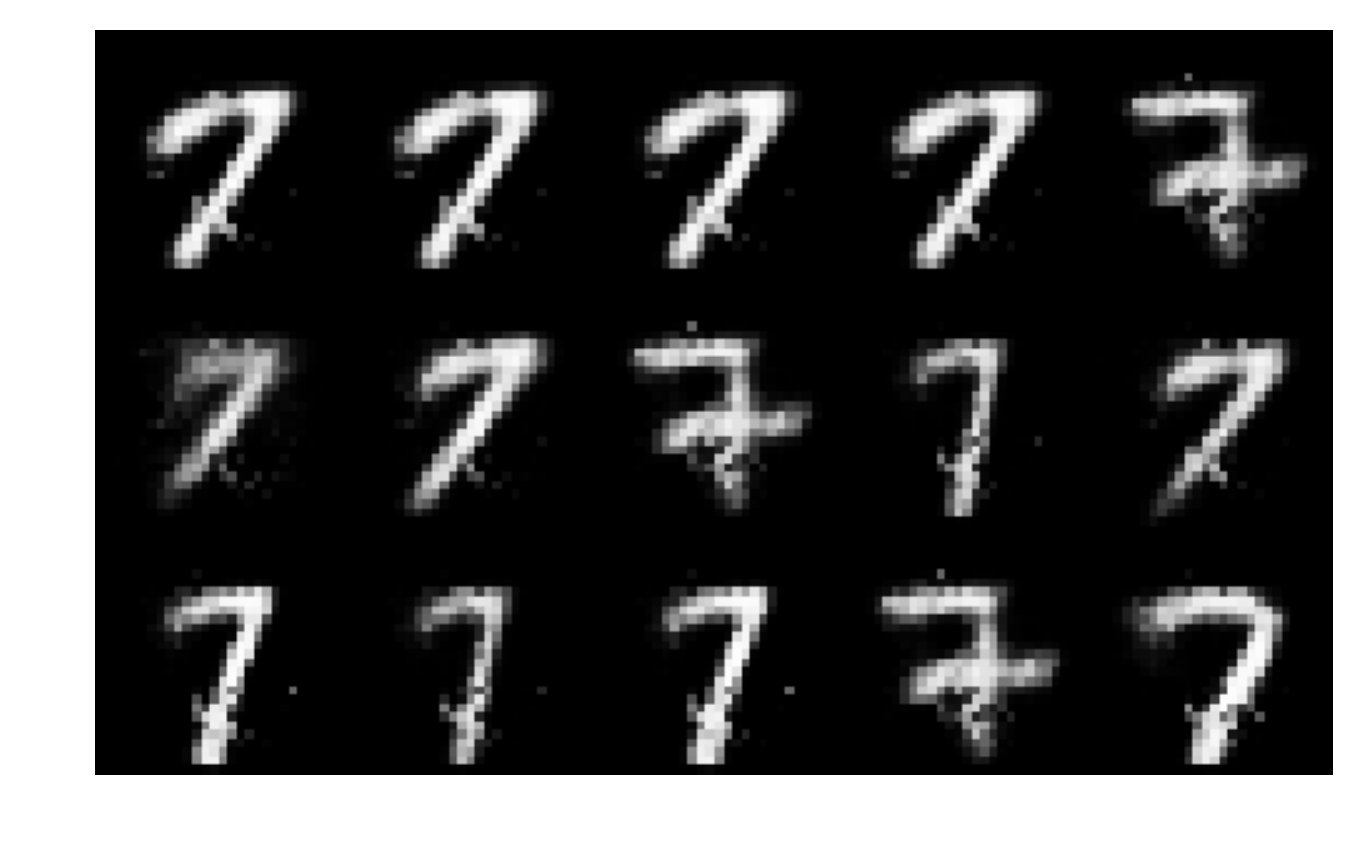}}&\makecell{\includegraphics[scale=0.12]{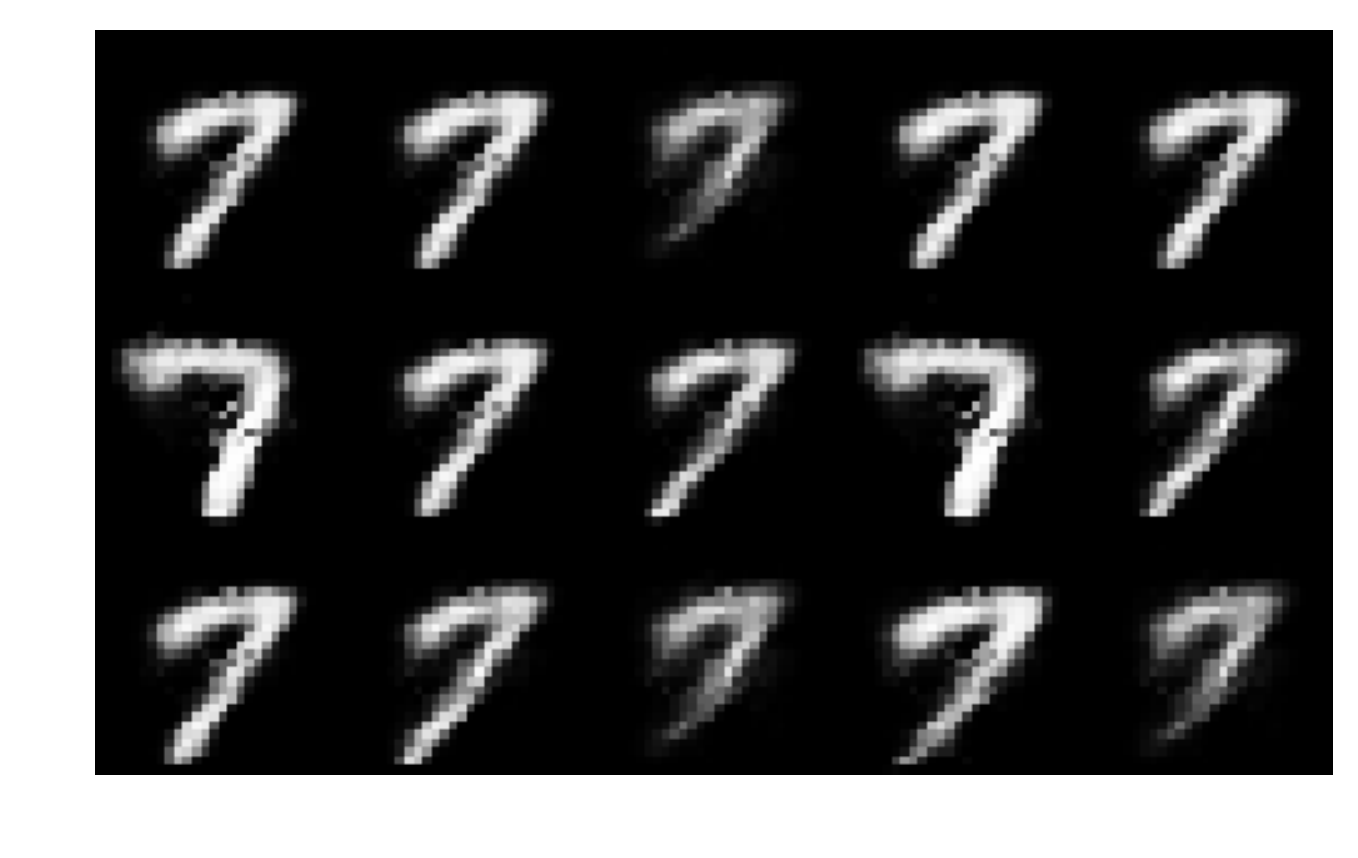}}&\makecell{\includegraphics[scale=0.12]{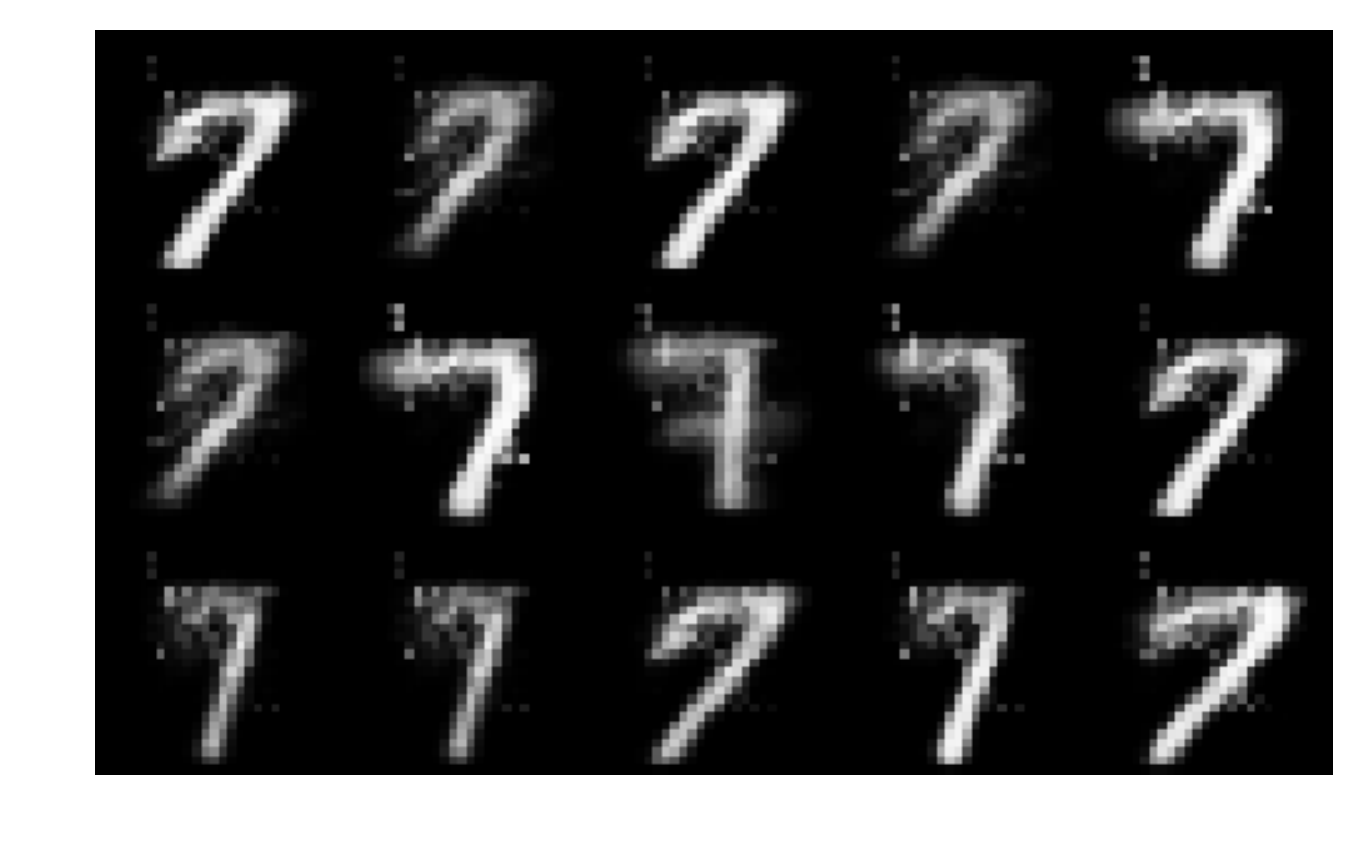}}&\makecell{\includegraphics[scale=0.12]{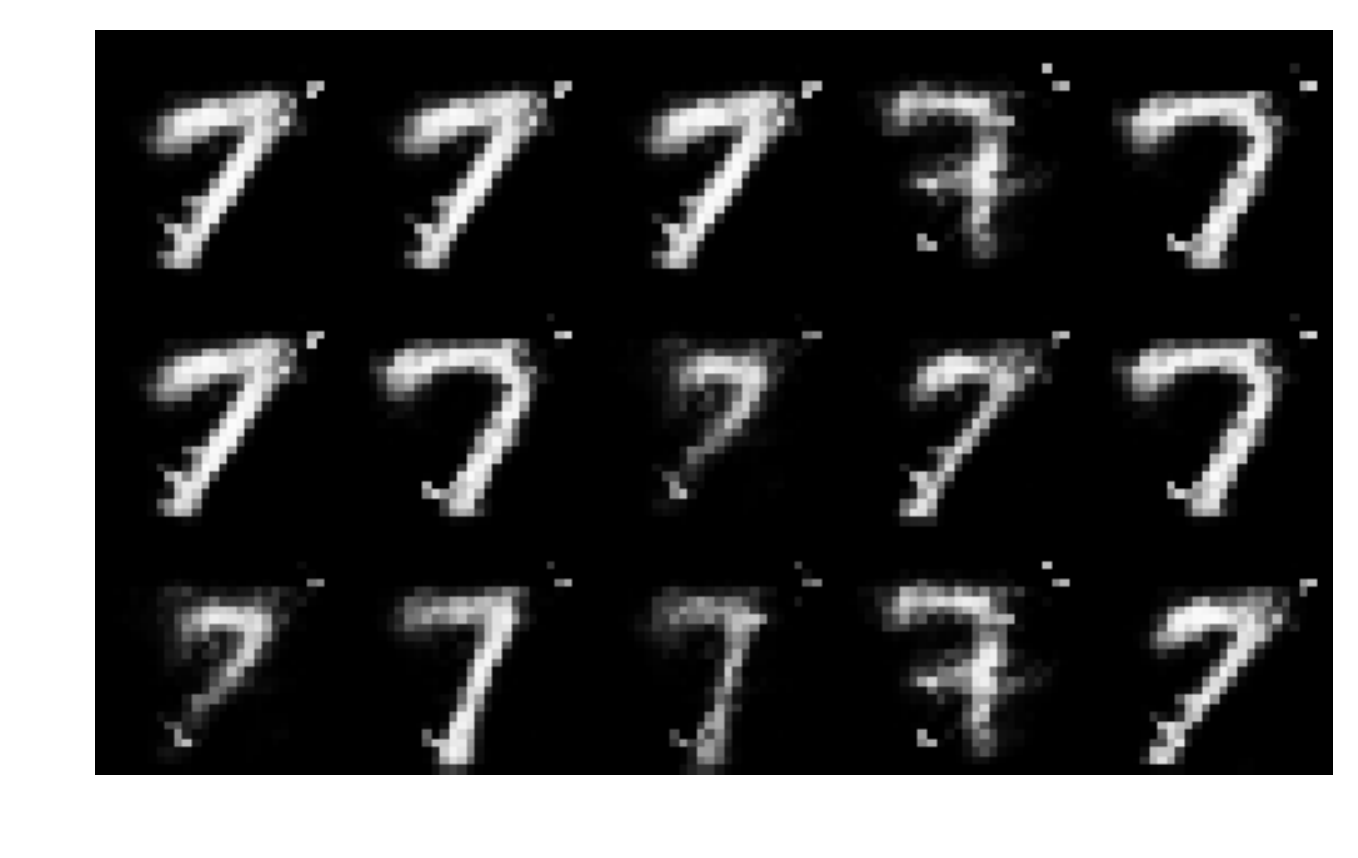}}&\makecell{\includegraphics[scale=0.12]{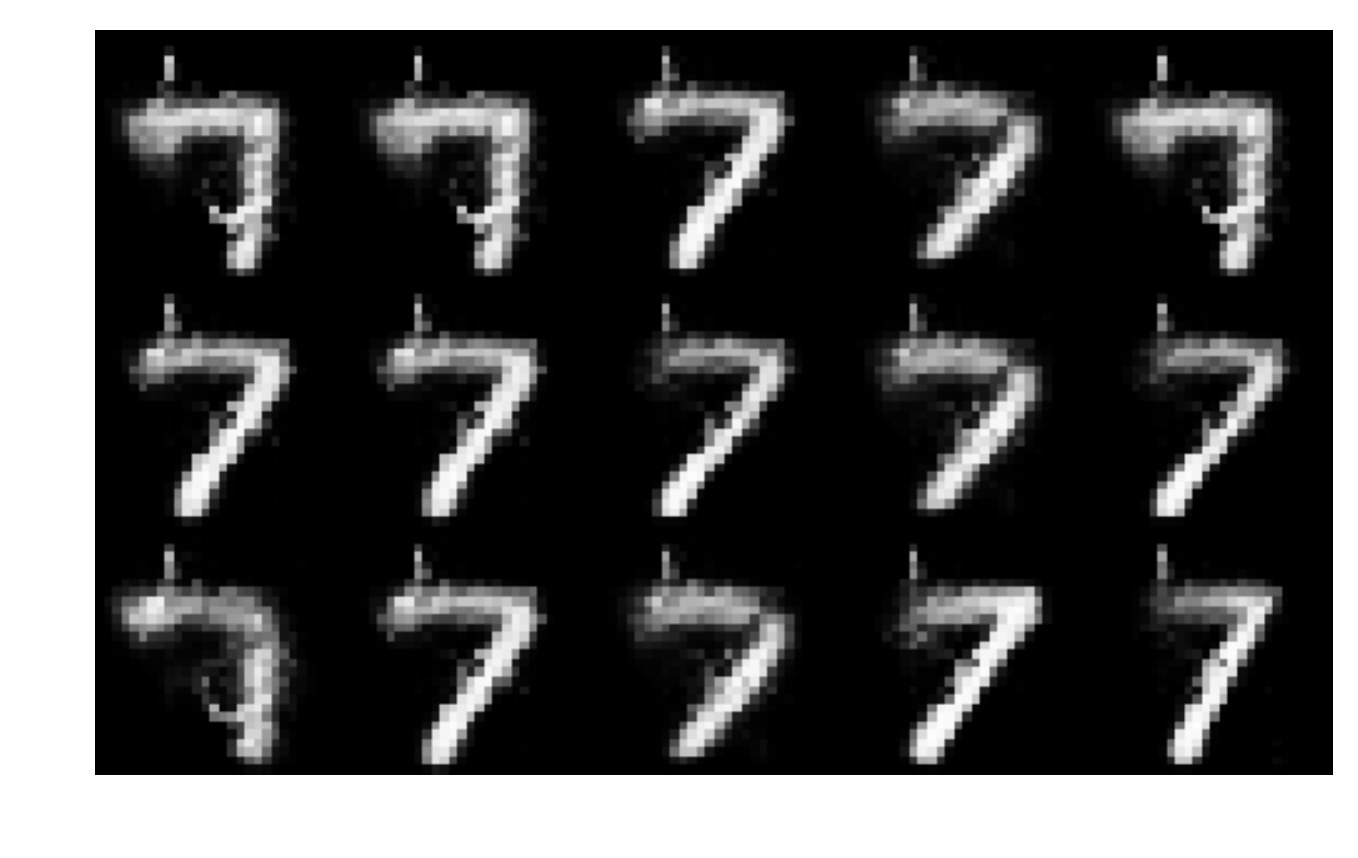}}&--&\makecell{\includegraphics[scale=0.12]{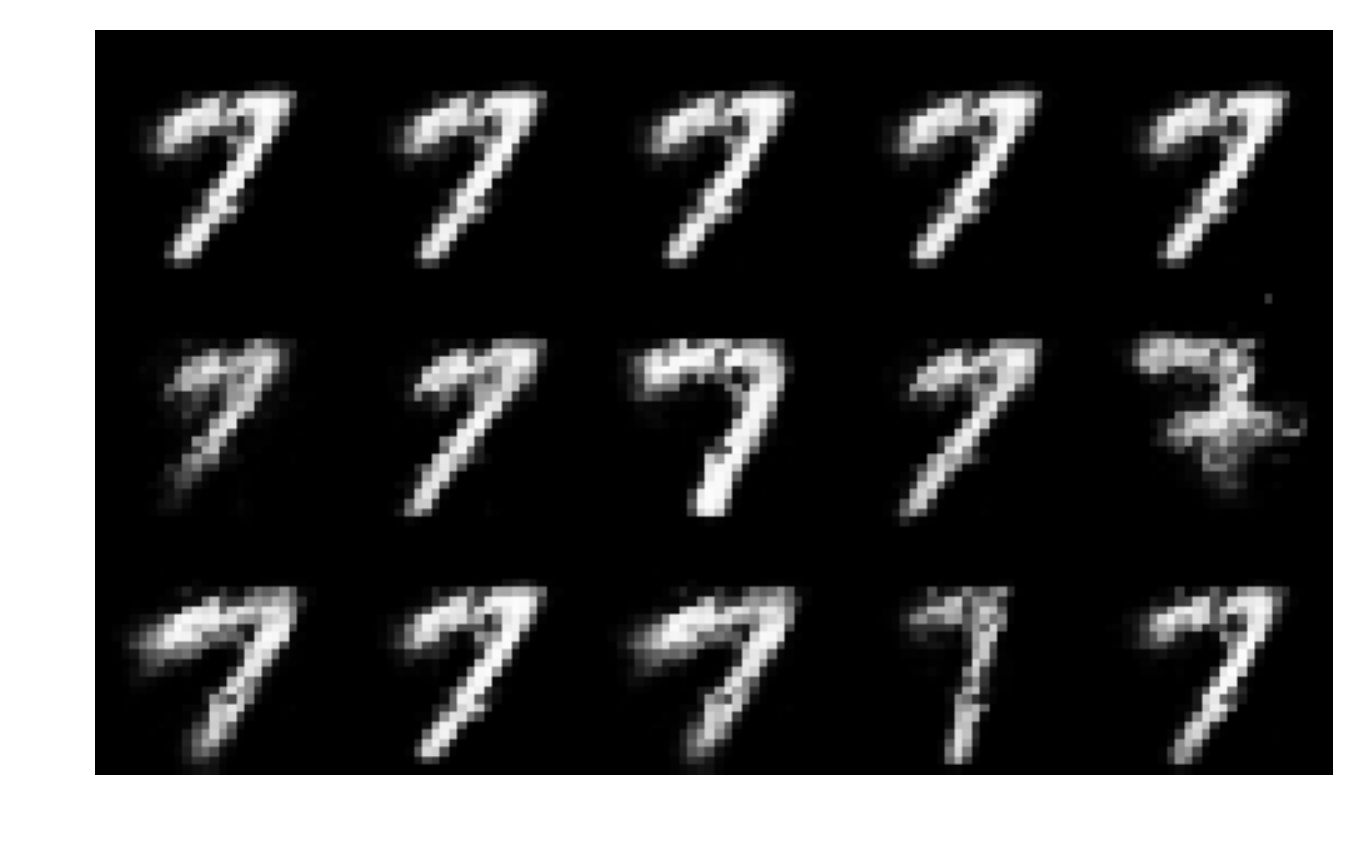}}&\makecell{\includegraphics[scale=0.12]{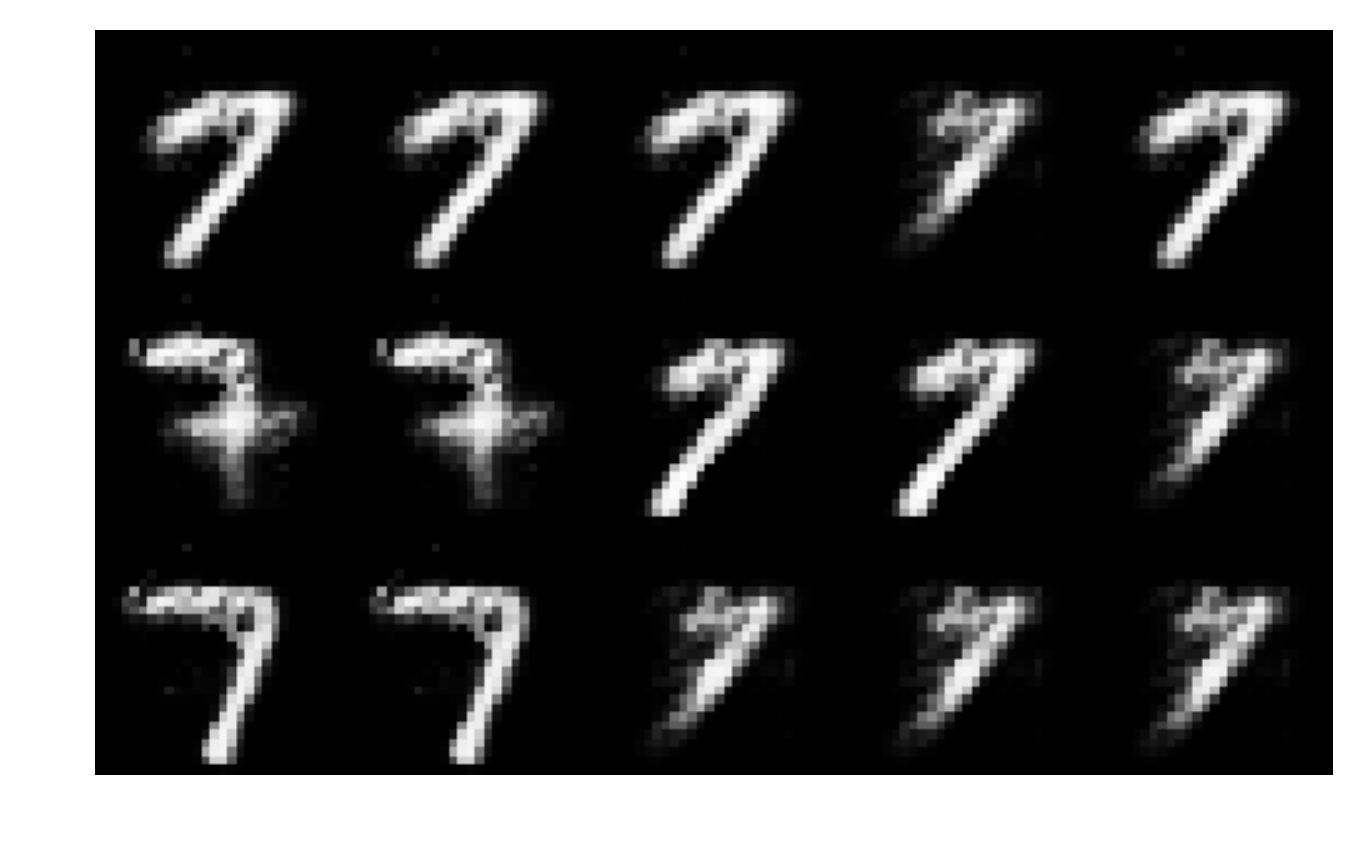}}\\ \hline
`8'&\makecell{\includegraphics[scale=0.12]{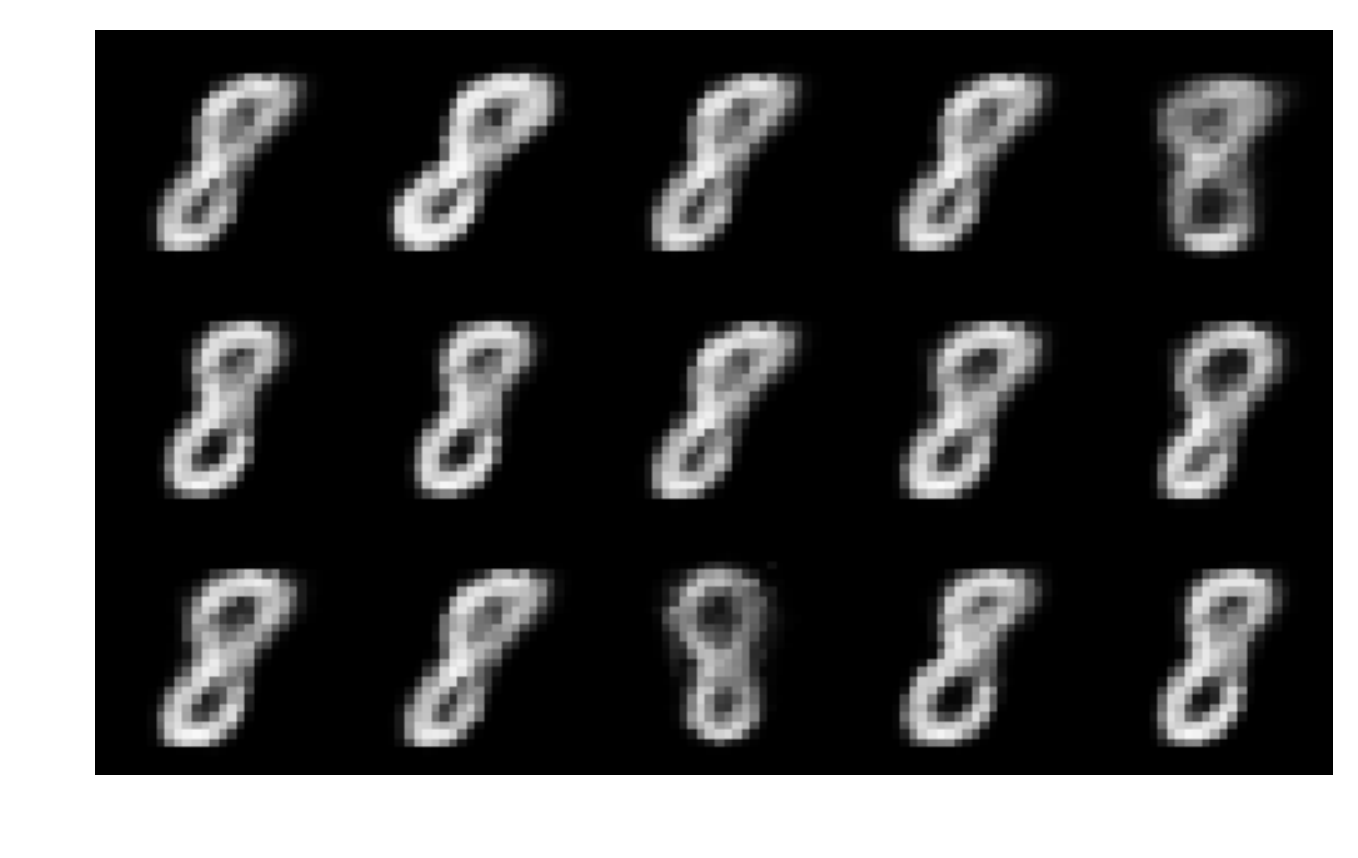}}&\makecell{\includegraphics[scale=0.12]{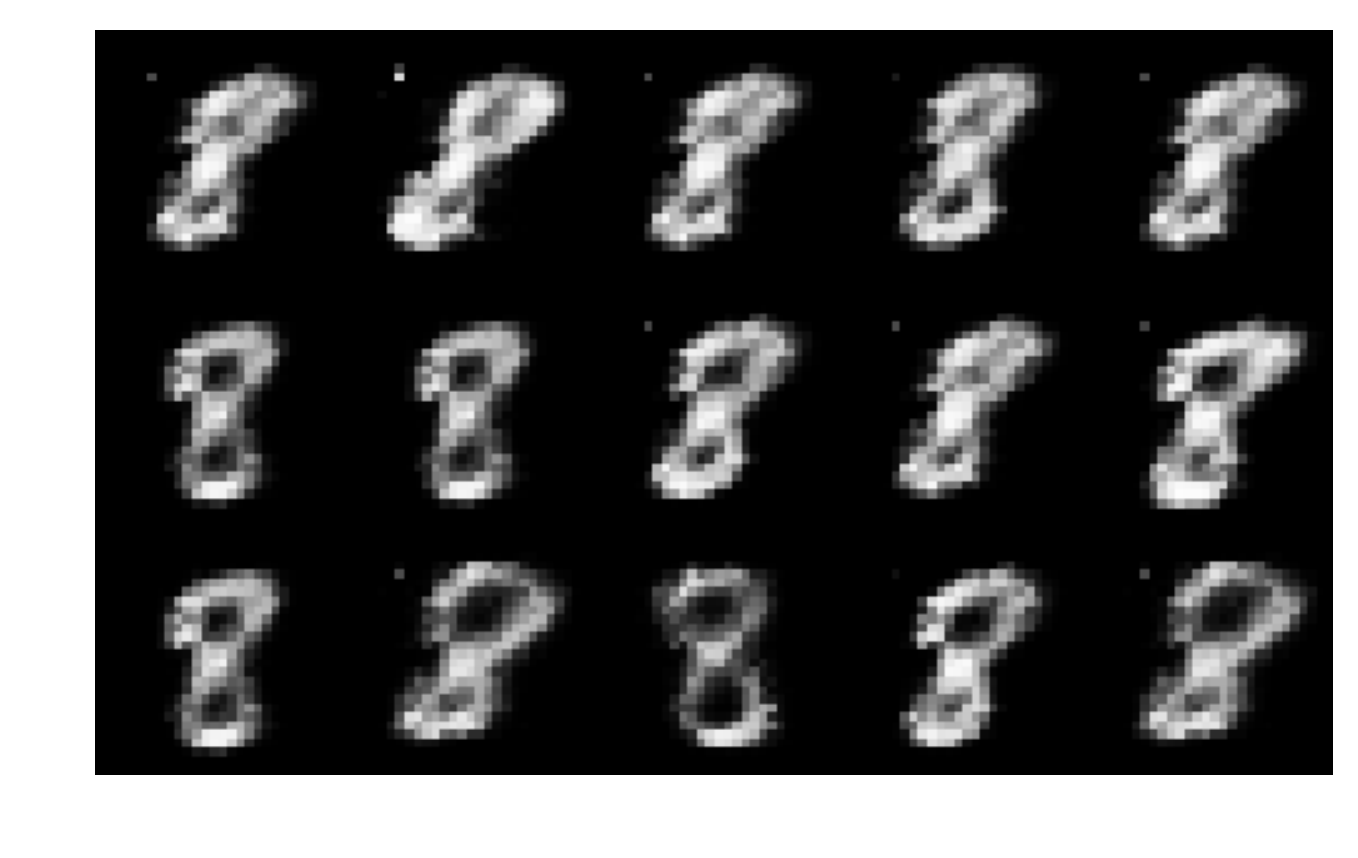}}&\makecell{\includegraphics[scale=0.12]{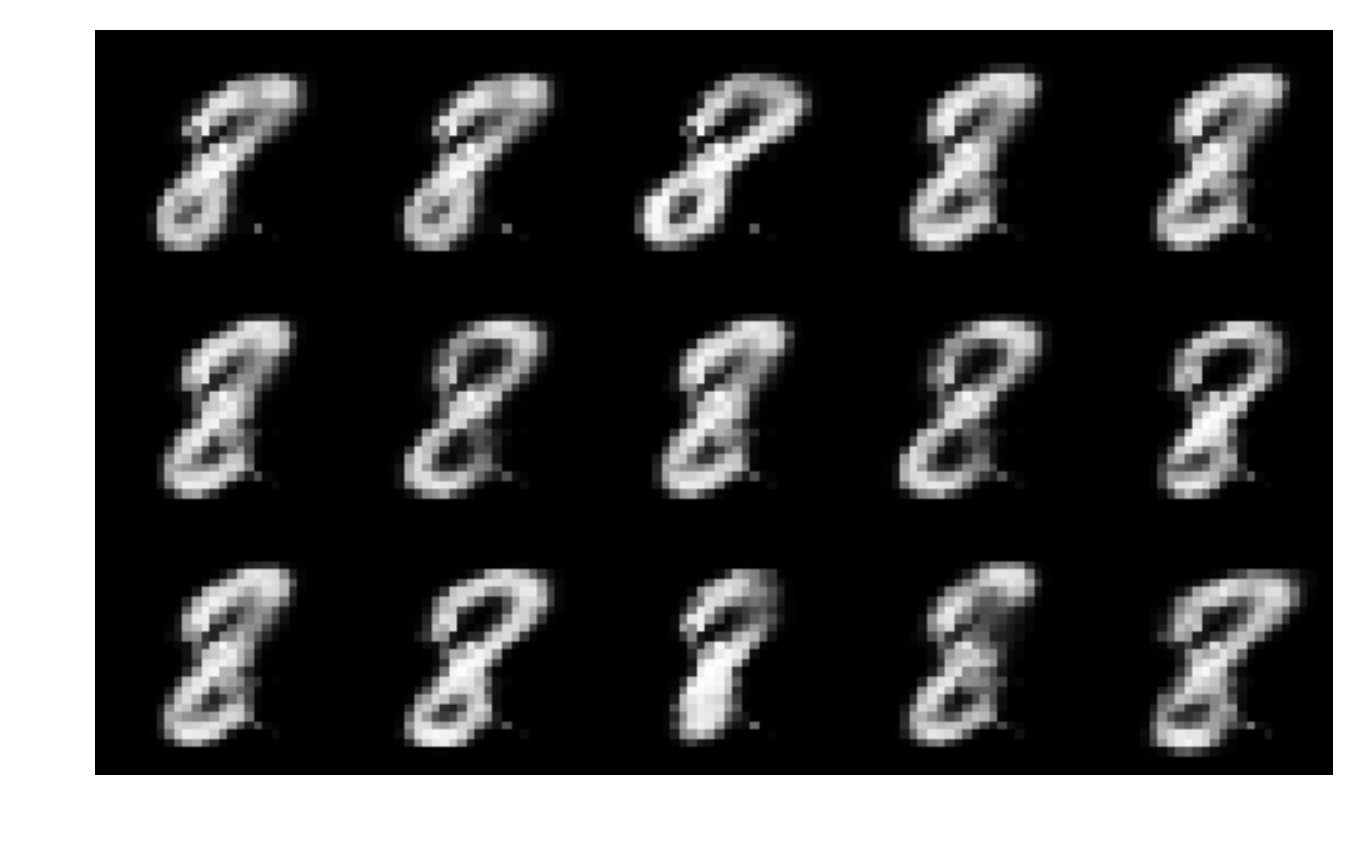}}&\makecell{\includegraphics[scale=0.12]{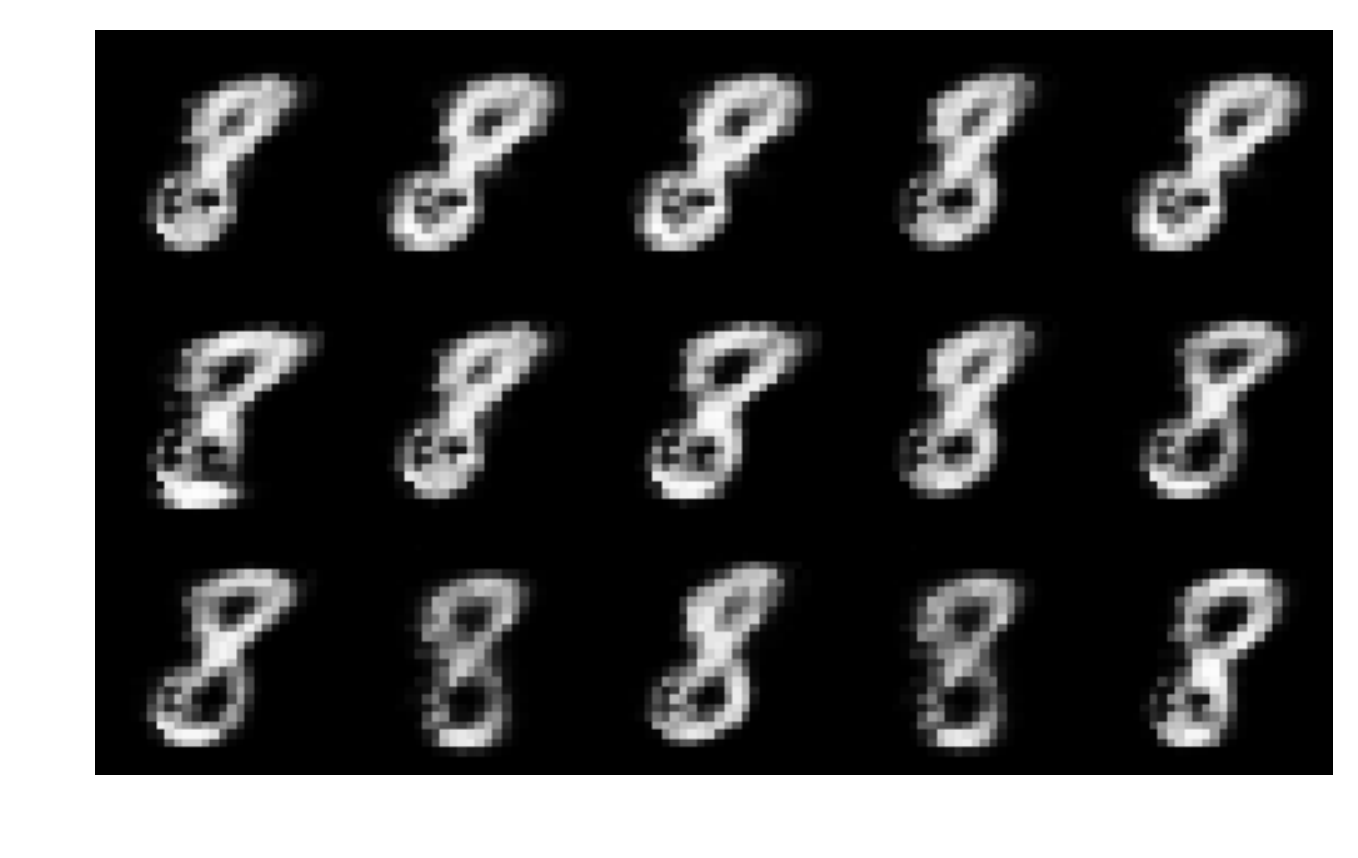}}&\makecell{\includegraphics[scale=0.12]{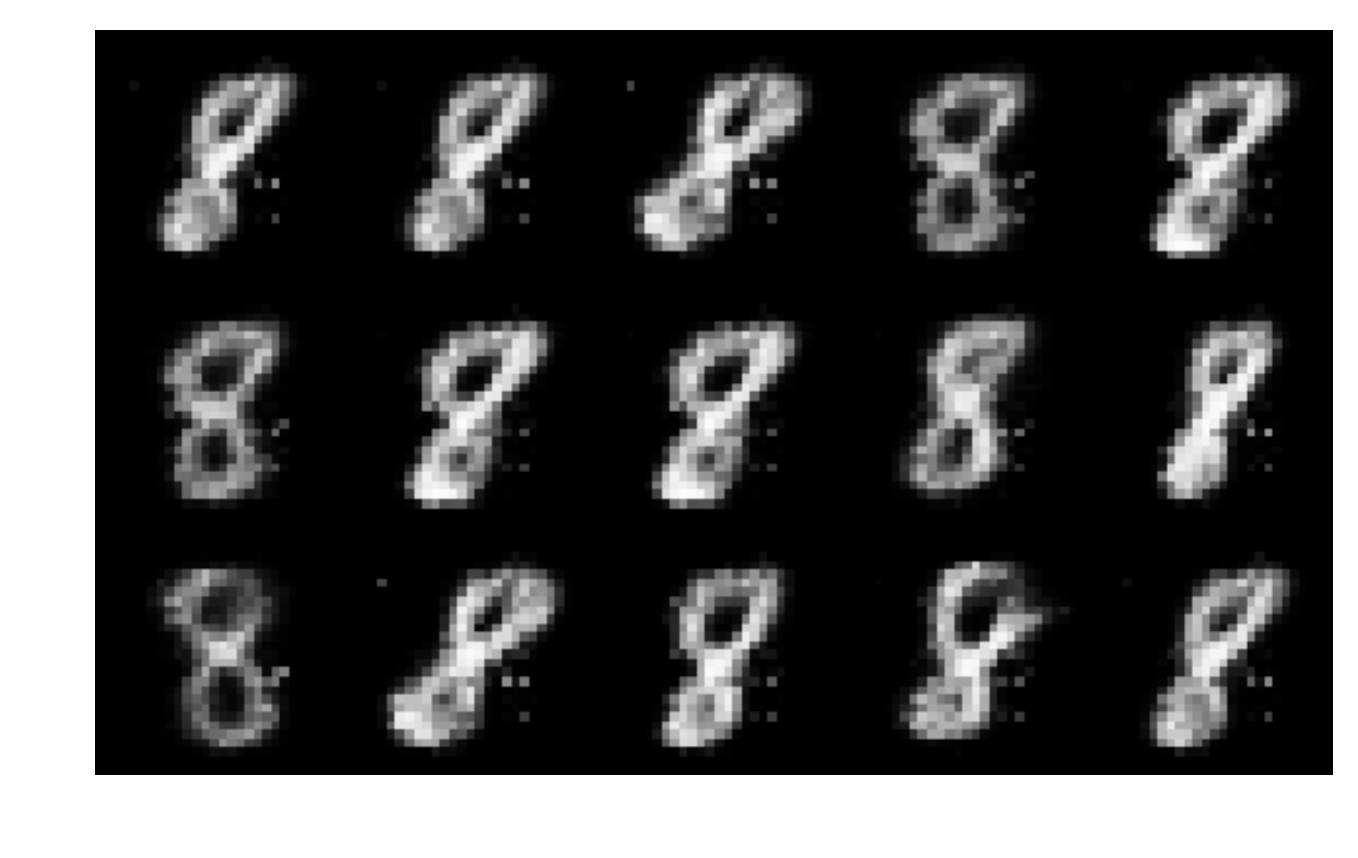}}&\makecell{\includegraphics[scale=0.12]{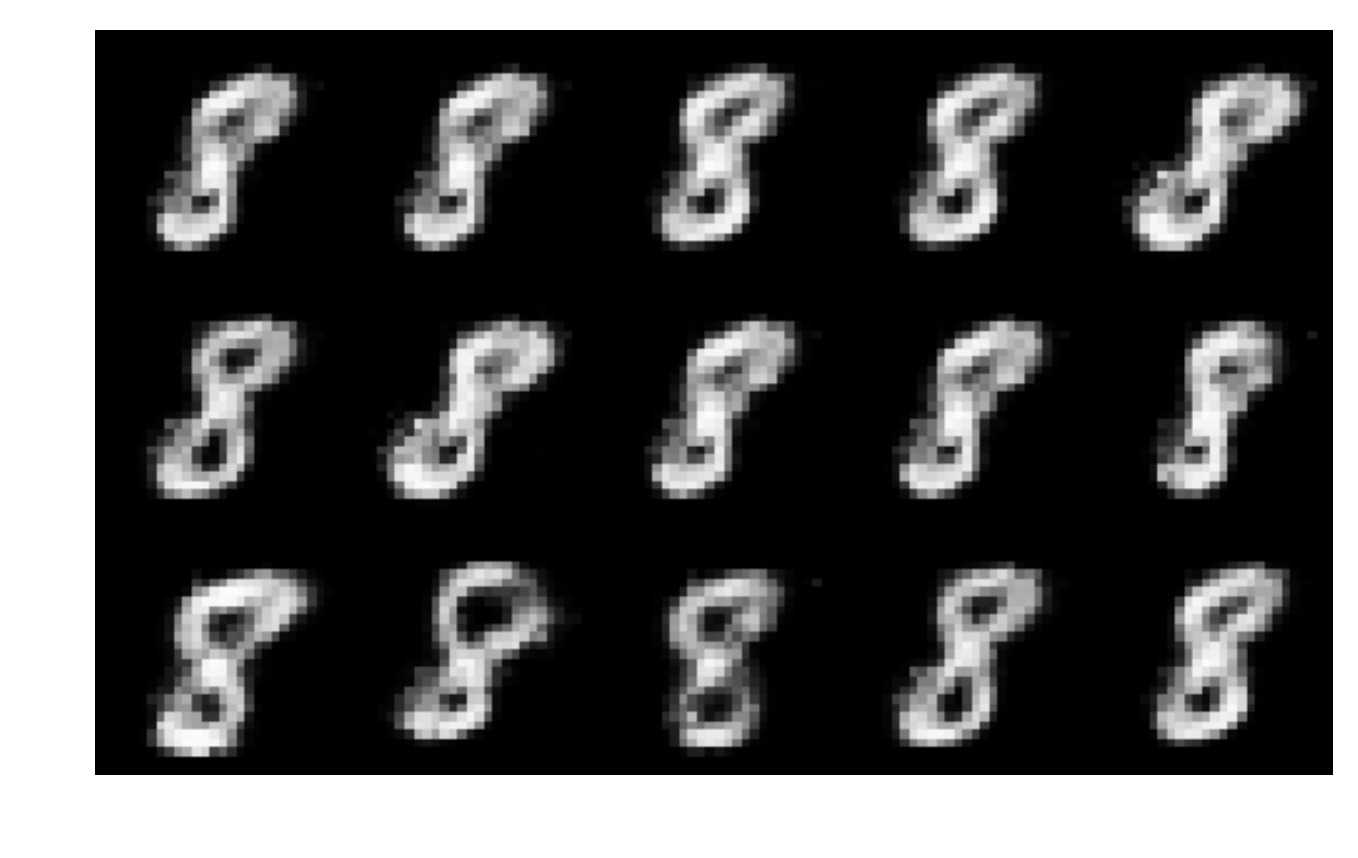}}&\makecell{\includegraphics[scale=0.12]{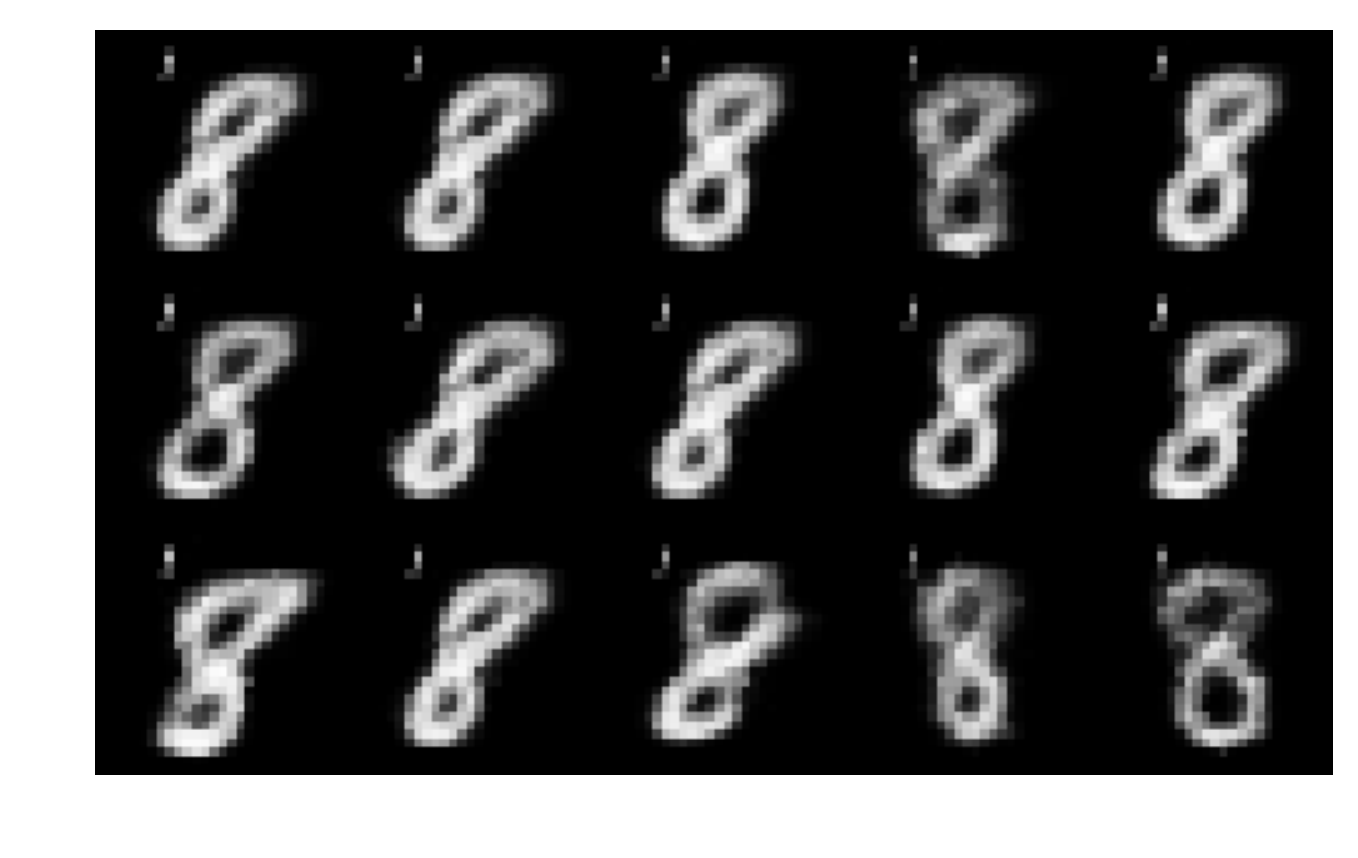}}&\makecell{\includegraphics[scale=0.12]{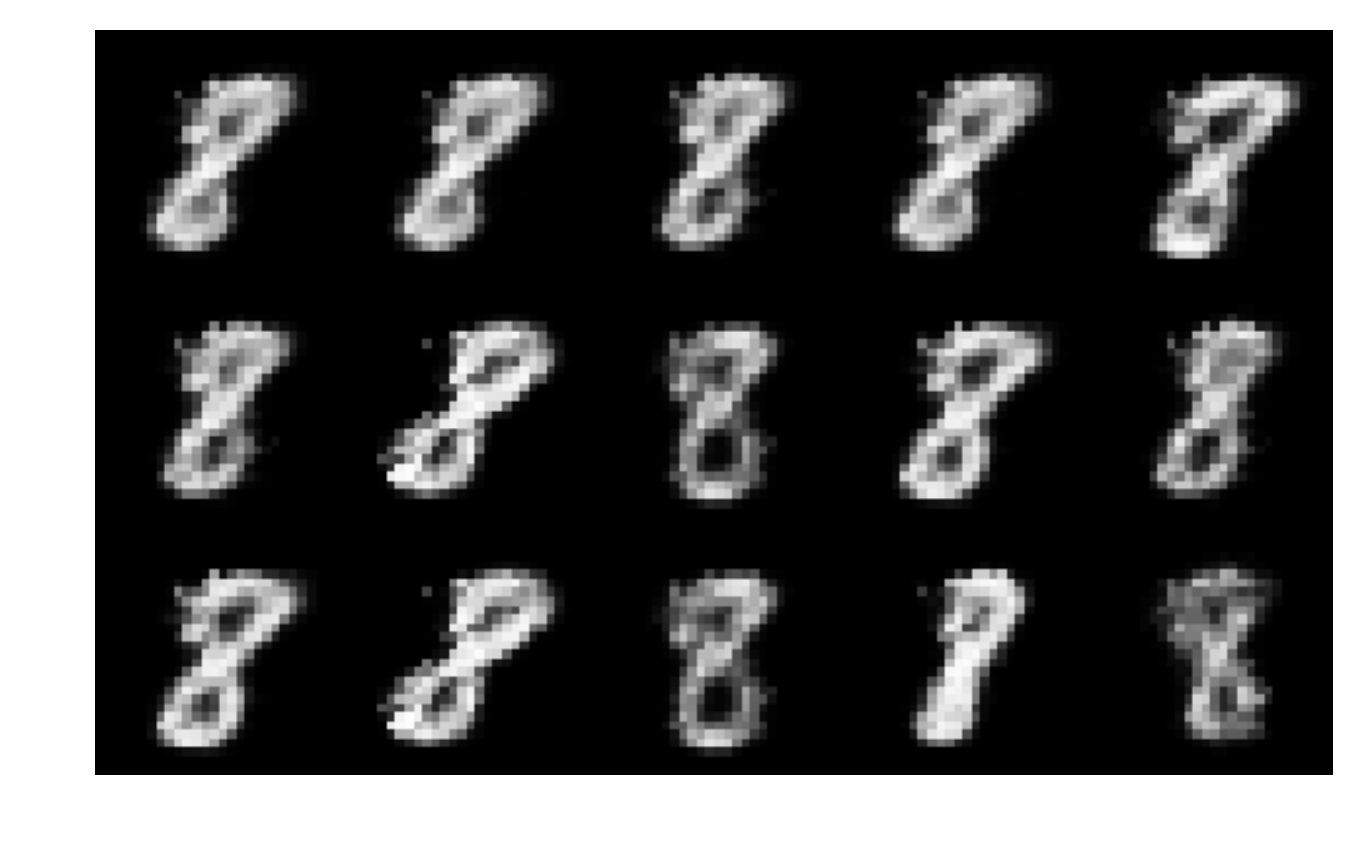}}&--&\makecell{\includegraphics[scale=0.12]{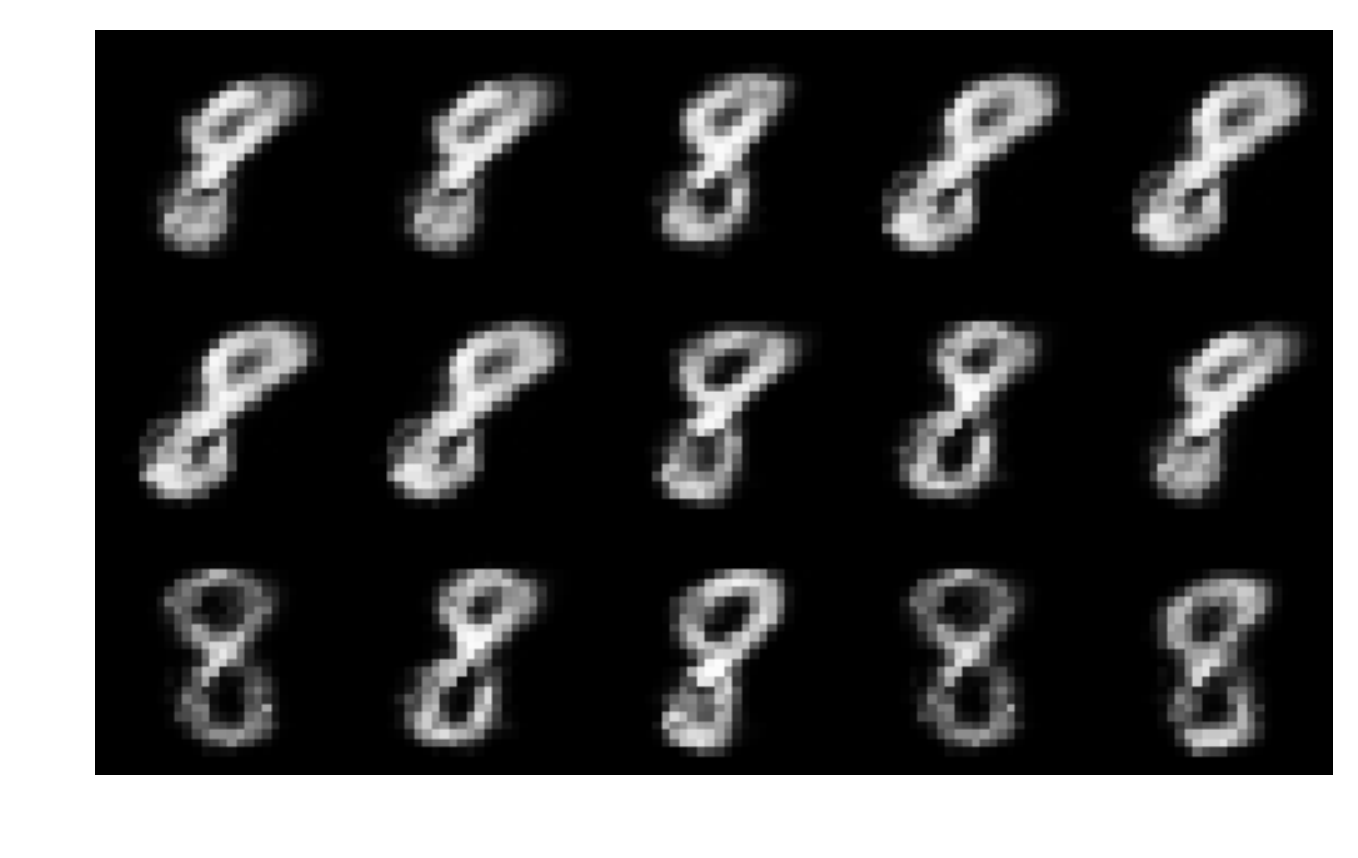}}\\ \hline
`9'&\makecell{\includegraphics[scale=0.12]{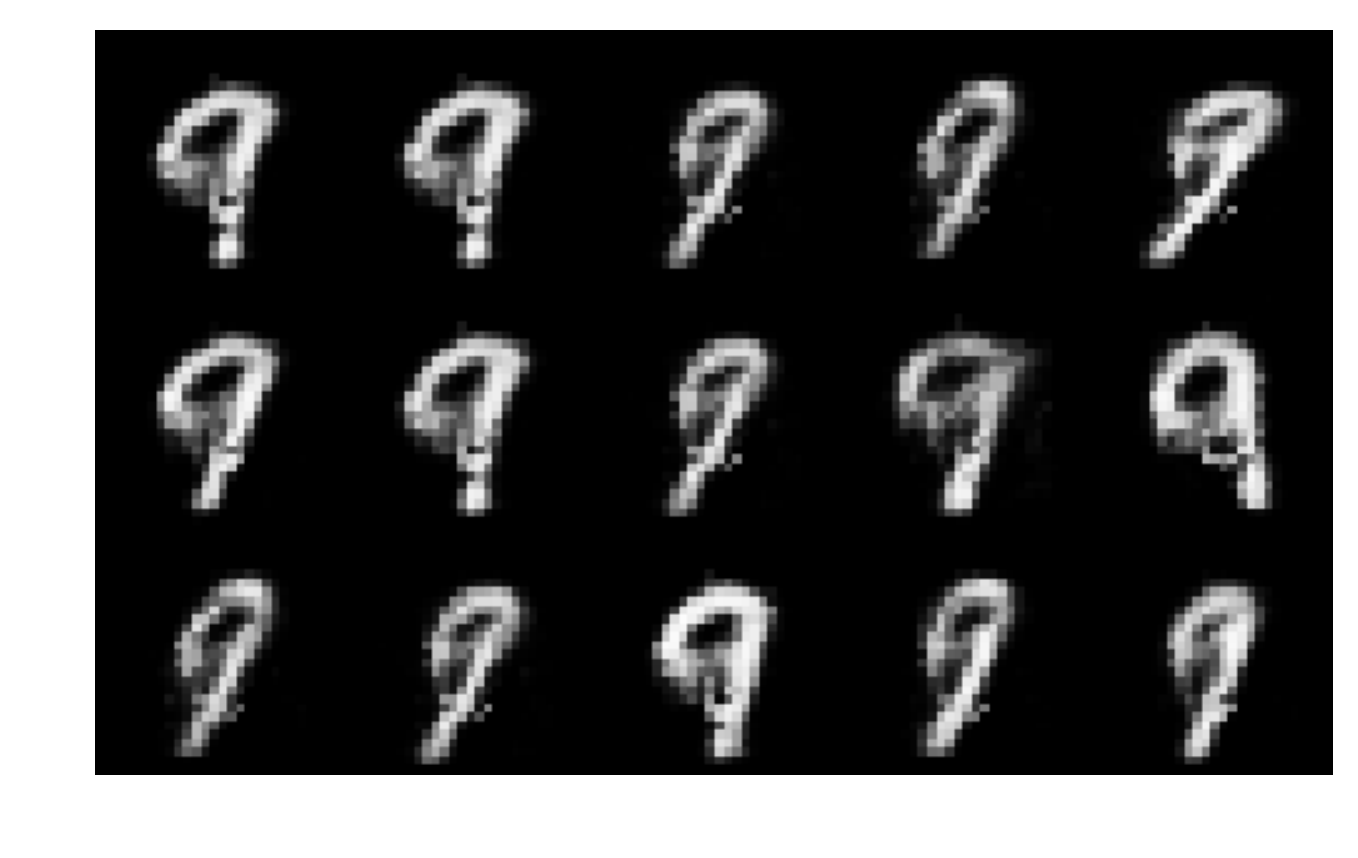}}&\makecell{\includegraphics[scale=0.12]{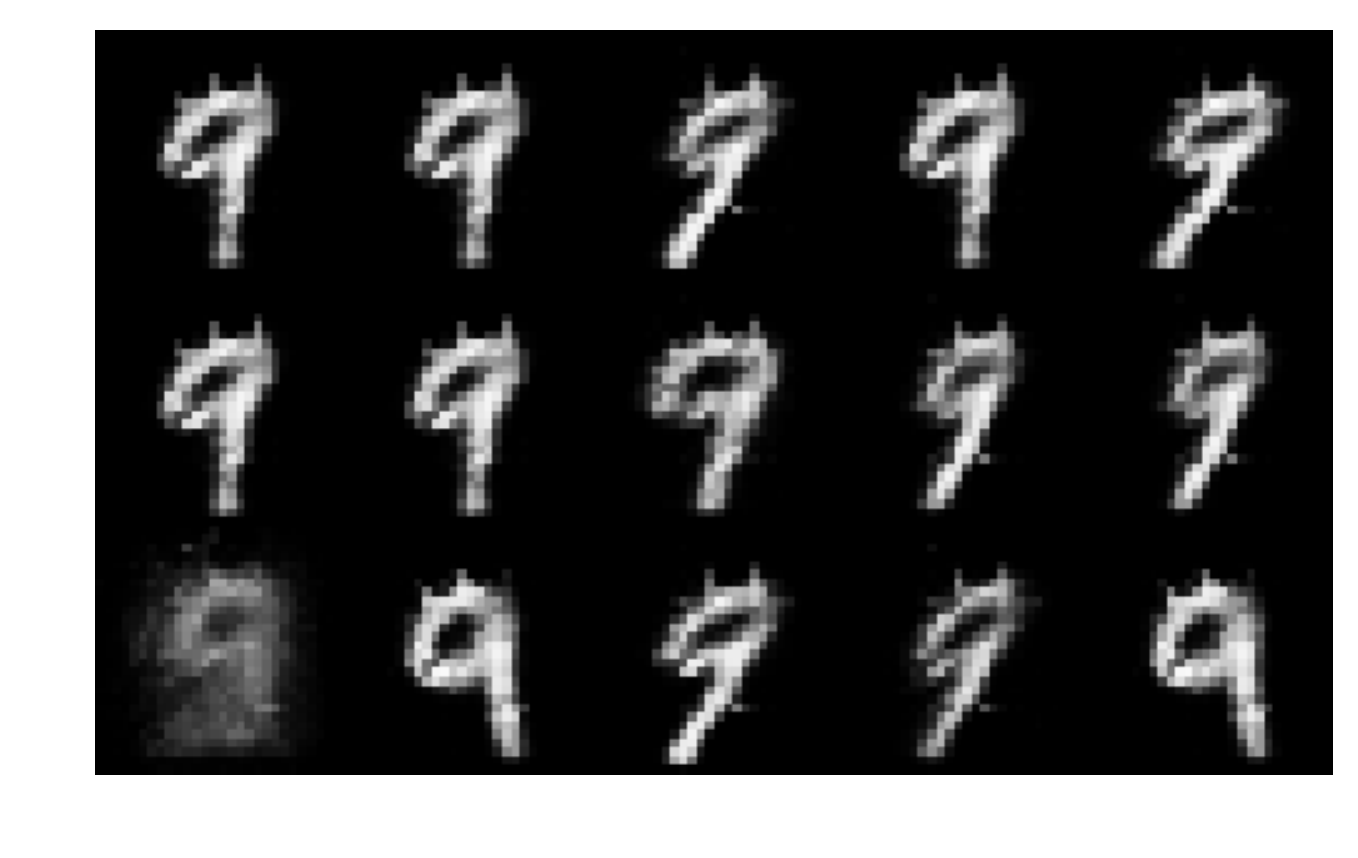}}&\makecell{\includegraphics[scale=0.12]{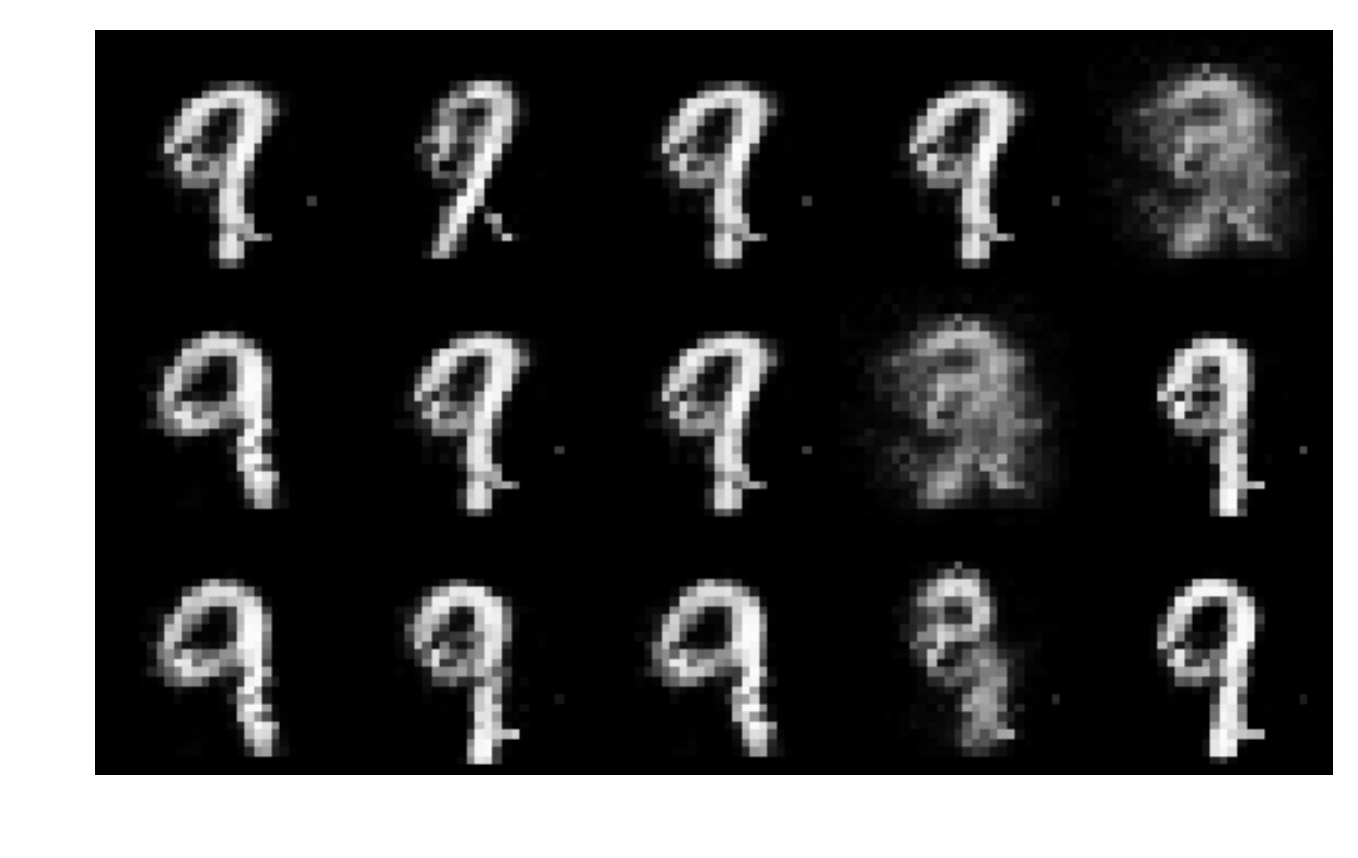}}&\makecell{\includegraphics[scale=0.12]{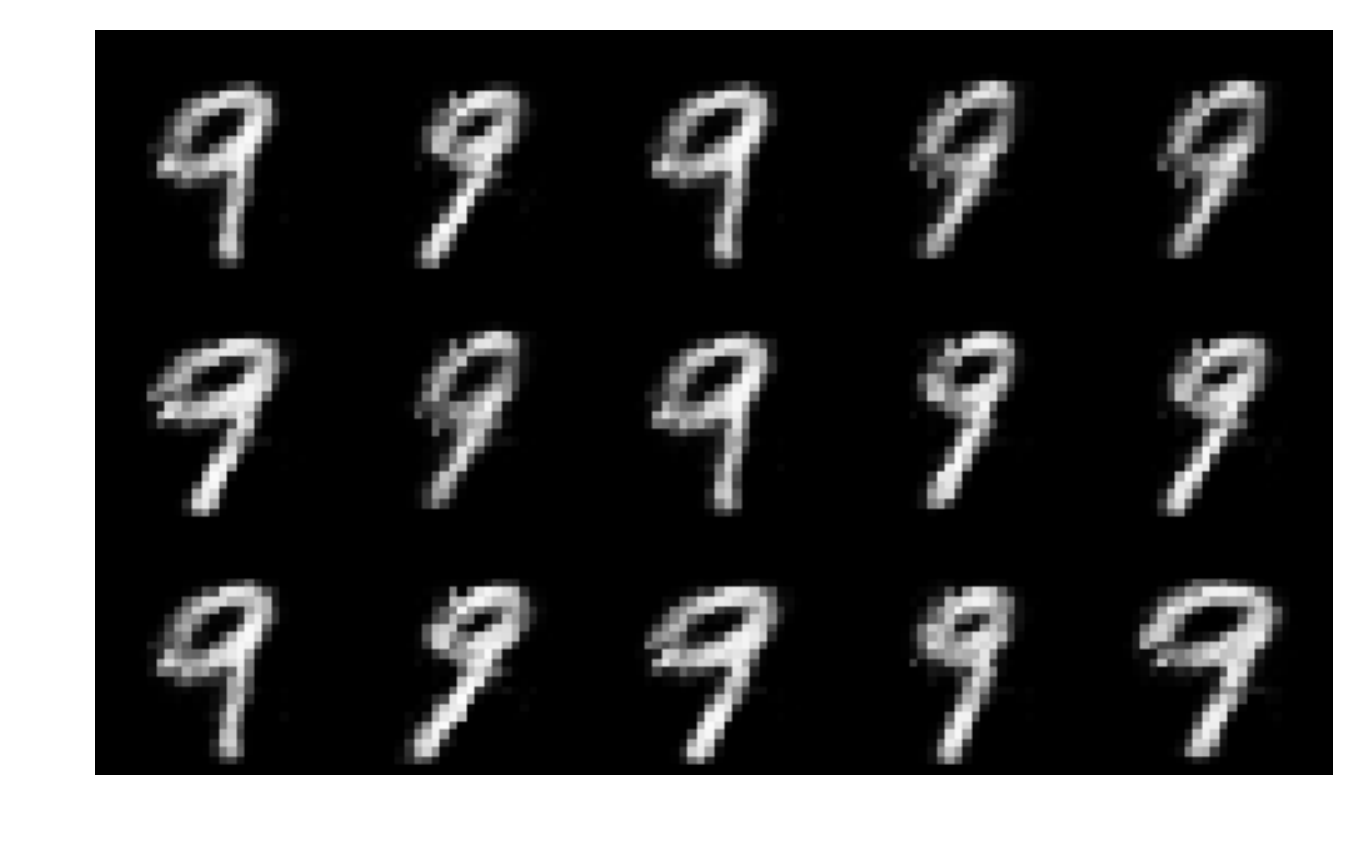}}&\makecell{\includegraphics[scale=0.12]{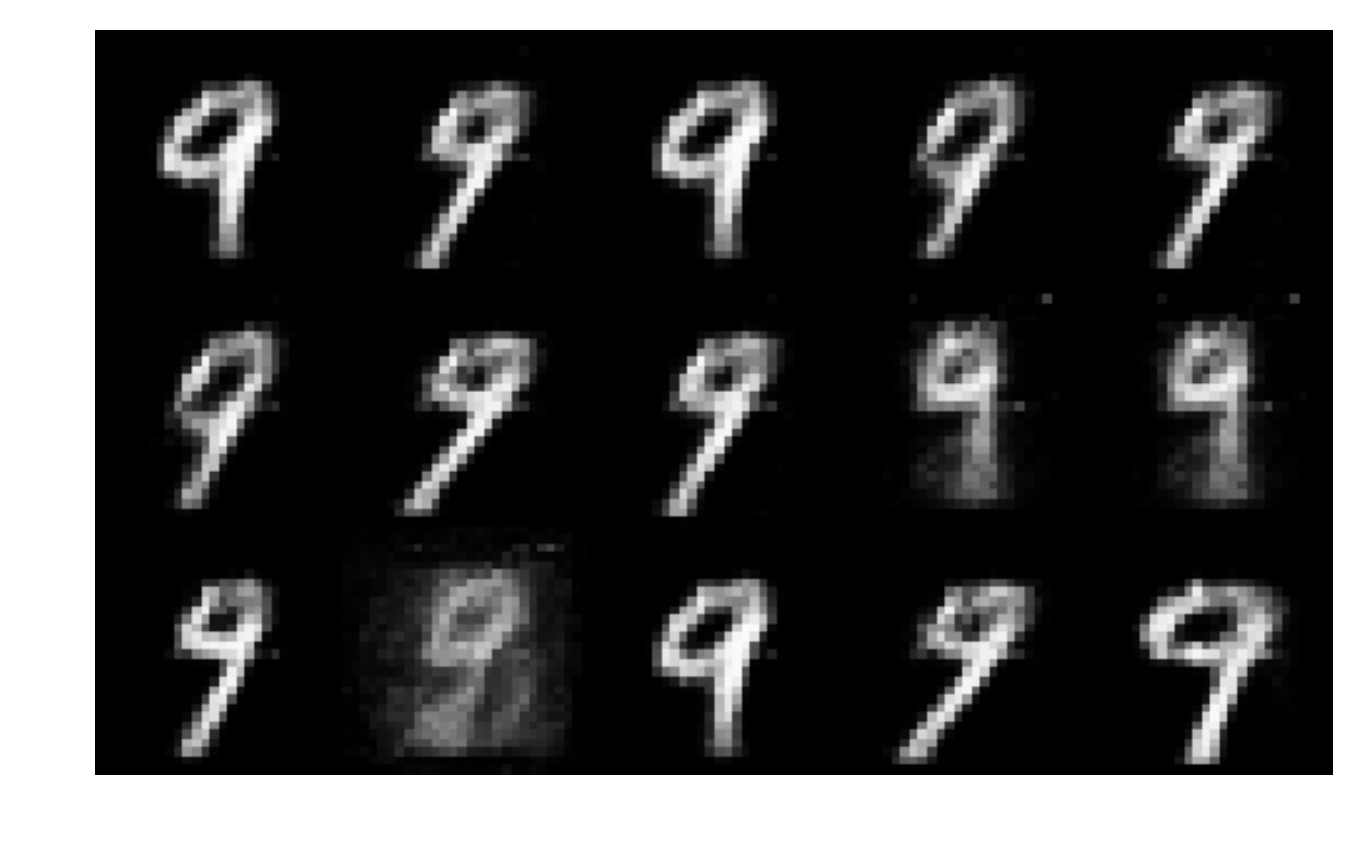}}&\makecell{\includegraphics[scale=0.12]{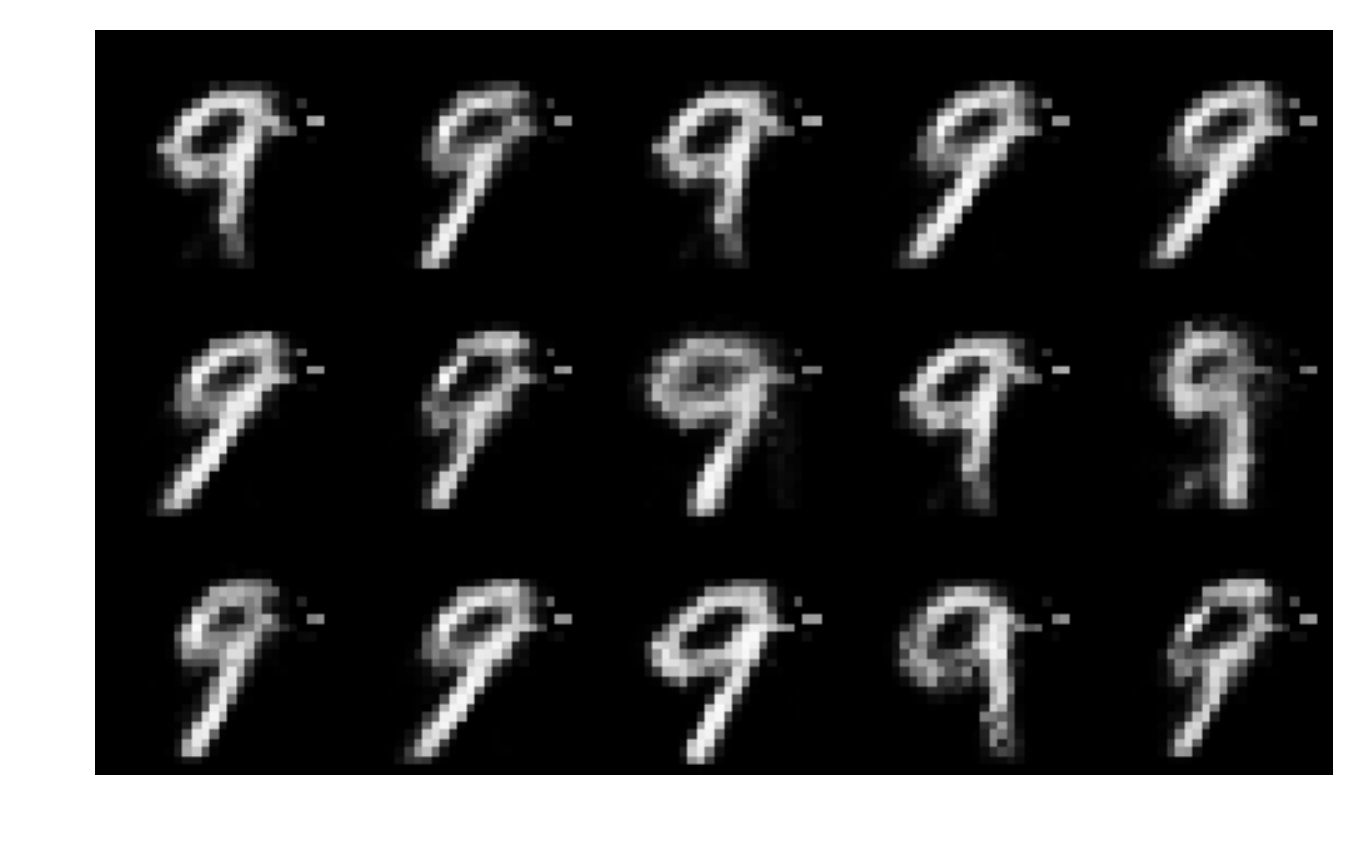}}&\makecell{\includegraphics[scale=0.12]{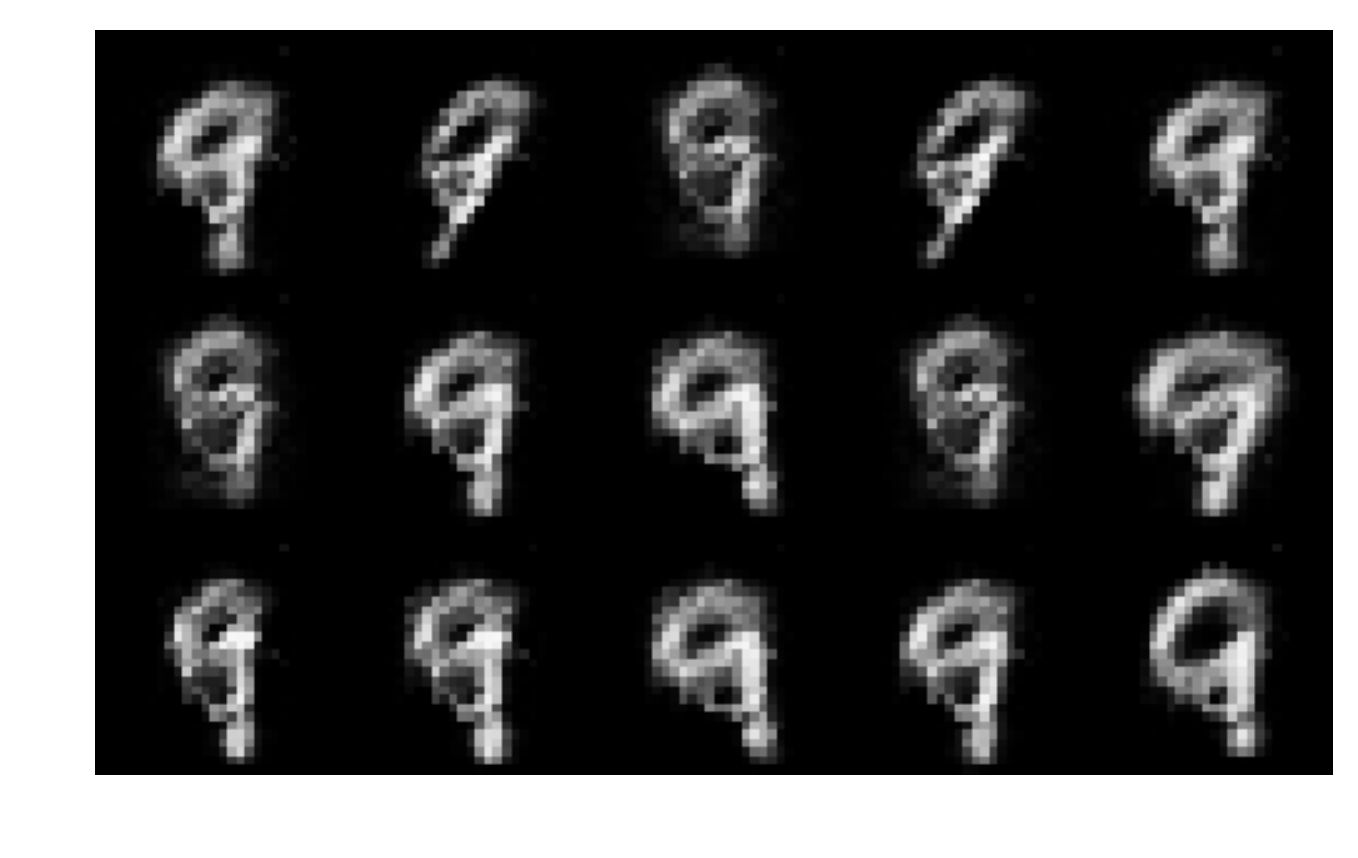}}&\makecell{\includegraphics[scale=0.12]{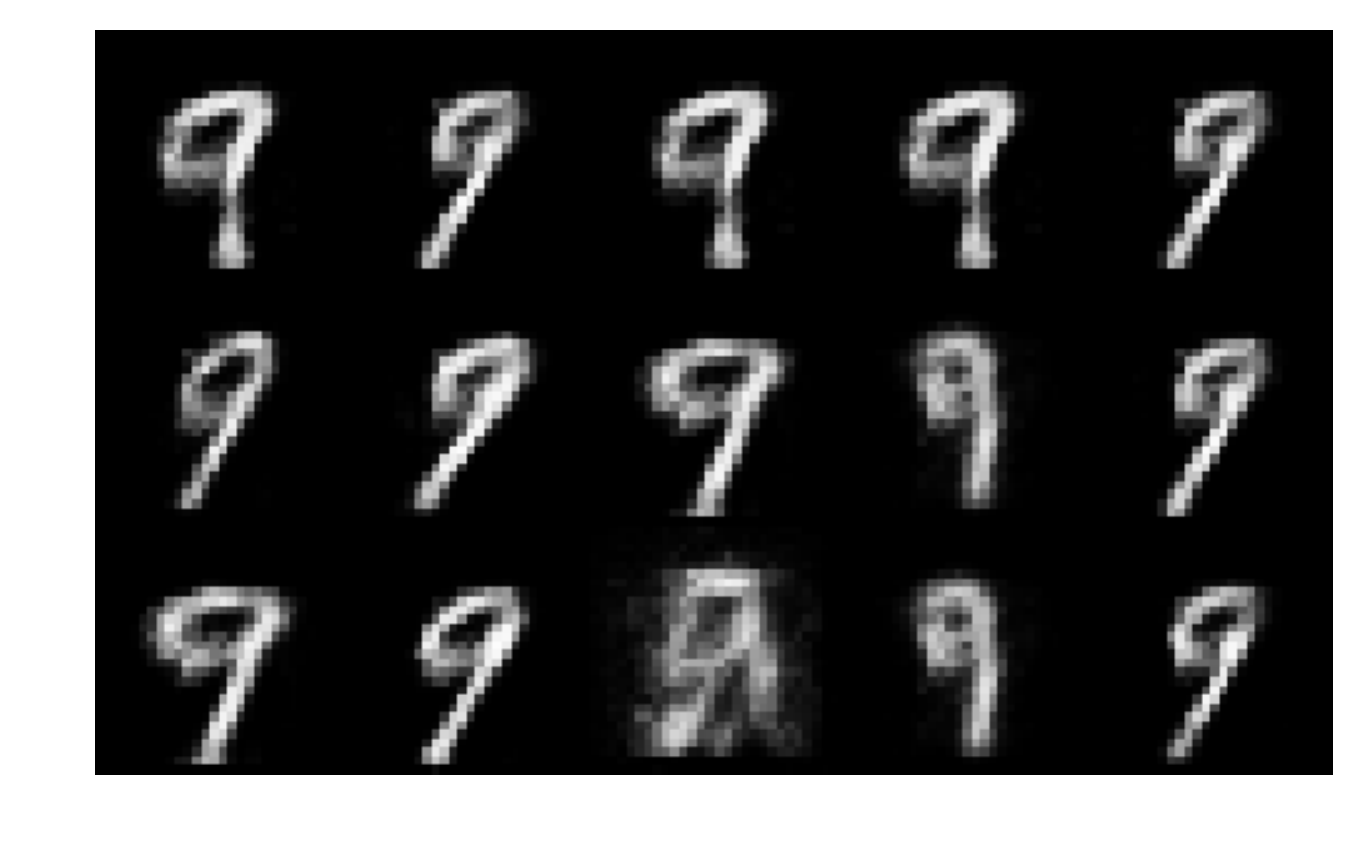}}&\makecell{\includegraphics[scale=0.12]{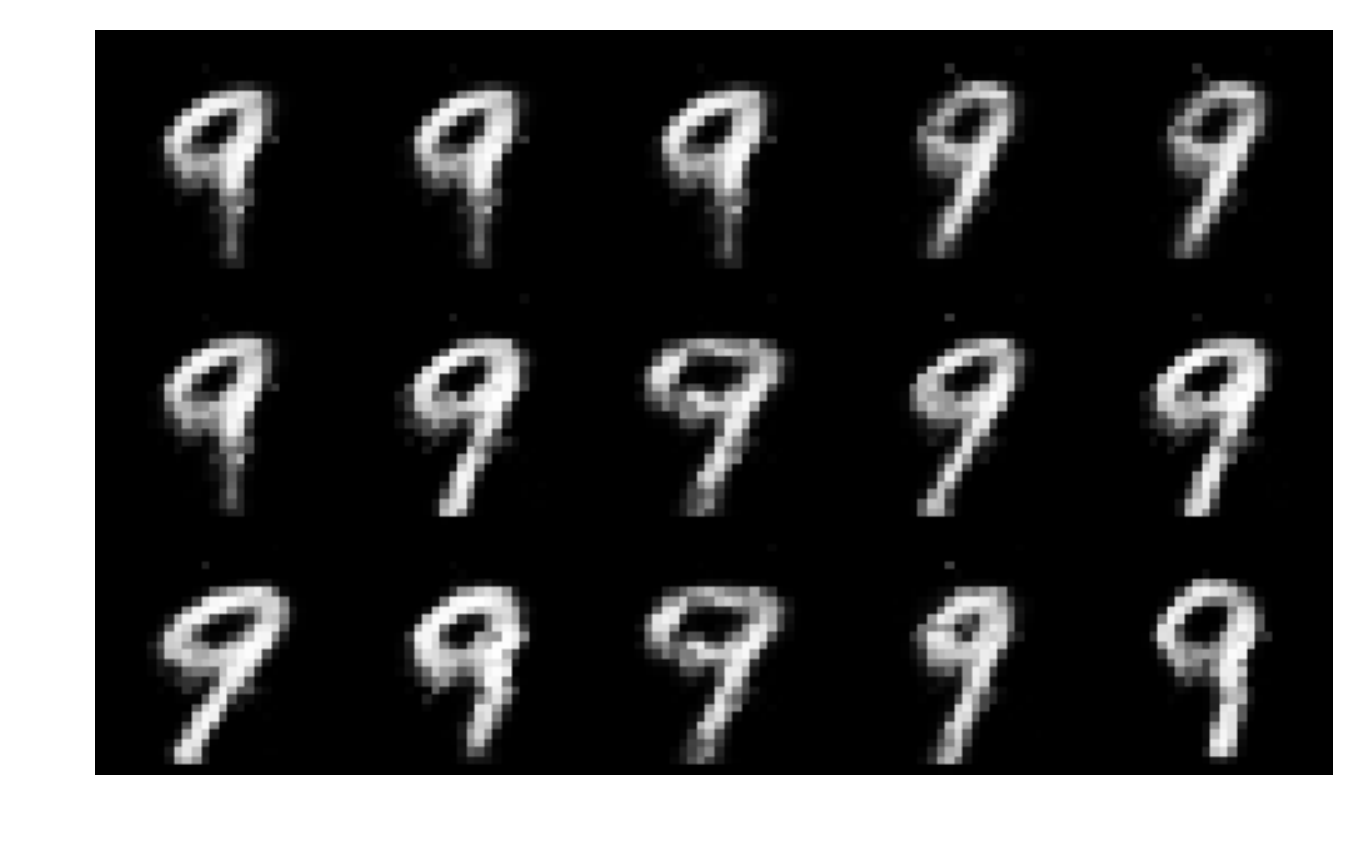}}&-- \\ \hline

    \end{tabular}
    \label{tab:intra-model-visualization}
\end{table*}{}

\begin{figure}[!htb]
    \centering
    \includegraphics[width=\columnwidth]{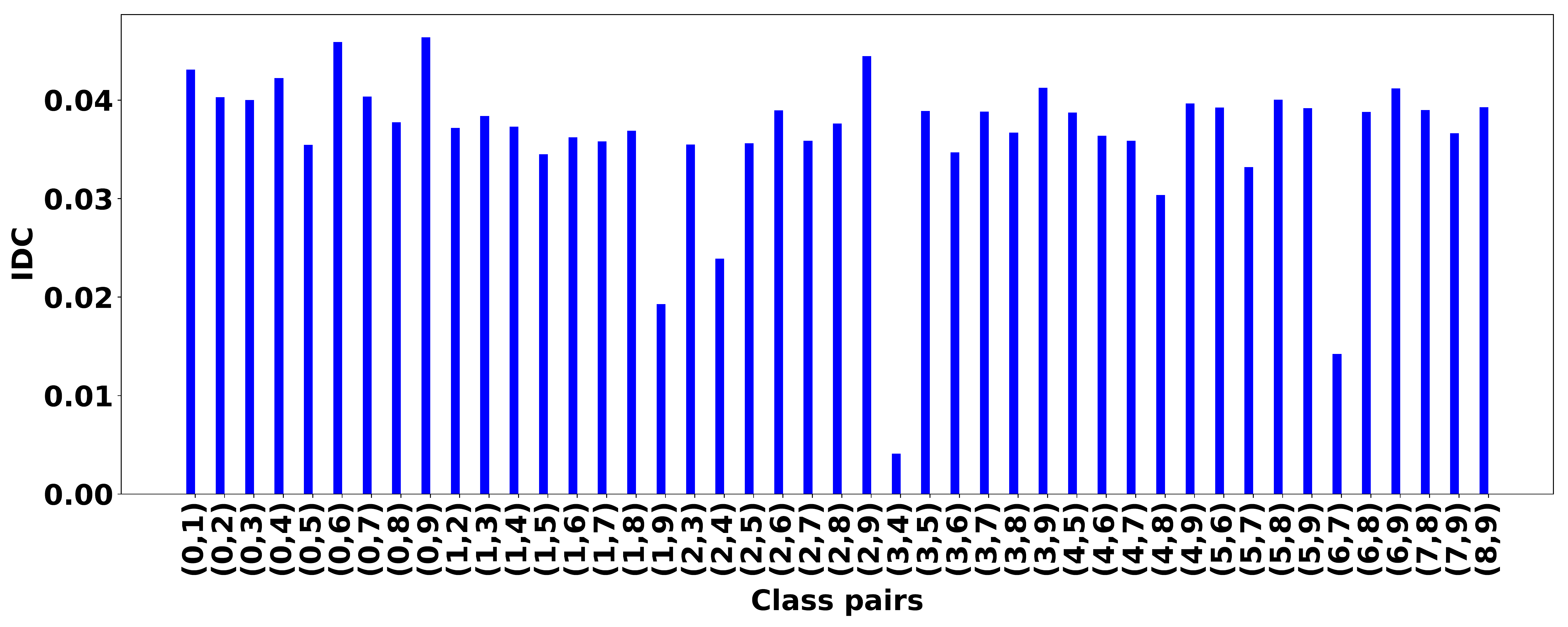}
    \caption{Input space Decision boundary Complexity (IDC) of model $\text{MNIST}_{\text{CNN}}$ according to the characteristic measure IDC discussed in Section~\ref{sec:geometrical} }
    \label{fig:intra-model-IDC}
\end{figure}

\begin{figure}[!htb]
    \centering
     \includegraphics[width=\columnwidth]{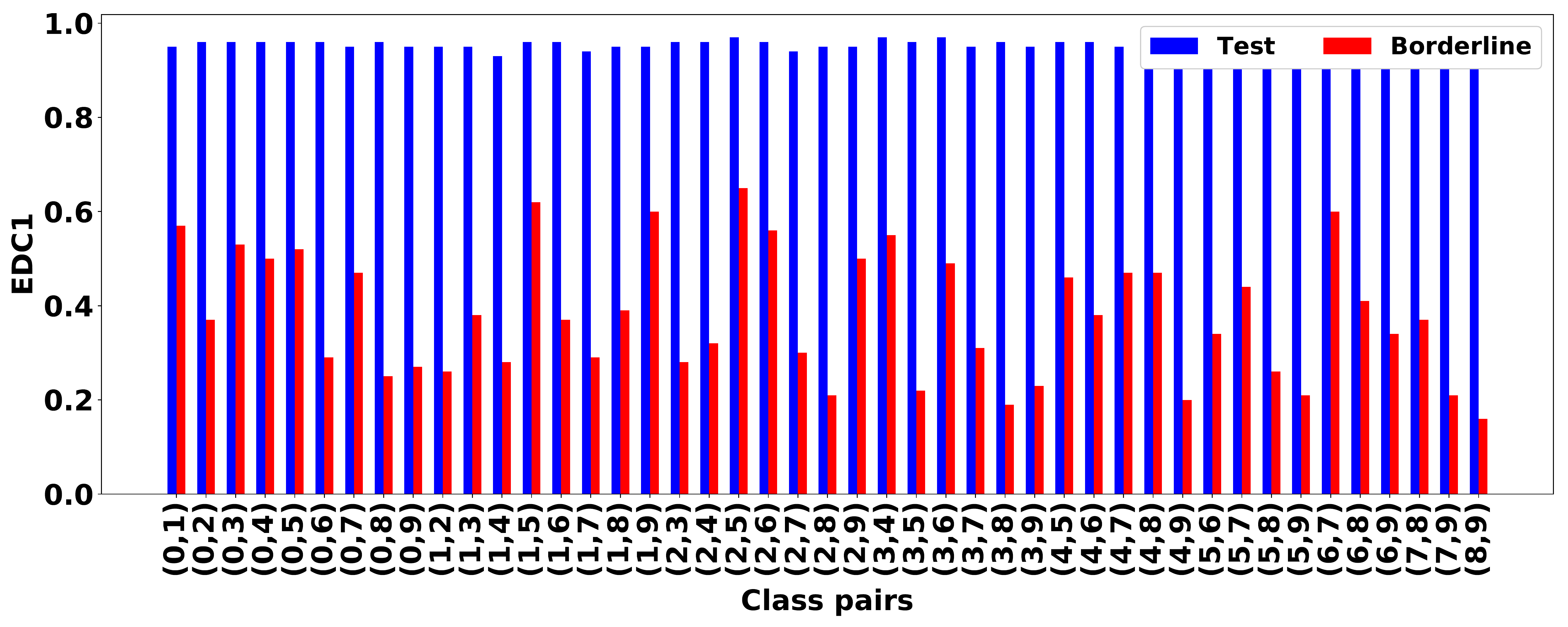}
    \caption{ Embedding space Decision boundary Complexity 1 (EDC1)  i.e., the distance from the separating hyperplane for all class pairs in model $\text{MNIST}_{\text{CNN}}$ according to the characteristic measure EDC1 discussed in Section~\ref{sec:linear}}
    \label{fig:intra-model-EDC1}
\end{figure}

\begin{figure}[!htb]
    \centering
     \includegraphics[width=\columnwidth]{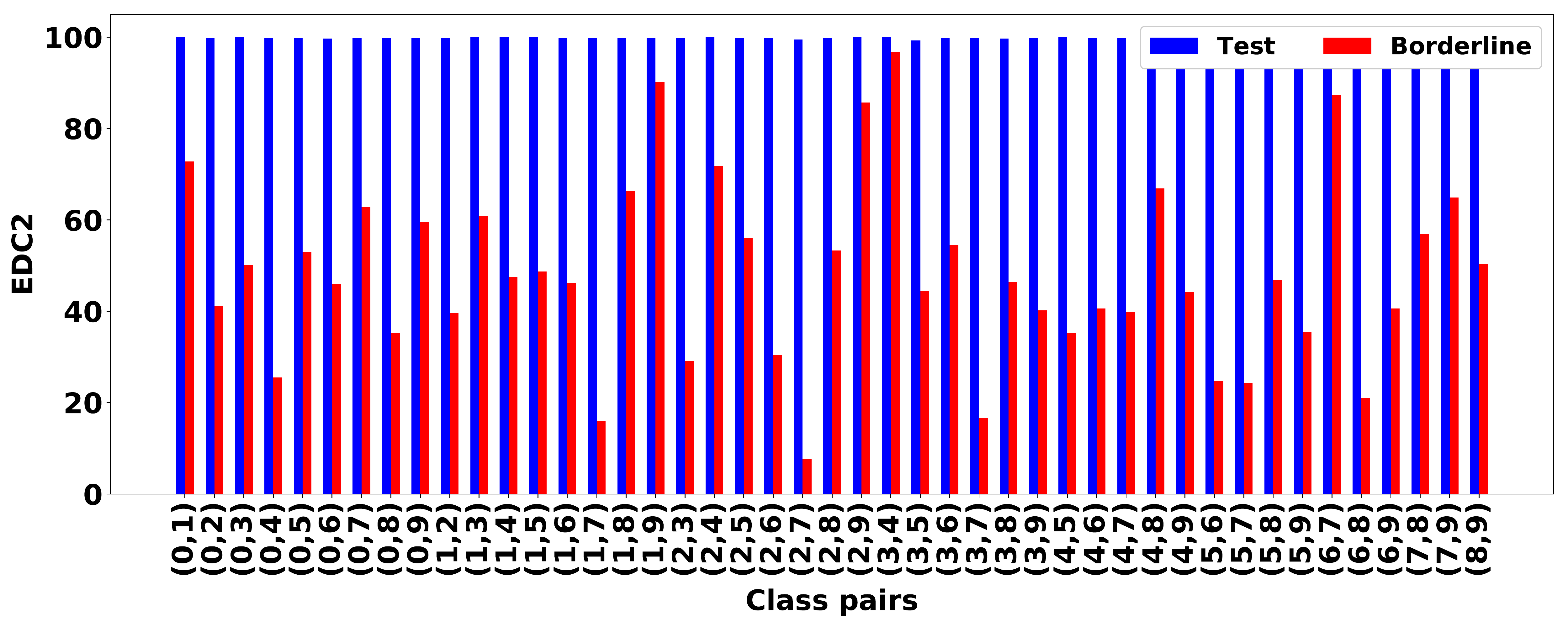}
    \caption{Embedding space Decision boundary Complexity 2 (EDC2) i.e., the accuracy against the linear SVM for all class pairs in model $\text{MNIST}_{\text{CNN}}$ according to the characteristic measure EDC2 discussed in Section~\ref{sec:linear}}
    \label{fig:intra-model-EDC2}
\end{figure}

In Section~\ref{sec:charac-dnns} we presented the decision boundary characteristics for several DNNs and a single pair of classes per each dataset. In this part, we investigate the decision boundary characterization for all pairs of classes. To this end, we focus on $\text{MNIST}_{\text{CNN}}$ as we achieved similar results for other DNNs. We apply DeepDIG to all pairs of classes in MNIST dataset, namely $\{(s,t) | s,t \in [\text{`0', `1'} \cdots \text{`9'}], s \neq t\}$. Note that there are $\binom{10}{2}=45$ decision boundaries for 10 classes in MNIST dataset. First, we visualize the generated borderline instances to make sure of their quality. Table~\ref{tab:intra-model-visualization} demonstrates some of the borderline instances (chosen randomly) for all pair-wise classes in model $\text{MNIST}_{\text{CNN}}$. As it is evident from this table, DeepDIG manages to generate borderline instances that are visibly very similar to real instances. Now we present the decision boundary characterization results for all pair-wise classes using the measures developed in Section~\ref{sec:characteristics}. Figure~\ref{fig:intra-model-IDC} shows the complexity measure IDC for all 45 decision boundaries. Figure~\ref{fig:intra-model-EDC1} shows the results for measure EDC1 i.e., the average distance from the separating hyperplane between two classes in the embedding space for both borderline and test samples.
Finally, Figure~\ref{fig:intra-model-EDC2} demonstrates the results for measure EDC2 i.e., the accuracy against the linear SVM in the embedding space for both borderline and test samples. Based on the results presented in these figures, we make the following observations.
{}

{}

{}
\begin{itemize}
    \item Similar to what observed in Section~\ref{sec:charac-dnns}, there is a correlation between IDC and $\text{EDC2}_{\text{Borderline}}$. The higher (lower) $\text{EDC2}_{\text{Borderline}}$ is, the lower (higher) IDC is. 
    \item We can observe that different class pairs have different degrees of input space decision boundary complexity (IDC). This depends on how samples are distributed in the input space and how a DNN (here $\text{MNIST}_{\text{CNN}}$) carves out decision regions in this space. Although the exact explanation for this subject (i.e., the sample distribution and model embedding learning) yet to be determined, using the results presented in Figure~\ref{fig:intra-model-IDC}, we can get some insights on how a DNN creates decision regions in the input space. As an example, see the IDC complexity for class `0' against all other classes. The higher value belongs to (`0', `9') while the lowest belongs to the pair (`0', `5'). This seems reasonable since our prior knowledge suggests that digits `0' and `9' have some common patterns e.g., a circle and probably are distributed in a common subspace in the input space whereas image patterns for `5' and `0' are distinct and probably their samples reside in a different subspace.  
    \item Similar reasoning with IDC can be applied to EDC measures as well. The difference in these measures for different class pairs informs us of the degree of difficulty for a DNN to distinguishably project  samples of the two classes in the embedding space. 
\end{itemize}{}

Based on the observations above, we point out a useful use-case of intra-model decision boundary characterization. Usually, deep learning practitioners need to look into the detail of a model and reinforce it against potential future failures. For instance, in some applications, one might be interested to know for which pair of classes a model is more likely to  mis-classify samples, so he/she can take actionable measures e.g., adding more samples. In this regard, intra-model decision boundary characterization can provide us with a full profile of the model strengths and weaknesses for all pair-wise classes and potentially can come useful in taking a more guided decision regarding model debugging/reinforcement.

\section{Related Work}
\label{sec:related}
In recent times, there has been an increasing effort in the machine learning community to propose methods to explain or interpret the results of deep neural networks. We believe one way to better understand DNNs is via taking them out of their ``comfort zone" i.e., where there are corner cases for which a DNNs is `confused' to make a decisive prediction. In this regard, investigating the decision boundary of deep neural networks is an interesting area. In this section, we review several existing papers that studied the decision boundary of DNNs.  

He et al.~\cite{He2018DecisionBA}, similar to our approach, utilized adversarial examples and investigated the decision boundary of DNNs. They considered a large neighborhood around adversarial examples and benign samples and then discovered that such neighborhoods have distinct proprieties e.g., in terms of the distance to the decision boundary. In an attempt to bridge theoretical properties and practical power of DNNs, the authors in~\cite{li2018decision} studied the decision boundary of DNNs. They proved and empirically showed that the last layer of a DNN behaves like a linear SVM. In Section~\ref{sec:linear}, we took advantage of this property of DNNs along with their linear separability property and characterized the decision boundary of DNNs in the embedding space. Fawzi et al.~\cite{fawzi2018empirical} studied topology and geometry of DNNs and showed that DNNs carve out complicated and \emph{connected} classification/decision regions. Furthermore, they investigated the curvature of the decision boundary through which they proposed a method to distinguish benign samples from adversarial ones. Authors of~\cite{moosavi2019robustness} made a connection between the adversarial training and decision boundary and further demonstrated that adversarial training helps in decreasing the curvature of the decision boundary. Yousefzadeh and O'Leary~\cite{yousefzadeh2019investigating} conducted a study to investigate the decision boundary of DNNs. Similar to our algorithm in component (III) (i.e., Algorithm~\ref{alg:middle}) they drew a trajectory between two samples at the opposite sides of the decision boundary and then tried to determine what they call ``flip points" i.e., borderline instances. Then they analyzed different patterns that emerge from connecting different points at the two sides of the decision boundary. In spirit, their method is similar to the first baseline method we considered in Section~\ref{sec:baselines} where we demonstrated that its success rate of generating proper borderline instances is very low for complex multi-class classification problems considered in our experiments. In a recent study, the authors in~\cite{alfarra2020on} introduced tropical geometry as a new perspective to study the decision boundary of DNNs. They used their mathematical findings of decision boundary of DNNs for two applications, namely adversarial examples generation and network pruning.

\section{Conclusion}
\label{sec:conclusion}

Although novel DNN architectures are continuously being developed to achieve better and better performance, the understanding of these models has primarily been ignored. One crucial aspect of DNNs that can help us deepen our understanding of their decision-making behavior is their decision boundaries.  However, this is fairly unexplored in the machine learning literature, and thus in this work, we embarked upon a research inquiry to study the decision boundary of DNNs and investigated their behaviors through the lens of their decision boundaries. To make this feasible, we proposed a new framework called \textbf{Deep} \textbf{D}ecision boundary \textbf{I}nstance \textbf{G}eneration (\textbf{DeepDIG}). DeepDIG utilized an approach based on adversarial example generation and generates two sets of adversarial examples at the opposite sides and near the decision boundary between two classes. Then, aiming at refining and discovering borderline samples, we proposed a method based on the binary search along a trajectory between the two sets of generated adversarial samples. To show the usefulness of DeepDIG, we utilized borderline instances and defined several important measures determining the complexity of the decision boundary between two classes in both input and embedding spaces. 

We conducted extensive experiments and demonstrated the working of DeepDIG. First, we showed that DeepDIG --with very high performance-- can generate borderline instances that are sufficiently close to the decision boundary. Moreover, we experimented on three datasets and two representative DNNs for each dataset and determined the behavior of different DNNs through the characterization of their decision boundaries. Notably, we bridged between the decision boundary in the input space and the embedding space learned by a DNN. Untimely, we applied DeepDIG on the full range of pair-wise classes of MNIST dataset for a DNN and showed how decision boundaries between different pairs of classes differ. There exist several important directions to follow up in the future:

\begin{itemize}
    \item[--] DeepDIG while being effective does not approach generating borderline instances in an end-to-end manner. We plan to formulate the borderline instance generation problem in a unified and end-to-end fashion. In particular, we intend to merge the optimization of DeepDIG into a single optimization formulation.
    \item[--] How to improve the robustness of DNNs against adversarial examples is an emerging and interesting research direction. In this regard, one can integrate borderline instances in the model training e.g., similar to the adversarial training method of~\citep{szegedy2013intriguing} and then investigate if the model is getting robust or not. Moreover, comparing decision boundary characteristics of a robust and non-robust model can potentially provide us with insights about the causes of non-robustness.
    \item[--] DeepDIG was optimized and tested against a continuous data type e.g., images. DeepDIG can be extended to discrete data types as well e.g., texts, graphs, and so on. This is a challenge and needs deliberated considerations. For instance, the distance metric capturing similarity of two samples --see Eq.~\ref{eq:deepdig1} and Eq.~\ref{eq:deepdig2}-- should change appropriately to properly quantify the  \emph{similarity} between two discrete data instances. Adapting ideas from adversarial examples generation methods for texts~\cite{zhang2019generating} seems like a proper avenue to extend DeepDIG to discrete data types.    
\end{itemize}{}

\balance 
\bibliographystyle{ACM-Reference-Format}
\bibliography{references}

\end{document}